\documentclass[11pt]{article}

\usepackage[table]{xcolor}

\usepackage[final]{acl}

\usepackage{times}
\usepackage{latexsym}

\usepackage[T1]{fontenc}

\usepackage{microtype}
\usepackage{url}
\usepackage{booktabs}
\usepackage{xspace}

\usepackage{lineno}

\definecolor{darkblue}{rgb}{0, 0, 0.5}

\usepackage{graphicx} 
\usepackage{subcaption}
\usepackage{booktabs}
\usepackage{amsmath}
\usepackage{amssymb}
\usepackage{amsfonts}
\usepackage{easyeqn}
\usepackage{bm}
\usepackage{natbib}
\usepackage{hyperref}
\usepackage{multirow}
\usepackage{multicol}
\usepackage{paralist}
\usepackage{placeins}

\newcommand{\xv}{\mathbf{x}}
\newcommand{\yv}{\mathbf{y}}

\newcommand{\DC}{\mathcal{D}}

\newcommand{\HC}{\mathcal{H}}

\newcommand{\methodname}{\textsc{Uncertainty-LINE}\xspace}

\newcommand{\xsum}{XSum\xspace}

\newcommand{\gsm}{GSM8K\xspace}

\newcommand{\llama}{Llama 3.1 8B\xspace}

\newcommand{\gemma}{Gemma 2 9B\xspace}

\newcommand{\eurollm}{EuroLLM 9B\xspace}

\hypersetup{colorlinks=true, citecolor=darkblue, linkcolor=darkblue, urlcolor=darkblue}
\title{\methodname: Length-Invariant Estimation of Uncertainty \\ for Large Language Models}
\author{Roman Vashurin\thanks{\hspace{0.2cm}These authors contributed equally.} \quad Maiya Goloburda$^{*}$ \quad Preslav Nakov \quad Maxim Panov \\
Mohamed bin Zayed University of Artificial Intelligence \\
{\small \texttt{\{Roman.Vashurin, Maiya.Goloburda, Preslav.Nakov, Maxim.Panov\}@mbzuai.ac.ae }} }

\begin{document}

\maketitle

\begin{abstract}
  Large Language Models (LLMs) have become indispensable tools across various applications, making it more important than ever to ensure the quality and the trustworthiness of their outputs. This has led to growing interest in uncertainty quantification (UQ) methods for assessing the reliability of LLM outputs. Many existing UQ techniques rely on token probabilities, which inadvertently introduces a bias with respect to the length of the output. While some methods attempt to account for this, we demonstrate that such biases persist even in length-normalized approaches. To address the problem, here we propose \methodname (Length-INvariant Estimation), a simple debiasing procedure that regresses uncertainty scores on output length and uses the residuals as corrected, length-invariant estimates. Our method is post-hoc, model-agnostic, and applicable to a range of UQ measures. Through extensive evaluation on machine translation, summarization, and question-answering tasks, we demonstrate that \methodname consistently improves over even nominally length-normalized UQ methods uncertainty estimates across multiple metrics and models. We release our code publicly.\footnote{\href{https://github.com/stat-ml/uncertainty-line}{https://github.com/stat-ml/uncertainty-line}}
\end{abstract}


\section{Introduction}
  Large Language Models (LLMs) have become essential in a wide range of applications. However, despite their impressive capabilities, LLMs can sometimes generate misleading or outright incorrect information. Given their widespread adoption in critical domains, ensuring the reliability of their responses has become a pressing concern. This has led to growing interest in uncertainty quantification (UQ) to measure the confidence in model-generated output~\citep{Gal2016Uncertainty, Hu2023UncertaintyIN, kotelevskii2025risk}.

  Many existing UQ methods for LLMs rely on token-level probabilities produced by LLM itself. However, the log-probabilities of the generated tokens in autoregressive models are summed over the sequence, meaning that the total sequence score becomes increasingly negative as length increases, leading to unreliable estimates~\citep{murray-chiang-2018-correcting, Braverman2019CalibrationER, zhao2023calibrating}. Some methods, such as perplexity and mean token entropy, have been proposed as length-normalized uncertainty measures~\cite{fomicheva-etal-2020-unsupervised}. Alternatively, some work has focused on calibrating model confidence scores using post-hoc methods or on reformulating uncertainty estimation at the token level~\citep{pmlr-v239-ren23a, zhao2023calibrating, 1019493}. While this yields improvements, it often requires additional supervision, architectural changes, or tuning for specific tasks.

  Here, we propose a simple and effective method for detrending uncertainty estimates with respect to output length, in both unsupervised and minimally supervised settings.
  It is post-hoc, model-agnostic, and applicable across a range of uncertainty measures. Our key contributions are as follows:
  \begin{itemize}
    \item We demonstrate that uncertainty estimation metrics exhibit length bias, even when length-normalization is applied; see Section~\ref{sec:methods:length_bias}.
    
    \item We propose Uncertainty-\underline{L}ength \underline{IN}variant \underline{E}stimation (\methodname), a simple unsupervised detrending approach that fits a regression between uncertainty scores and output length, and uses the residuals as uncertainty estimates. We also formulate a supervised extension for cases where the output length correlates with quality; see Section~\ref{sec:method}.
    
    
    \item We evaluate our approach on machine translation, summarization, and mathematical reasoning tasks, showing improved performance of the uncertainty estimates; see Section~\ref{sec:experiments}.
  \end{itemize}


\section{Related Work}

\paragraph{Length Bias in Sequence Likelihood.}
  Early work in sequence generation noted that sequence-level likelihood (the joint probability of an output) is biased with regards to the output length. Neural models often assign disproportionately low probabilities to longer outputs, causing shorter sequences to appear ``more likely''~\citep{murray-chiang-2018-correcting, adiwardana2020humanlikeopendomainchatbot}. More recently, \citet{santilli-etal-2025-revisiting} demonstrate that such length effects also impact uncertainty evaluation, highlighting the need for explicit length bias removal. 

\paragraph{Length-Normalized Uncertainty Measures.}
  A common remedy is to use the average per token confidence score (i.e. normalize overall confidence by sequence length) instead of the raw sum. For example, \textit{Perplexity} or mean log-probability per token is often used as a sequence-level confidence score instead of raw joint probability~\citep{fomicheva-etal-2020-unsupervised}. Another measure is \textit{Mean Token Entropy}: the average entropy of the model's predictive distribution at each time step of the output~\citep{fomicheva-etal-2020-unsupervised}. 
 On the other hand, \textit{Monte Carlo Sequence Entropy} (MCSE) and its length-normalized variant \textit{Monte Carlo Normalzied Sequence Entropy} (MCNSE) measure uncertainty by estimating the entropy of predictive distribution using multiple outputs sampled via stochastic decoding~\citep{malinin2020uncertainty, kuhn2023semantic}. Lastly, TokenSAR provides a length-normalized measure that reweights token log-probabilities based on their importance to the meaning of an output~\citep{duan-etal-2024-shifting}. However, naively dividing by length can overcorrect: it can overly penalize shorter sequences, flipping the bias in the opposite direction~\citep{1019493}. These findings show that while length normalization is a useful tool, it needs careful consideration to avoid introducing a new bias.

\paragraph{Uncertainty Calibration.}
  Recent work~\citep{pmlr-v239-ren23a} shows that sequence-level confidence scores remain poorly calibrated with output quality, even after applying length normalization. To address this problem, various methods have been proposed, including token-level self-evaluation, sequence likelihood calibration, language model cascades that leverage token-level uncertainty and post-hoc correctors for uncertainty estimates~\citep{pmlr-v239-ren23a, zhao2023calibrating, 1019493, li-etal-2025-towards}. 
  
  While such techniques improve the alignment between model confidence and human judgment, they often require additional supervision or task-specific tuning and, crucially, do not directly target the problem of length bias.

\paragraph{Non-token likelihood based uncertainty measures.}
  Consistency uncertainty measures have emerged as a way to bypass token-level scores entirely~\citep{fomicheva-etal-2020-unsupervised, lin2023generating, kuhn2023semantic}. By evaluating uncertainty as a level of agreement between sampled outputs, they provide a length-agnostic confidence estimate. However, while these measures are designed to be length-invariant, their practical application encounters several challenges. First, these methods rely on sampling multiple outputs from the language model to capture the distribution of possible generations. This sampling process is computationally intensive, especially for large models. Secondly, implementations often depend on pre-trained models, to assess semantic similarity between generated outputs. While effective for certain tasks, these models are frequently trained on shorter texts, leading to less reliable estimates for long generations. 
  
  Another approach that does not rely on token likelihoods is verbalized uncertainty, where models express their confidence in natural language. However, studies have shown that LLMs often exhibit overconfidence in their verbalized uncertainty assessments~\citep{xiong2024can}. This overconfidence suggests that without proper calibration or fine-tuning, verbalized uncertainty may not reliably reflect true predictive uncertainty~\citep{liu-etal-2024-llms-learn-uncertainty}. Therefore, while verbalized uncertainty offers a promising, length-invariant alternative, its practical utility is limited unless accompanied by effective calibration strategies. 

  These works highlight the limitations of existing UQ methods, including length-normalized measures in addressing the length bias problem.


  \begin{figure*}[t!]
    \centering
    
    \begin{subfigure}{0.326\textwidth}
      \includegraphics[width=\linewidth]{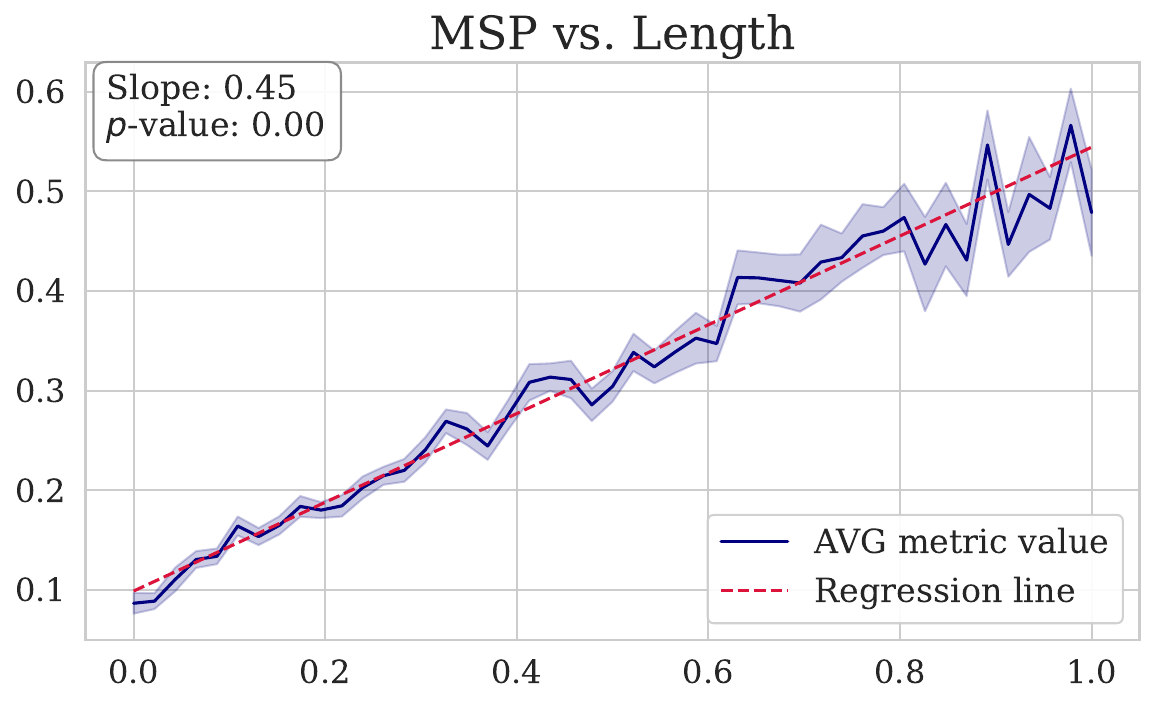}
    \end{subfigure}
    \begin{subfigure}{0.326\textwidth}
      \includegraphics[width=\linewidth]{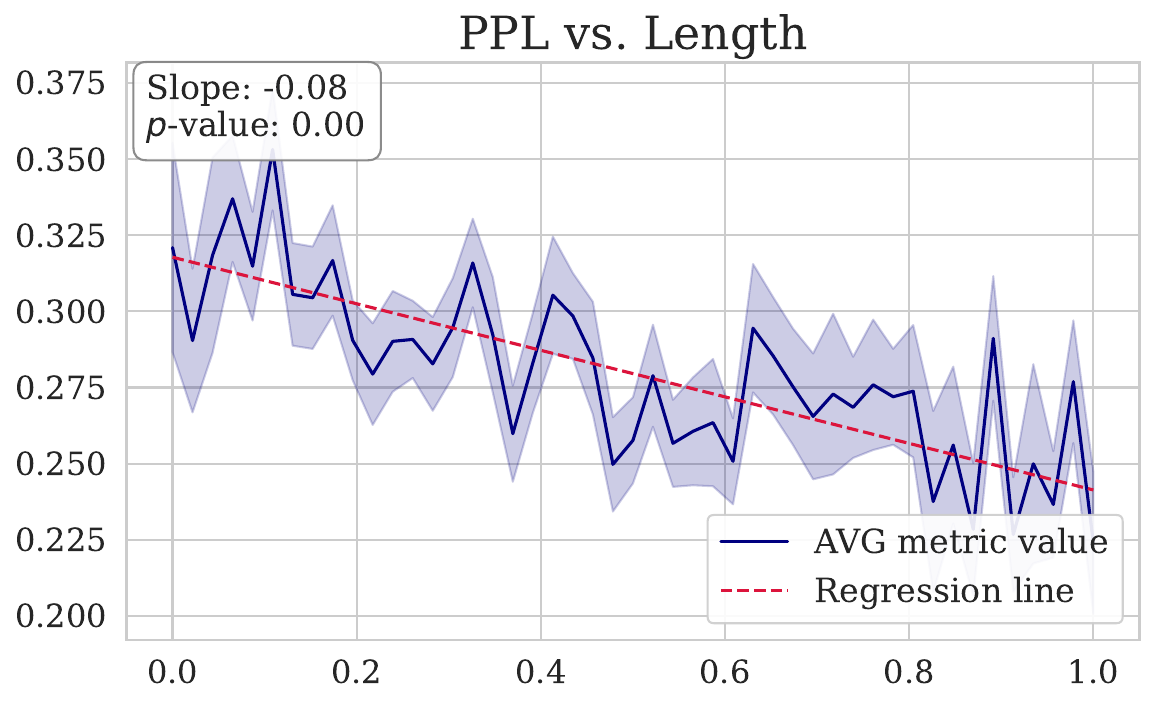}
    \end{subfigure}
    \begin{subfigure}{0.326\textwidth}
      \includegraphics[width=\linewidth]{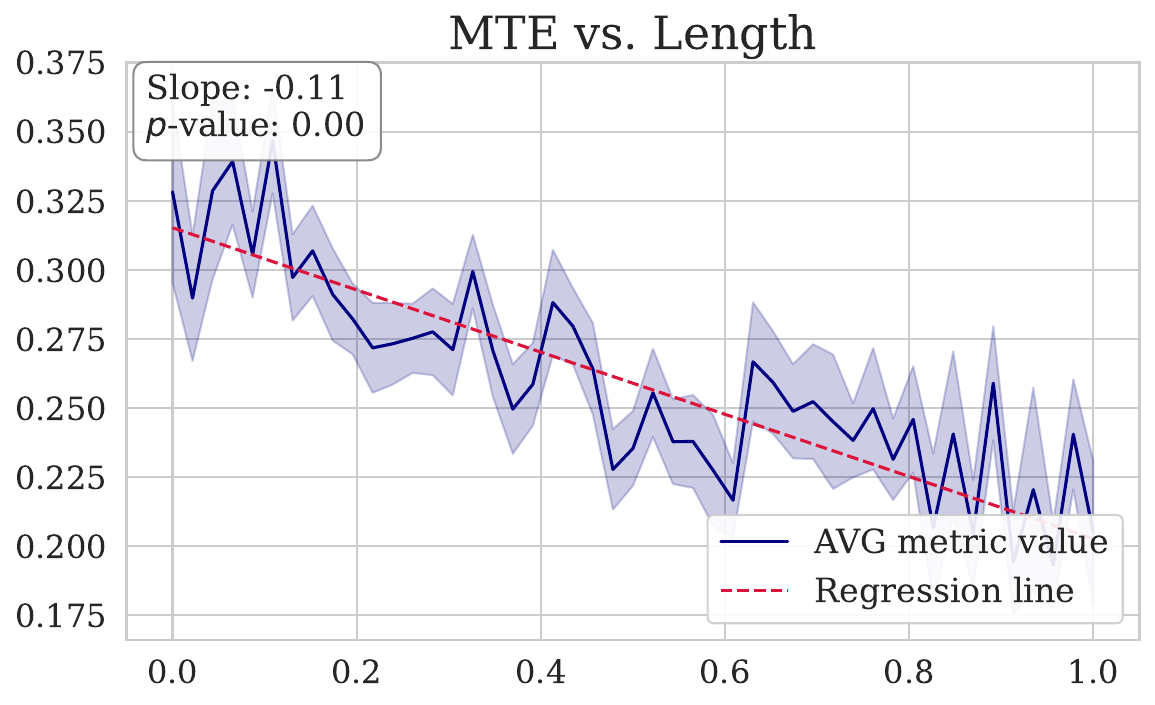}
    \end{subfigure}
    
    \begin{subfigure}{0.326\textwidth}
      \includegraphics[width=\linewidth]{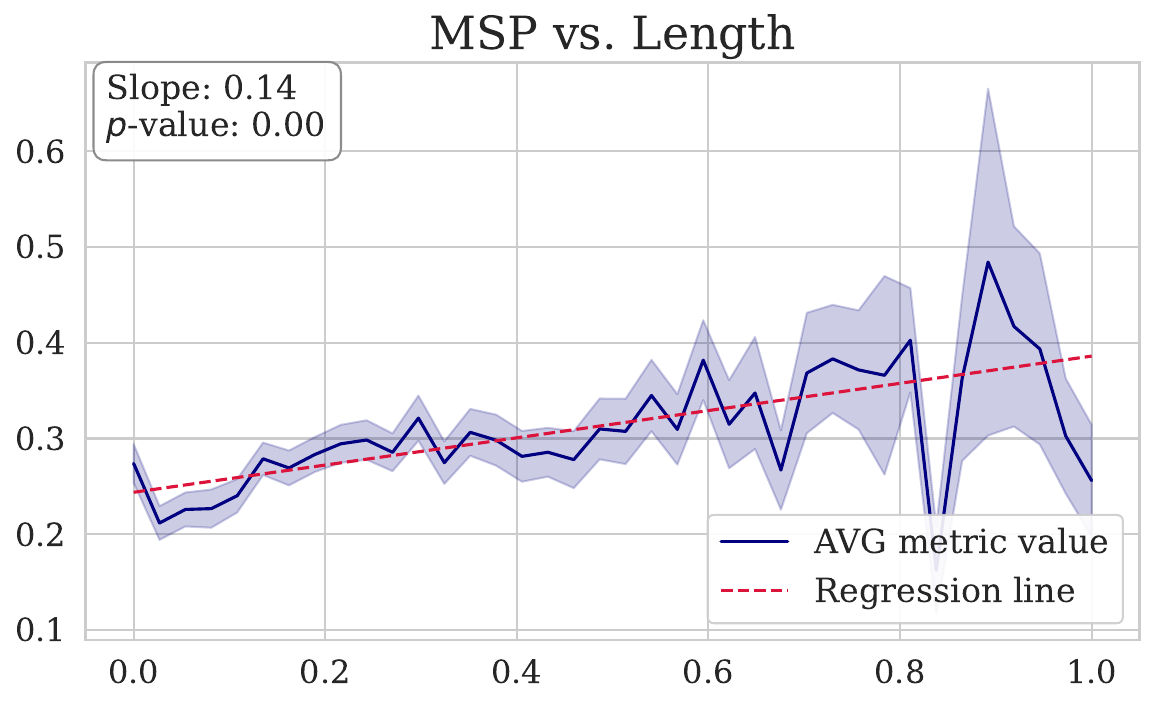}
    \end{subfigure}
    \begin{subfigure}{0.326\textwidth}
      \includegraphics[width=\linewidth]{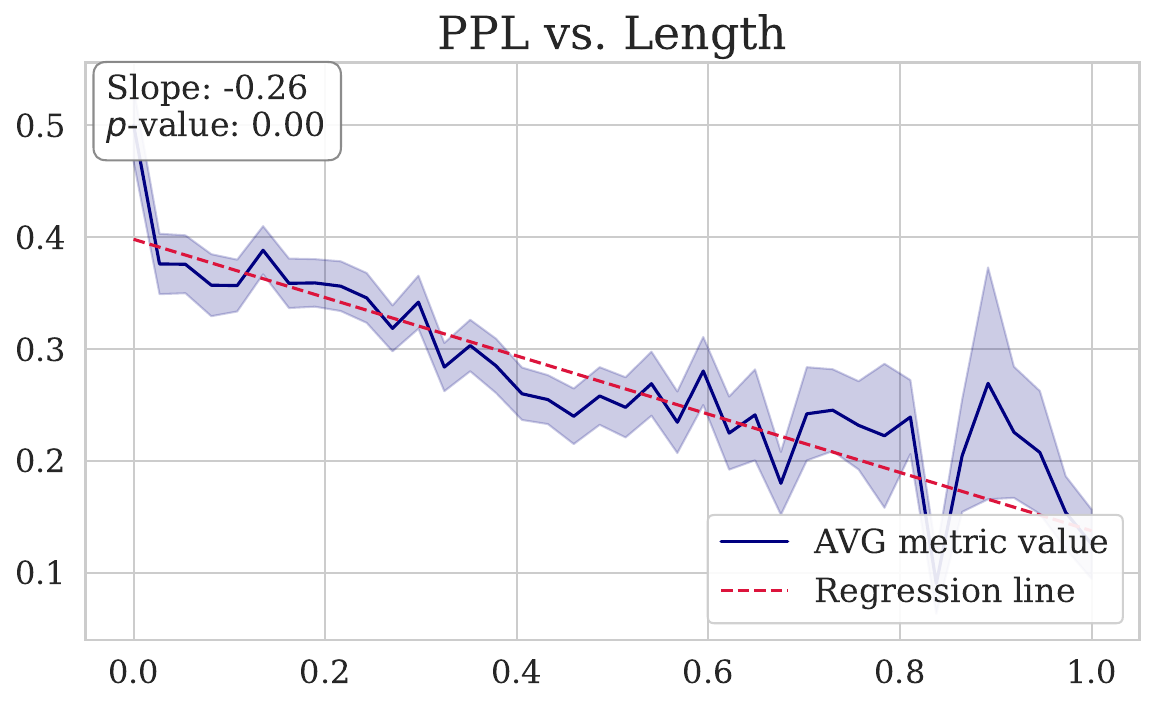}
    \end{subfigure}
    \begin{subfigure}{0.326\textwidth}
      \includegraphics[width=\linewidth]{figures/final_plots_for_main_part/xsum_Perplexity_llama_train.pdf}
    \end{subfigure}

    \begin{subfigure}{0.326\textwidth}
      \includegraphics[width=\linewidth]{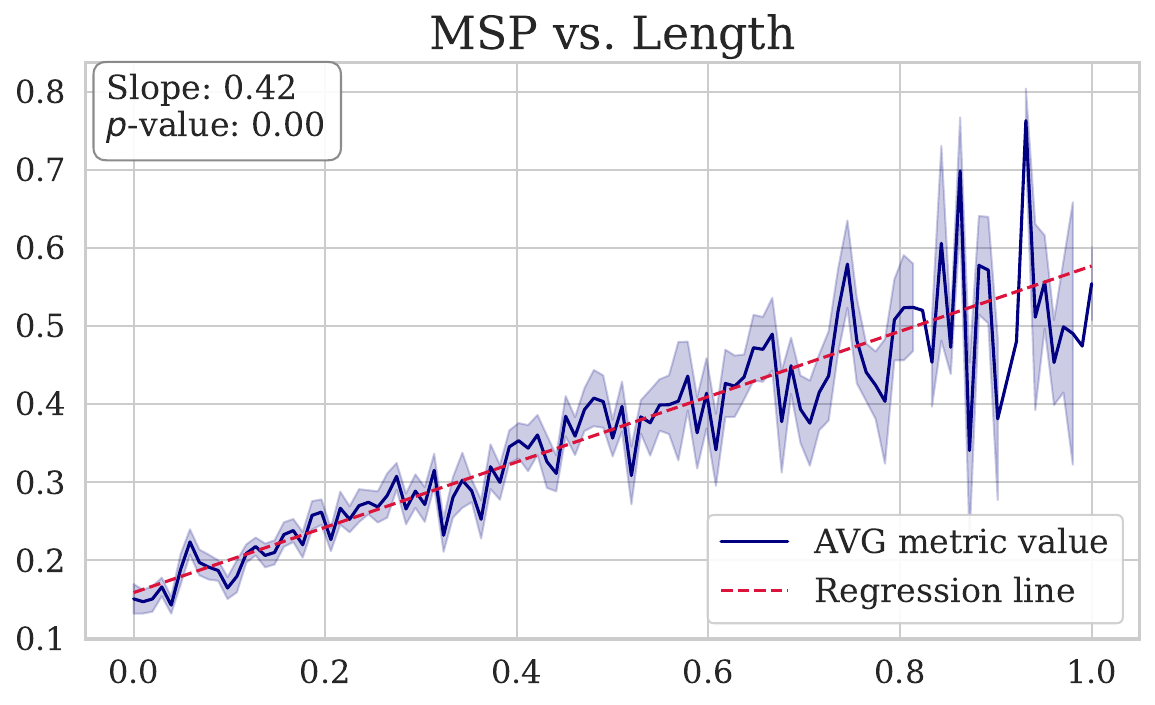}
    \end{subfigure}
    \begin{subfigure}{0.326\textwidth}
      \includegraphics[width=\linewidth]{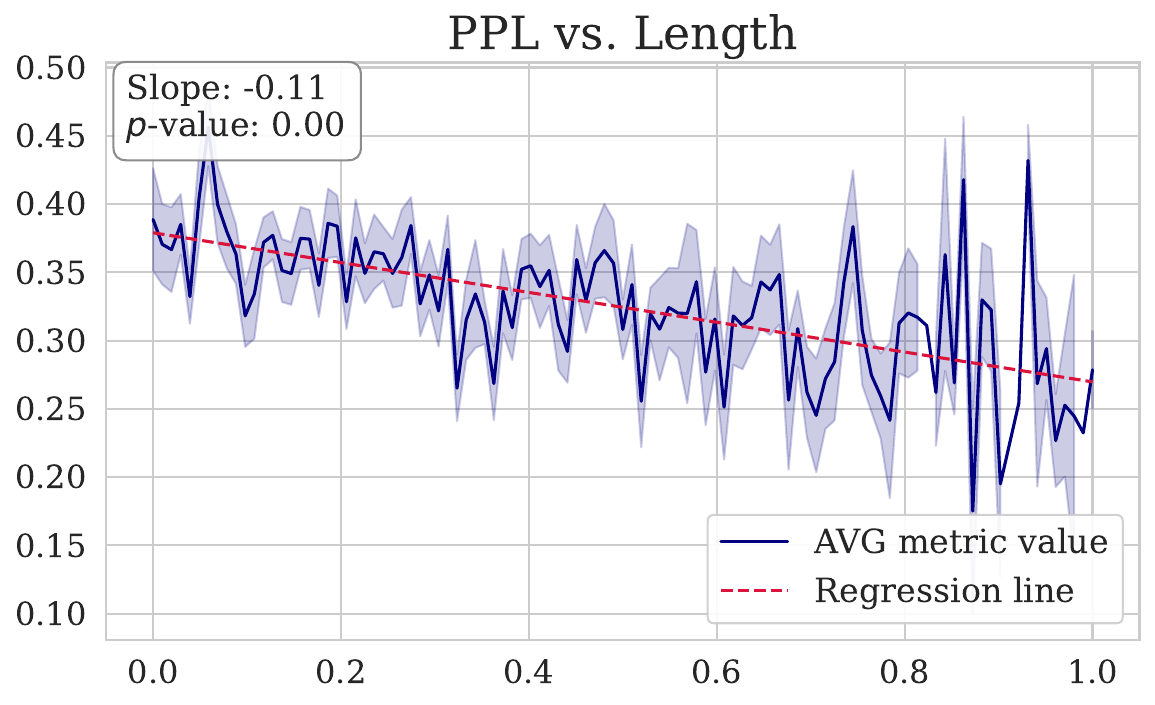}
    \end{subfigure}
    \begin{subfigure}{0.326\textwidth}
      \includegraphics[width=\linewidth]{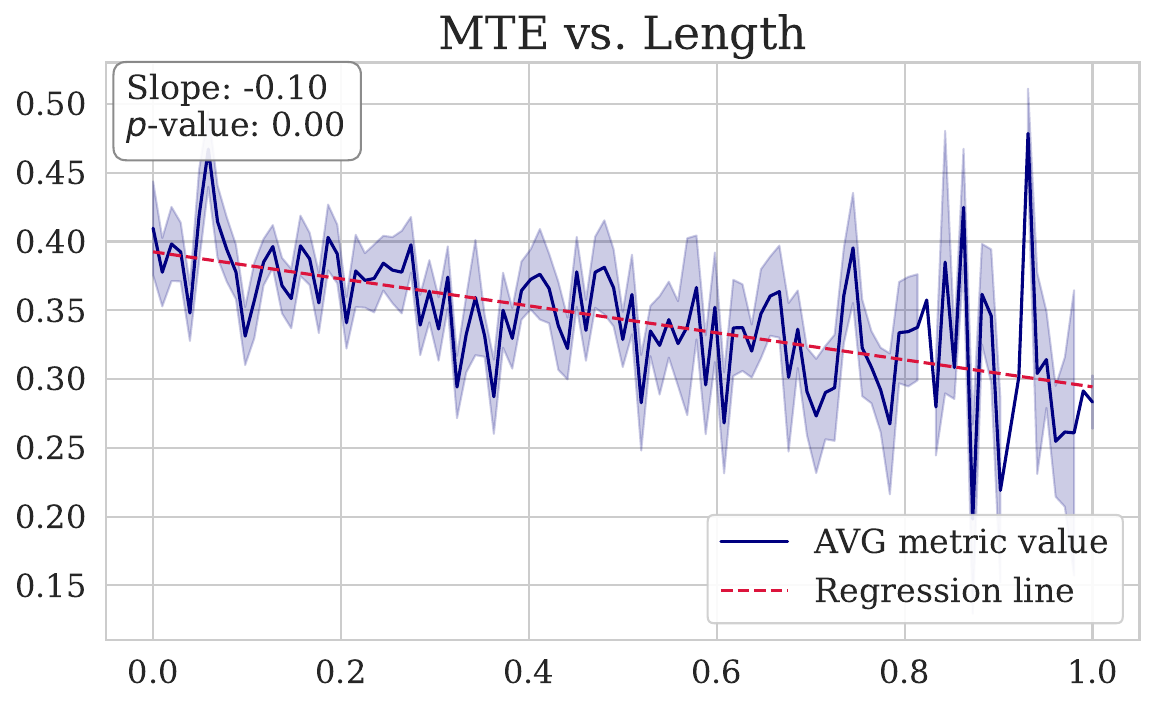}
    \end{subfigure}
    \caption{Trends in UQ scores with respect to normalized sequence length for WMT14 De-En, \gsm and \xsum (Model -- \llama). Each subplot shows a linear regression fit over binned scores. The \textbf{slope} indicates the strength and direction of correlation, \textbf{p-value} reflects statistical significance.}
  \label{fig:ue_metrics}
  \end{figure*}

\section{Length Bias in Uncertainty Measures and Quality Metrics}
\label{sec:methods:length_bias}
  We start by quantifying the degree to which uncertainty quantification (UQ) measures applied to the output of \llama~\citep{grattafiori2024llama3herdmodels} model exhibit dependency on the length of generated sequences. We do that on a comprehensive set of tasks that imply varying length of the generated response: neural machine translation (NMT), mathematical reasoning, and abstractive summarization. 
  
  For NMT, we used four language pairs from WMT14 and WMT19, \gsm for mathematical reasoning, and \xsum for ATS~\citep{bojar-etal-2014-findings, barrault-etal-2019-findings, cobbe2021gsm8k, Narayan2018DontGM}. We performed estimation on the random subset of 2000 points from each dataset.

  The level of dependency is expressed as the slope of a linear regression fit by ordinary least squares (OLS), with the UQ values as the response variable and the generated sequence length as a predictor. The significance of the obtained linear trend was assessed using the Wald test and corresponding $p$-value was calculated along with the slope. 

\paragraph{Uncertainty Measures are Strongly Length-Dependent.}
  Figure~\ref{fig:ue_metrics} presents results on one of the machine translation datasets (WMT14 De-En), \xsum and \gsm. The UQ measures show clear and significant trends. For the Maximum Sequence Probability (MSP), the average UQ score increases with sequence length, indicating that longer generations are assigned lower model confidence, which is potentially misleading, as longer outputs may simply reflect more confident token-level predictions. Both Perplexity (PPL) and Mean Token Entropy (MTE) exhibit the opposite trend, with average uncertainty decreasing as length increases. 
  
  This is notable, as both measures were designed to normalize for length-related effects, yet the trends persist. In the majority of cases, the $p$-values of regression coefficients are below 0.05, confirming that the observed relationships are statistically significant. This highlights a key concern: although these UQ measures are widely used, they may conflate uncertainty with sequence length in practice.

  Results for the rest of the datasets, models and UQ methods can be found in Appendix~\ref{sec:length_effects_ue}.
  
  \begin{figure*}[ht!]
    \centering
    \begin{subfigure}{0.329\textwidth}
      \includegraphics[width=\linewidth]{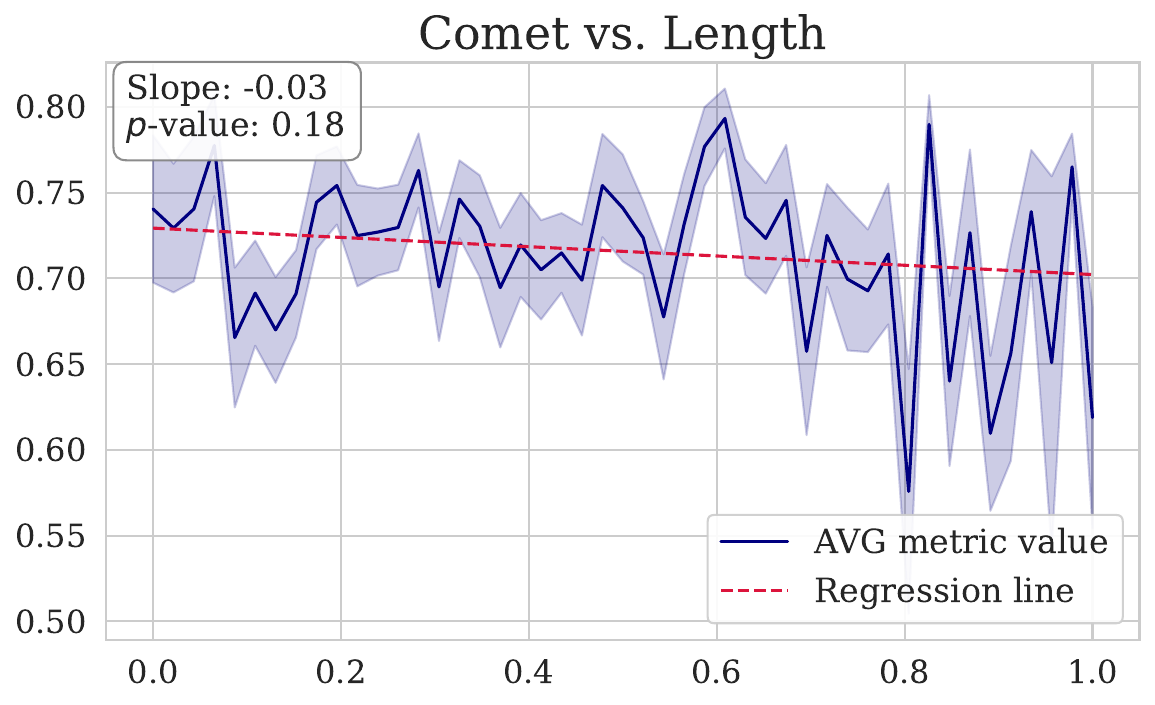}
    \end{subfigure}
    \begin{subfigure}{0.329\textwidth}
      \includegraphics[width=\linewidth]{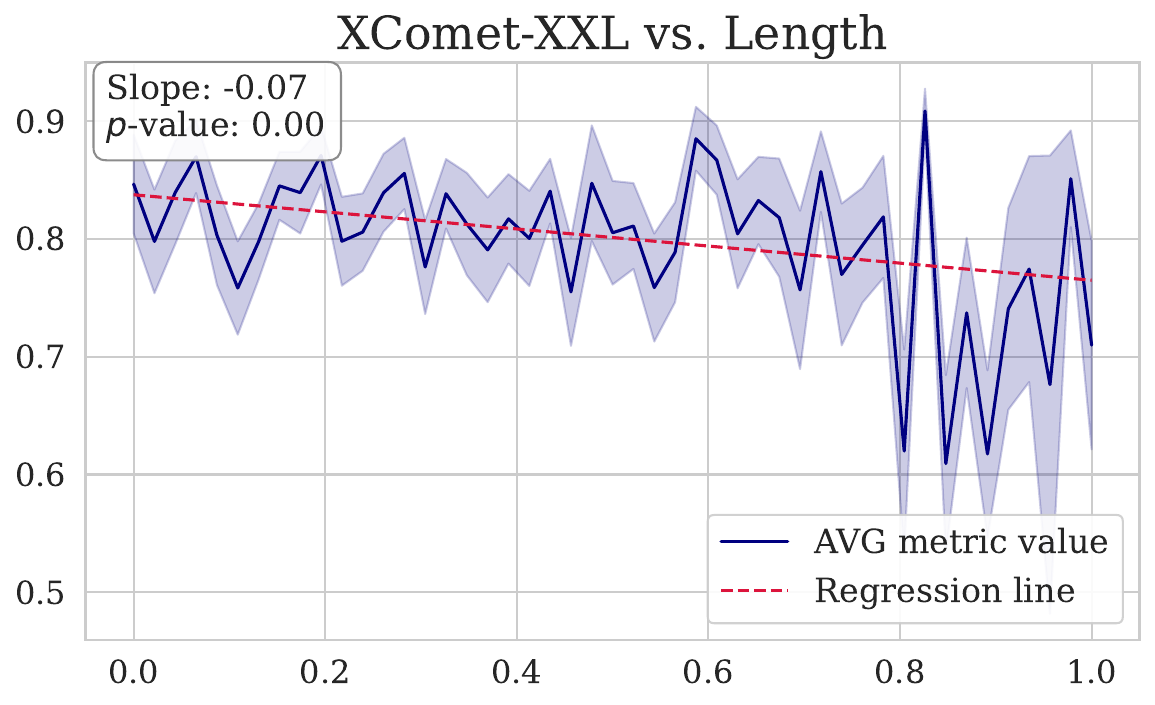}
    \end{subfigure}
    \begin{subfigure}{0.329\textwidth}
      \includegraphics[width=\linewidth]{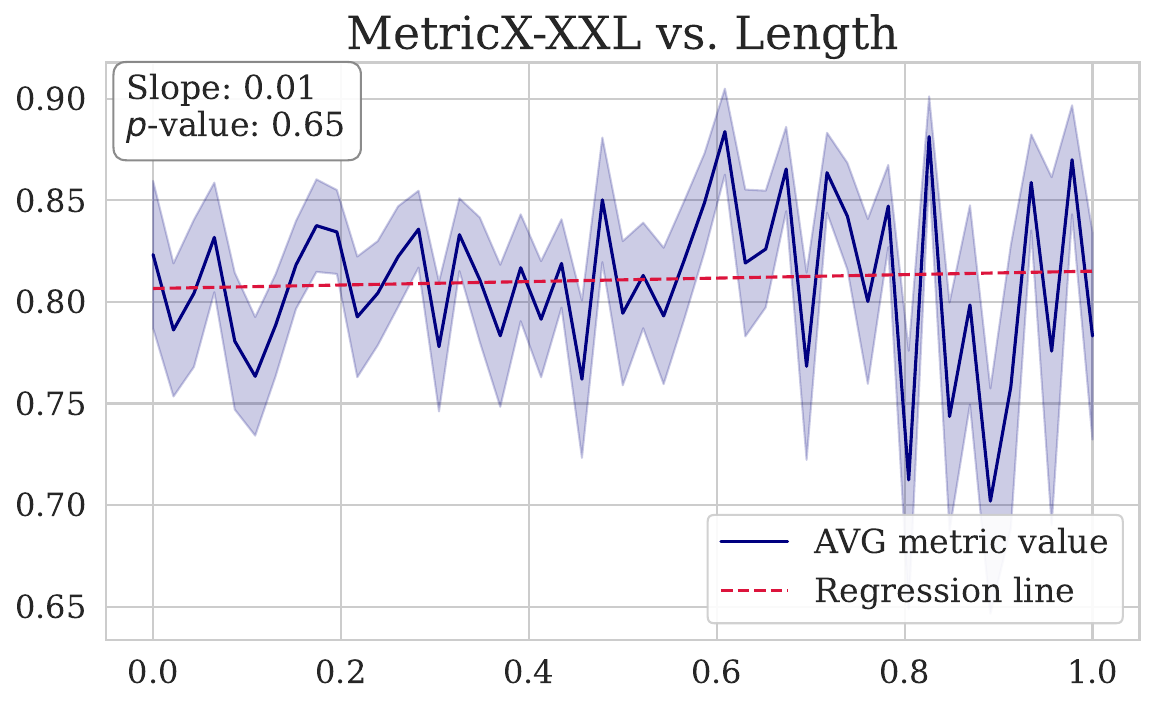}
    \end{subfigure}
    \caption{Trends for Comet WMT 22, XComet XXL and Metric X XXL scores with respect to normalized sequence length for WMT14 De-En machine translation dataset (Model - \llama). Each subplot shows a linear regression fit over binned scores. The \textbf{slope} indicates the strength and direction of correlation, \textbf{p-value} reflects statistical significance.}
  \label{fig:nmt_trends}
  \end{figure*}

  \begin{figure*}[ht!]
    \centering
    \begin{subfigure}{0.495\textwidth}
      \includegraphics[width=\linewidth]{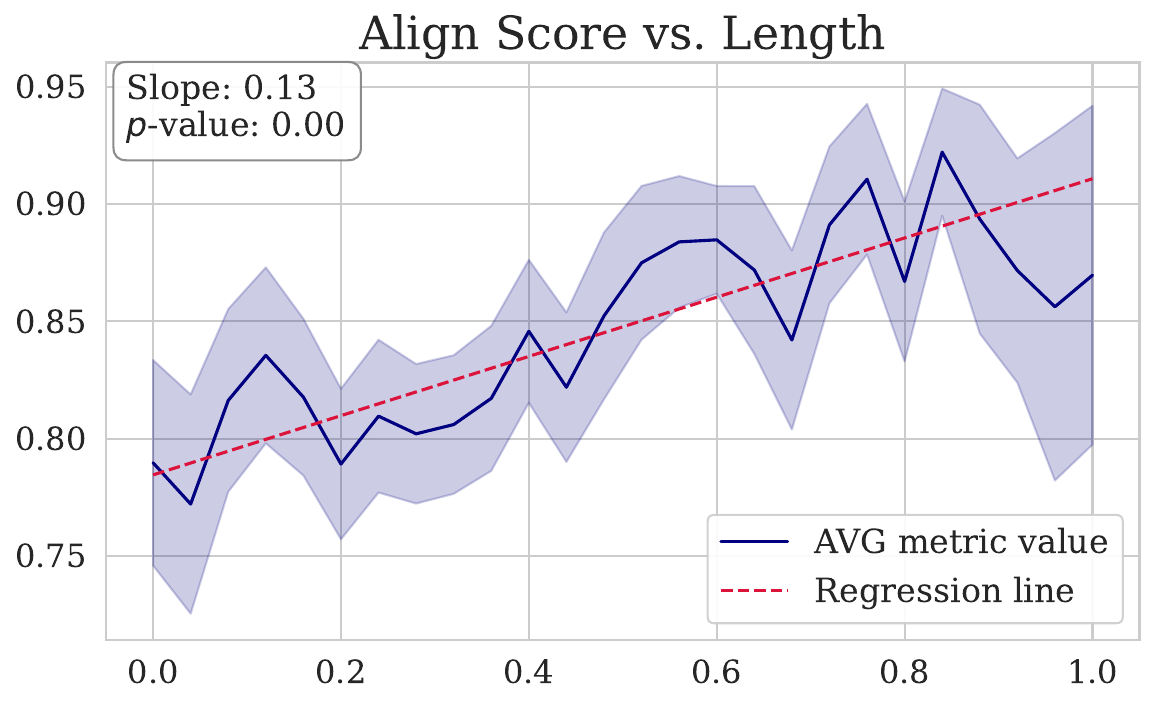}
    \end{subfigure}
    \begin{subfigure}{0.495\textwidth}
      \includegraphics[width=\linewidth]{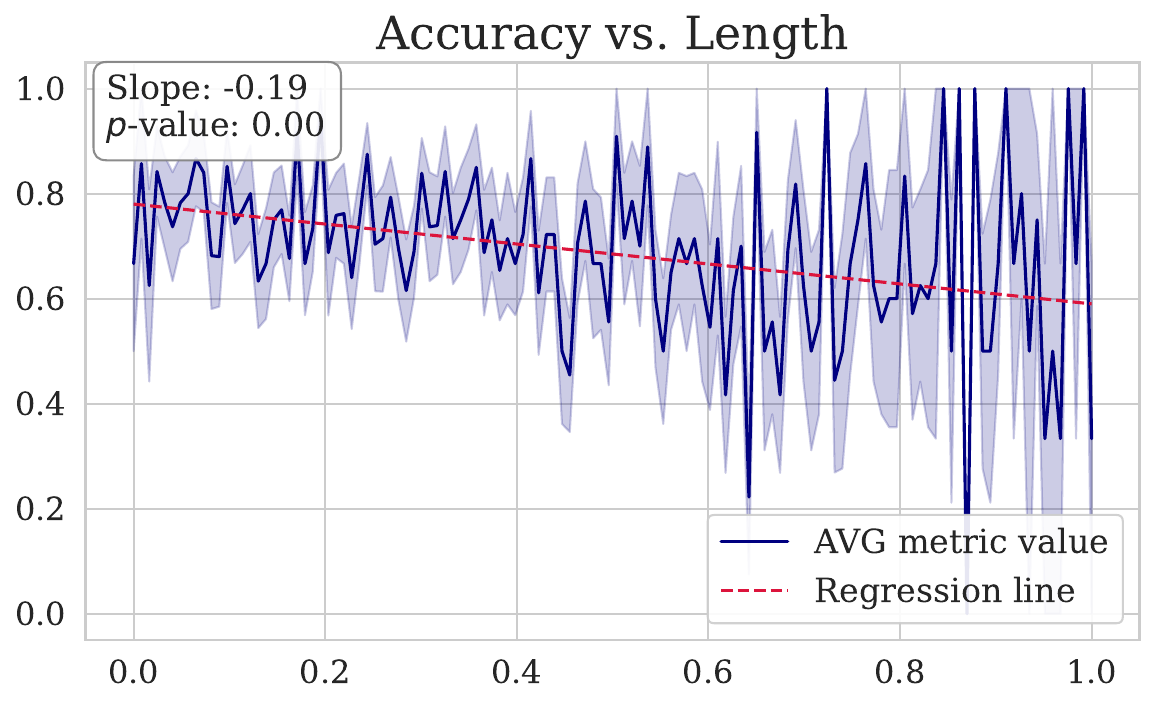}
    \end{subfigure}
    
    \caption{Trends in AlignScore and Accuracy scores with respect to normalized sequence length for \xsum and \gsm datasets respectively (Model -- \gemma). Each subplot shows a linear regression fit over binned scores. The \textbf{slope} indicates the strength and direction of correlation, \textbf{p-value} reflects statistical significance. }
  \label{fig:other_metrics_trends}
  \end{figure*}

\paragraph{Quality Metrics for Machine Translation are Largely Length-Agnostic.}
  Performance of the UQ method is largely defined by the extent of its correlation with some meaningful measure of prediction quality. Thus, having assessed relationship of UQ with generation length, we perform a similar analysis of the behavior of several quality metrics for the same selection of tasks.

  For NMT, Figure~\ref{fig:nmt_trends} shows Comet WMT 22, XComet XXL and Metric X XXL scores over normalized generation lengths~\citep{comet, xcomet, metricx}. Across datasets, the fitted linear regression lines exhibit near-zero slopes, and the associated $p$-values exceed the standard 0.05 threshold in the majority of cases. 
  
  This suggests that there is no statistically significant correlation between generation length and quality scores. In the few cases where a trend is observed, the magnitude of the slope is quite small, especially when compared to the slopes observed in uncertainty measures. This is expected: in machine translation, the length of the output is strongly determined by the length of the input sentence, and thus variation in output length does not reflect variation in task difficulty or translation quality. We conclude that translation quality metrics are effectively length-invariant for given datasets and robust to variations in output length.

\paragraph{Quality Metrics for Summarization and Mathematical Reasoning are Length-Dependent.}
  Figure~\ref{fig:other_metrics_trends} reports quality metric trends for \xsum (summarization) and \gsm (arithmetic question answering). Unlike machine translation, these tasks show noticeable correlations between quality scores and output length: quality (measured by Accuracy) tends to decrease with longer outputs in \gsm, while in summarization (\xsum), quality (measured by AlignScore~\citep{zha-etal-2023-alignscore}) slightly increases with length. 
  
  This observation aligns with intuition: in \gsm, more complex problems often require longer reasoning chains, increasing the likelihood of errors; in \xsum, longer summaries may better capture essential content, improving quality scores. These findings suggest that, unlike for translation task, it is important to take into account that the quality is correlated with the length of an output, even if not as strongly as uncertainty measures.

  For detailed results on the relationship between quality metrics and generation length, refer to Appendix~\ref{sec:length_effects_quality}.


\section{Method}
\label{sec:method}
  Building on our analysis of length‐dependent trends in Section~\ref{sec:methods:length_bias}, we introduce a simple post‐hoc correction method \methodname designed to remove spurious correlations between uncertainty scores and output sequence length. Our approach consists of two main stages: fitting a bias model on unlabeled data and debiasing uncertainty estimates at inference time.

\subsection{Problem Setup}
\label{sec:method:setup}
  Consider a dataset
  \[
    \DC_{\mathrm{train}} = \{\xv_i, \yv_i, u_i = u(\yv_i)\}_{i = 1}^N,
  \]
  where \(\xv\) is the input sequence, \(\yv\) is the model output, and \(u(\yv)\) is the associated uncertainty score. We consider two settings. In the first one, there is no systematic correlation between output length and quality (e.g., in machine translation). In the second one, quality is length-dependent (e.g., in summarization or reasoning tasks), and we use quality labels $\yv_i^*$ for \(\DC_{\mathrm{train}}\) to fit a quality–length model. In this case, labels are used for estimating the task-specific length-induced quality trend.

  Our goal is to adjust uncertainty estimates \(u(\yv)\) such that they are no longer spuriously correlated with the length \(\lvert \yv \rvert\) of the generated sequence.

\subsection{Debiasing of UQ Scores}
\label{sec:method:modeling}
  On the training set \(\DC_{\mathrm{train}}\), we model the relationship between uncertainty scores and output lengths by fitting a simple linear regression:
  \begin{equation}
    \hat{u}(\yv) = \alpha \lvert \yv \rvert + \beta.
  \label{eq:debiasing:fit}
  \end{equation} 
  We adopt a linear regression model for debiasing based on the empirical observations in Section~\ref{sec:methods:length_bias}. As shown in Figure~\ref{fig:ue_metrics}, UQ scores such as MSP, PPL, and MTE exhibit strong linear trends with respect to output length. In addition, Appendix~\ref{sec:poly_detr} presents the results of experiments using second- and third-degree polynomials. As shown, these do not provide any significant or consistent improvement over the linear fit. All of this suggests that a linear correction is both sufficient and preferable to avoid overfitting. This model captures the systematic trend between uncertainty scores and sequence length, which we aim to remove. 

  At inference time, we apply the learned linear model to debias raw uncertainty scores on the test set. To achieve this, for each test example, we compute a \emph{length-debiased} uncertainty score by subtracting the length‐predicted component from its raw score:
  \begin{equation}
    u^{\mathrm{deb}}(\yv) = u(\yv) - \hat{u}(\yv).
  \label{eq:debiasing:apply}
  \end{equation}
  This subtraction step is equivalent to computing the residuals from the fitted regression, a standard approach in statistics for removing systematic linear trends from data. 

\paragraph{Preserving Quality-Based Trends.}
  However, not all length effects are spurious. As demonstrated in Section~\ref{sec:methods:length_bias}, for tasks like summarization or QA, quality of an output is correlated with its length and final uncertainty score should reflect this. To preserve this meaningful length-dependence, we explicitly model how quality varies with length.

  Let \(\yv^*\) be the gold-standard reference for the input \(\xv\) and let us consider the quality score \(q(\yv, \yv^*)\) between \(\yv^*\) and model generation \(\yv\). We treat the negated quality score \(-q(\yv, \yv^*)\) as a proxy for ground-truth uncertainty, assuming that higher-quality outputs are less uncertain. We then fit a second linear model on an extended dataset \(\DC_{\mathrm{train}}^* = \{\xv_i, \yv_i, q_i = q(\yv_i, \yv_i^*)\}_{i = 1}^N\):
  \begin{equation}
    \hat{q}(\yv) = \delta \lvert \yv \rvert + \gamma,
    \label{eq:debiasing:qualitytrend}
  \end{equation}
  where \(\delta\) and \(\gamma\) describe the quality-induced length effect.

  At test time, we debias the uncertainty score by subtracting the spurious trend \(\hat{u}(\yv)\) and adding back the quality-based trend \(-\hat{q}(\yv)\):
  \begin{equation}
    u^{\mathrm{deb}}(\yv) = u(\yv) - \hat{u}(\yv) - \hat{q}(\yv).
  \label{eq:debiasing:final}
  \end{equation}
  This procedure retains task-relevant length dependencies while removing confounding biases unrelated to quality. While this method requires reference-based quality scores at training time, it can be applied to unlabeled data at inference, making it practical for real-world settings.


\section{Experiments}
\label{sec:experiments}
\subsection{Experimental Setup}
  To perform our evaluation, we extended the \texttt{LM-Polygraph} library~\citep{fadeeva2023lm,vashurin2025benchmarking} by integrating our debiasing approach into its evaluation framework. The library provides built-in implementations of various uncertainty metrics, making it a convenient foundation for conducting experiments and ensuring consistent comparisons across methods.

\paragraph{Datasets.}
  Tasks and dataset selection was based on the need for long-form generation tasks, in order to meaningfully analyze the relationship between generation length, uncertainty, and quality in tasks where length varies naturally. We conduct our experiments on a set of machine translation benchmarks from the WMT14 and WMT19 shared tasks~\citep{bojar-etal-2014-findings, barrault-etal-2019-findings}. 
  
  Specifically, we evaluate on four language pairs from each benchmark: Cs–En, De–En, Fr–En, and Ru–En from WMT14; and De–En, Fi–En, Lt–En, and Ru–En from WMT19. Each dataset includes source inputs, model-generated translations, and reference outputs for evaluation. In addition to translation, we include two open-ended generation tasks: \xsum for abstractive summarization and \gsm for arithmetic question answering~\citep{Narayan2018DontGM, cobbe2021gsm8k}. \xsum consists of document-summary pairs focused on single-sentence summaries of BBC articles. \gsm contains grade-school-level math word problems requiring multi-step reasoning to generate a final numerical answer. 

  For translation, we apply the debiasing formulation described in equation~\eqref{eq:debiasing:apply}. For summarization and mathematical reasoning, we use equation~\eqref{eq:debiasing:final}. These formulations are chosen to align with the specific characteristics of each task: as translation quality is largely independent of output length, it is sufficient to simply remove the length-induced trend. In contrast, for tasks like summarization and mathematical reasoning, where quality often correlates with length, we retain the quality-associated component and eliminate only the spurious bias.

\paragraph{Models.}
  We use three base versions of multilingual generative language models to generate outputs for all datasets: \llama~\citep{grattafiori2024llama3herdmodels}, \gemma~\citep{Riviere2024Gemma2I}, and \eurollm~\citep{MARTINS202553}. These models were selected to represent a diversity of open-source architectures. While \llama and \gemma are used across all tasks, \eurollm is only evaluated on translation datasets due to its limited support for an open-ended generation tasks such as summarization and mathematical reasoning.

\paragraph{UQ measures.}
  We evaluate the following uncertainty quantification (UQ) measures, commonly used in sequence generation tasks: Maximum Sequence Probability (MSP), Perplexity (PPL), Mean Token Entropy (MTE), Monte Carlo Sequence Entropy (MCSE), Monte Carlo Normalized Sequence Entropy (MCNSE), Lexical Similarity with Rouge L as similarity function (LSRL) and TokenSAR~\cite{duan-etal-2024-shifting}. They capture different aspects of model uncertainty: MSP and MCSE reflect aggregate confidence in the full sequence and are not length-normalized, while PPL, MTE and MCNSE explicitly normalize for output length. 
  
  TokenSAR also normalizes for output length, but while accounting for each token's relevance weight, i.e.,~how important the token is to the overall meaning of the generation. LSRL is based on sample diversity, providing non-likelihood-based perspective on length bias in uncertainty. Detailed description of each measure is given in Appendix~\ref{sec:appendix_methods}.

  \begin{figure}[t]
    \centering
    \includegraphics[trim={0.cm 0.cm 0.cm 0.cm},clip,width=\linewidth]{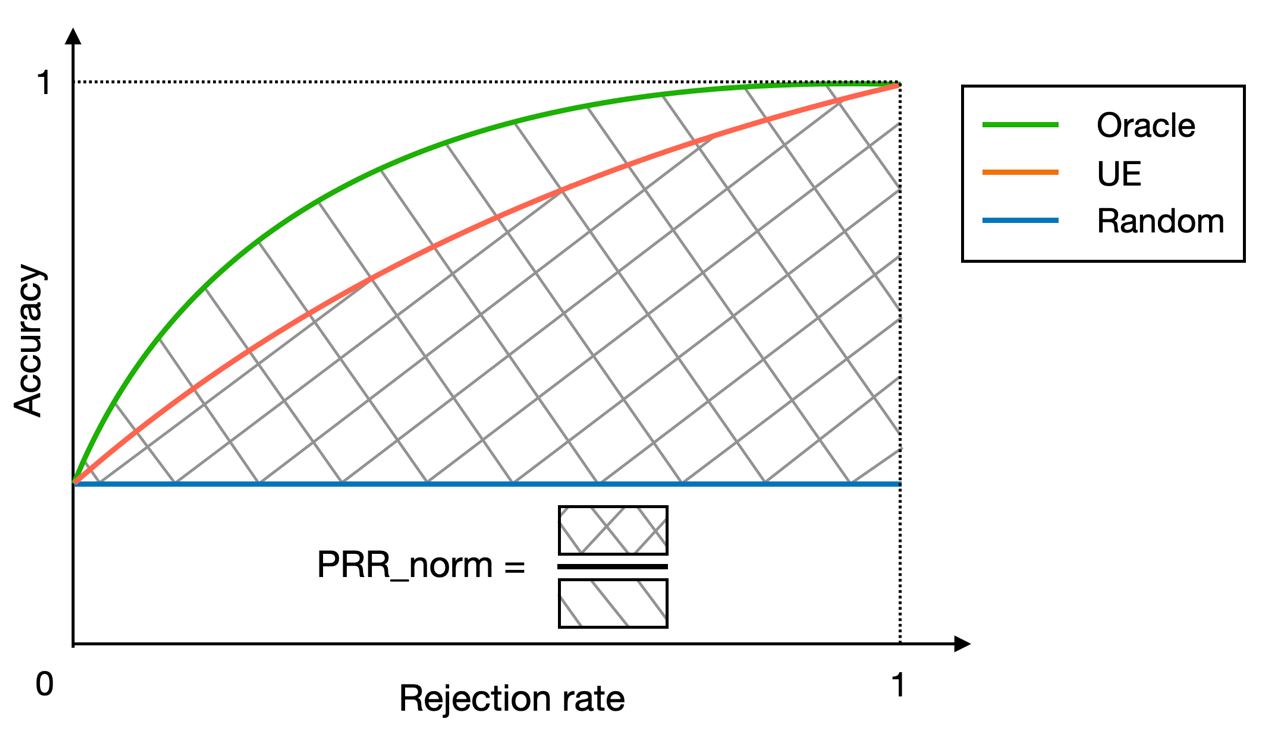}
    \caption{Illustration of the Prediction-Rejection Ratio (PRR). The PR curve plots the output quality against the rejection rate. The \textbf{oracle} curve ranks outputs perfectly by quality, while the \textbf{random} curve represents a random uncertainty score. The area between the UQ curve and random baseline (numerator) is normalized by the area between the oracle and random curves (denominator), yielding a PRR score between 0 and 1.}
    \label{fig:prr}
  \end{figure}

  \begin{table*}[h!]
\centering
\small
\scalebox{0.81}{
    \begin{tabular}{lcccccccccccccccc}
\toprule
&\multicolumn{8}{c}{\textbf{WMT14}}&\multicolumn{8}{c}{\textbf{WMT19}}\\
\cmidrule(lr){2-9}
\cmidrule(lr){10-17}
Model & \multicolumn{2}{c}{Cs-En} & \multicolumn{2}{c}{De-En} & \multicolumn{2}{c}{Ru-En} & \multicolumn{2}{c}{Fr-En} & \multicolumn{2}{c}{De-En} & \multicolumn{2}{c}{Fi-En} & \multicolumn{2}{c}{Lt-En} & \multicolumn{2}{c}{Ru-En} \\
  & Base & LINE & Base & LINE & Base & LINE & Base & LINE & Base & LINE & Base & LINE & Base & LINE & Base & LINE \\
\midrule
\multicolumn{17}{c}{\textbf{MetricX XXL}} \\
\midrule
Llama 3.1 8B & 0.47 & 0.54$\uparrow$ & 0.48 & 0.51$\uparrow$ & 0.46 & 0.54$\uparrow$ & 0.39 & 0.43$\uparrow$ & 0.43 & 0.47$\uparrow$ & 0.52 & 0.51 & 0.49 & 0.49 & 0.36 & 0.45$\uparrow$ \\
Gemma 2 9B & 0.45 & 0.46$\uparrow$ & 0.47 & 0.49$\uparrow$ & 0.42 & 0.46$\uparrow$ & 0.36 & 0.37$\uparrow$ & 0.44 & 0.47$\uparrow$ & 0.45 & 0.45 & 0.34 & 0.37$\uparrow$ & 0.38 & 0.41$\uparrow$ \\
EuroLLM 9B & 0.54 & 0.55$\uparrow$ & 0.54 & 0.55$\uparrow$ & 0.48 & 0.47 & 0.46 & 0.46 & 0.50 & 0.51$\uparrow$ & 0.49 & 0.47 & 0.42 & 0.47$\uparrow$ & 0.36 & 0.42$\uparrow$ \\
\midrule
\multicolumn{17}{c}{\textbf{XComet XXL}} \\
\midrule
Llama 3.1 8B & 0.40 & 0.48$\uparrow$ & 0.37 & 0.47$\uparrow$ & 0.41 & 0.53$\uparrow$ & 0.33 & 0.42$\uparrow$ & 0.34 & 0.44$\uparrow$ & 0.51 & 0.49 & 0.53 & 0.52 & 0.37 & 0.51$\uparrow$ \\
Gemma 2 9B & 0.35 & 0.37$\uparrow$ & 0.35 & 0.38$\uparrow$ & 0.39 & 0.48$\uparrow$ & 0.27 & 0.34$\uparrow$ & 0.34 & 0.38$\uparrow$ & 0.42 & 0.40 & 0.30 & 0.32$\uparrow$ & 0.35 & 0.37$\uparrow$ \\
EuroLLM 9B & 0.43 & 0.46$\uparrow$ & 0.42 & 0.46$\uparrow$ & 0.39 & 0.51$\uparrow$ & 0.35 & 0.42$\uparrow$ & 0.43 & 0.47$\uparrow$ & 0.45 & 0.42 & 0.39 & 0.38 & 0.36 & 0.44$\uparrow$ \\
\midrule
\multicolumn{17}{c}{\textbf{Comet WMT22}} \\
\midrule
Llama 3.1 8B & 0.48 & 0.58$\uparrow$ & 0.48 & 0.56$\uparrow$ & 0.45 & 0.59$\uparrow$ & 0.37 & 0.48$\uparrow$ & 0.46 & 0.55$\uparrow$ & 0.54 & 0.56$\uparrow$ & 0.52 & 0.56$\uparrow$ & 0.43 & 0.53$\uparrow$ \\
Gemma 2 9B & 0.44 & 0.49$\uparrow$ & 0.50 & 0.54$\uparrow$ & 0.43 & 0.53$\uparrow$ & 0.37 & 0.44$\uparrow$ & 0.49 & 0.53$\uparrow$ & 0.49 & 0.49 & 0.35 & 0.36$\uparrow$ & 0.40 & 0.41$\uparrow$ \\
EuroLLM 9B & 0.52 & 0.57$\uparrow$ & 0.52 & 0.55$\uparrow$ & 0.46 & 0.56$\uparrow$ & 0.47 & 0.52$\uparrow$ & 0.52 & 0.58$\uparrow$ & 0.51 & 0.52$\uparrow$ & 0.37 & 0.45$\uparrow$ & 0.43 & 0.45$\uparrow$ \\
\bottomrule
\end{tabular}
}
\caption{Comparison between best raw and detrended PRR scores across all metrics and models for translation datasets. Arrows indicate improvements in detrended over raw method.}
\label{tab:nmt}
\end{table*}

\paragraph{Evaluation.}
  We evaluate uncertainty estimates using the Prediction Rejection (PR) curve, which measures how the average output quality \( Q(f(\xv_i), \yv_i) \) changes as uncertain examples are rejected~\citep{malinin-etal-2017-incorporating, malinin2020uncertainty}. For a given uncertainty threshold \( a \), it shows the average quality over all instances where the uncertainty \( U(\xv_i) < a \). To quantify the effectiveness of an uncertainty measure, we use the Prediction-Rejection Ratio (PRR). It compares the area under the PR curve (AUC) to that of a random baseline and an oracle that ranks instances perfectly by their actual output quality (see Figure~\ref{fig:prr}):
  \begin{equation}
  \label{eq:prr}
    \text{PRR} = \frac{\text{AUC}_{\text{unc}} - \text{AUC}_{\text{rnd}}}{\text{AUC}_{\text{oracle}} - \text{AUC}_{\text{rnd}}}.
  \end{equation}

  A higher PRR indicates better alignment between the uncertainty estimate and the actual model quality. We use PRR as our primary evaluation measure, as it captures the utility of uncertainty scores for selective prediction in generation tasks. PRR measures how well uncertainty estimates rank outputs by quality, and is more appropriate than classification or calibration measures for continuous-valued evaluation~\citep{fadeeva2023lm, vashurin2025benchmarking}.

\paragraph{Quality Metrics.}
  To evaluate the output quality, we use the following measures: \textit{COMET}, \textit{XComet-XXL} and \textit{MetricX-XXL}~\citep{comet, xcomet, metricx}, which represent a diverse set of neural quality estimation models. We use \textit{Accuracy} for \textsc{\gsm}, and \textit{AlignScore}~\citep{zha-etal-2023-alignscore} to measure semantic alignment between input and output for summarization task.

\subsection{Results}

  \begin{table}[t]
\centering
\begin{tabular}{lcccc}
\toprule
 & \multicolumn{2}{c}{\textbf{XSum}} & \multicolumn{2}{c}{\textbf{GSM8k}} \\
\cmidrule(lr){2-3} \cmidrule(lr){4-5} \\
Model               & Base & LINE & Base & LINE \\
\midrule
Llama 3.1 8B   & 0.37 & 0.37 & 0.36 & 0.40↑ \\
Gemma 2 9B     & 0.35 & 0.38↑ & 0.39 & 0.40↑ \\
\bottomrule
\end{tabular}
\caption{Comparison between best raw and detrended PRR scores for summarization and mathematical reasoning tasks. Arrows indicate improvements in detrended over raw method.}
\label{tab:ats_qa}
\end{table}

  For each dataset, Tables~\ref{tab:nmt} and~\ref{tab:ats_qa} present the PRR score of the best-performing UQ method and best-performing \methodname variation. Across both translation and open-ended generation tasks, we consistently observe improvements in PRR scores when applying our detrending procedure, demonstrating its effectiveness in mitigating length-related bias and enhancing the reliability of uncertainty estimates. However, it is important to note that there are a few cases -- particularly in \xsum for \llama or WMT 19 Fi-En for all considered models -- where the gains are marginal or the detrended score does not outperform the raw variant. 
  
  This suggests that while length-induced bias is a prevalent issue, the extent of its impact can vary by task and model, and in some settings, additional sources of uncertainty may dominate. Detailed  experimental results with breakdown of PRR scores before and after \methodname transformation for each of the UQ methods are provided in Appendix~\ref{sec:experimental_results}.

  \begin{table*}[th!]
    \centering
    \begin{tabular}{lccccc}
      \toprule
       \textbf{Method} & \multicolumn{3}{c}{\textbf{WMT}} & \textbf{\xsum} & \textbf{\gsm} \\
      \cmidrule(lr){2-4} \cmidrule(lr){5-5} \cmidrule(lr){6-6} \\
       & Comet WMT22 & XComet XXL & MetricX-XXL & Align Score & Accuracy \\
      \midrule
      MSP & 0.09 $\pm$ 0.02 & 0.09 $\pm$ 0.02 & 0.18 $\pm$ 0.01 & 0.03 $\pm$ 0.00 & 0.00 $\pm$ 0.01 \\
      PPL & 0.05 $\pm$ 0.01 & 0.05 $\pm$ 0.01 & 0.02 $\pm$ 0.00 & 0.01 $\pm$ 0.01 & 0.09 $\pm$ 0.02 \\
      MTE & 0.08 $\pm$ 0.01 & 0.07 $\pm$ 0.01 & 0.03 $\pm$ 0.01 & 0.01 $\pm$ 0.01 & 0.08 $\pm$ 0.02 \\
      MCSE & 0.07 $\pm$ 0.02 & 0.07 $\pm$ 0.02 & 0.16 $\pm$ 0.01 & 0.02 $\pm$ 0.01 & 0.00 $\pm$ 0.00 \\
      MCNSE & 0.02 $\pm$ 0.01 & 0.02 $\pm$ 0.01 & 0.00 $\pm$ 0.00 & 0.01 $\pm$ 0.00 & 0.02 $\pm$ 0.00 \\
      LSRL & 0.00 $\pm$ 0.01 & 0.01 $\pm$ 0.01 & 0.00 $\pm$ 0.00 & 0.03 $\pm$ 0.02 & 0.00 $\pm$ 0.00 \\
      TokenSAR & 0.05 $\pm$ 0.01 & 0.05 $\pm$ 0.01 & 0.02 $\pm$ 0.01 & 0.00 $\pm$ 0.01	 & 0.09 $\pm$ 0.02 \\
      \bottomrule
    \end{tabular}
    \caption{Average improvement across datasets and models in PRR scores after detrending for three tasks: WMT (machine translation), \xsum (summarization), and \gsm (mathematical reasoning). Values are reported as mean improvements with their associated standard error of the mean (SEM).}
  \label{tab:avg_improvements}
  \end{table*}

  \begin{figure*}[th!]
    \begin{subfigure}{0.33\textwidth}
        \includegraphics[width=\linewidth]{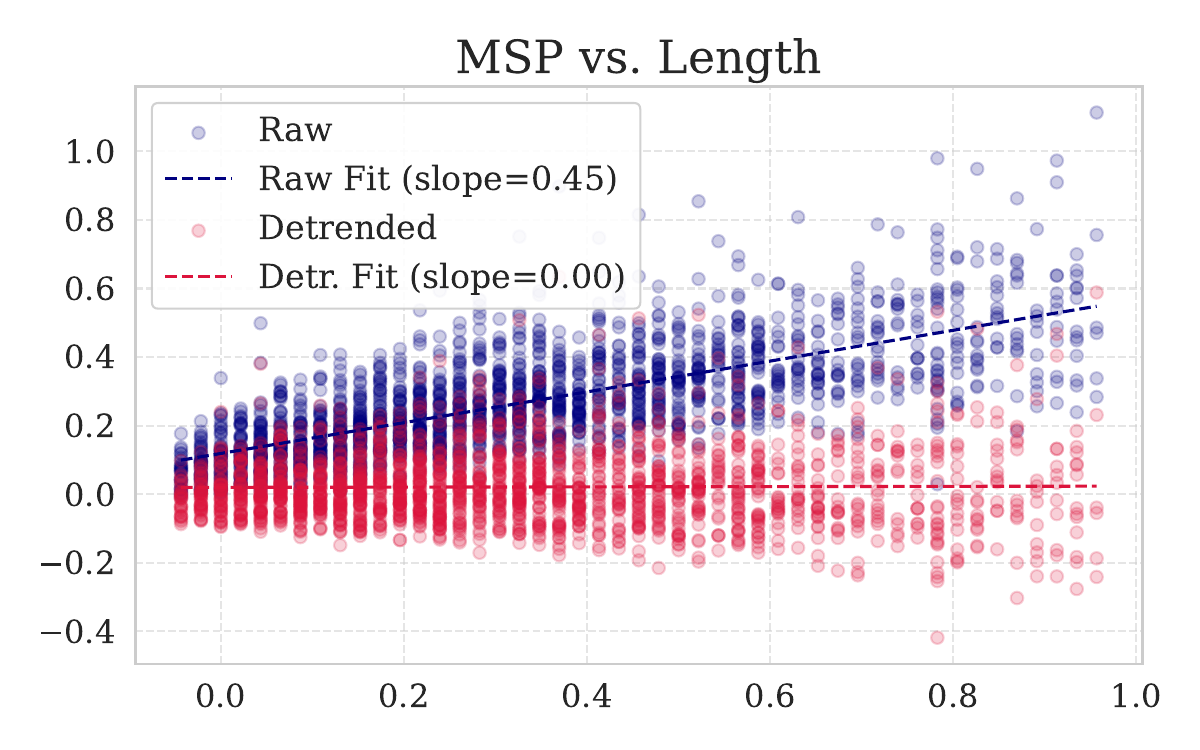}
    \end{subfigure}
    \begin{subfigure}{0.33\textwidth}
        \includegraphics[width=\linewidth]{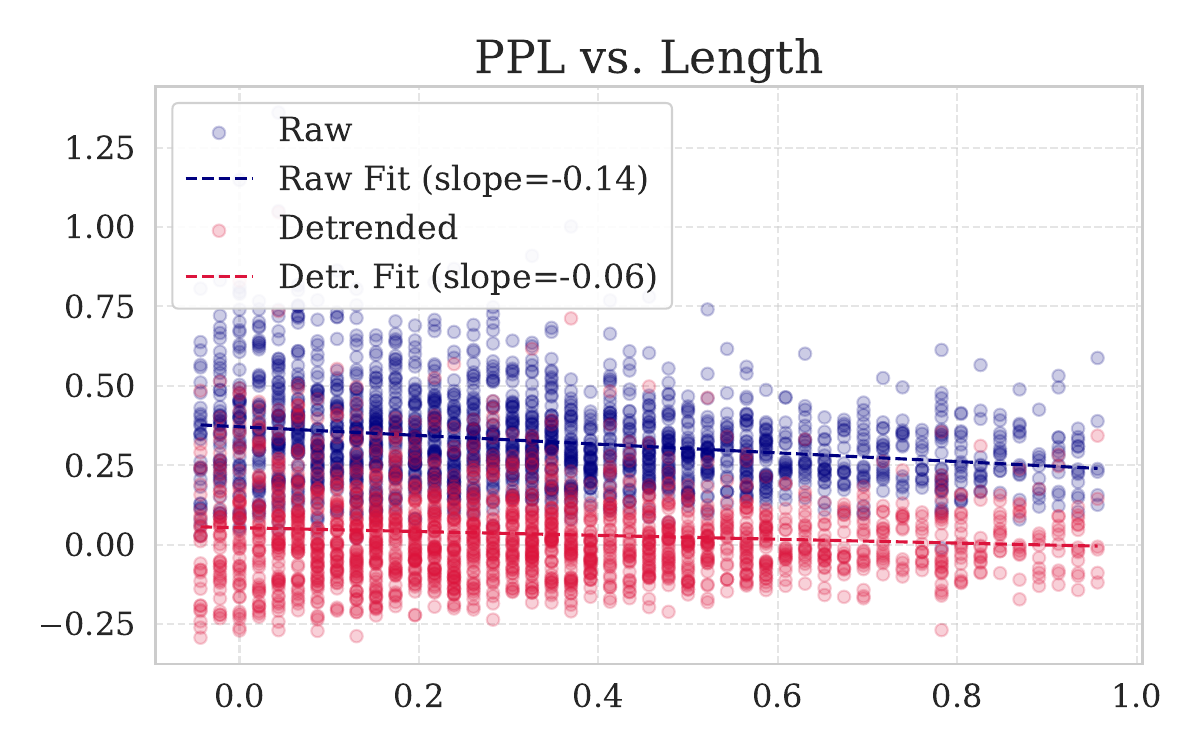}
    \end{subfigure}
    \begin{subfigure}{0.33\textwidth}
        \includegraphics[width=\linewidth]{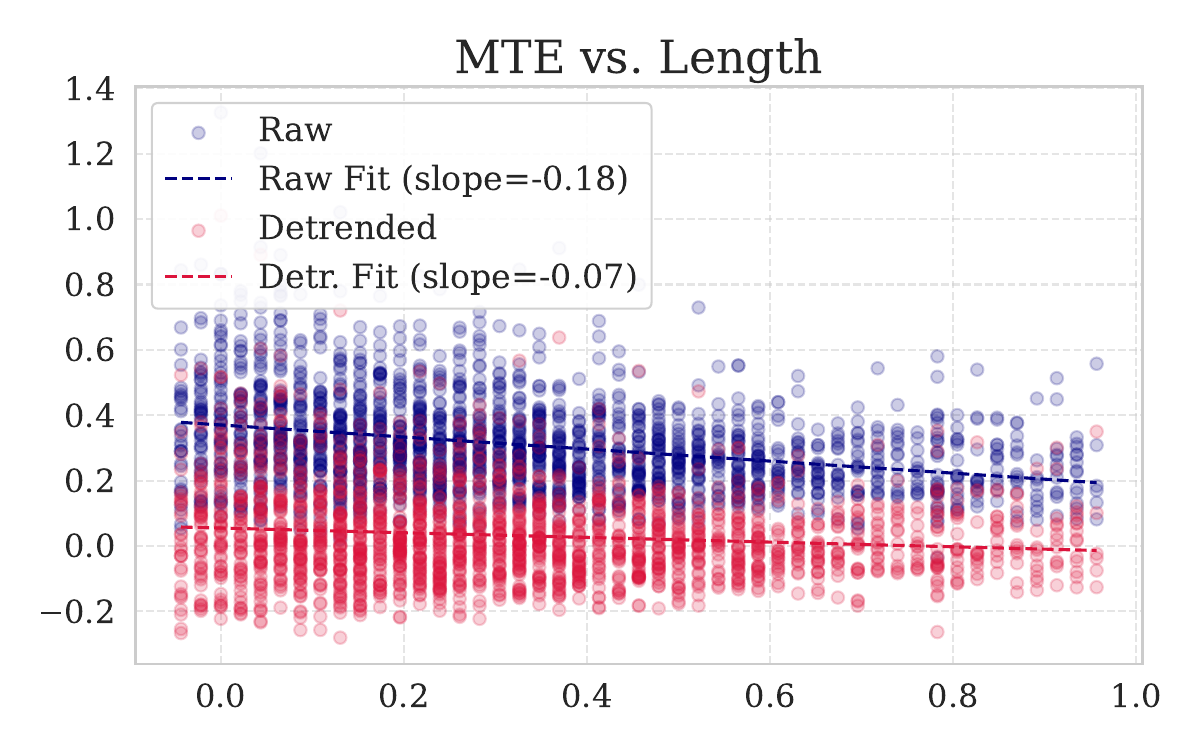}
    \end{subfigure}
    \caption{Example of how detrending removes length-related bias from uncertainty estimates. Shown for MSP, PPL, and MTE on WMT14 De–En (Model - \llama), the raw scores exhibit clear length dependency, which is largely reduced after detrending.}
    \label{fig:detrending}
  \end{figure*}

  Table~\ref{tab:avg_improvements} reports improvements in PRR scores after \methodname transformation for each UQ method, average over all tasks. As evident from the table, the most substantial gains occur in uncertainty estimation methods that are highly sensitive to sequence length, such as MSP and PPL. Detrending enhances their ability to discriminate between high- and low-quality generations. In contrast, methods like LSRL, which estimate uncertainty based on the semantic similarity of sampled outputs, exhibit far smaller improvements, if any. This is expected, as can be seen in Appendix~\ref{sec:length_effects_ue}, LSRL exhibits the smallest trends with respect to length.

  Figure~\ref{fig:detrending} offers an illustration of the impact of our detrending procedure on three uncertainty estimation scores for translation tasks. For MSP, PPL and MTE, we observe a strong correlation between sequence length and raw uncertainty scores, indicating a clear length-induced bias. We can see that, after detrending, these trends are largely eliminated, as shown by the near-zero slopes. On the other hand, after applying the equation~\eqref{eq:debiasing:final} for the summarization and mathematical reasoning tasks, the detrended uncertainty scores exhibit a length bias that is comparable to that of the quality evaluation measure itself.


\section{Conclusion}
  We introduced \methodname, a simple yet effective framework for removing generation length effects from uncertainty estimates. Through extensive analysis across tasks (translation, summarization, and mathematical reasoning), we demonstrated that uncertainty scores are often confounded by output length, and that correcting for this bias improves the reliability of uncertainty-based rejection. Our method is lightweight, model-agnostic, and requires minimal supervision only when known quality-length correlation is present. While certain assumptions limit its applicability in more complex settings, \methodname offers a strong foundation for more interpretable and trustworthy uncertainty estimation in text generation. 
  
  Future work could explore addressing length bias directly during model training, rather than correcting it in a post-hoc manner. 


\section{Limitations}
  While \methodname offers a simple and effective correction for length-induced bias, several important considerations remain.

  We assume a linear relationship between uncertainty scores and output length, as well as quality scores and length. While this simplifies both implementation and interpretation, it may not fully capture the complexity of interactions between length and uncertainty. 
  However, as demonstrated in Section~\ref{sec:poly_detr}, linear approximation is a reasonable and effective first-order correction. Nonetheless, in tasks such as multi-step reasoning, where uncertainty may follow phase-specific patterns, a linear fit may be insufficient. 

  Our method requires a small number of quality-labeled examples to estimate the quality-length relationship. However, this only applies when there is a known or observed correlation between output length and quality. Moreover, in Appendix~\ref{sec:small_quality_labels} we demonstrate that the quality trend can be estimated using as little as 500 generations. In tasks where quality is largely length-independent, our method can be applied without any quality annotations. 
  
  This leads to another consideration - we assume prior knowledge of quality-length relationship. In tasks where this relationship is unclear or poorly understood, effectiveness may be reduced.

\section*{Ethical Considerations}
  \methodname improves uncertainty quantification by removing spurious correlations between output length and estimated uncertainty. This helps to prevent misleading high- or low-uncertainty signals that often affect longer generations. However, it does not prevent the generation of incorrect or harmful content, and low uncertainty scores do not guarantee factuality. Moreover, it does not address factors such as prompt phrasing or out-of-domain data. 
  
  Reliable uncertainty estimates are crucial for enabling selective generation, abstention, or human-in-the-loop review, especially in tasks where correctness cannot be easily verified. \methodname improves robustness to length-related bias, but it should be used as part of a broader reliability strategy, especially in critical applications.

\section*{Acknowledgments}
We would like to thank Evgenii Tsymbalov for the valuable advice and feedback during this work.

\bibliography{main}

\newpage

\appendix
\onecolumn 

\section{Length Bias Analysis}
\label{sec:length_effects}

\subsection{Response Quality vs Generation Length}
\label{sec:length_effects_quality}
\begin{figure*}[h!]
    \centering

\vspace{-0.5em}
{\centering \textbf{\small \llama} \par}
\vspace{0.3em}

    \begin{subfigure}{0.24\textwidth}
        \includegraphics[width=\linewidth]{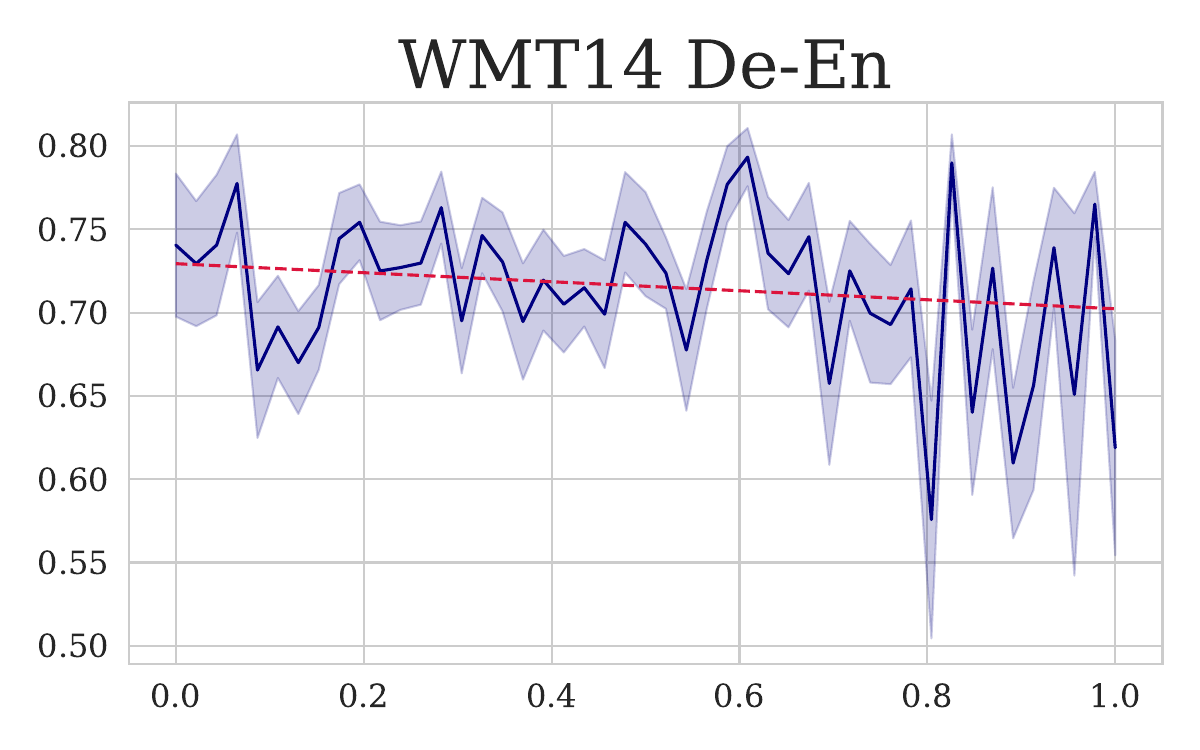}
    \end{subfigure}
    \begin{subfigure}{0.24\textwidth}
        \includegraphics[width=\linewidth]{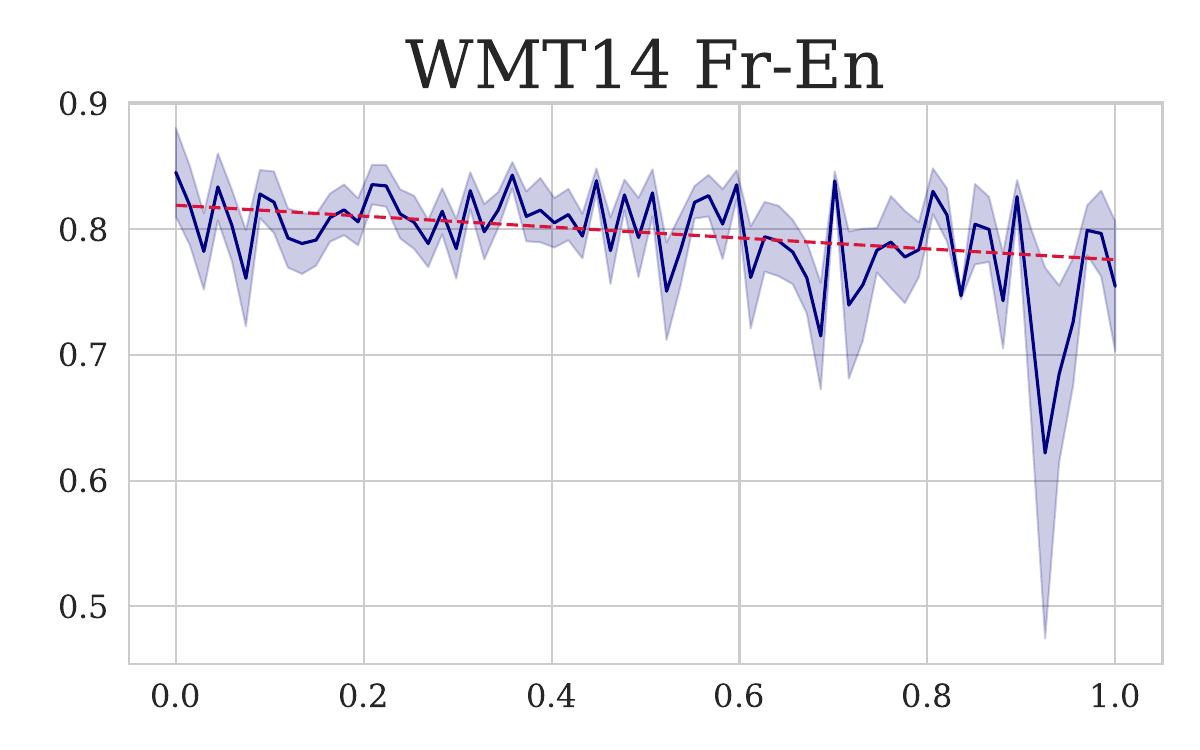}
    \end{subfigure}
    \begin{subfigure}{0.24\textwidth}
        \includegraphics[width=\linewidth]{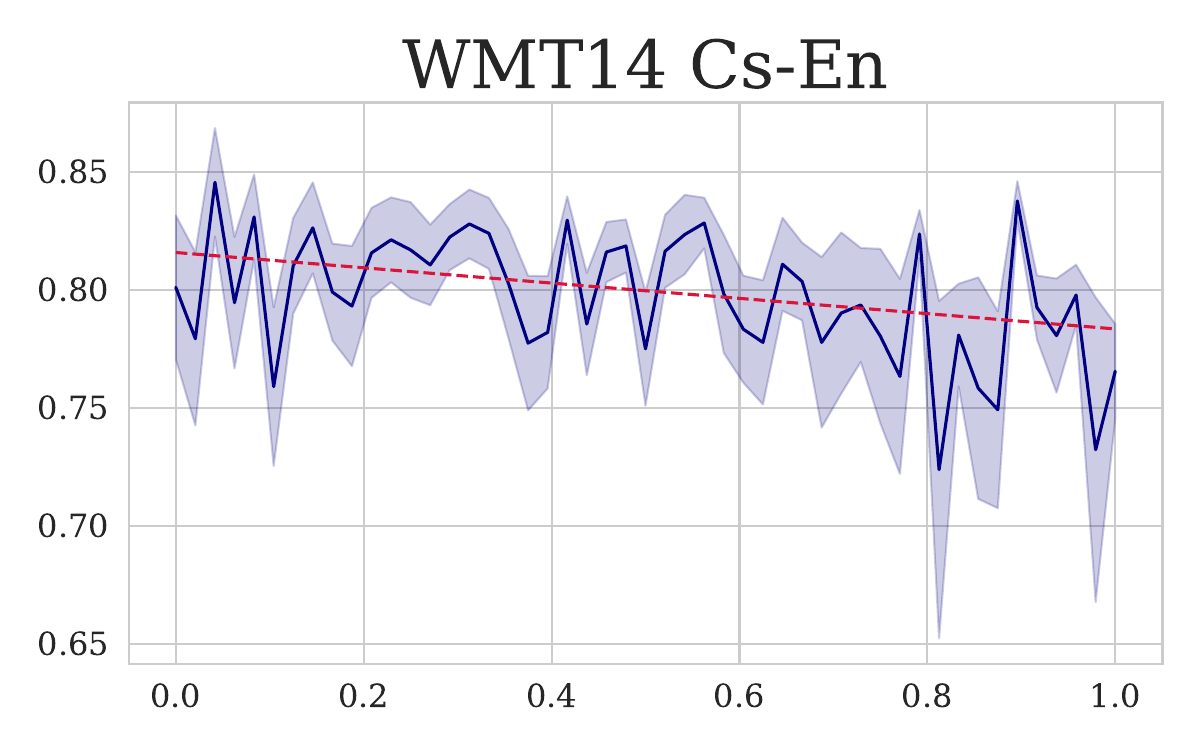}
    \end{subfigure}
    \begin{subfigure}{0.24\textwidth}
        \includegraphics[width=\linewidth]{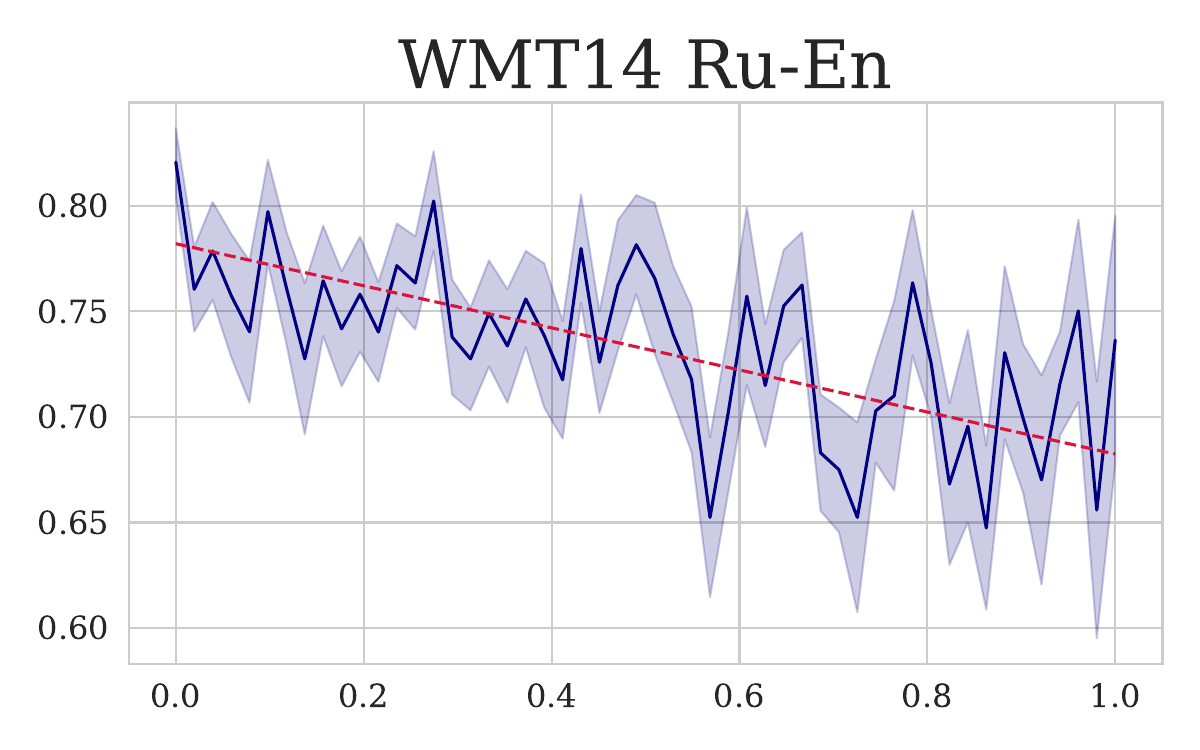}
    \end{subfigure}

    \begin{subfigure}{0.24\textwidth}
        \includegraphics[width=\linewidth]{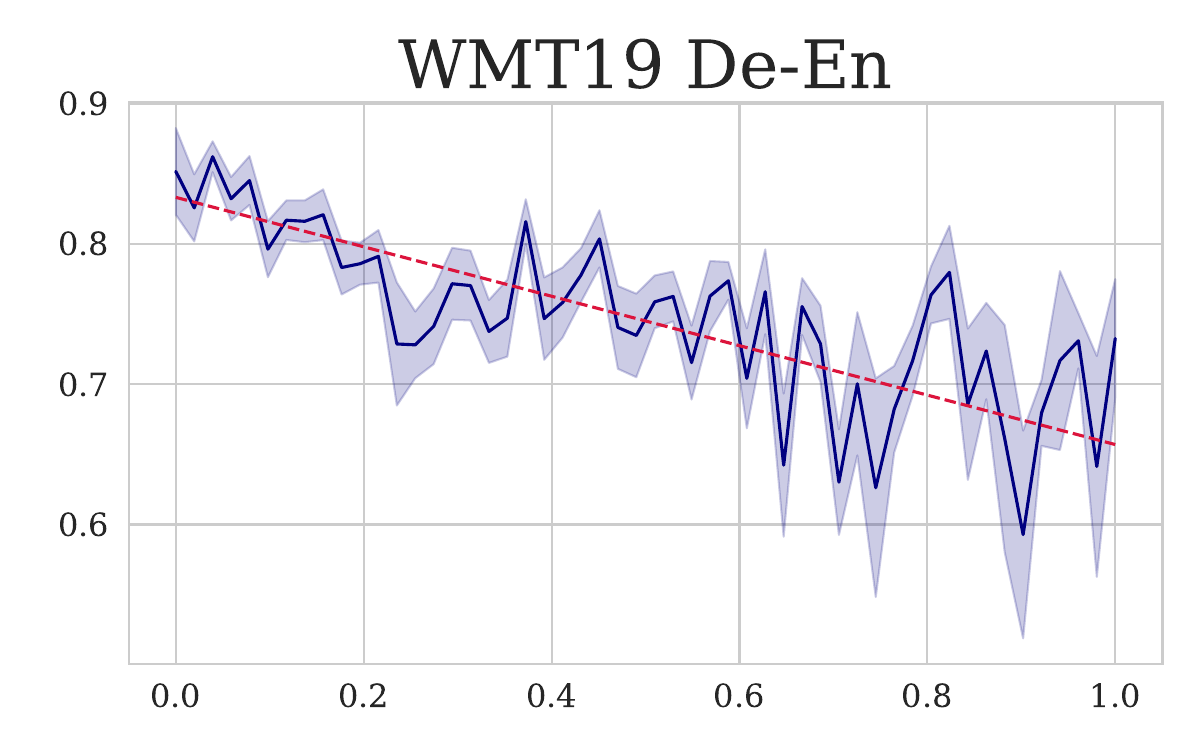}
    \end{subfigure}
    \begin{subfigure}{0.24\textwidth}
        \includegraphics[width=\linewidth]{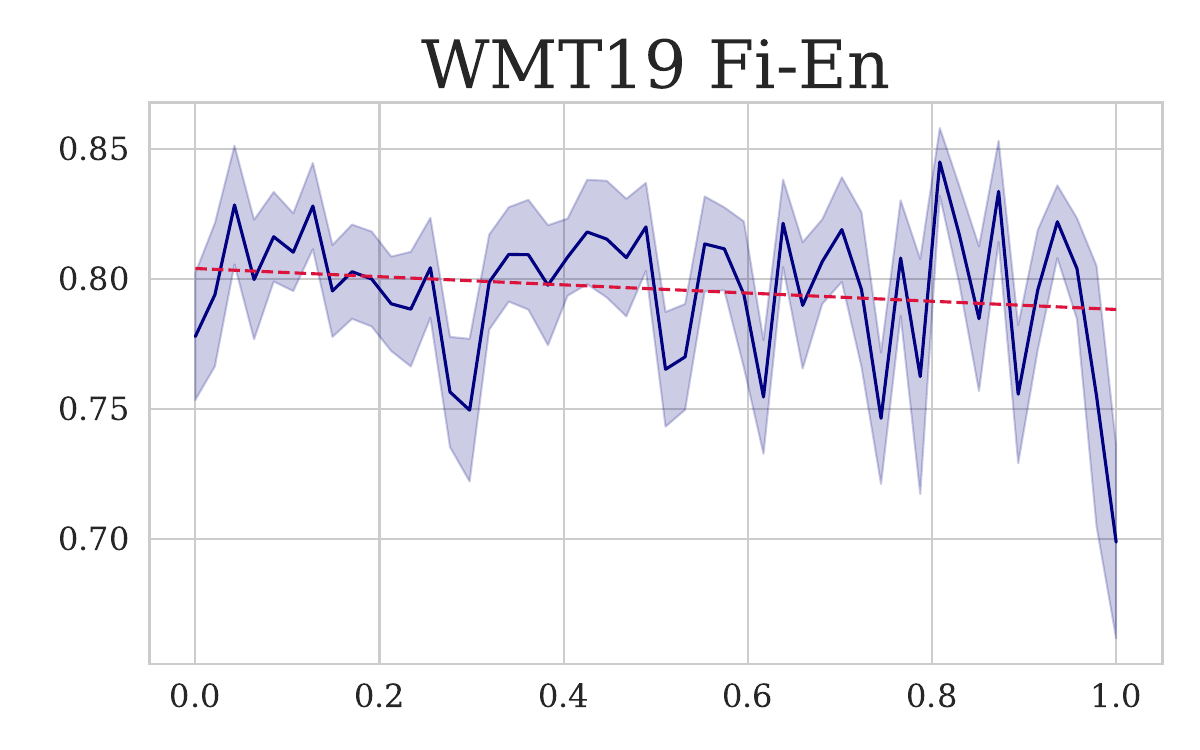}
    \end{subfigure}
    \begin{subfigure}{0.24\textwidth}
        \includegraphics[width=\linewidth]{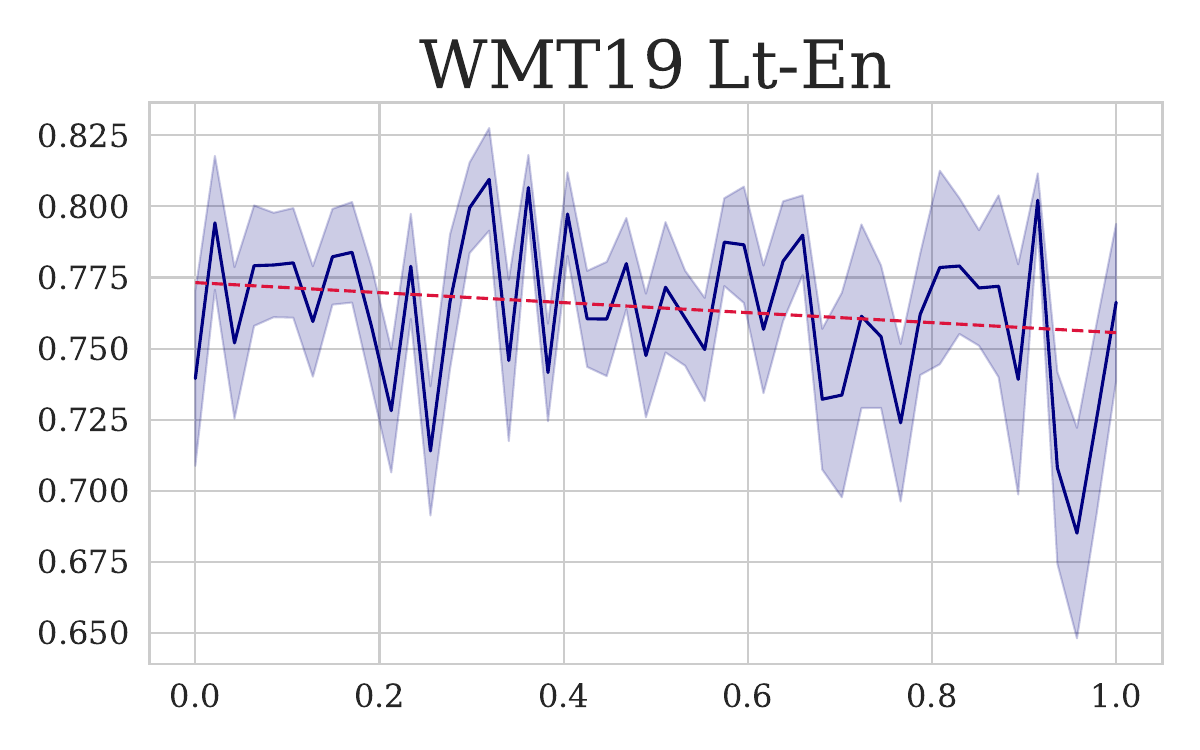}
    \end{subfigure}
    \begin{subfigure}{0.24\textwidth}
        \includegraphics[width=\linewidth]{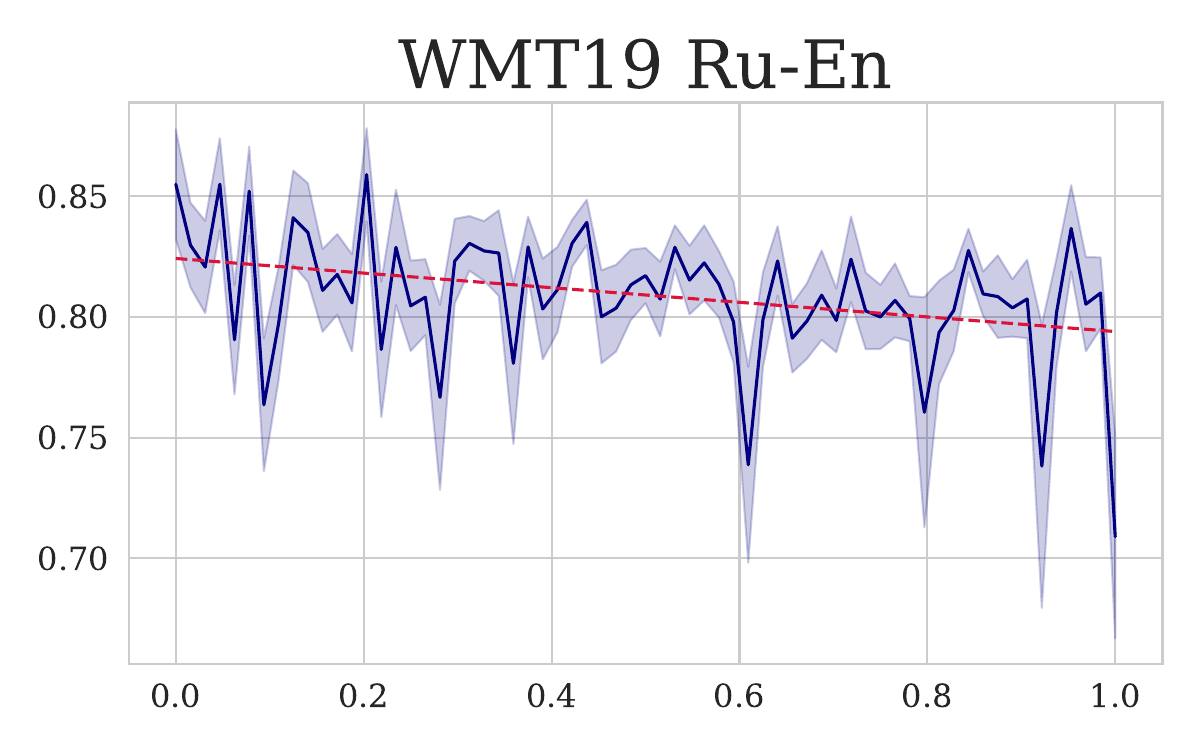}
    \end{subfigure}

\vspace{-0.5em}
{\centering \textbf{\small \gemma} \par}
\vspace{0.3em}

 \begin{subfigure}{0.24\textwidth}
        \includegraphics[width=\linewidth]{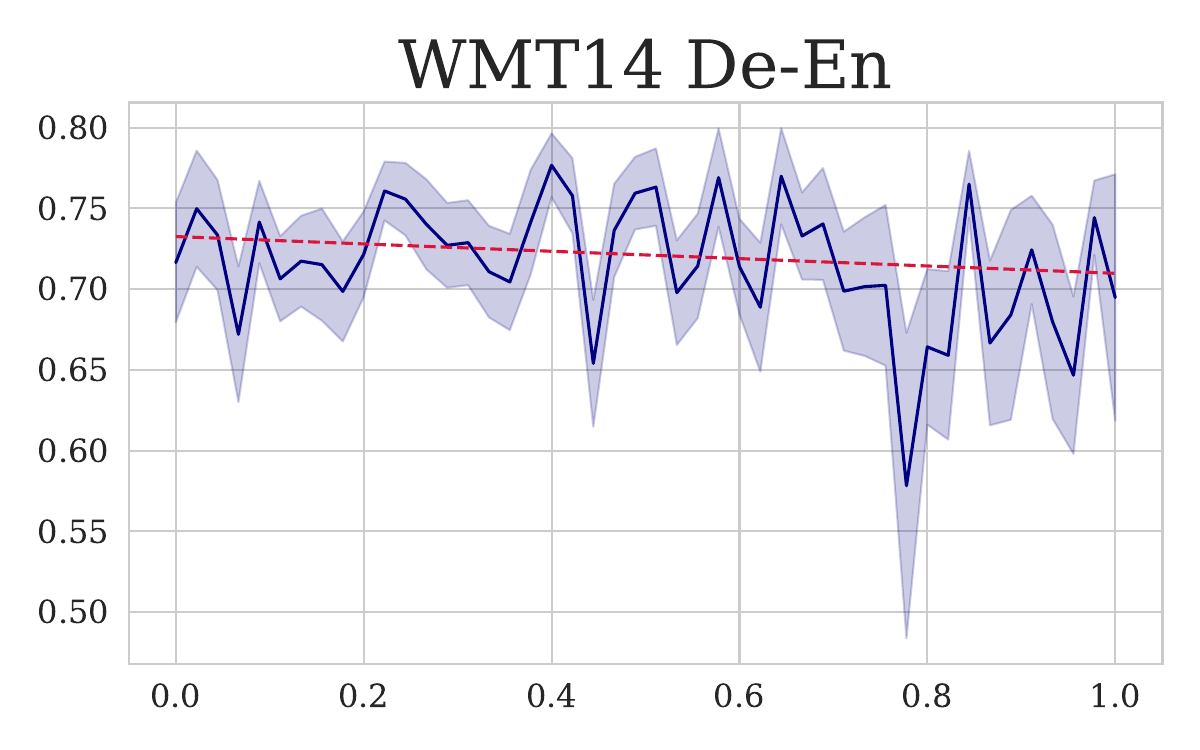}
    \end{subfigure}
    \begin{subfigure}{0.24\textwidth}
        \includegraphics[width=\linewidth]{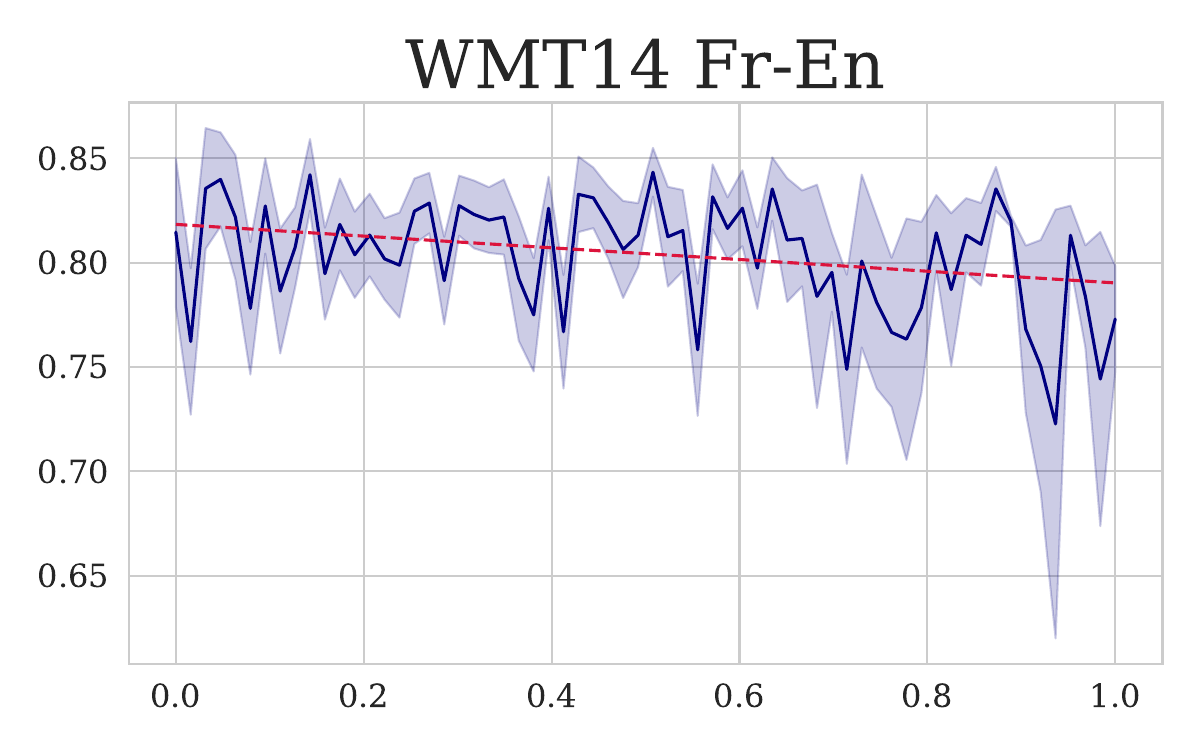}
    \end{subfigure}
    \begin{subfigure}{0.24\textwidth}
        \includegraphics[width=\linewidth]{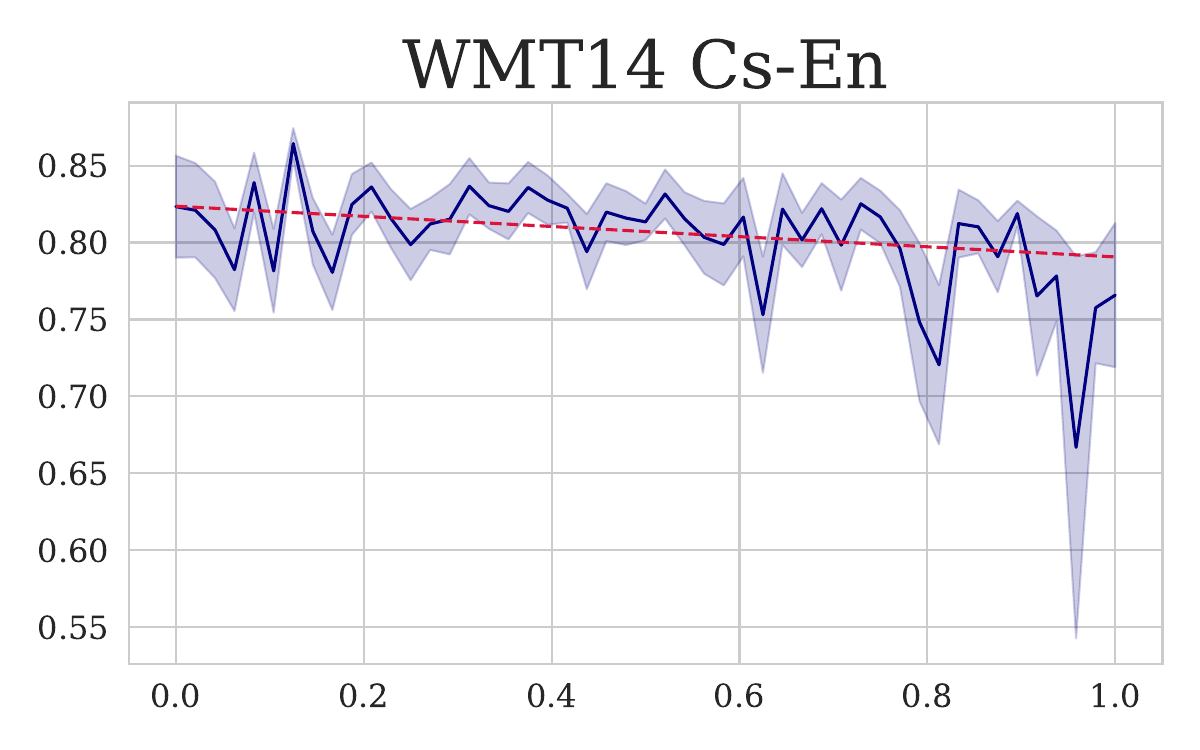}
    \end{subfigure}
    \begin{subfigure}{0.24\textwidth}
        \includegraphics[width=\linewidth]{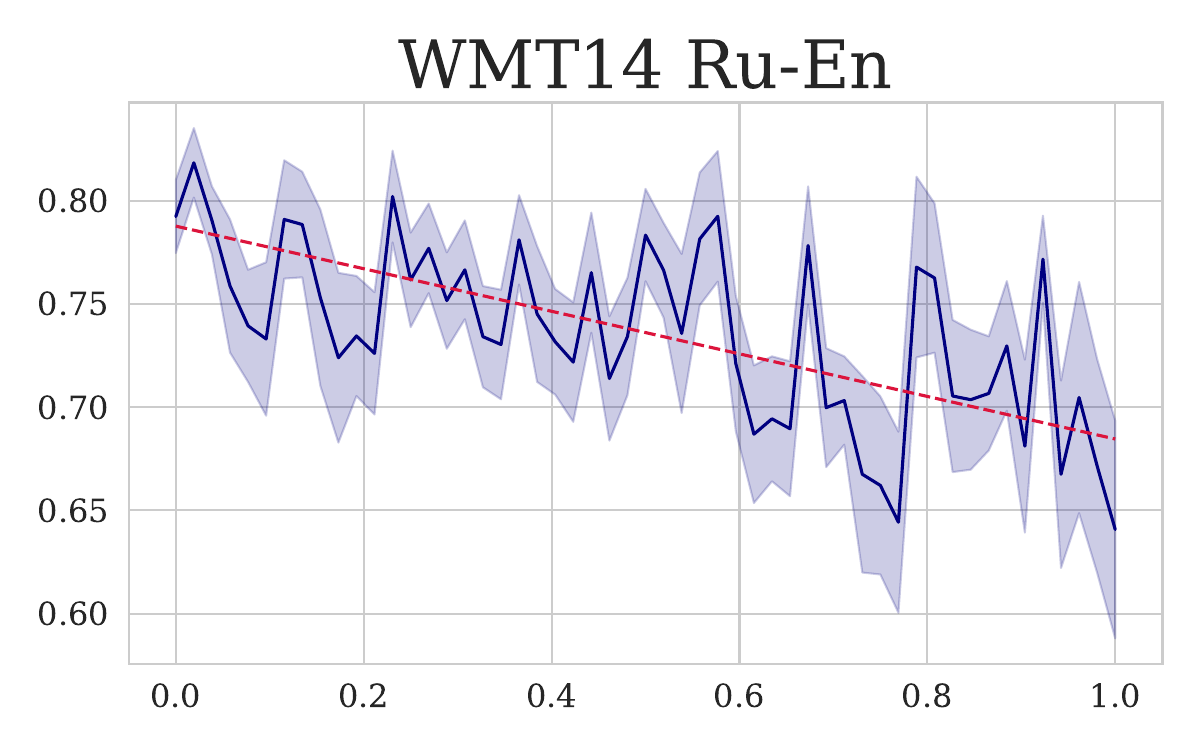}
    \end{subfigure}

    \begin{subfigure}{0.24\textwidth}
        \includegraphics[width=\linewidth]{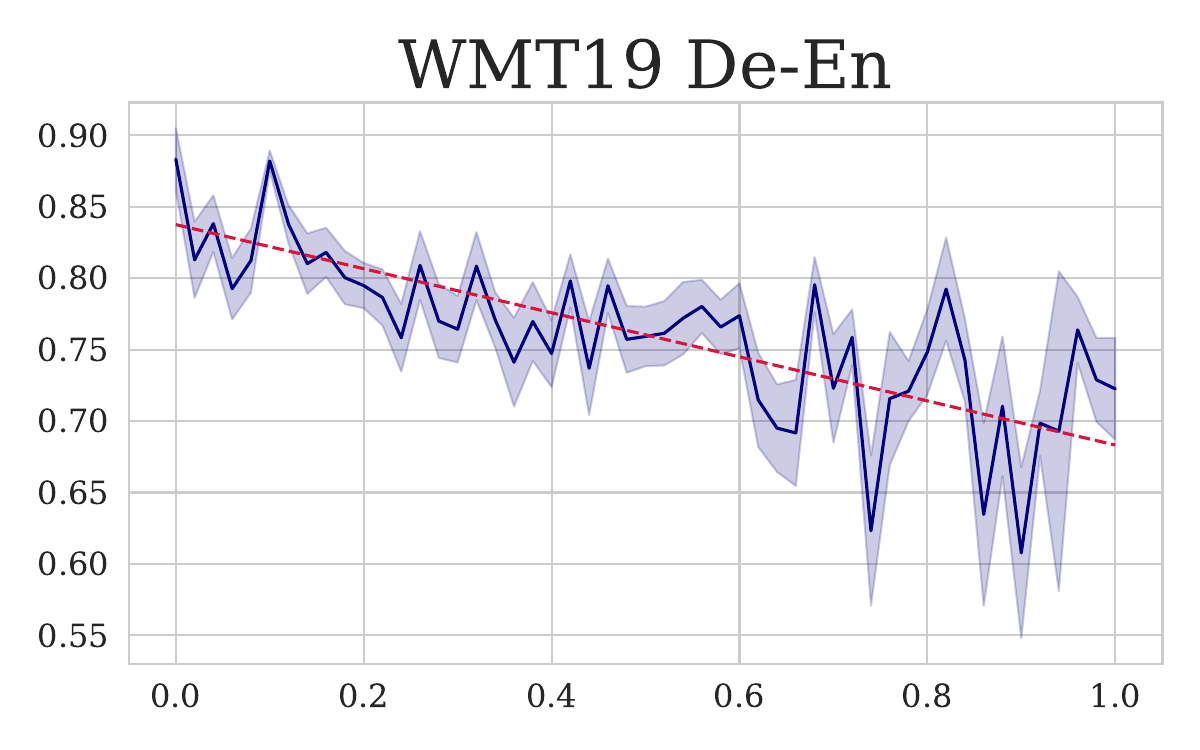}
    \end{subfigure}
    \begin{subfigure}{0.24\textwidth}
        \includegraphics[width=\linewidth]{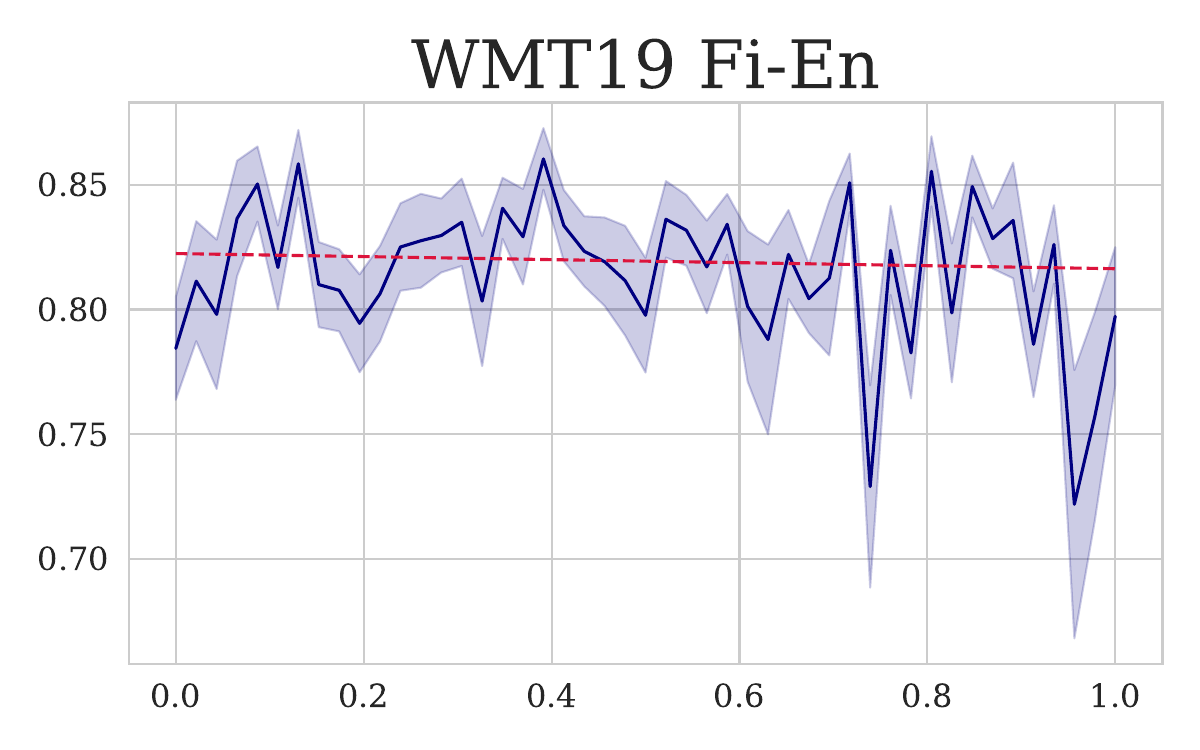}
    \end{subfigure}
    \begin{subfigure}{0.24\textwidth}
        \includegraphics[width=\linewidth]{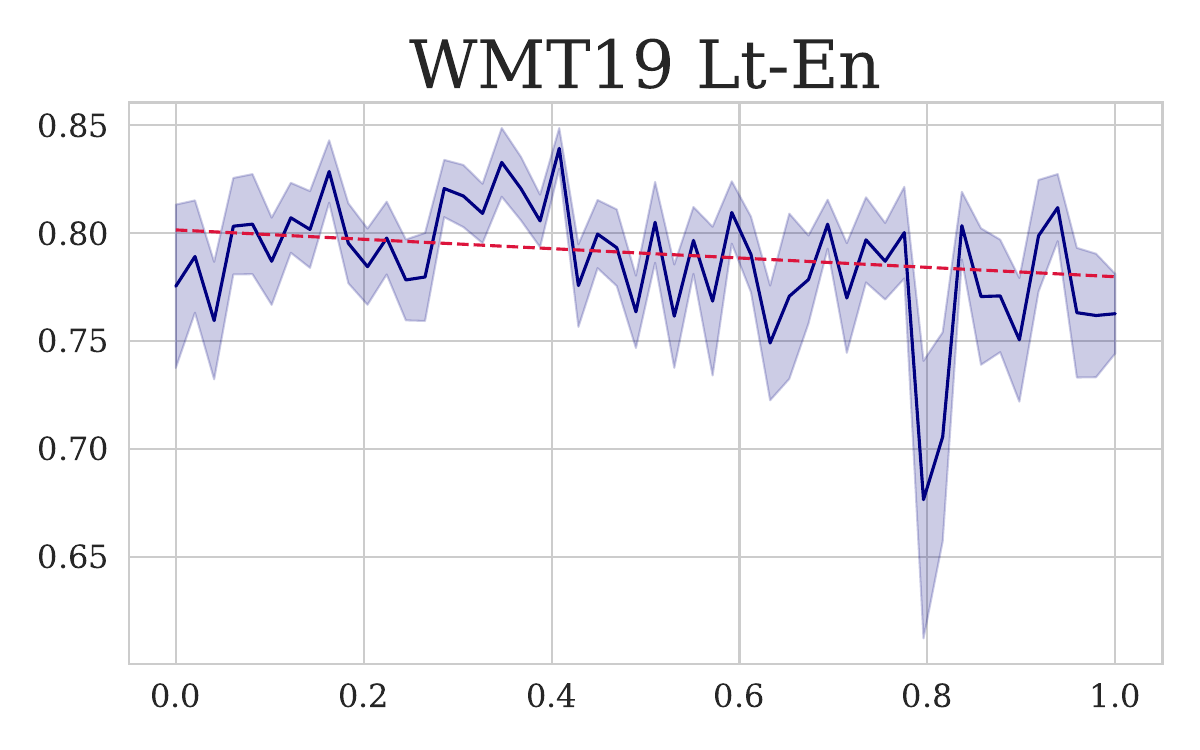}
    \end{subfigure}
    \begin{subfigure}{0.24\textwidth}
        \includegraphics[width=\linewidth]{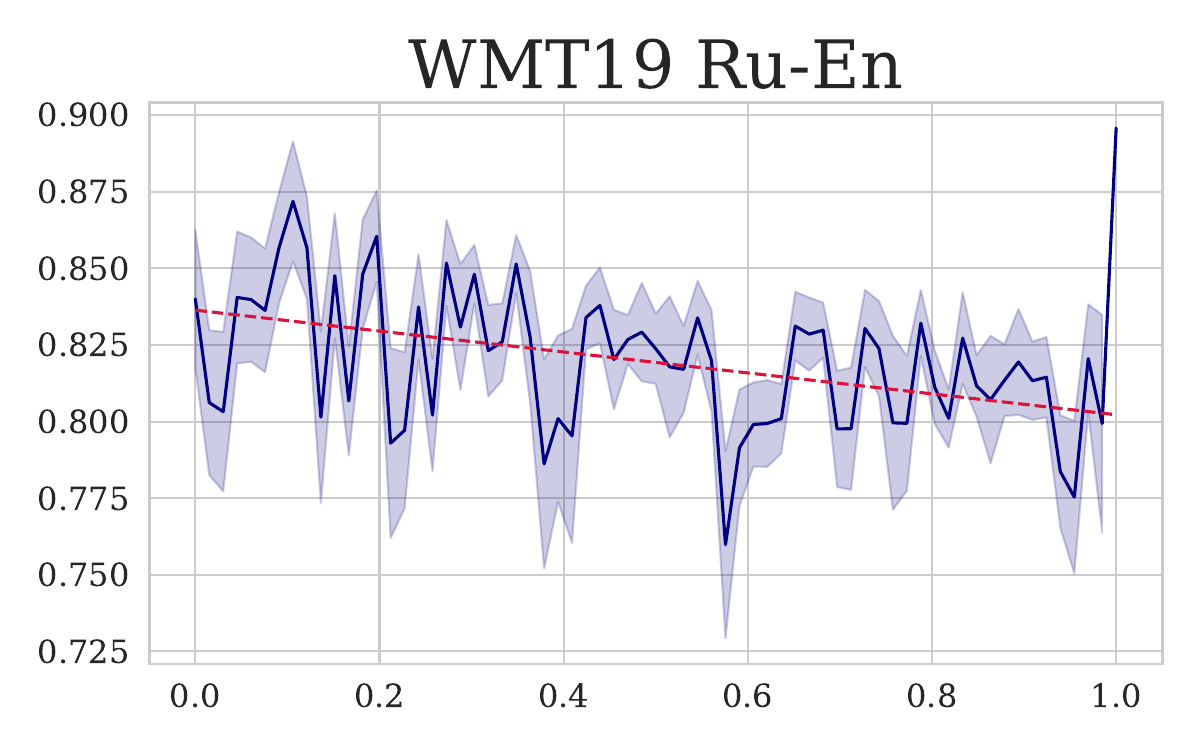}
    \end{subfigure}

\vspace{-0.5em}
{\centering \textbf{\small \eurollm} \par}
\vspace{0.3em}
 \begin{subfigure}{0.24\textwidth}
        \includegraphics[width=\linewidth]{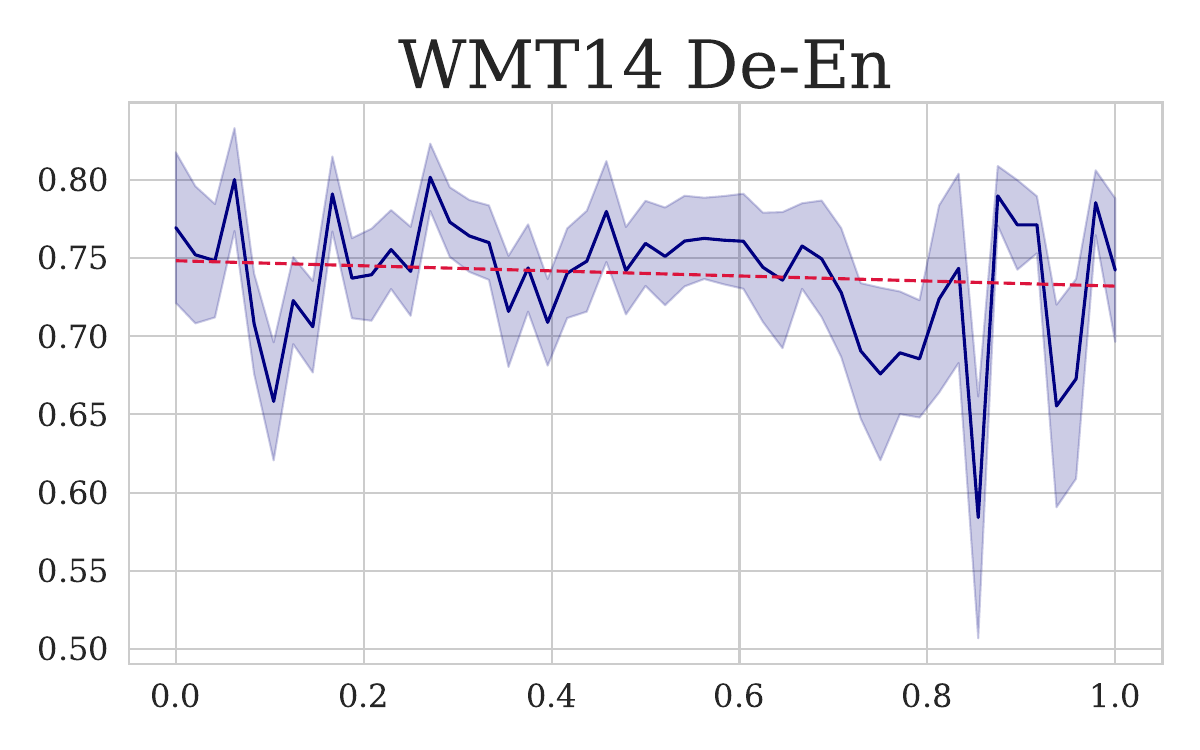}
    \end{subfigure}
    \begin{subfigure}{0.24\textwidth}
        \includegraphics[width=\linewidth]{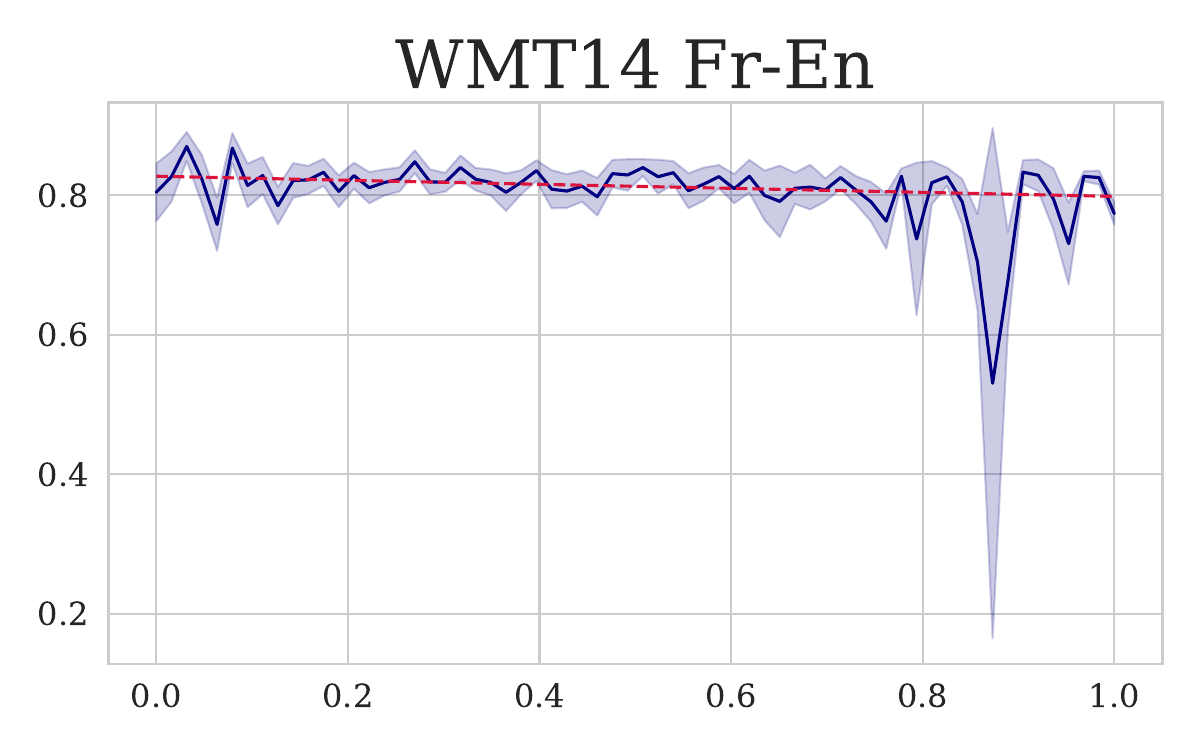}
    \end{subfigure}
    \begin{subfigure}{0.24\textwidth}
        \includegraphics[width=\linewidth]{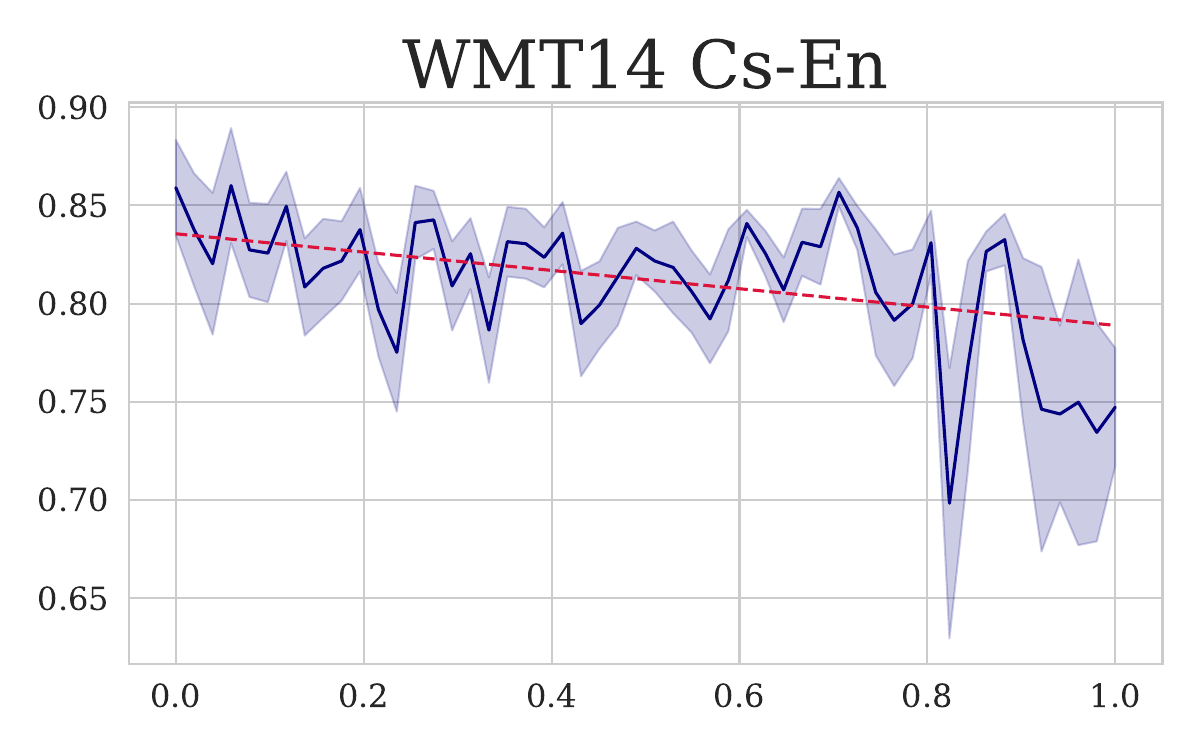}
    \end{subfigure}
    \begin{subfigure}{0.24\textwidth}
        \includegraphics[width=\linewidth]{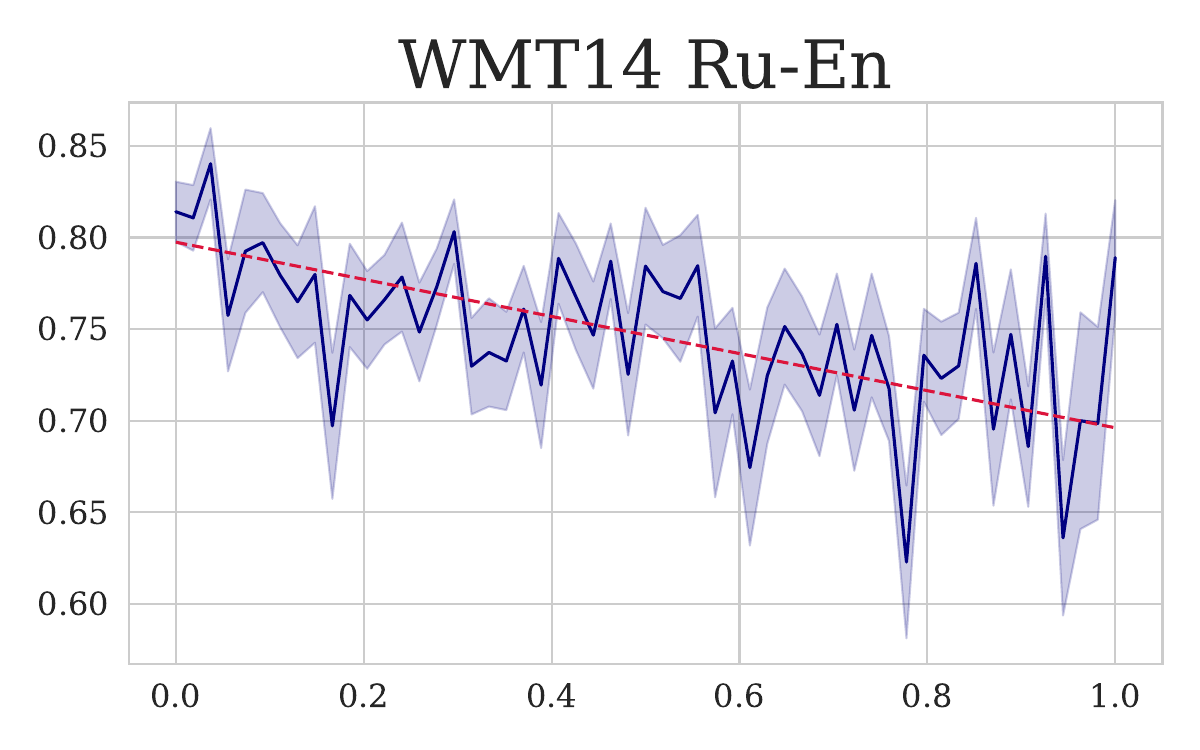}
    \end{subfigure}

    \begin{subfigure}{0.24\textwidth}
        \includegraphics[width=\linewidth]{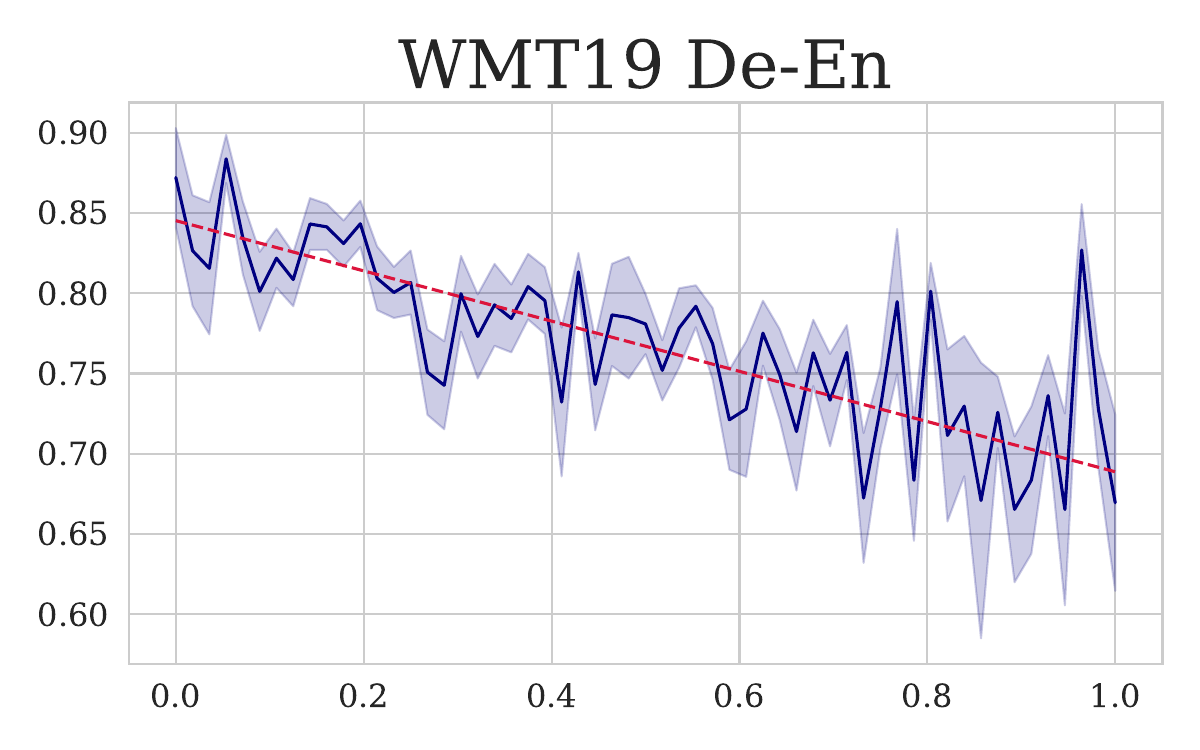}
    \end{subfigure}
    \begin{subfigure}{0.24\textwidth}
        \includegraphics[width=\linewidth]{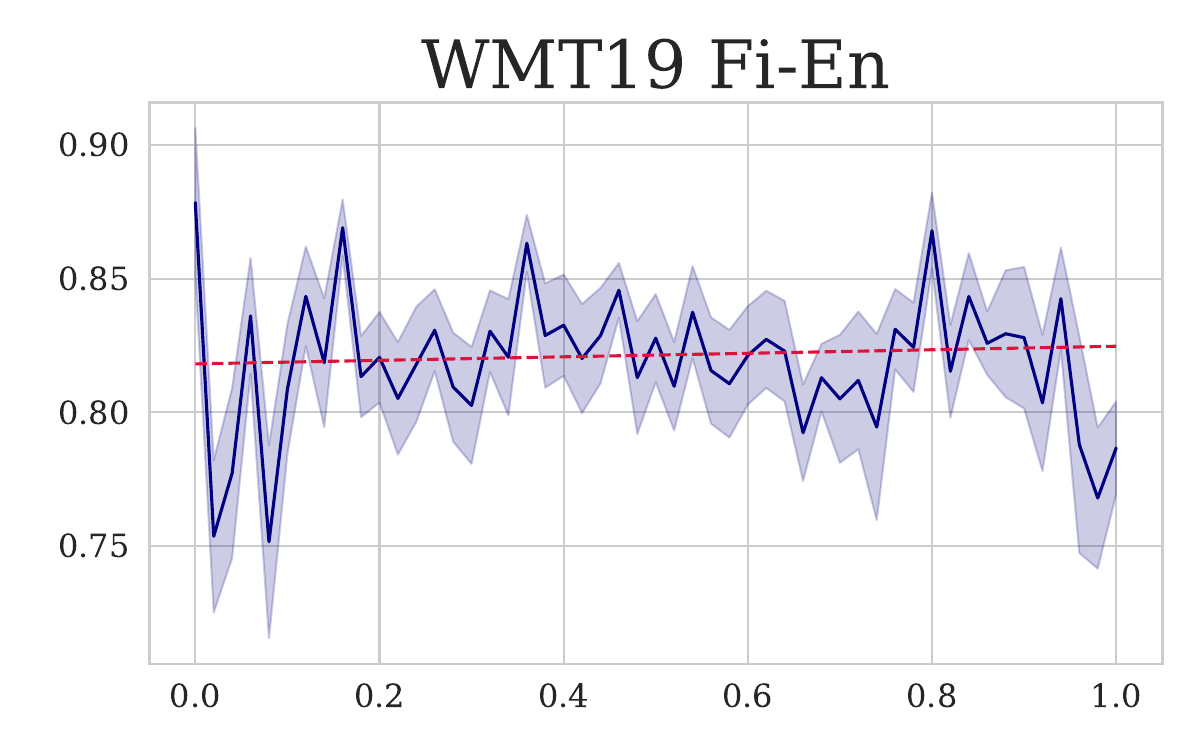}
    \end{subfigure}
    \begin{subfigure}{0.24\textwidth}
        \includegraphics[width=\linewidth]{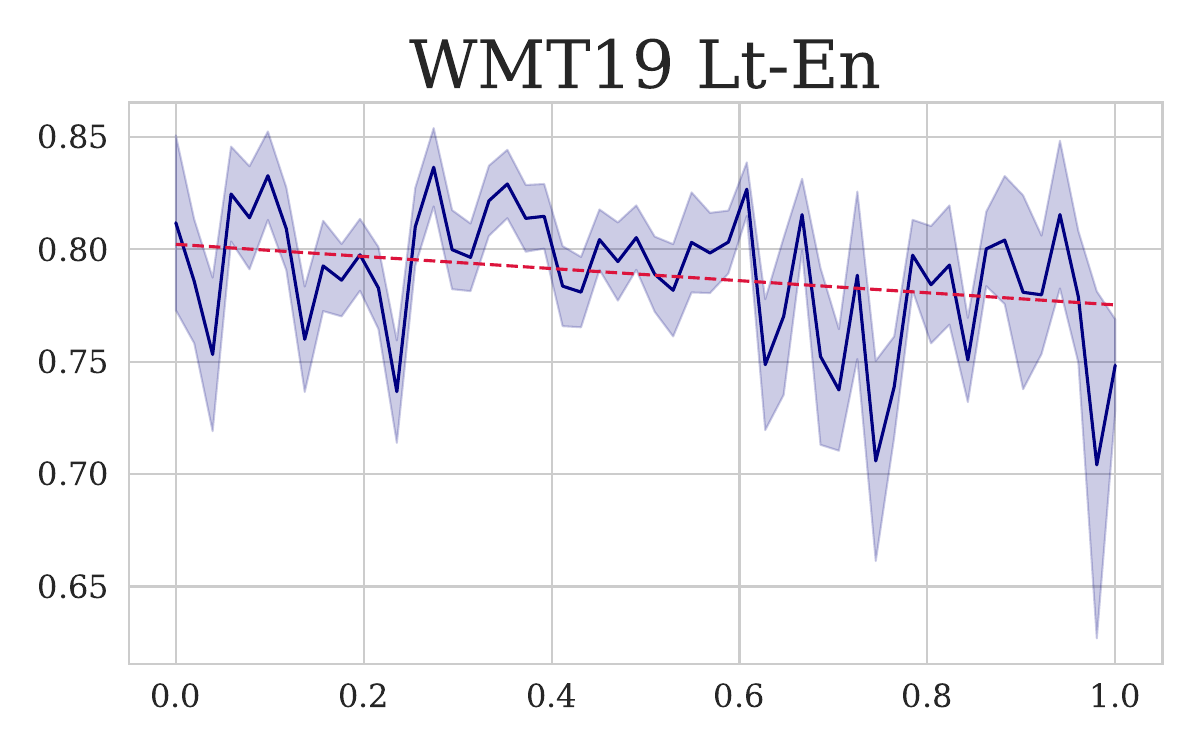}
    \end{subfigure}
    \begin{subfigure}{0.24\textwidth}
        \includegraphics[width=\linewidth]{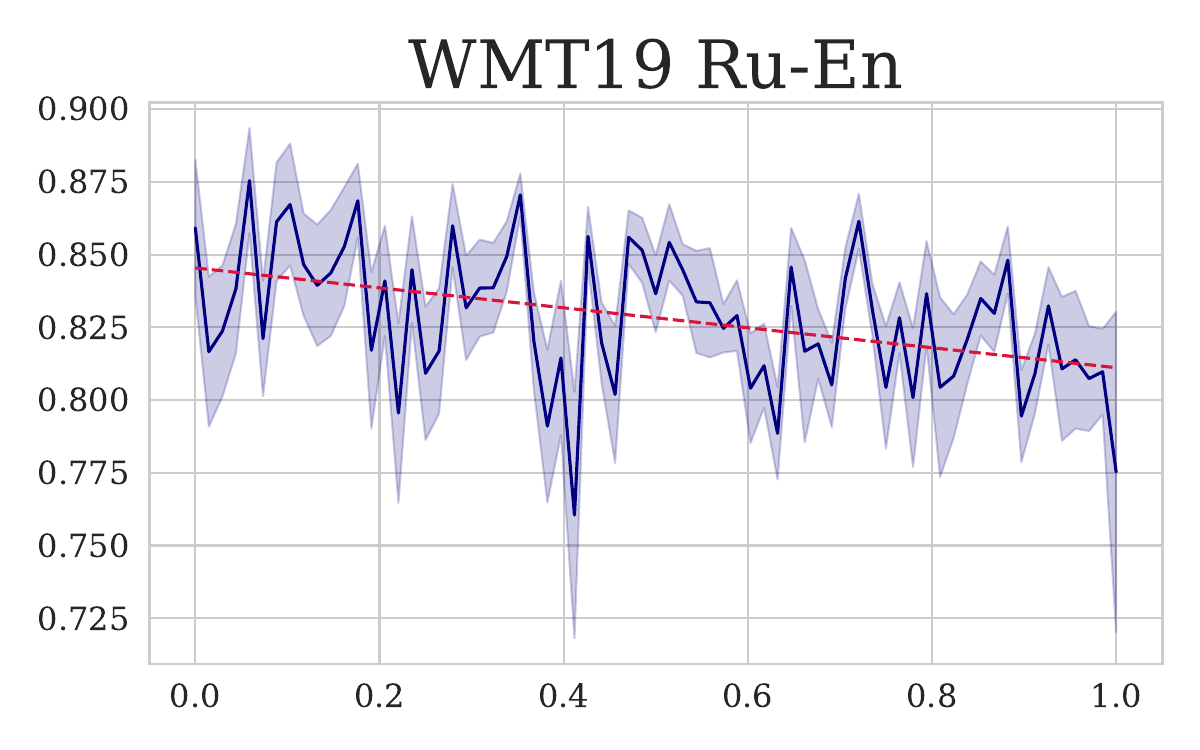}
    \end{subfigure}

    \caption{Comet score trends with respect to normalized generated sequence length across four machine translation datasets. Each subplot shows a linear regression fit over binned Comet scores.}
    \label{fig:comet_trends}
\end{figure*}

\begin{figure*}[ht!]
    \centering
\vspace{-0.5em}
{\centering \textbf{\small \llama} \par}
\vspace{0.3em}

    \begin{subfigure}{0.24\textwidth}
        \includegraphics[width=\linewidth]{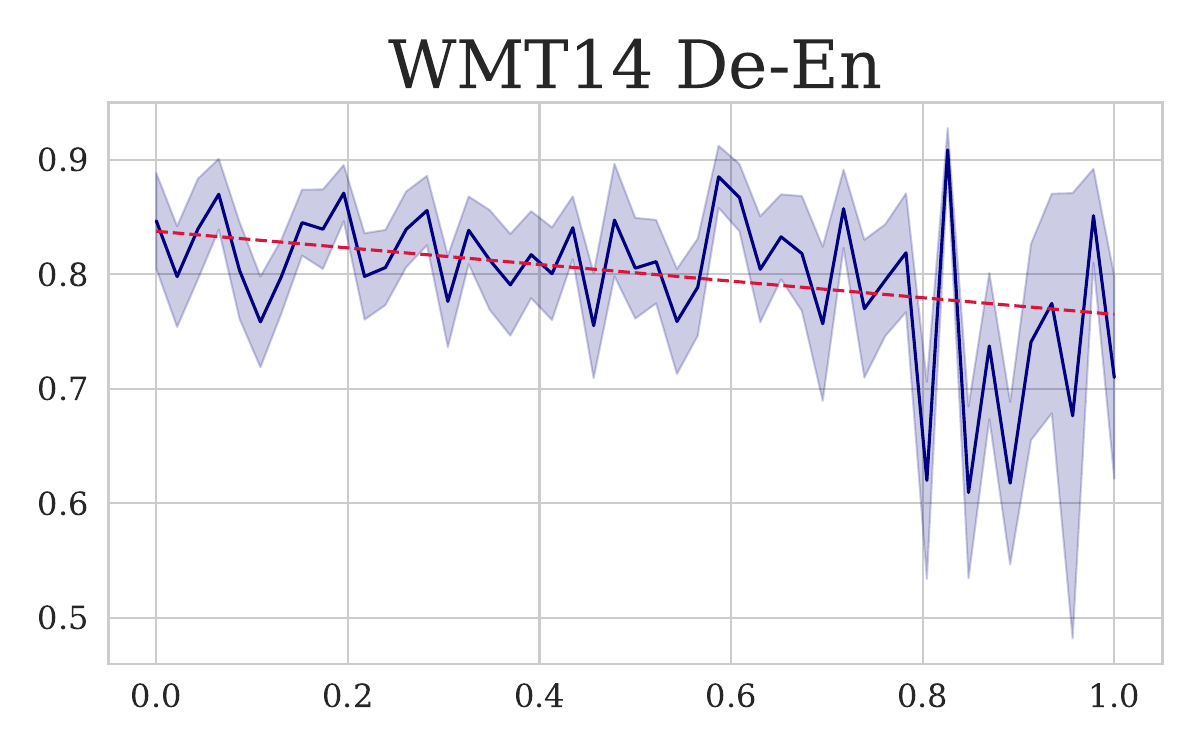}
    \end{subfigure}
    \begin{subfigure}{0.24\textwidth}
        \includegraphics[width=\linewidth]{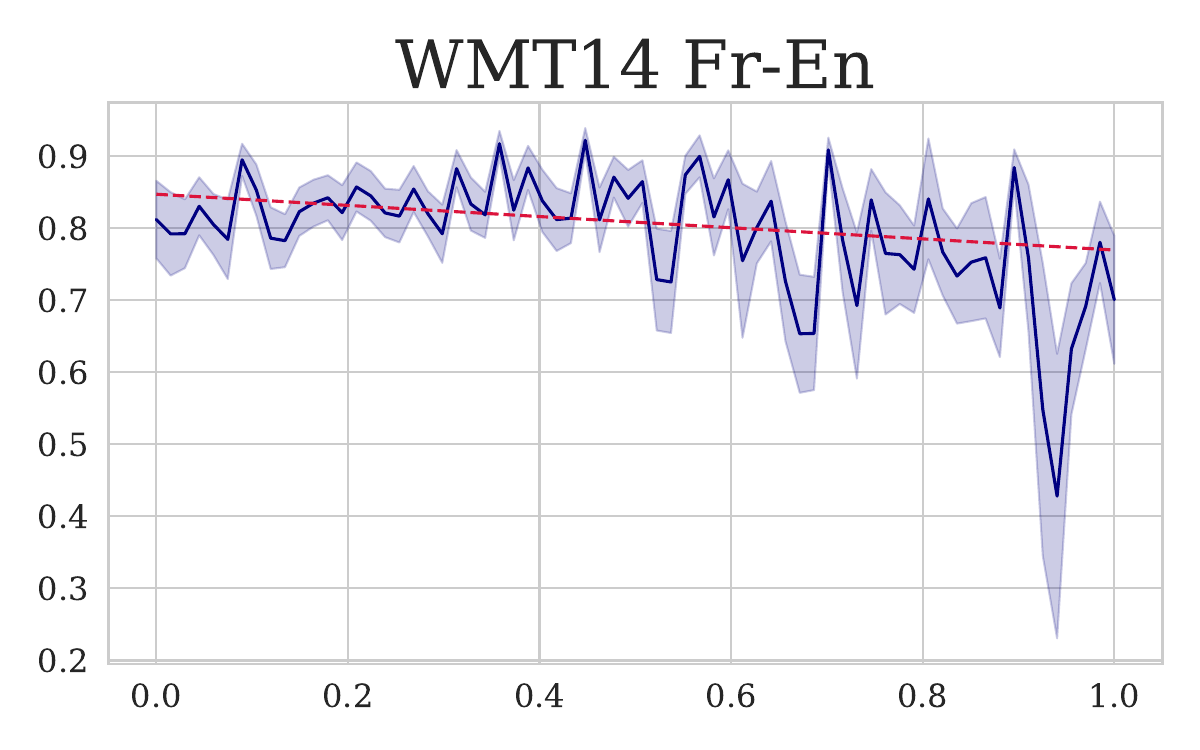}
    \end{subfigure}
    \begin{subfigure}{0.24\textwidth}
        \includegraphics[width=\linewidth]{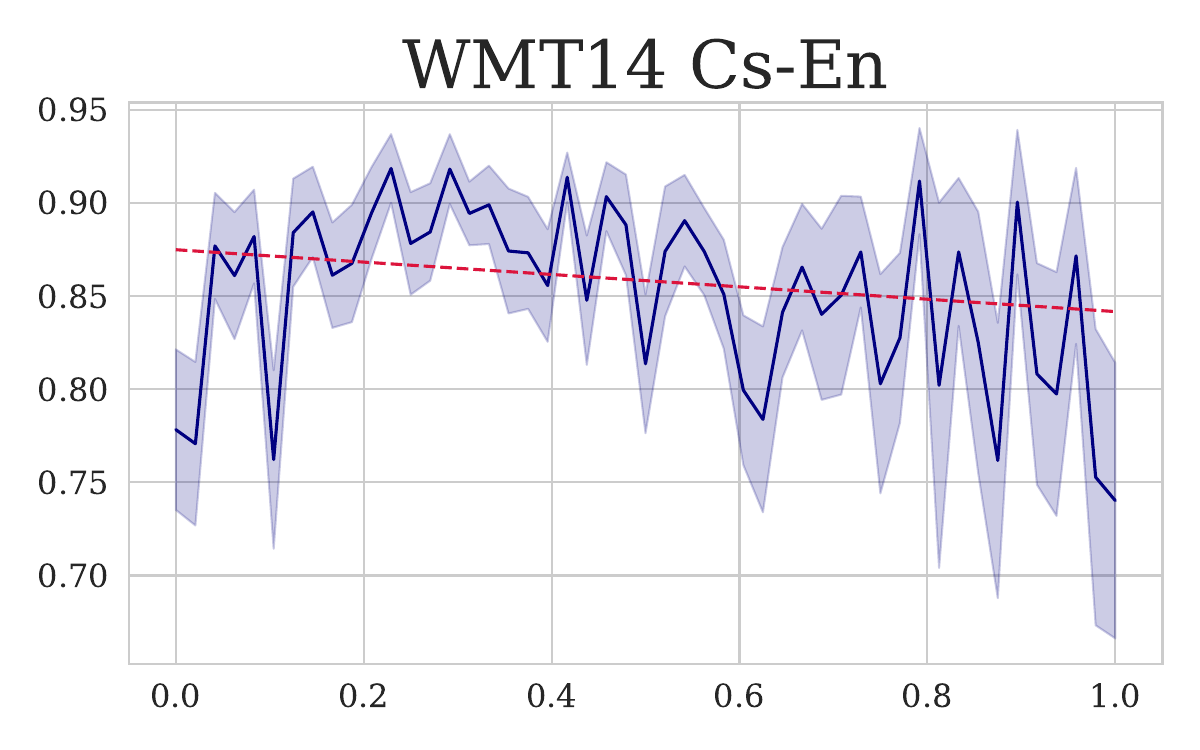}
    \end{subfigure}
    \begin{subfigure}{0.24\textwidth}
        \includegraphics[width=\linewidth]{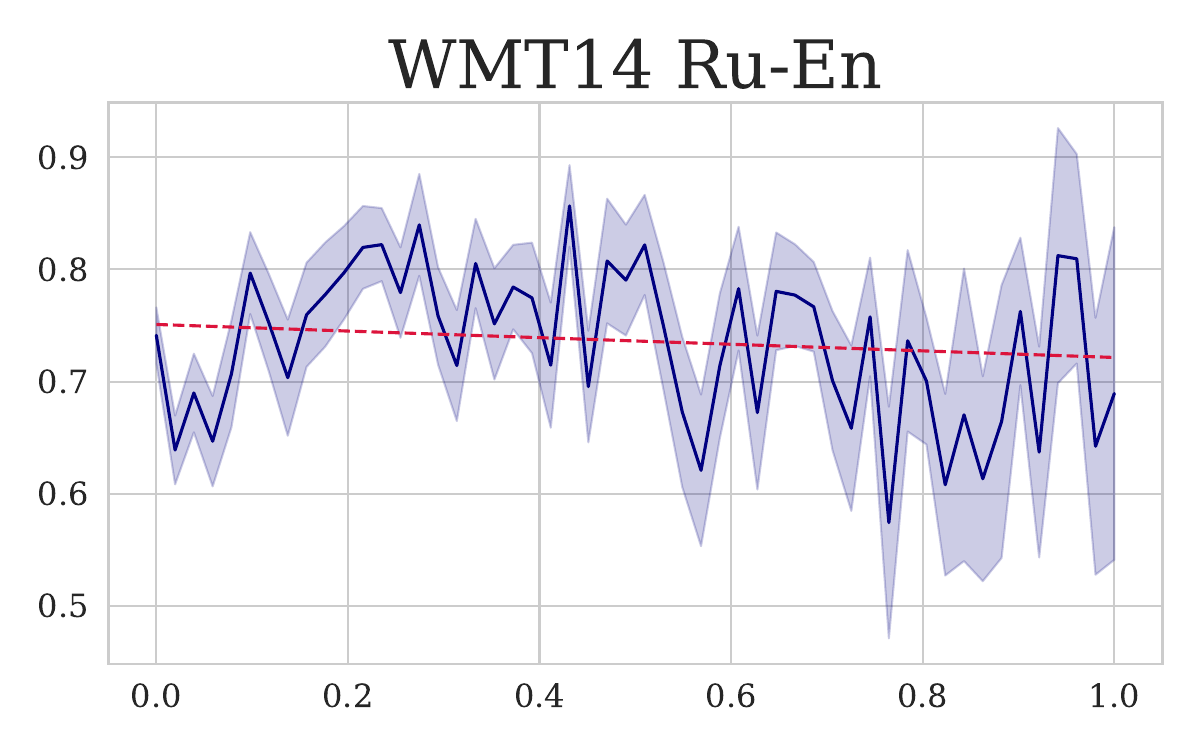}
    \end{subfigure}

    \begin{subfigure}{0.24\textwidth}
        \includegraphics[width=\linewidth]{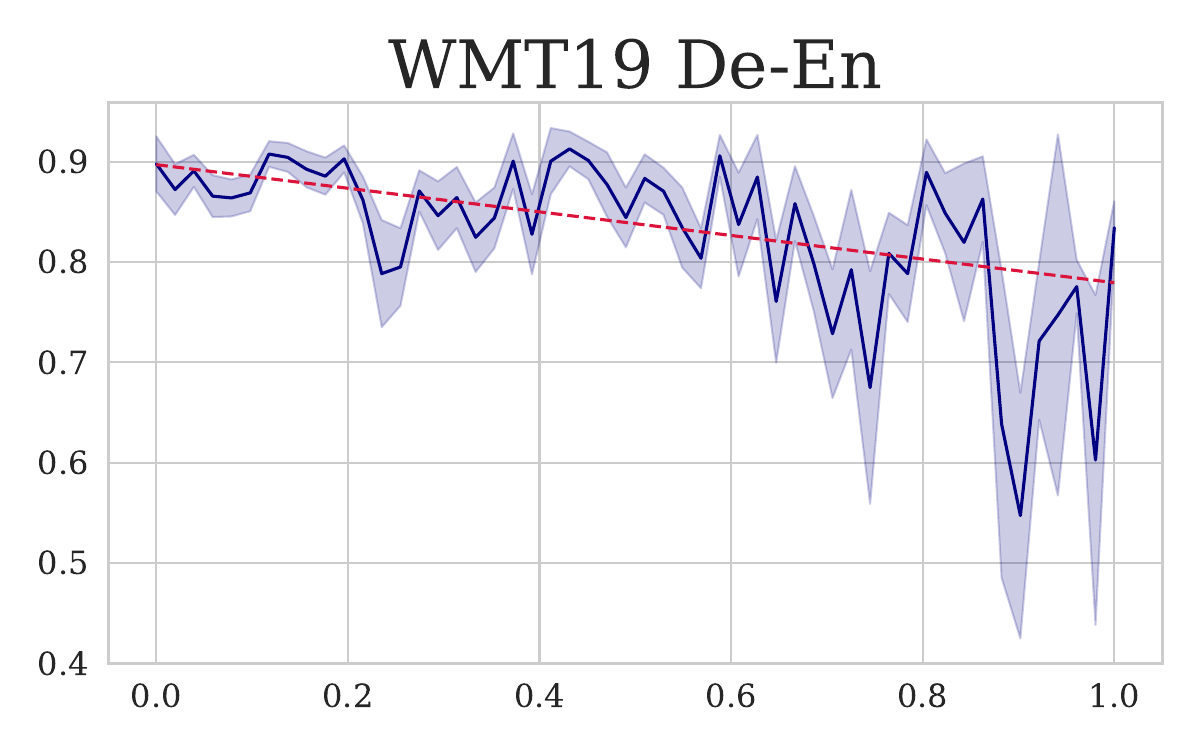}
    \end{subfigure}
    \begin{subfigure}{0.24\textwidth}
        \includegraphics[width=\linewidth]{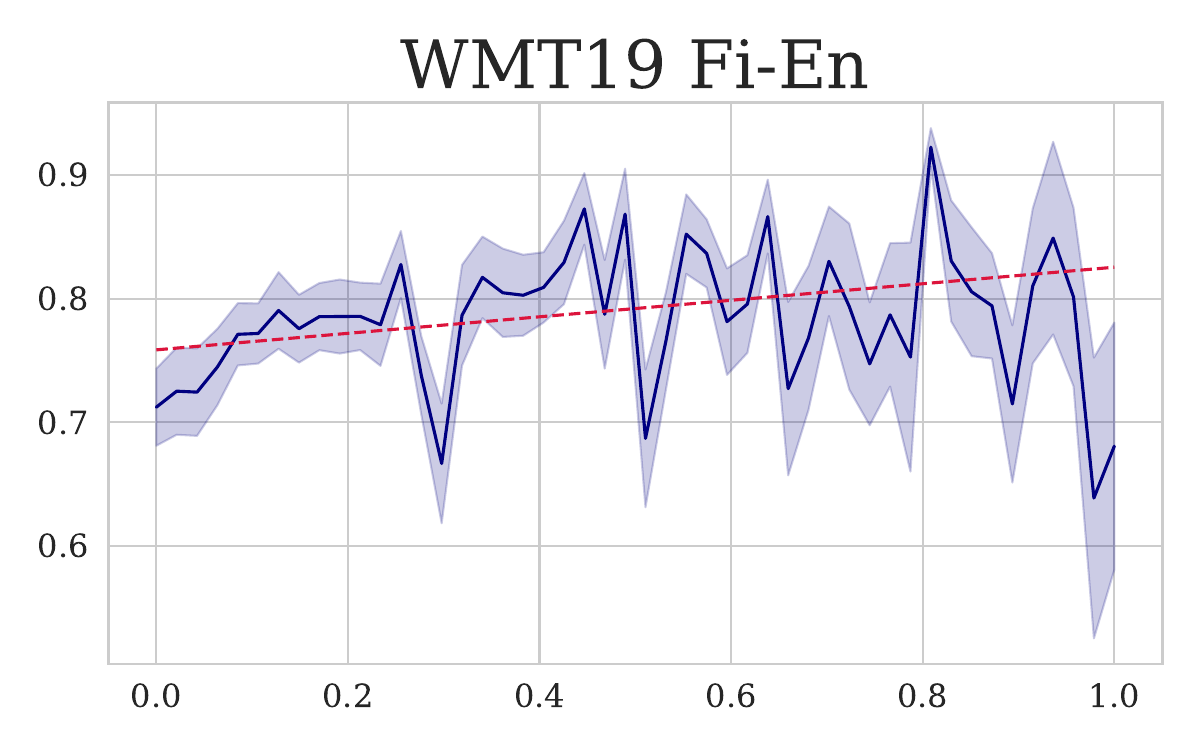}
    \end{subfigure}
    \begin{subfigure}{0.24\textwidth}
        \includegraphics[width=\linewidth]{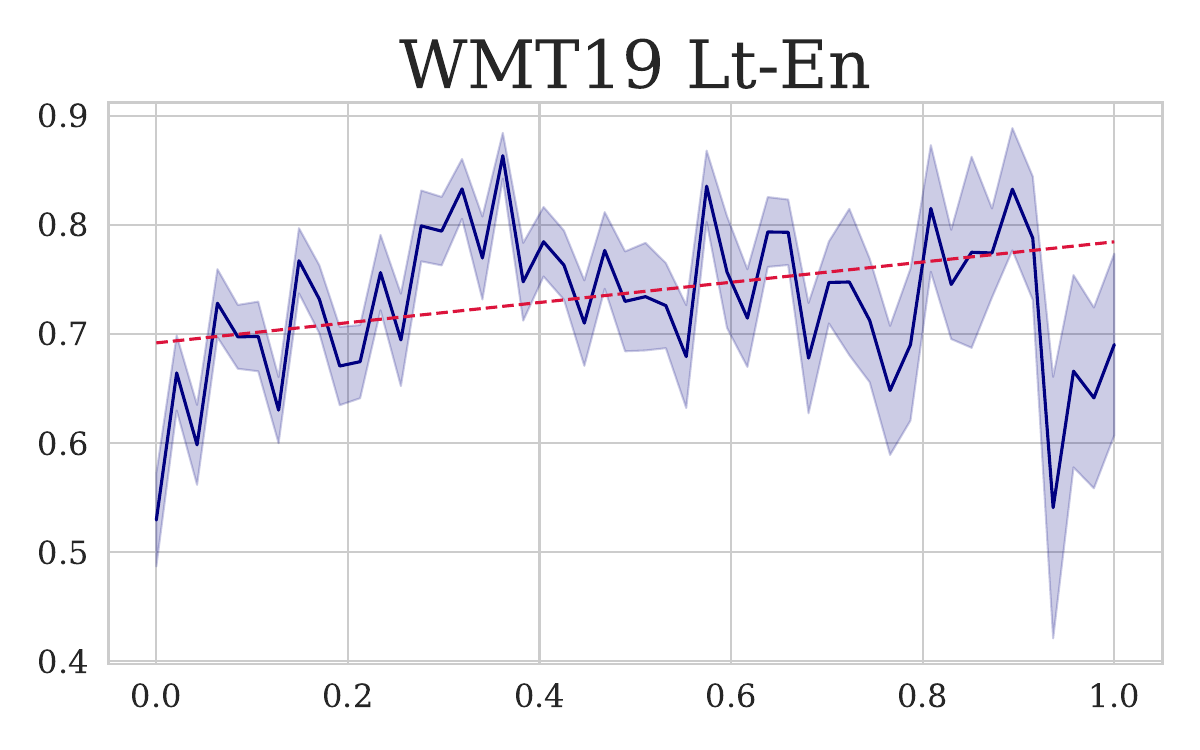}
    \end{subfigure}
    \begin{subfigure}{0.24\textwidth}
        \includegraphics[width=\linewidth]{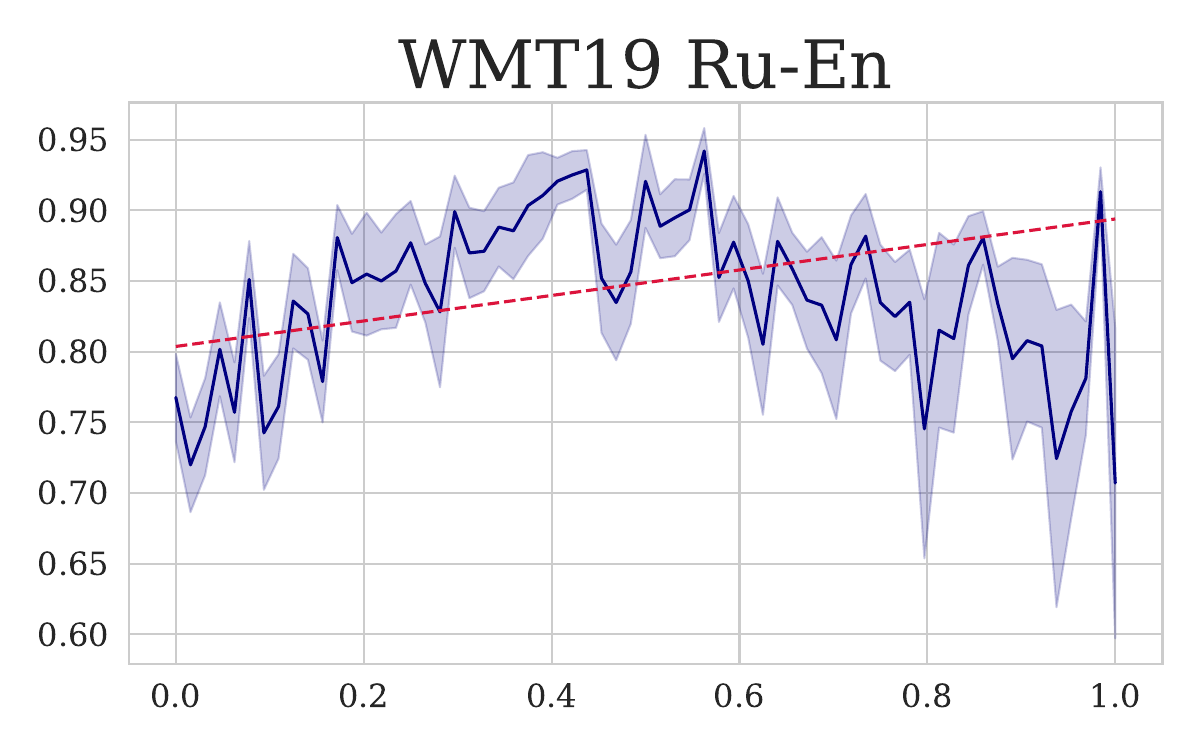}
    \end{subfigure}

\vspace{-0.5em}
{\centering \textbf{\small \gemma} \par}
\vspace{0.3em}

 \begin{subfigure}{0.24\textwidth}
        \includegraphics[width=\linewidth]{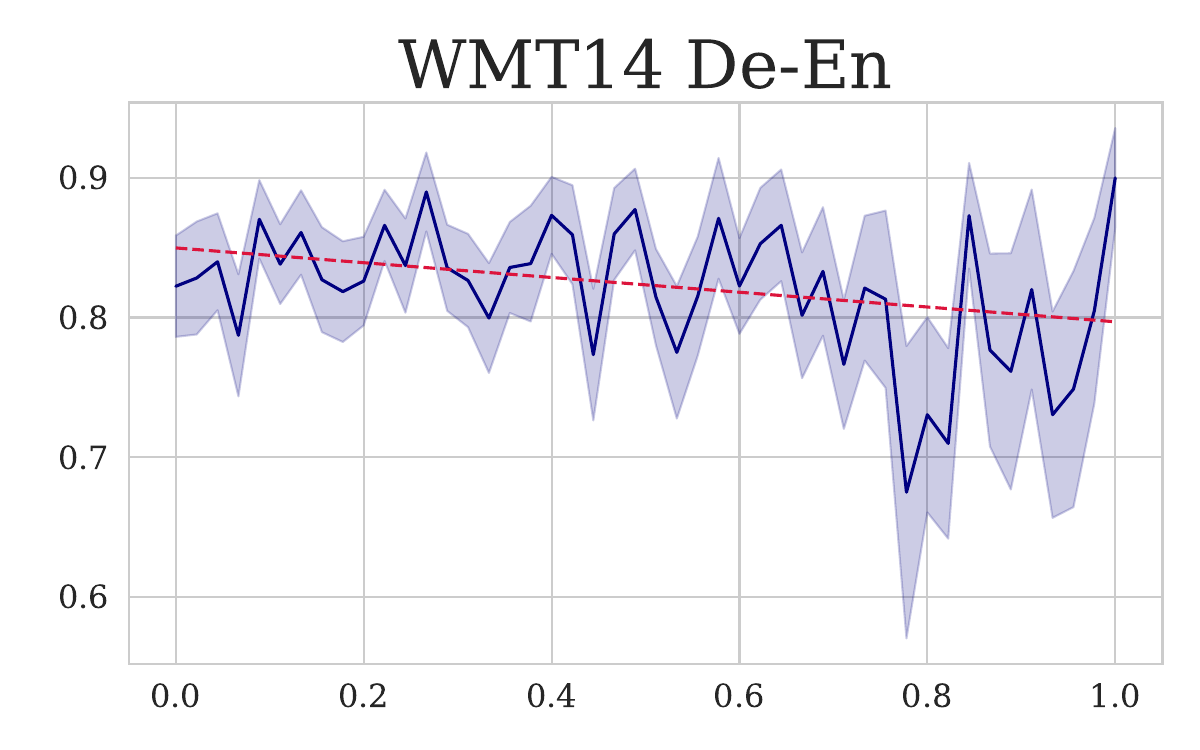}
    \end{subfigure}
    \begin{subfigure}{0.24\textwidth}
        \includegraphics[width=\linewidth]{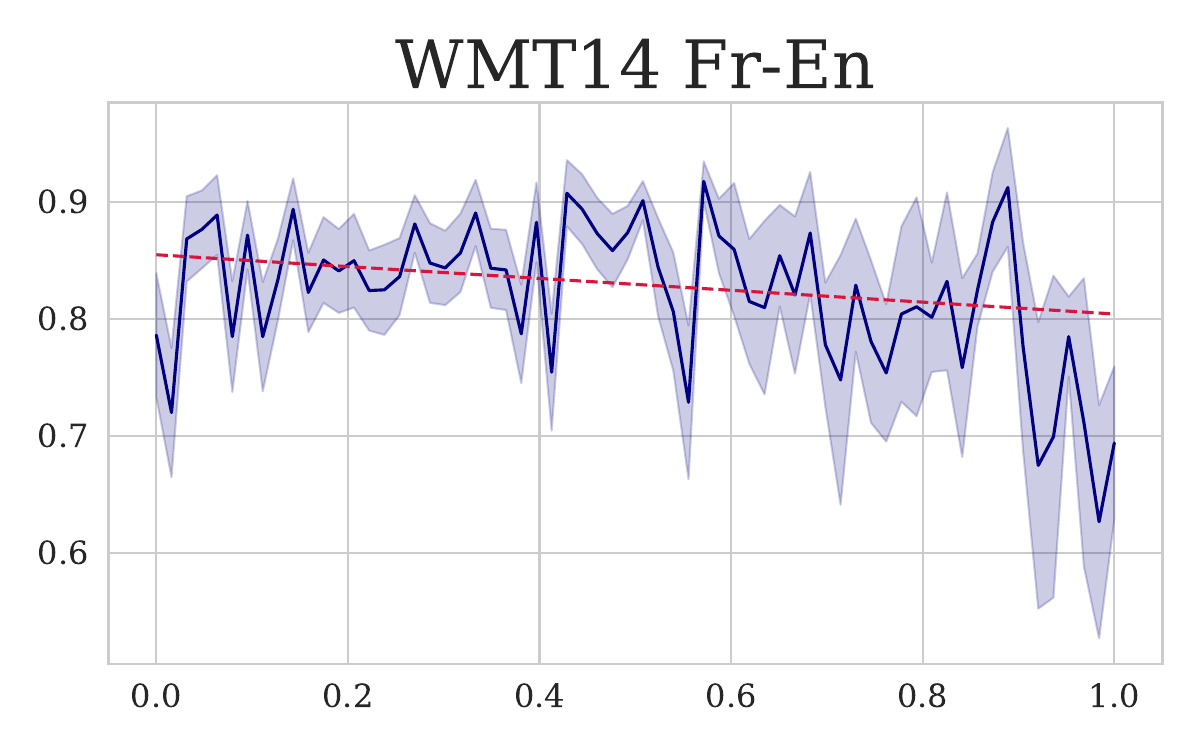}
    \end{subfigure}
    \begin{subfigure}{0.24\textwidth}
        \includegraphics[width=\linewidth]{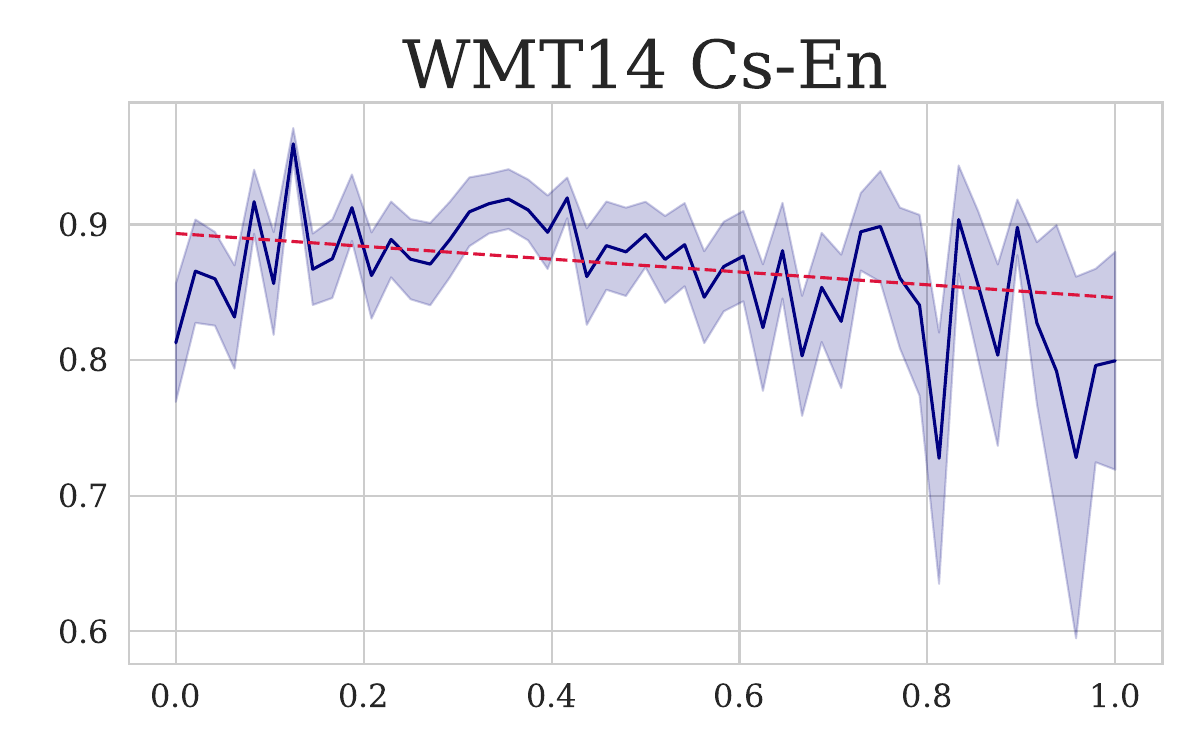}
    \end{subfigure}
    \begin{subfigure}{0.24\textwidth}
        \includegraphics[width=\linewidth]{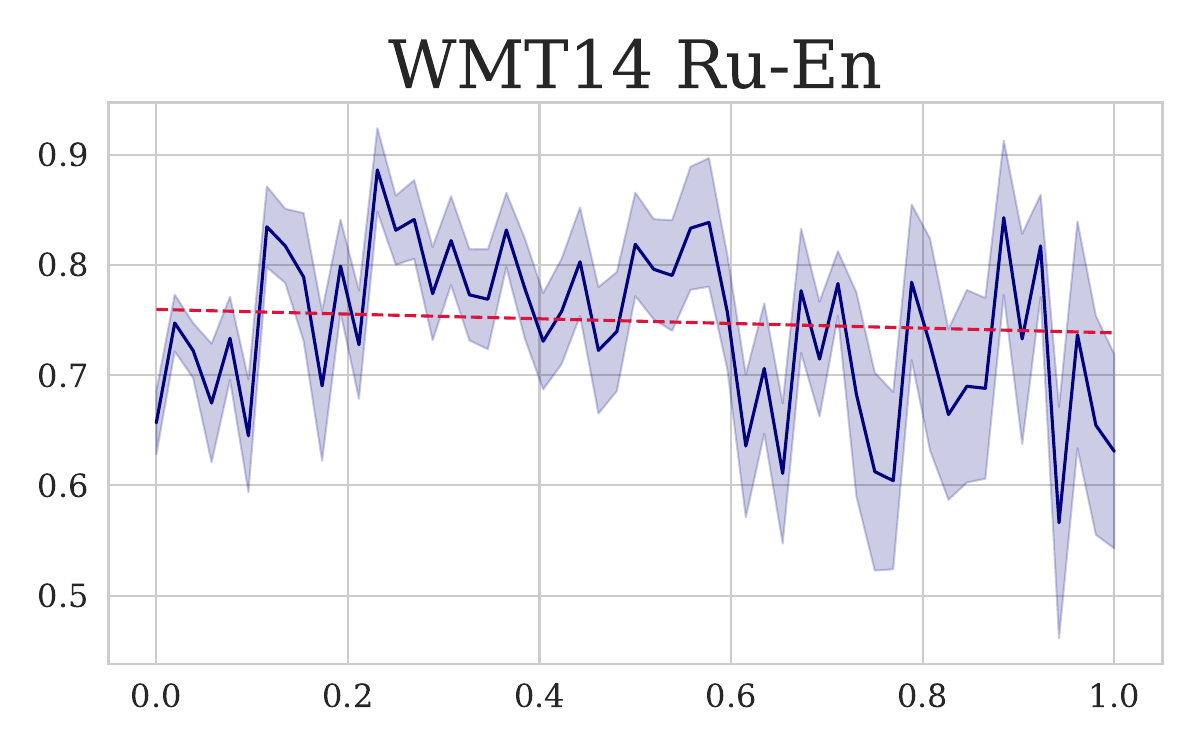}
    \end{subfigure}

    \begin{subfigure}{0.24\textwidth}
        \includegraphics[width=\linewidth]{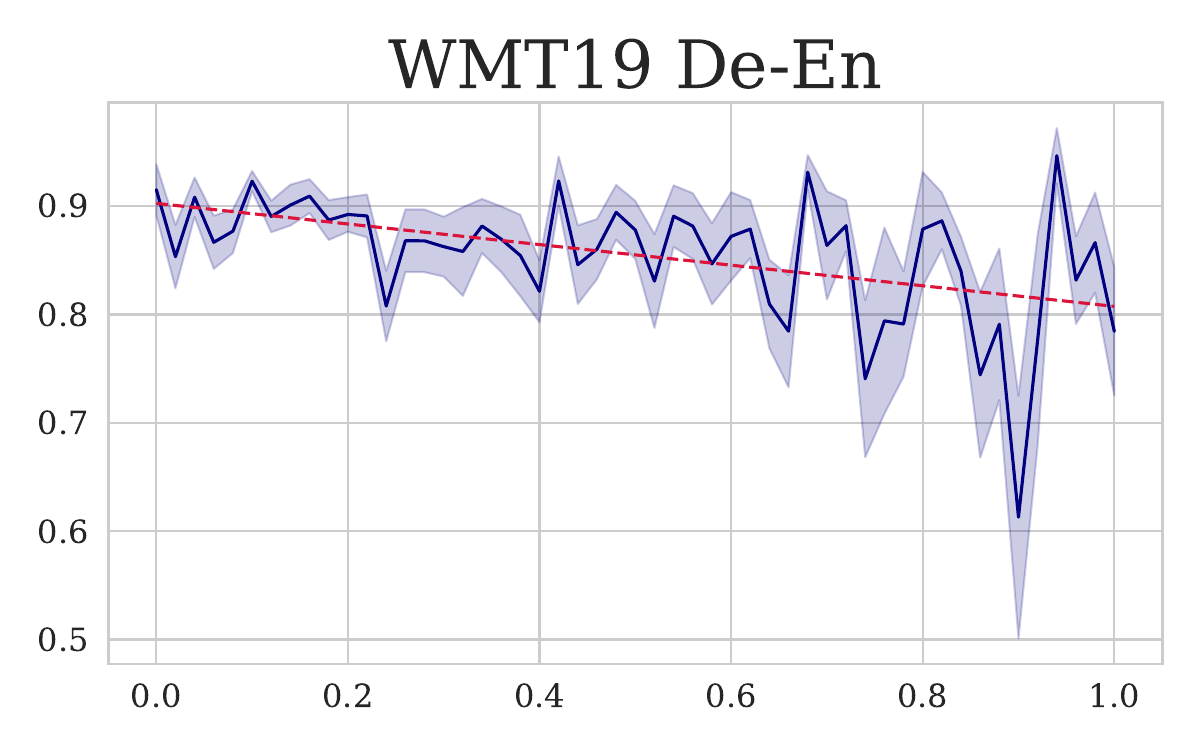}
    \end{subfigure}
    \begin{subfigure}{0.24\textwidth}
        \includegraphics[width=\linewidth]{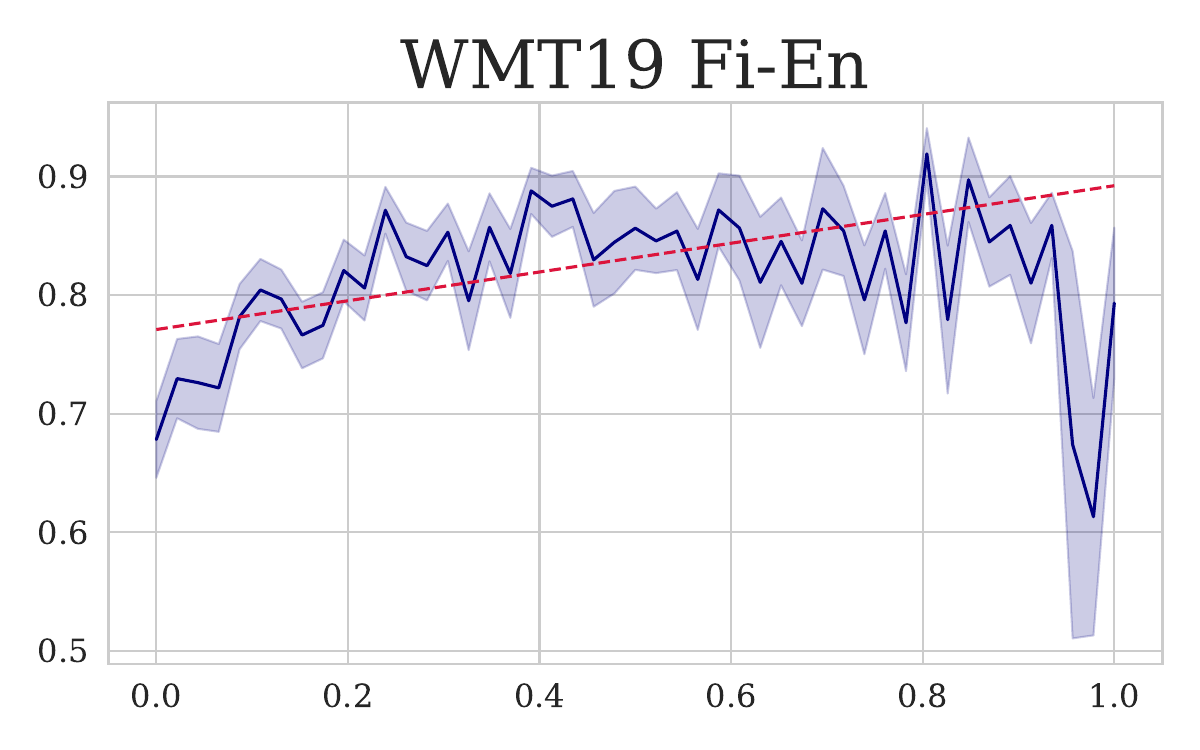}
    \end{subfigure}
    \begin{subfigure}{0.24\textwidth}
        \includegraphics[width=\linewidth]{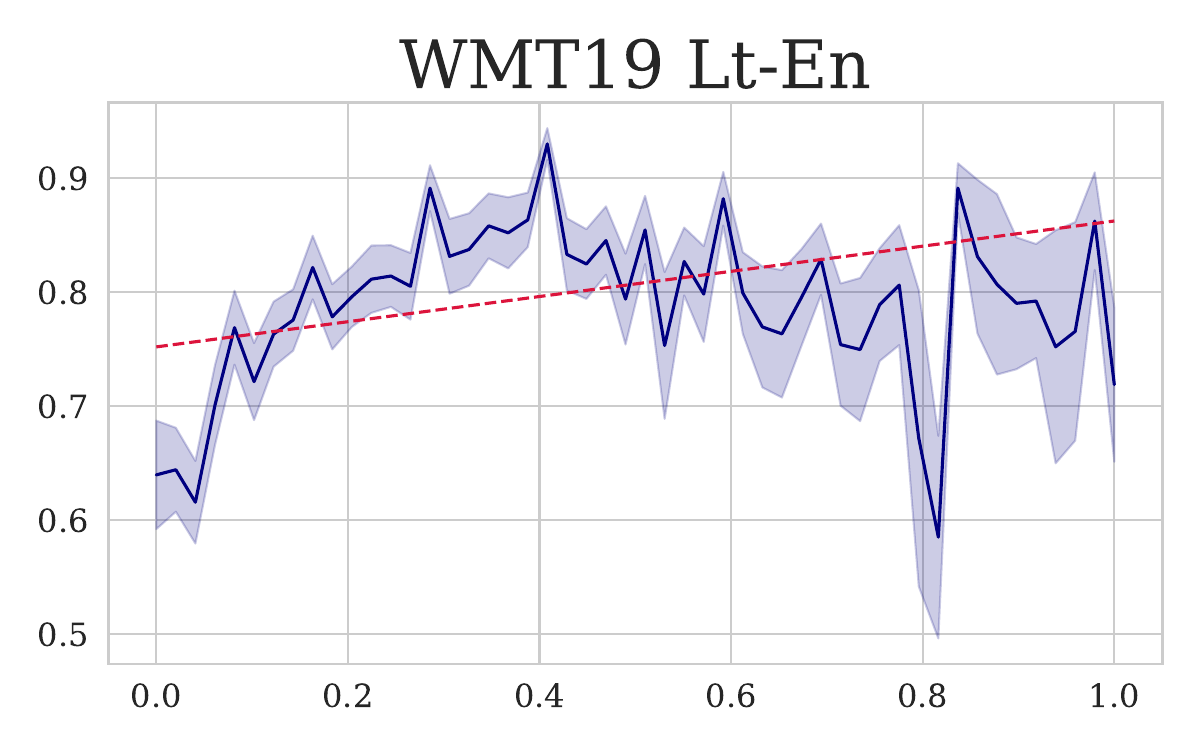}
    \end{subfigure}
    \begin{subfigure}{0.24\textwidth}
        \includegraphics[width=\linewidth]{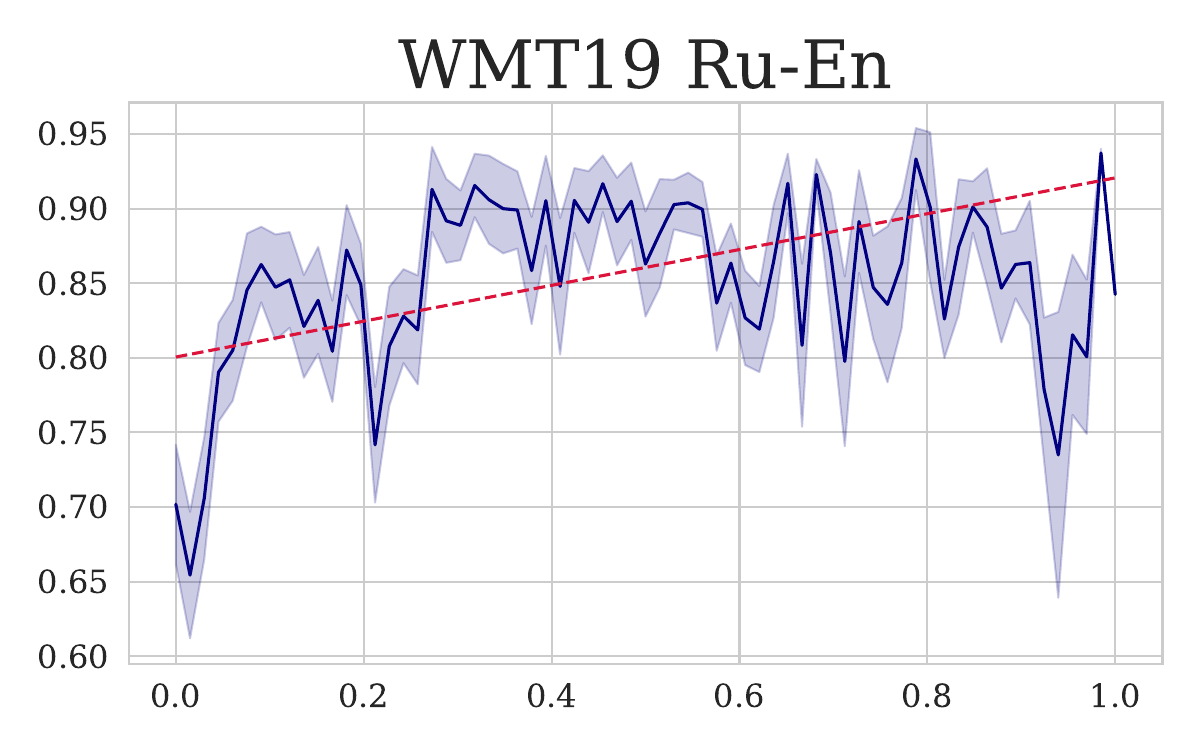}
    \end{subfigure}

\vspace{-0.5em}
{\centering \textbf{\small \eurollm} \par}
\vspace{0.3em}
 \begin{subfigure}{0.24\textwidth}
        \includegraphics[width=\linewidth]{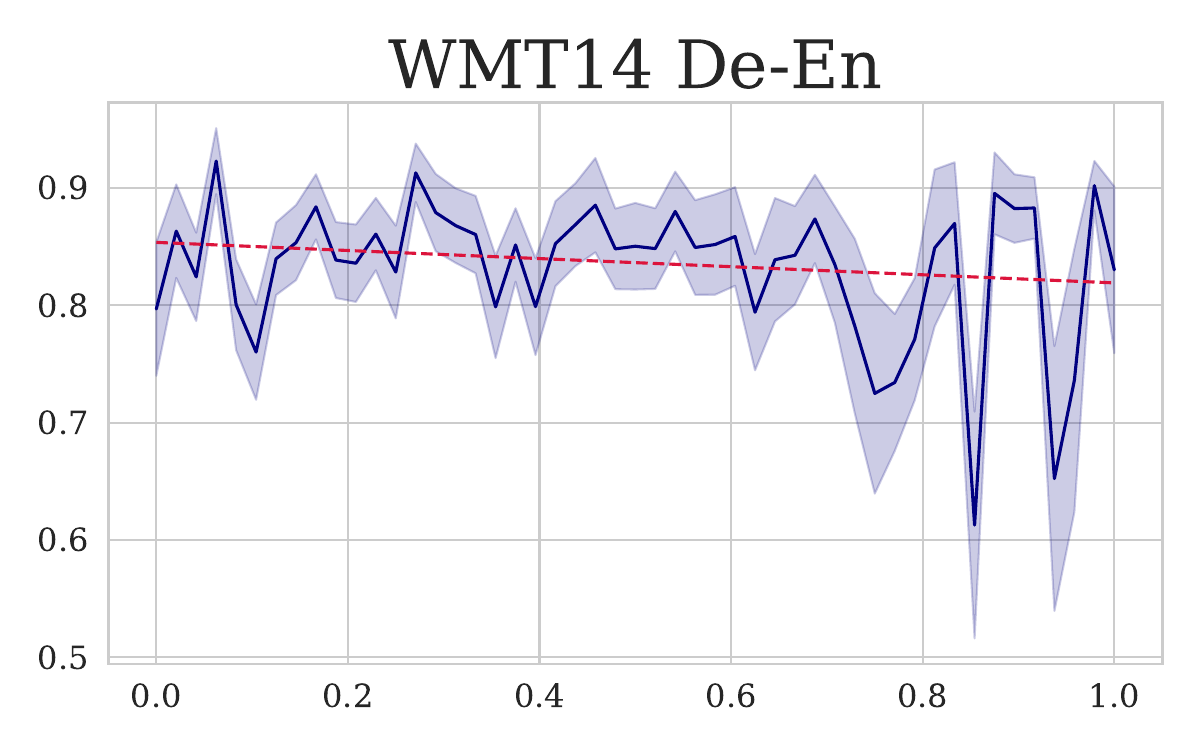}
    \end{subfigure}
    \begin{subfigure}{0.24\textwidth}
        \includegraphics[width=\linewidth]{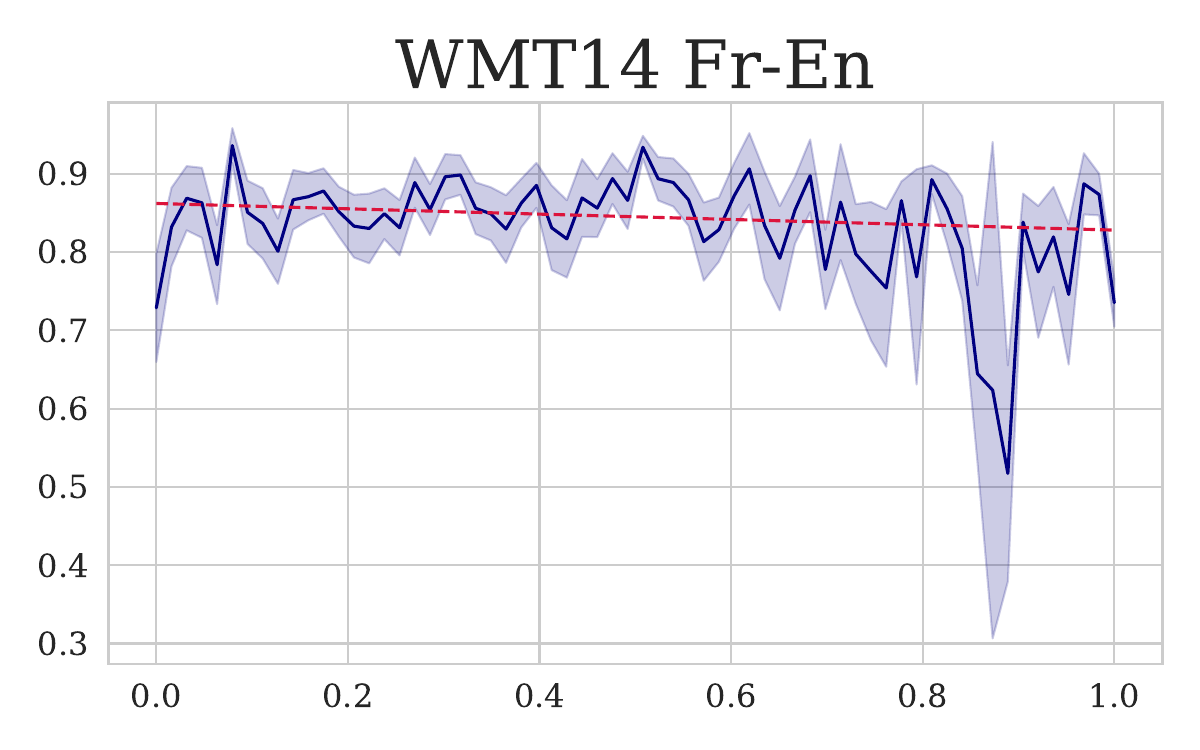}
    \end{subfigure}
    \begin{subfigure}{0.24\textwidth}
        \includegraphics[width=\linewidth]{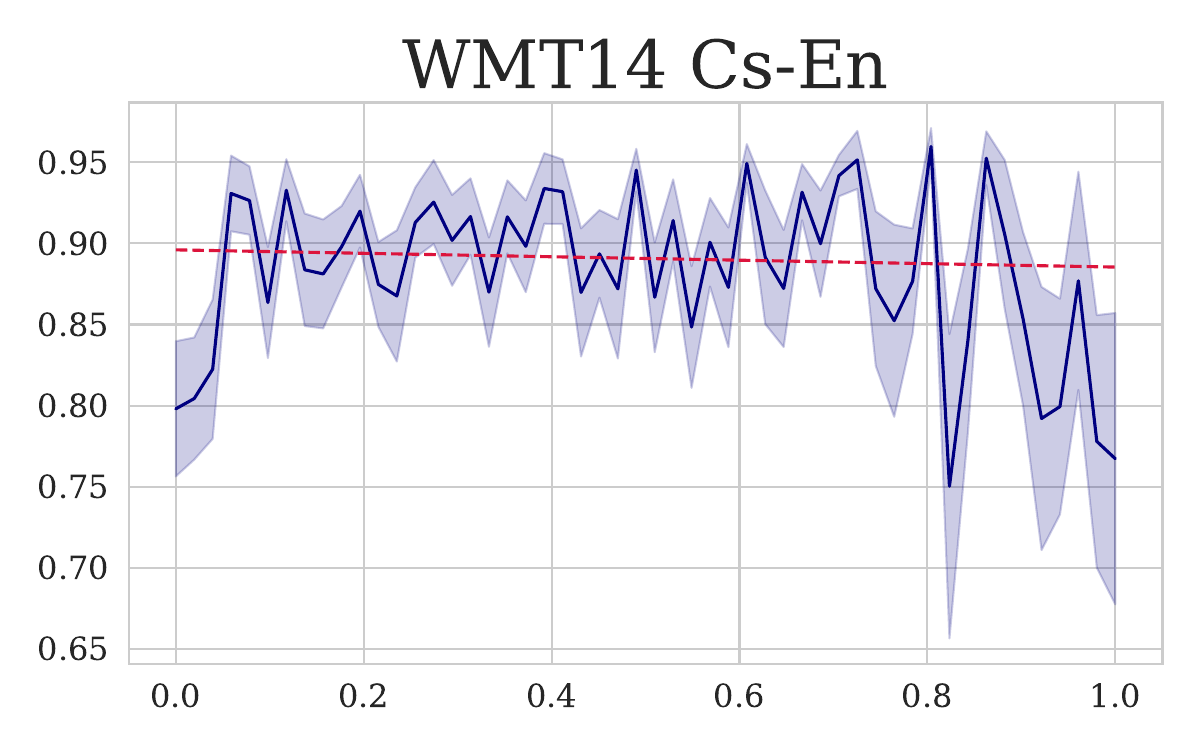}
    \end{subfigure}
    \begin{subfigure}{0.24\textwidth}
        \includegraphics[width=\linewidth]{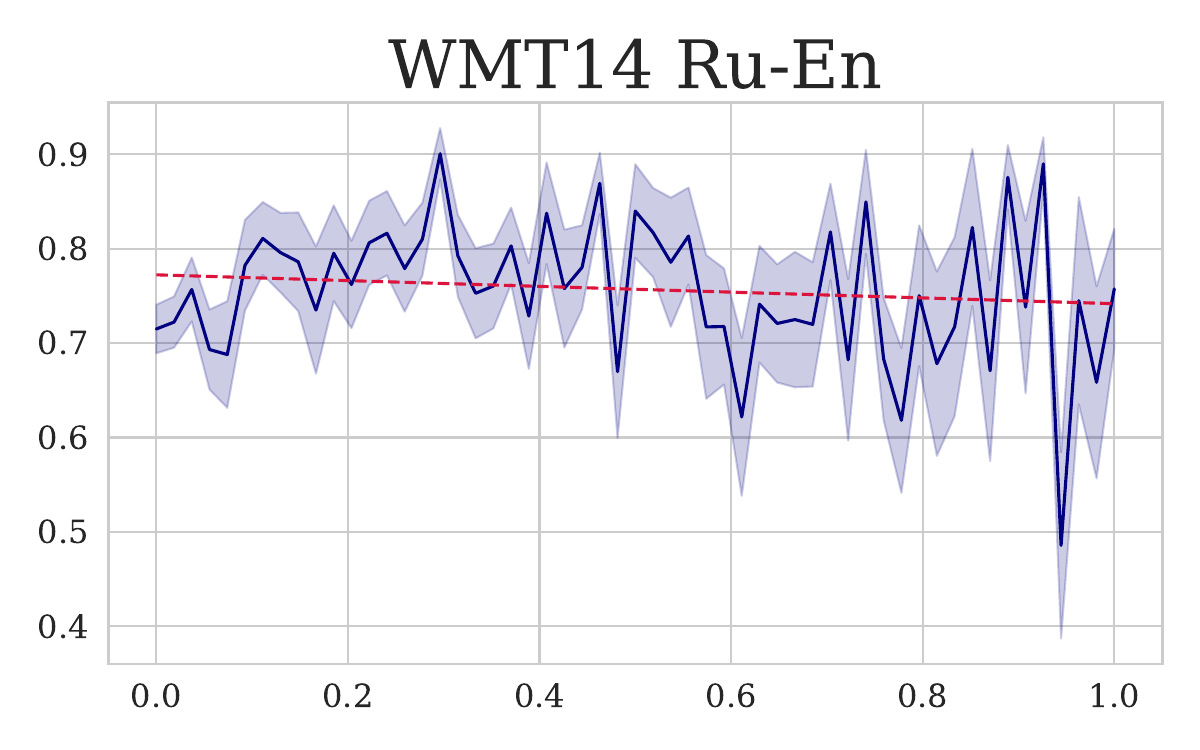}
    \end{subfigure}

    \begin{subfigure}{0.24\textwidth}
        \includegraphics[width=\linewidth]{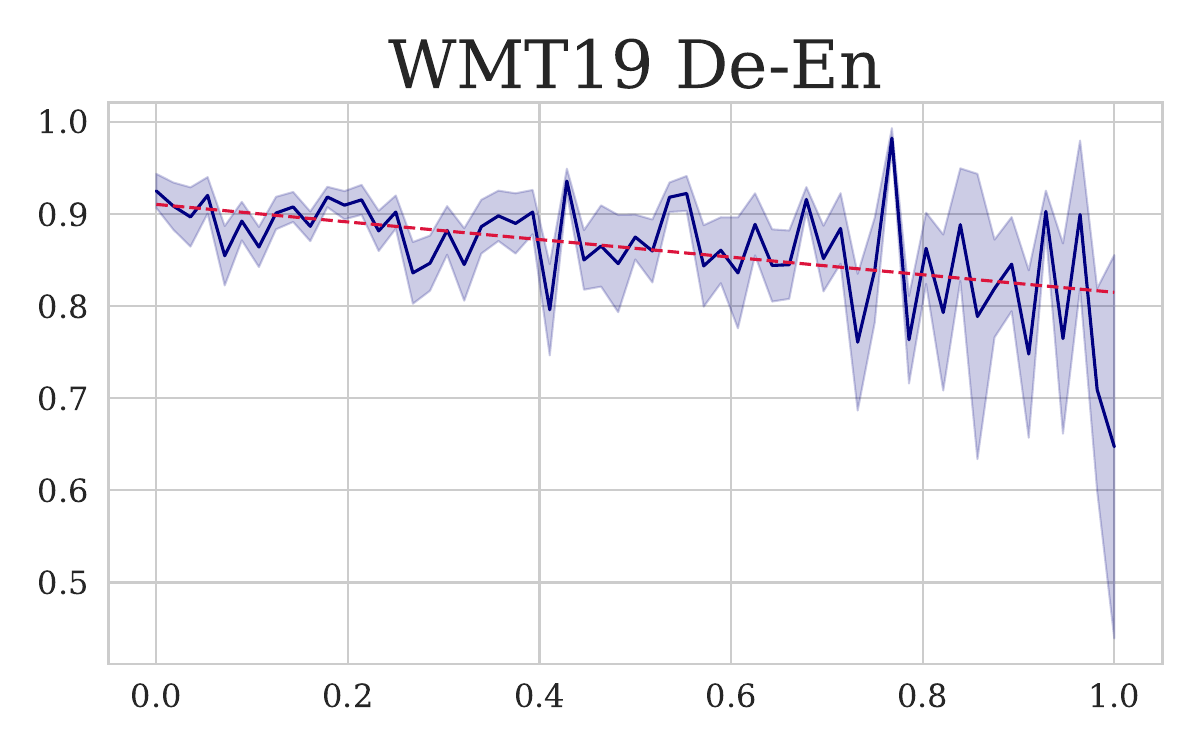}
    \end{subfigure}
    \begin{subfigure}{0.24\textwidth}
        \includegraphics[width=\linewidth]{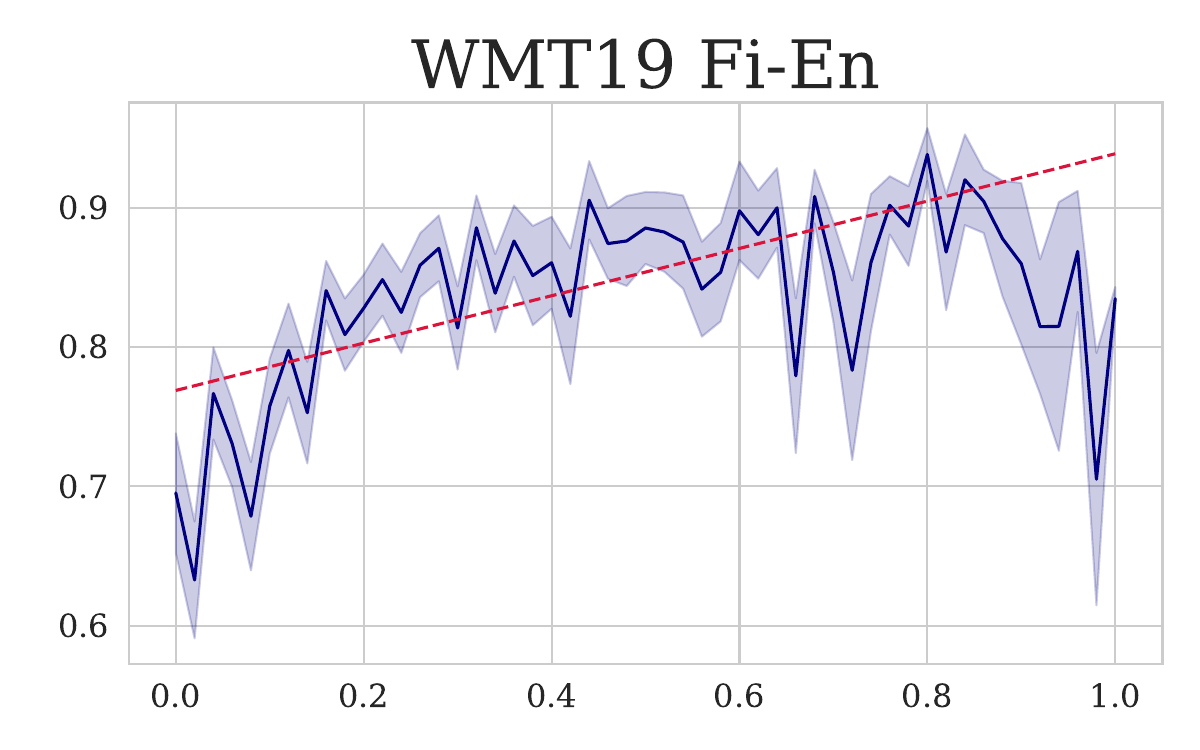}
    \end{subfigure}
    \begin{subfigure}{0.24\textwidth}
        \includegraphics[width=\linewidth]{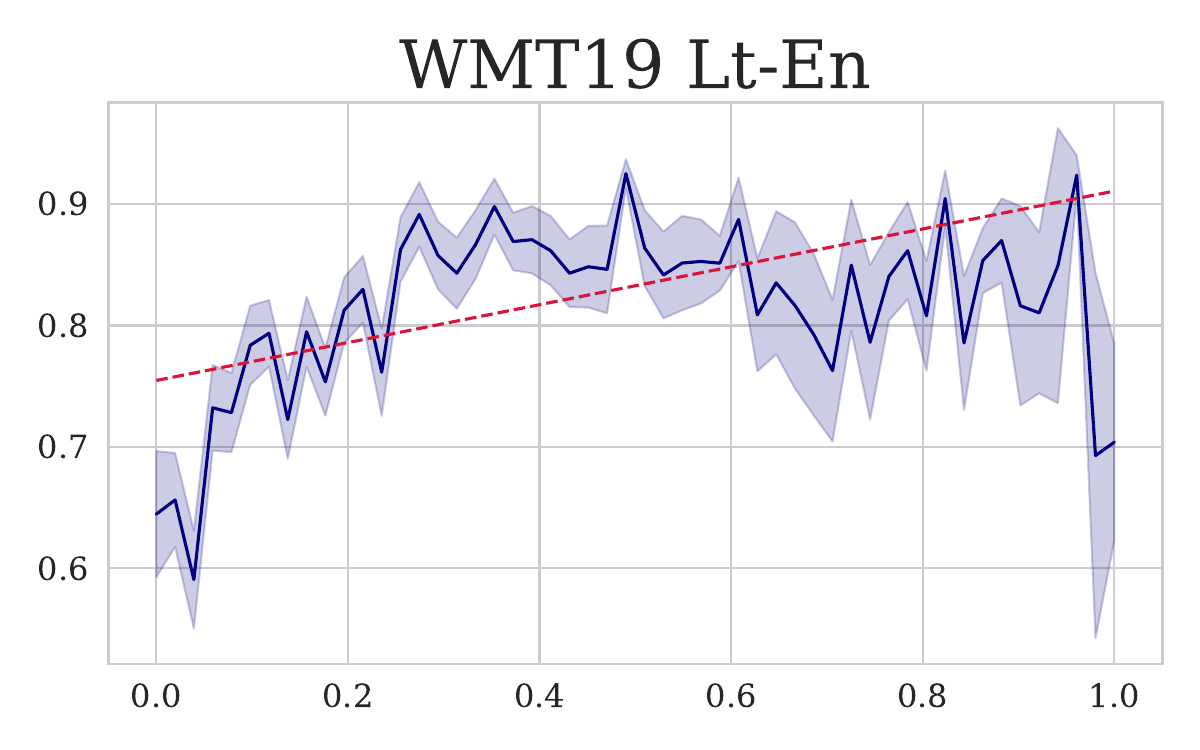}
    \end{subfigure}
    \begin{subfigure}{0.24\textwidth}
        \includegraphics[width=\linewidth]{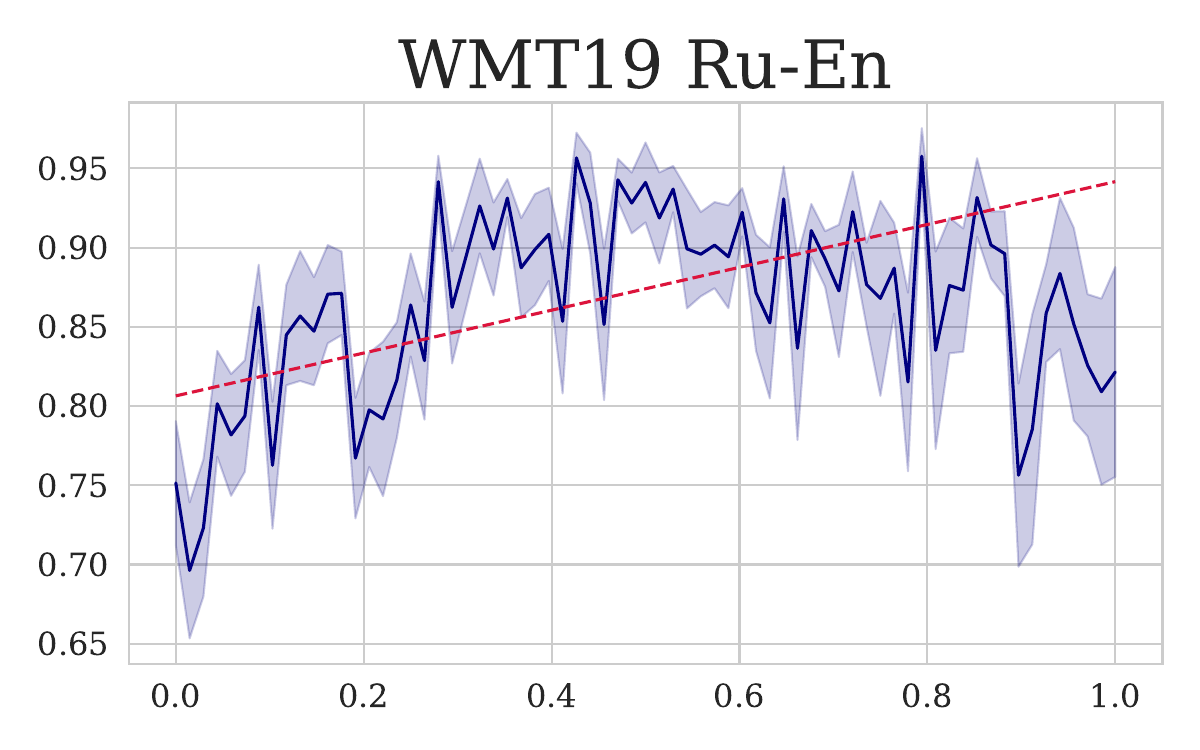}
    \end{subfigure}

    \caption{XComet-XXL score trends with respect to normalized generated sequence length across four machine translation datasets. Each subplot shows a linear regression fit over binned XComet-XXL scores.}
    \label{fig:xcometxxl_trends}
\end{figure*}

\begin{figure*}[h!]
    \centering

\vspace{-0.5em}
{\centering \textbf{\small \llama} \par}
\vspace{0.3em}

    \begin{subfigure}{0.24\textwidth}
        \includegraphics[width=\linewidth]{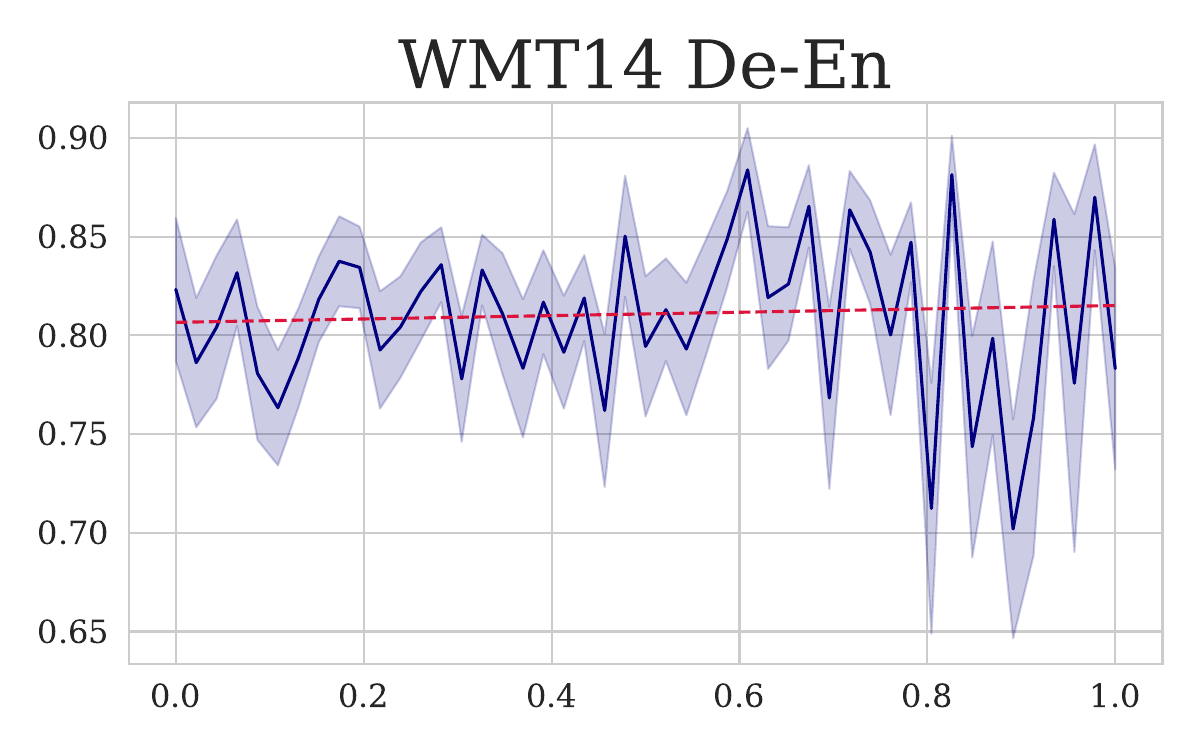}
    \end{subfigure}
    \begin{subfigure}{0.24\textwidth}
        \includegraphics[width=\linewidth]{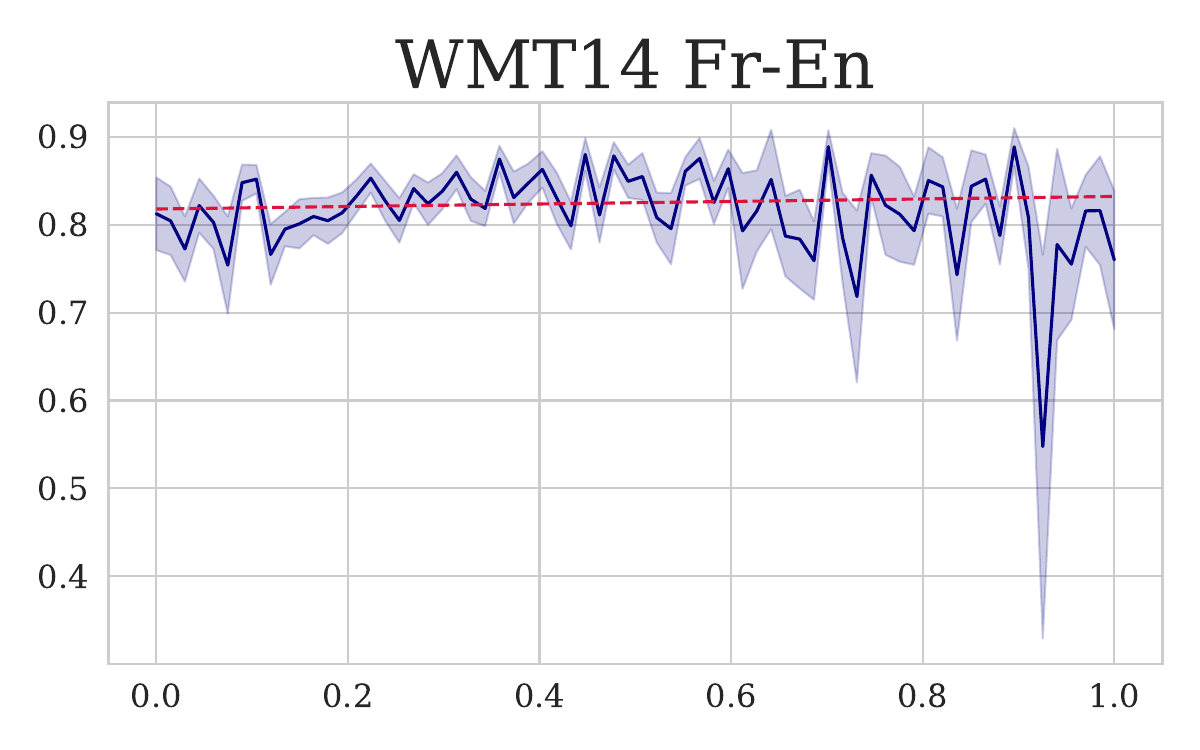}
    \end{subfigure}
    \begin{subfigure}{0.24\textwidth}
        \includegraphics[width=\linewidth]{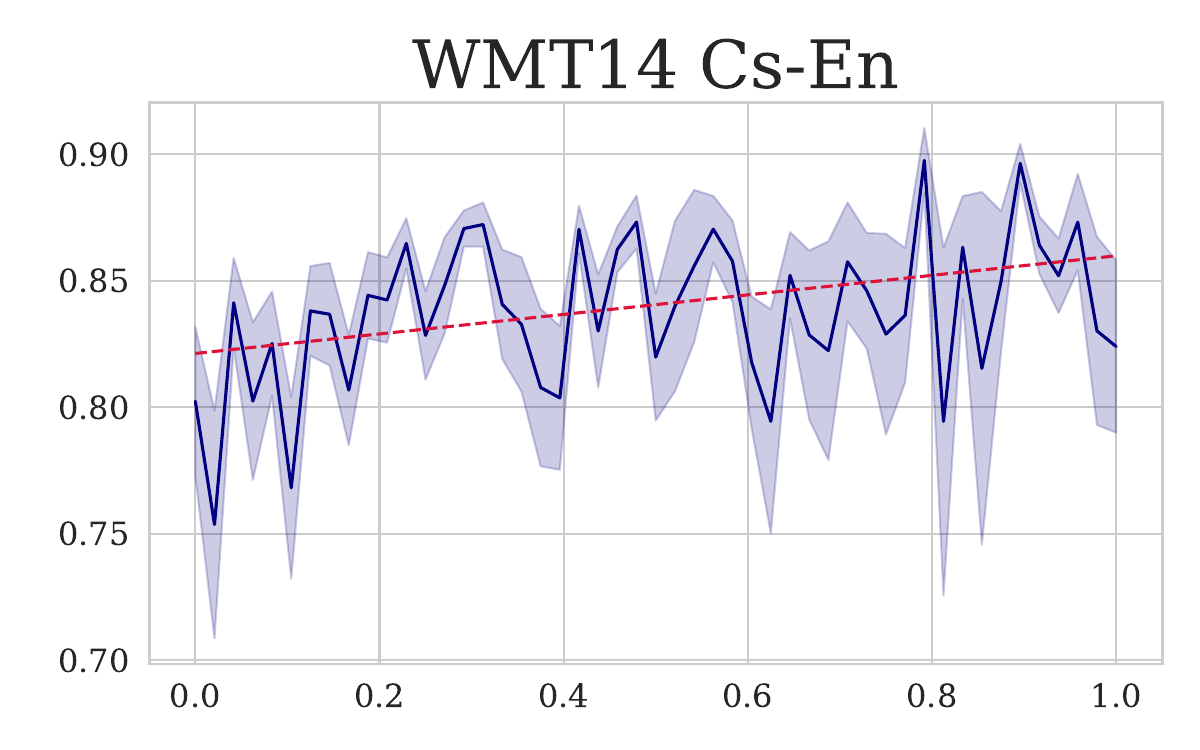}
    \end{subfigure}
    \begin{subfigure}{0.24\textwidth}
        \includegraphics[width=\linewidth]{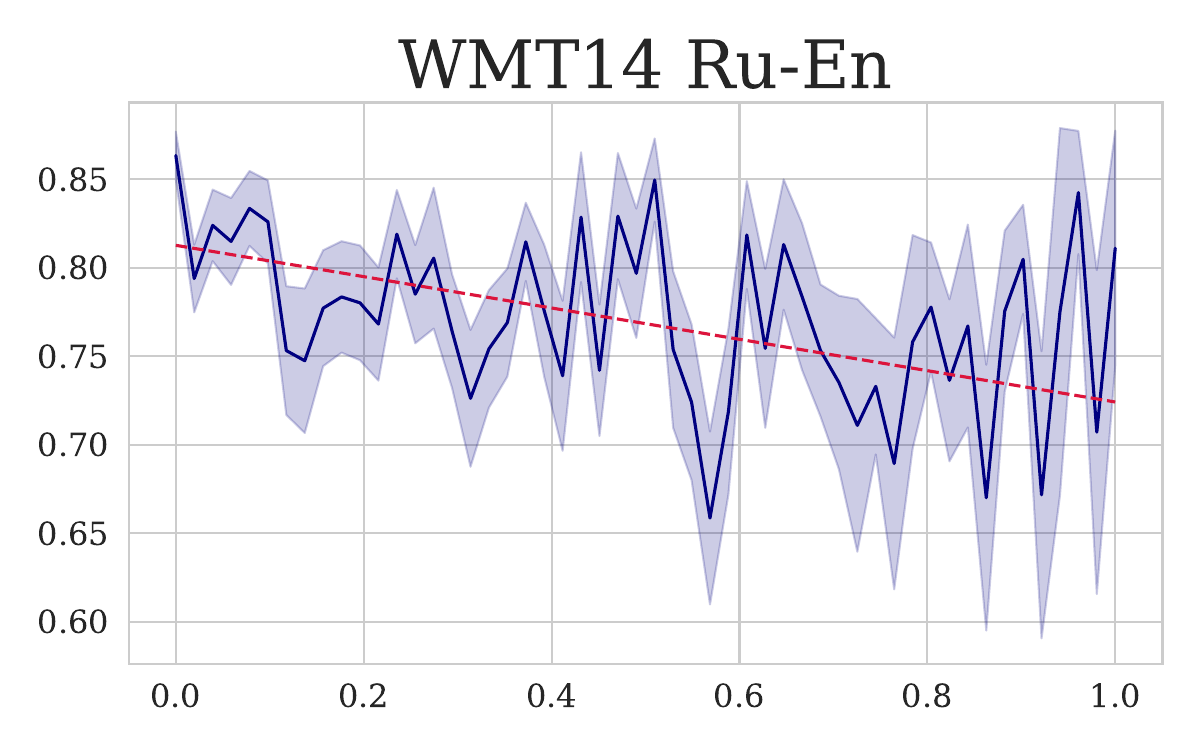}
    \end{subfigure}

    \begin{subfigure}{0.24\textwidth}
        \includegraphics[width=\linewidth]{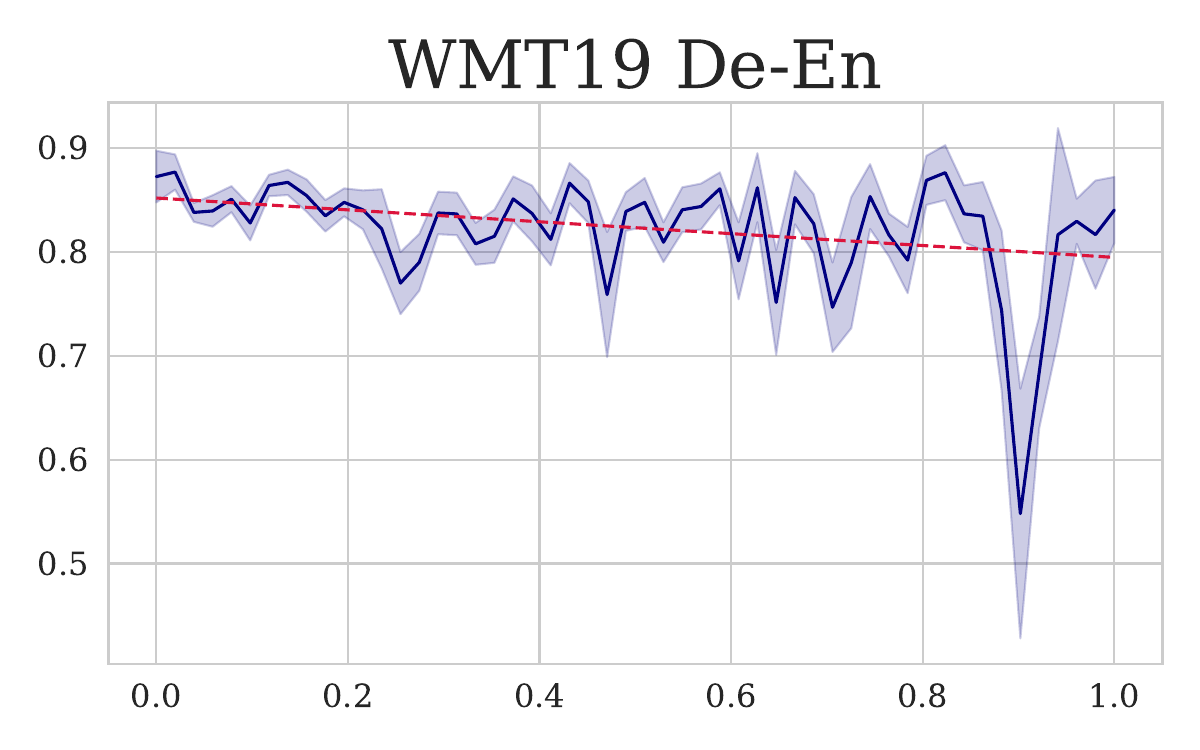}
    \end{subfigure}
    \begin{subfigure}{0.24\textwidth}
        \includegraphics[width=\linewidth]{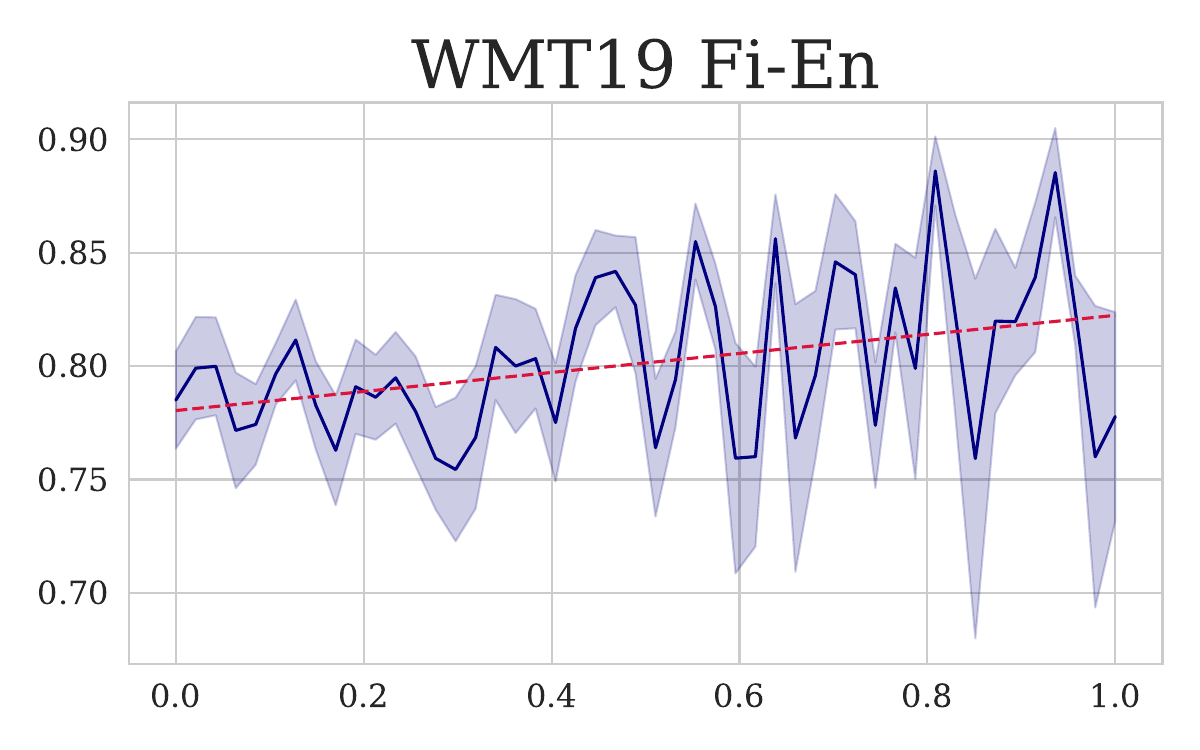}
    \end{subfigure}
    \begin{subfigure}{0.24\textwidth}
        \includegraphics[width=\linewidth]{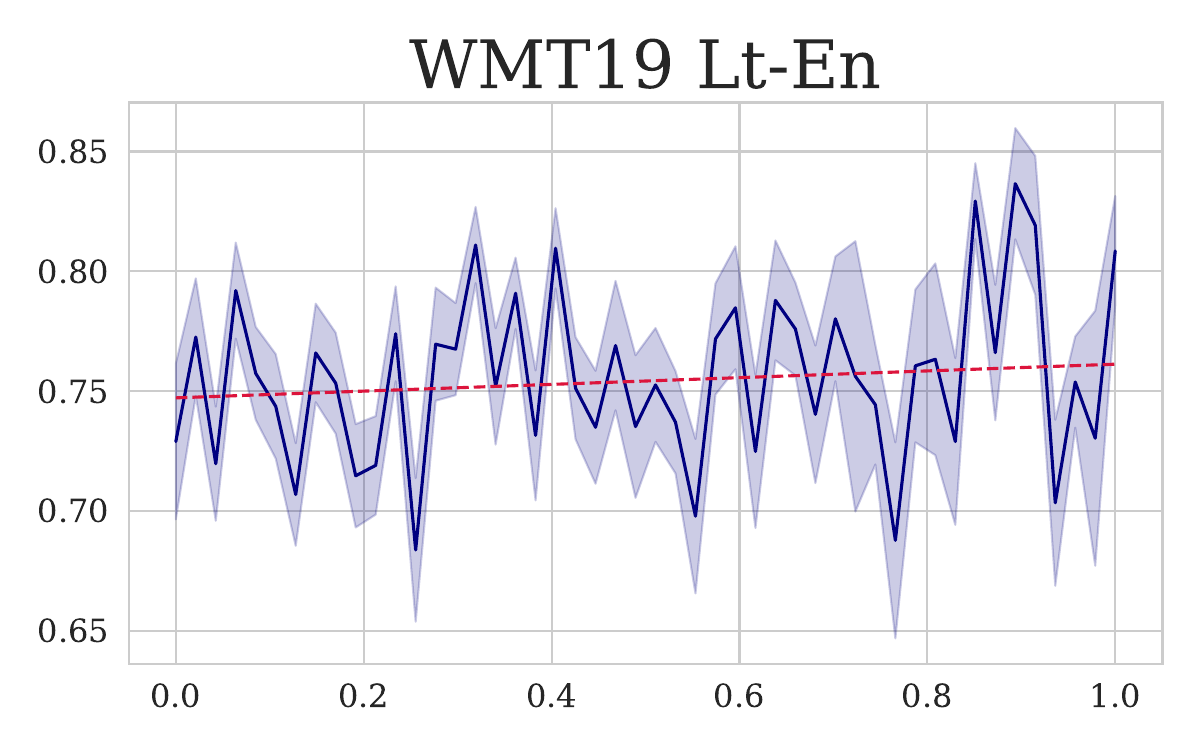}
    \end{subfigure}
    \begin{subfigure}{0.24\textwidth}
        \includegraphics[width=\linewidth]{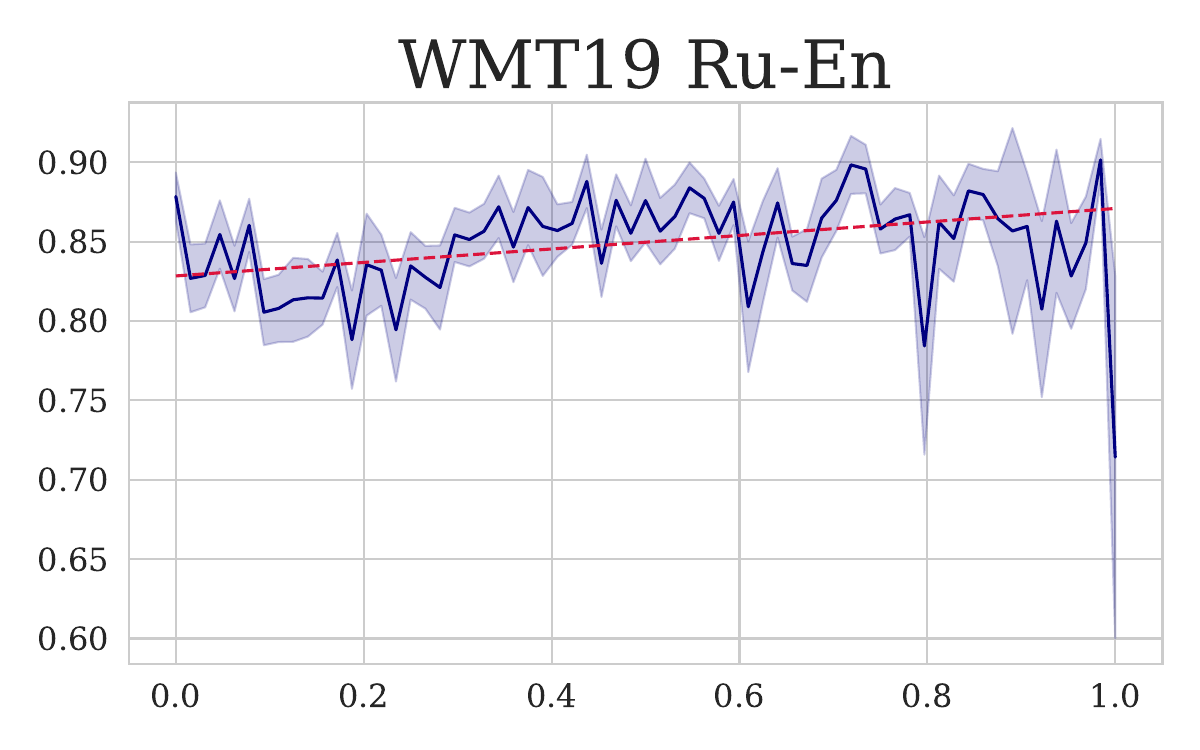}
    \end{subfigure}

\vspace{-0.5em}
{\centering \textbf{\small \gemma} \par}
\vspace{0.3em}

 \begin{subfigure}{0.24\textwidth}
        \includegraphics[width=\linewidth]{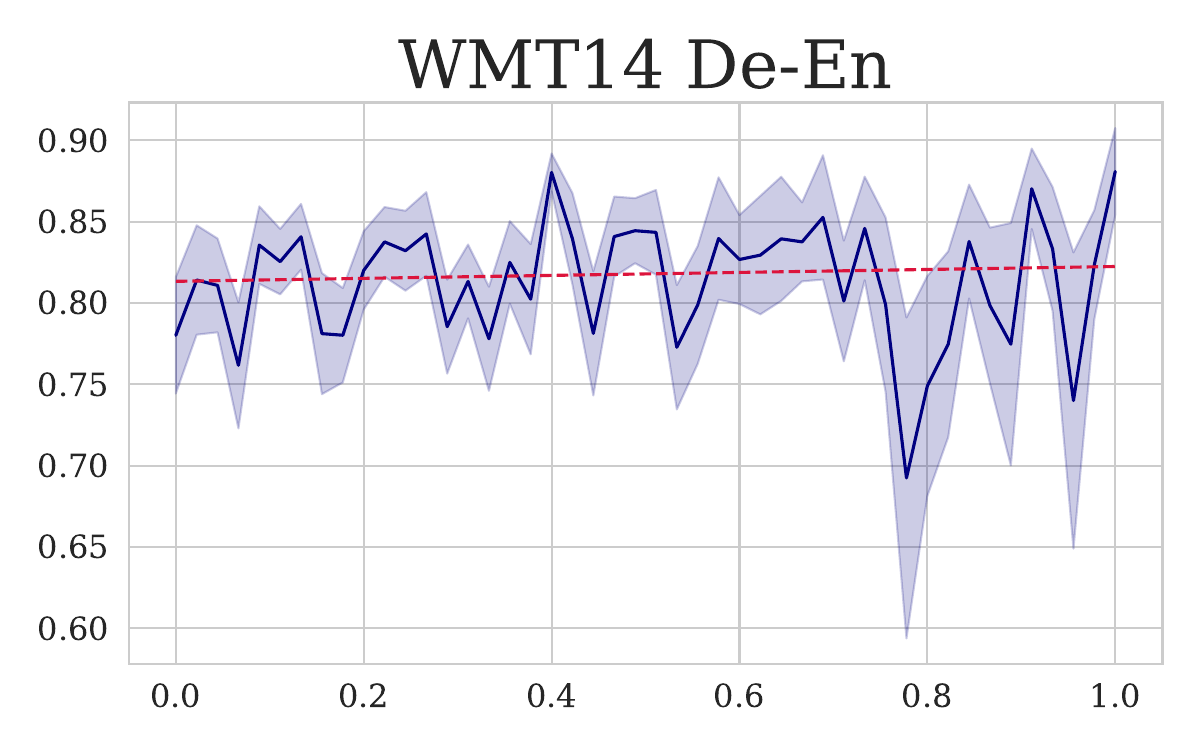}
    \end{subfigure}
    \begin{subfigure}{0.24\textwidth}
        \includegraphics[width=\linewidth]{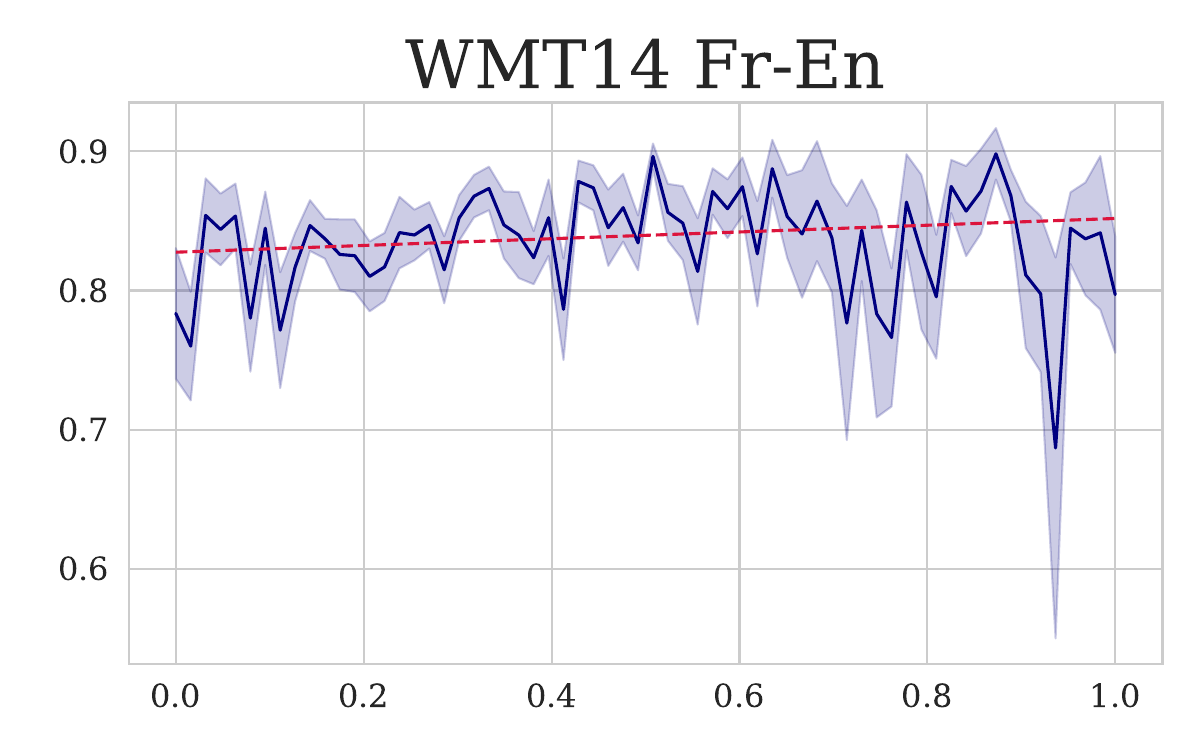}
    \end{subfigure}
    \begin{subfigure}{0.24\textwidth}
        \includegraphics[width=\linewidth]{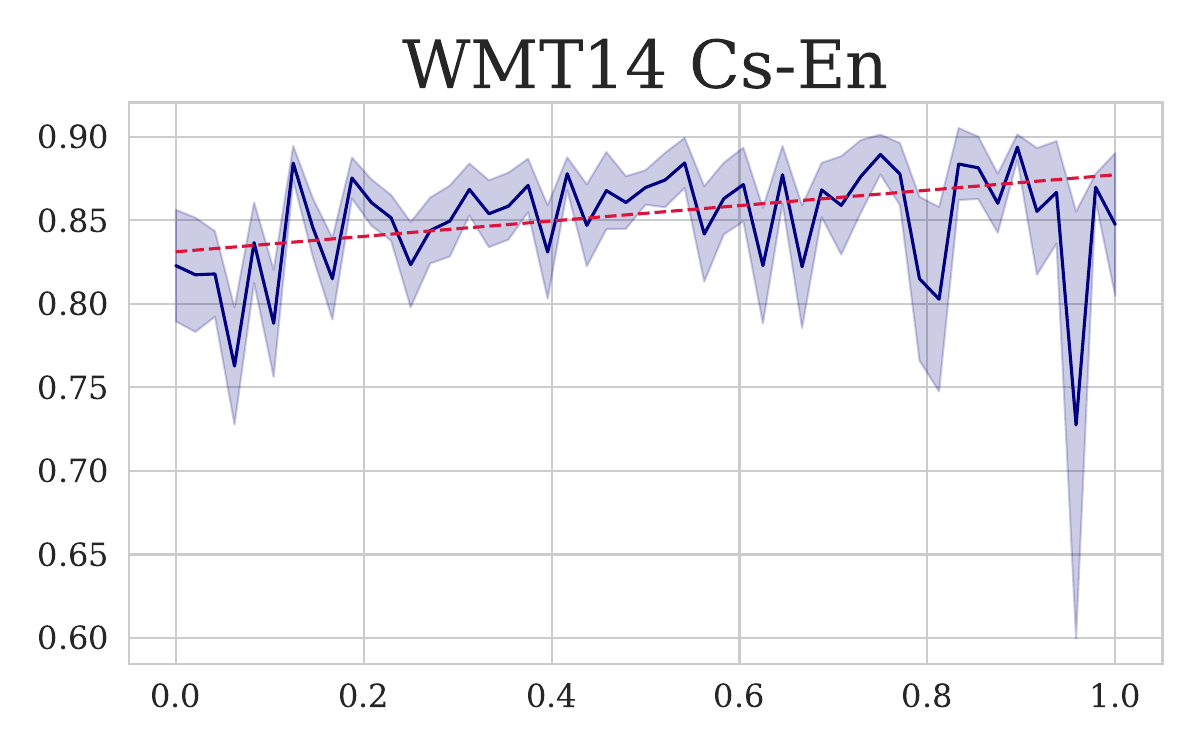}
    \end{subfigure}
    \begin{subfigure}{0.24\textwidth}
        \includegraphics[width=\linewidth]{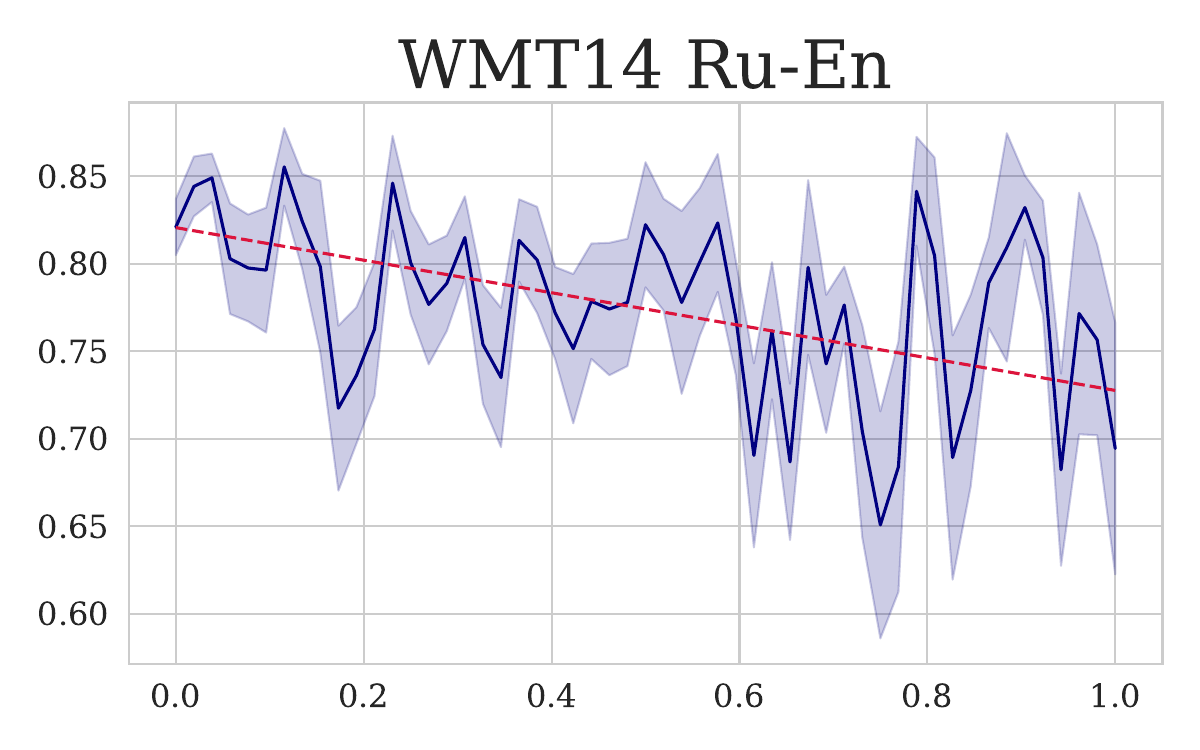}
    \end{subfigure}

    \begin{subfigure}{0.24\textwidth}
        \includegraphics[width=\linewidth]{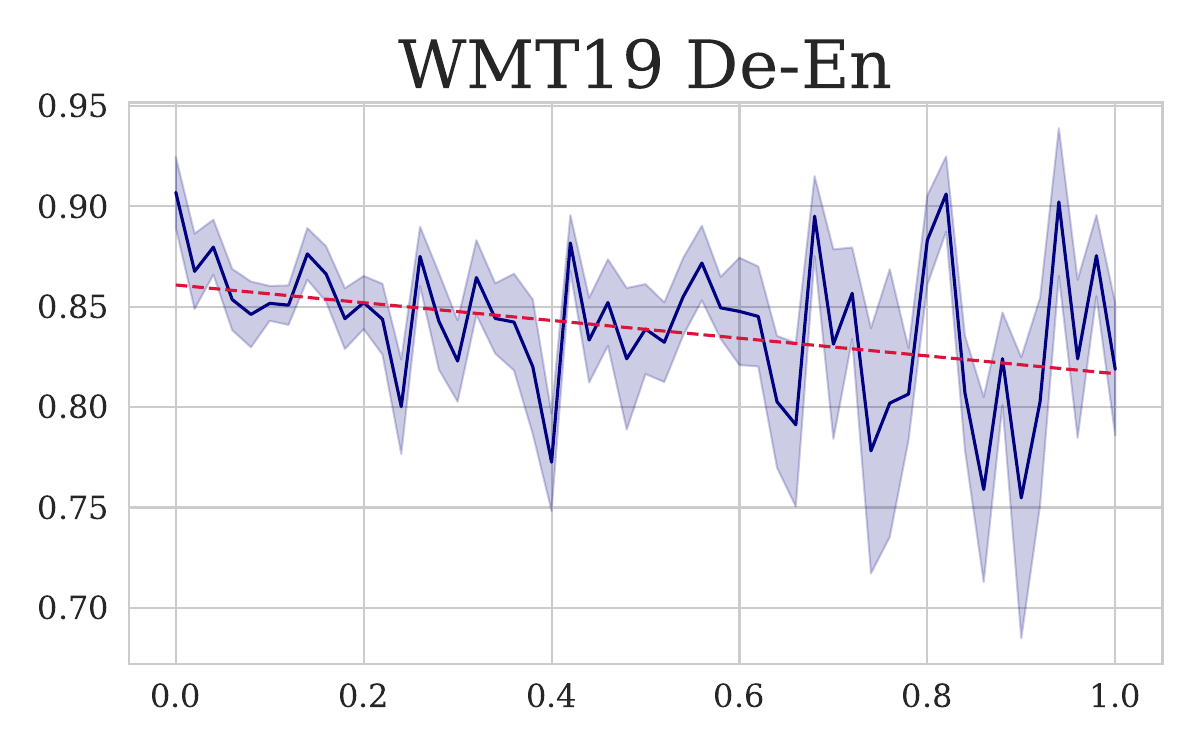}
    \end{subfigure}
    \begin{subfigure}{0.24\textwidth}
        \includegraphics[width=\linewidth]{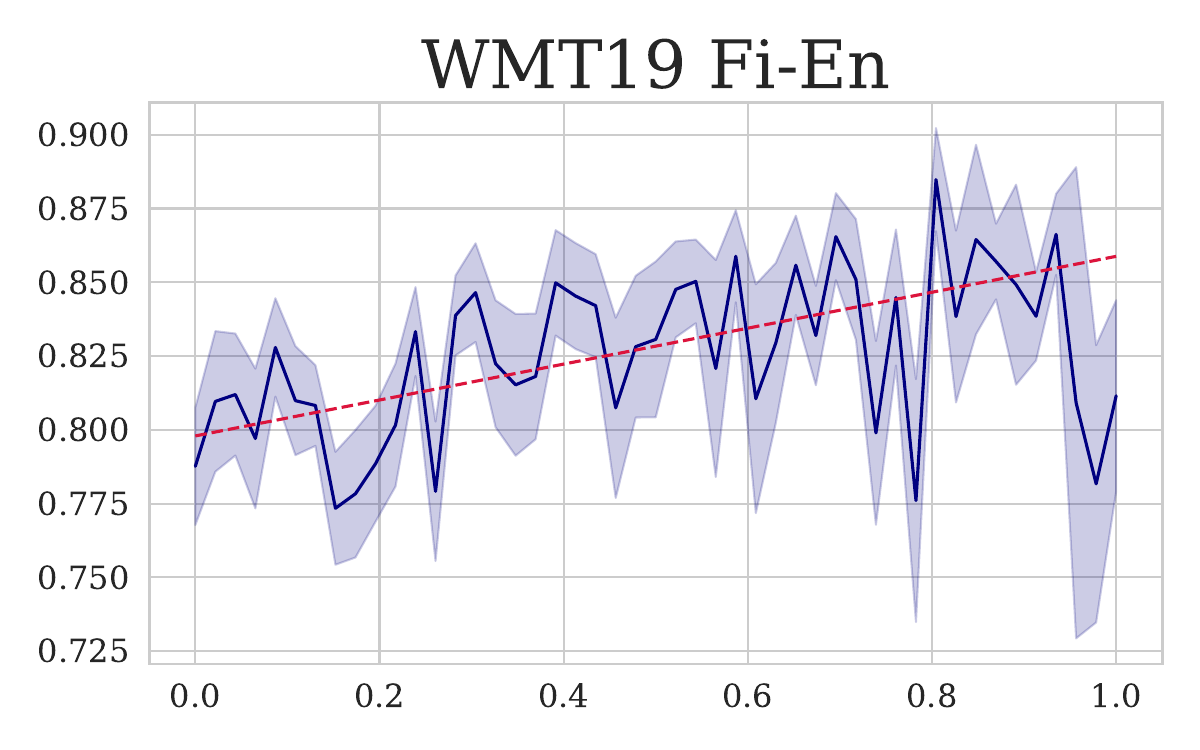}
    \end{subfigure}
    \begin{subfigure}{0.24\textwidth}
        \includegraphics[width=\linewidth]{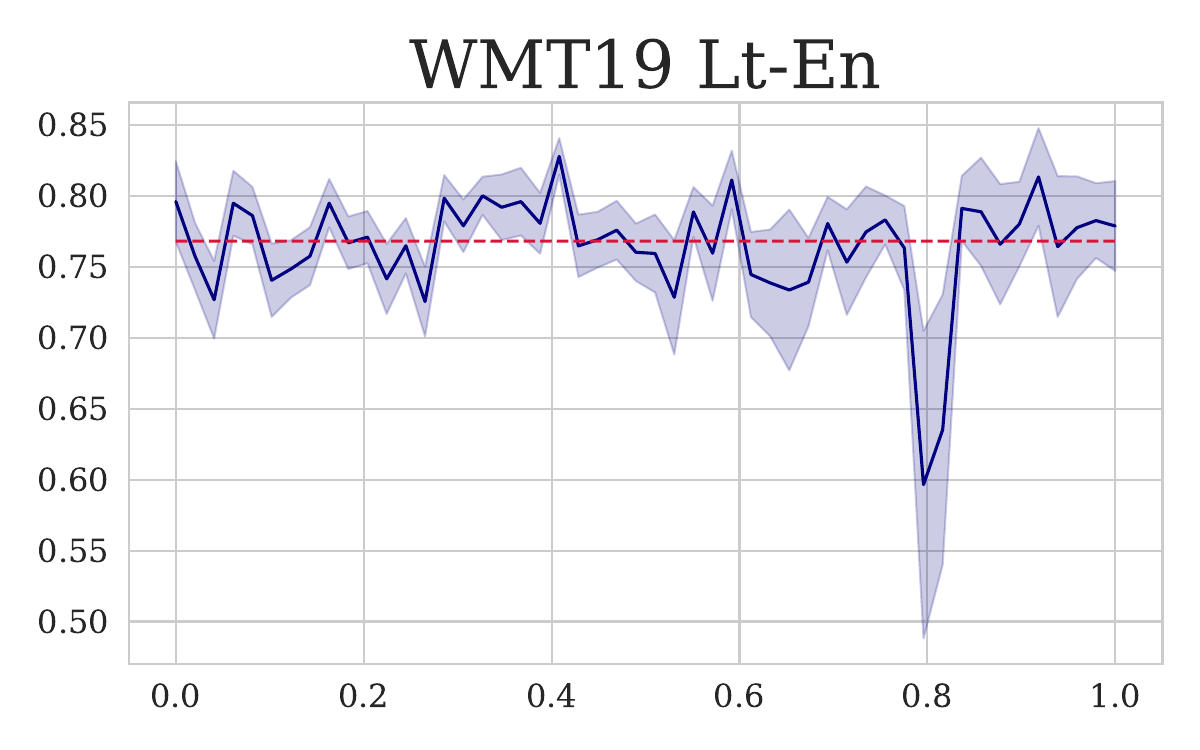}
    \end{subfigure}
    \begin{subfigure}{0.24\textwidth}
        \includegraphics[width=\linewidth]{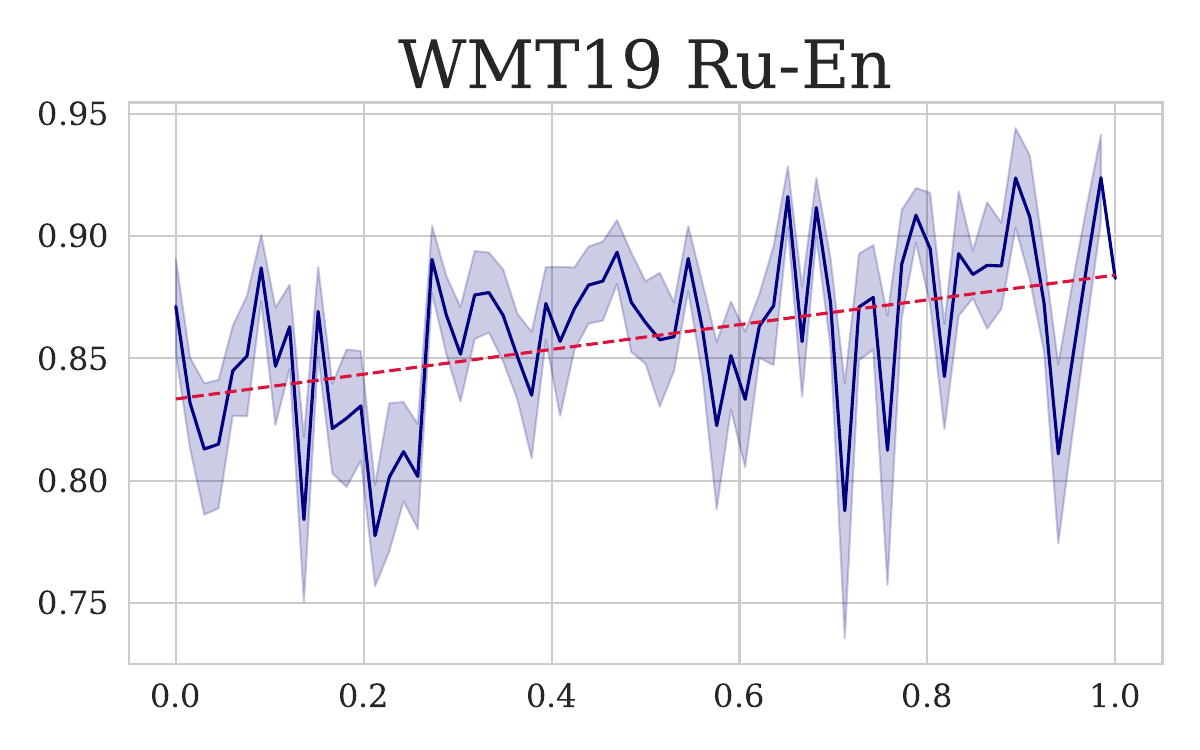}
    \end{subfigure}

\vspace{-0.5em}
{\centering \textbf{\small \eurollm} \par}
\vspace{0.3em}
 \begin{subfigure}{0.24\textwidth}
        \includegraphics[width=\linewidth]{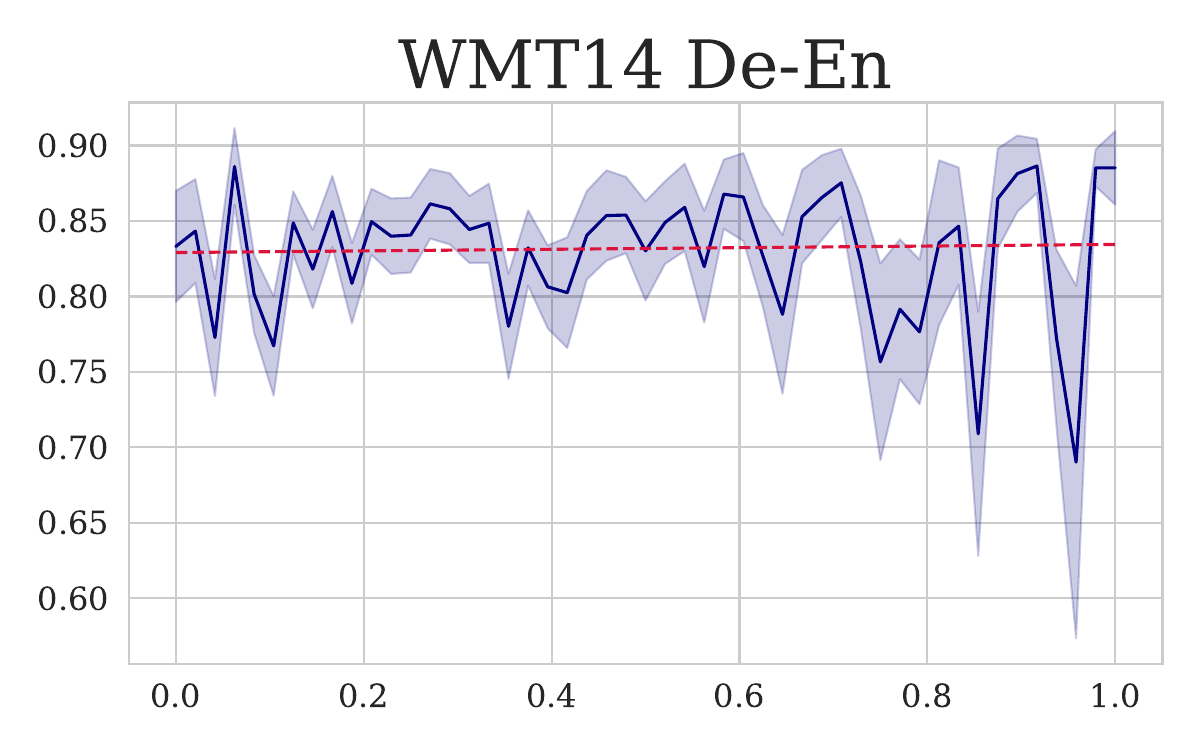}
    \end{subfigure}
    \begin{subfigure}{0.24\textwidth}
        \includegraphics[width=\linewidth]{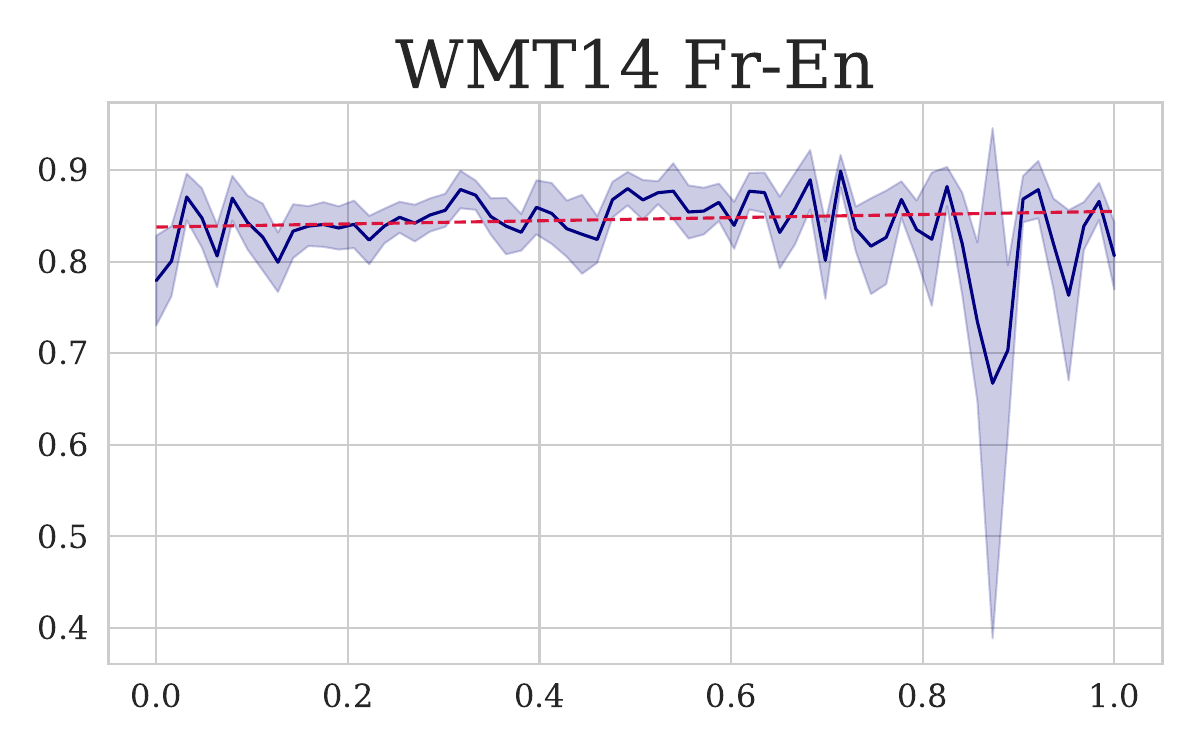}
    \end{subfigure}
    \begin{subfigure}{0.24\textwidth}
        \includegraphics[width=\linewidth]{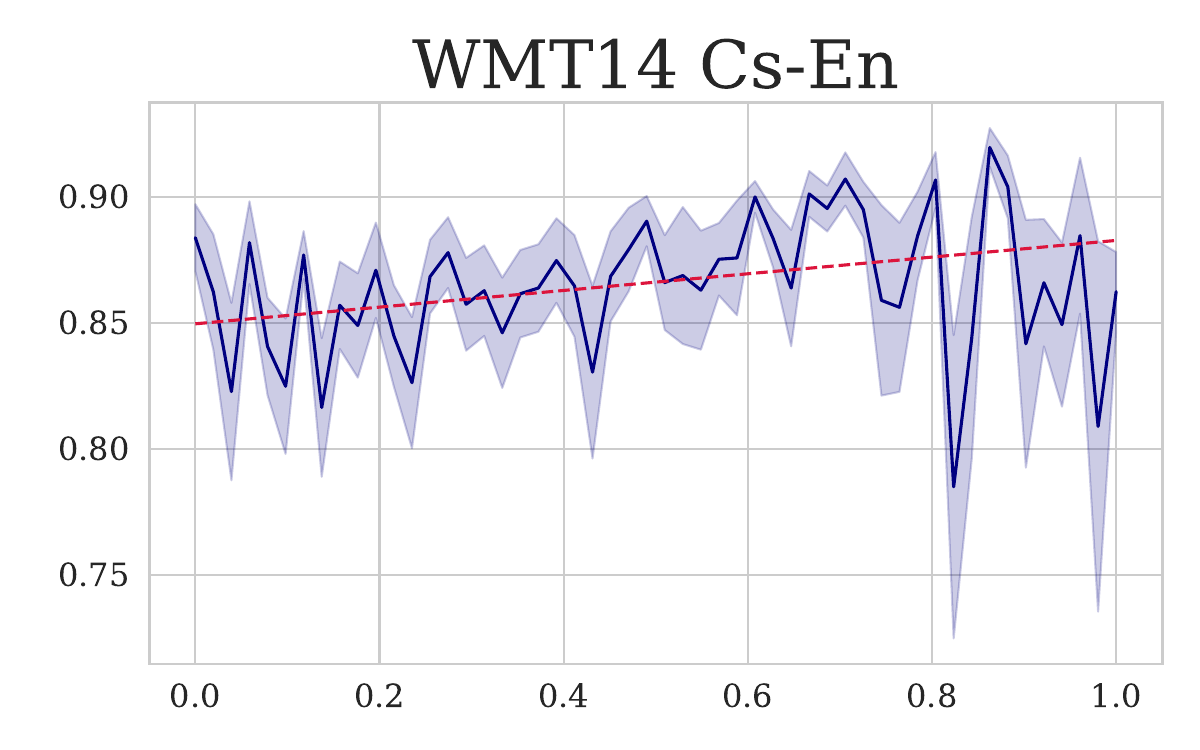}
    \end{subfigure}
    \begin{subfigure}{0.24\textwidth}
        \includegraphics[width=\linewidth]{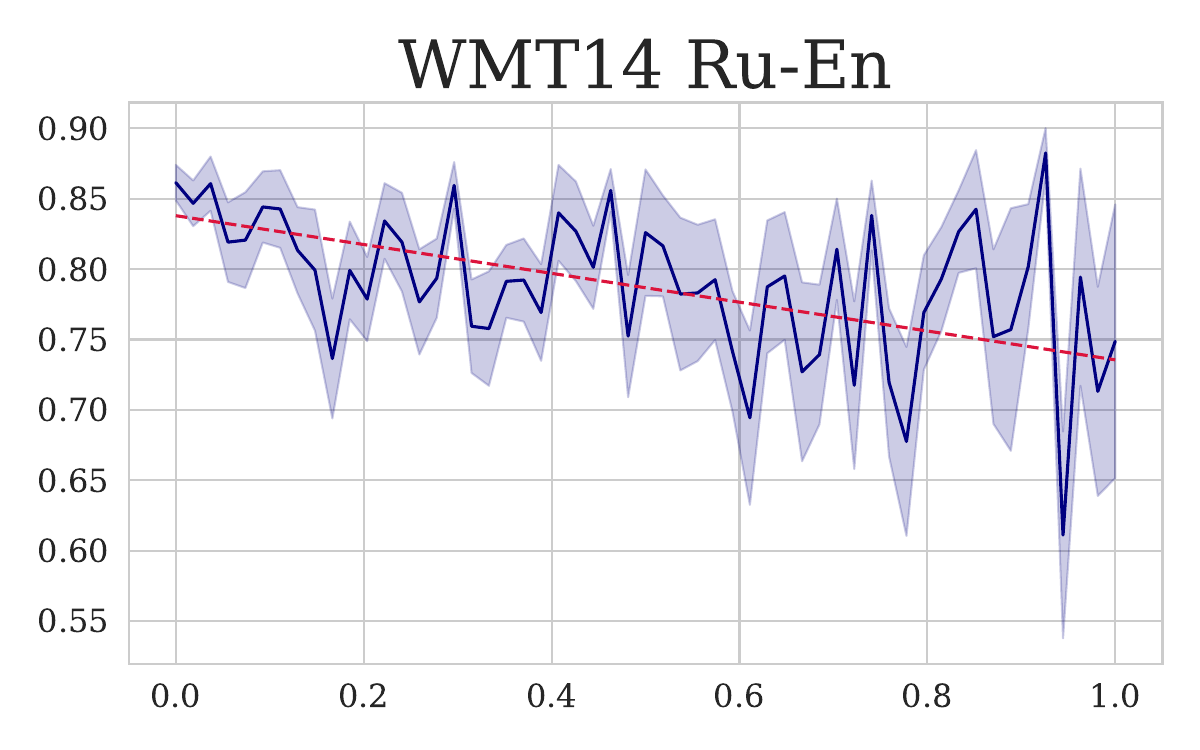}
    \end{subfigure}

    \begin{subfigure}{0.24\textwidth}
        \includegraphics[width=\linewidth]{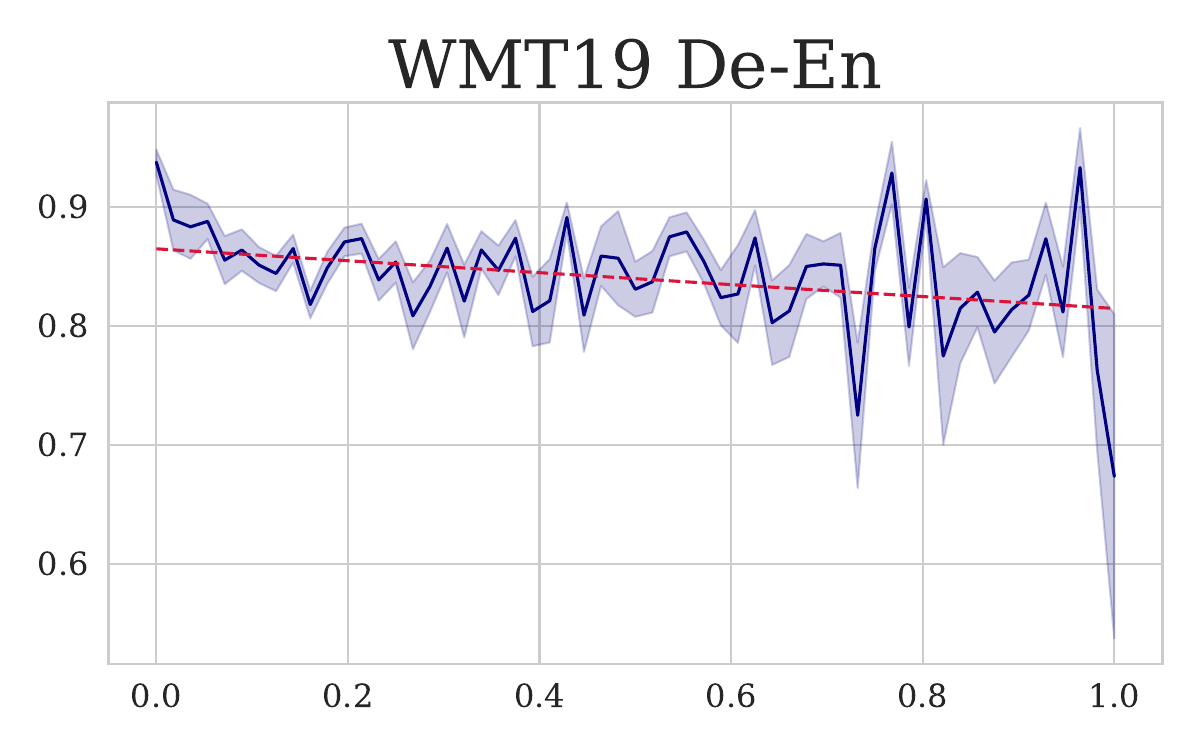}
    \end{subfigure}
    \begin{subfigure}{0.24\textwidth}
        \includegraphics[width=\linewidth]{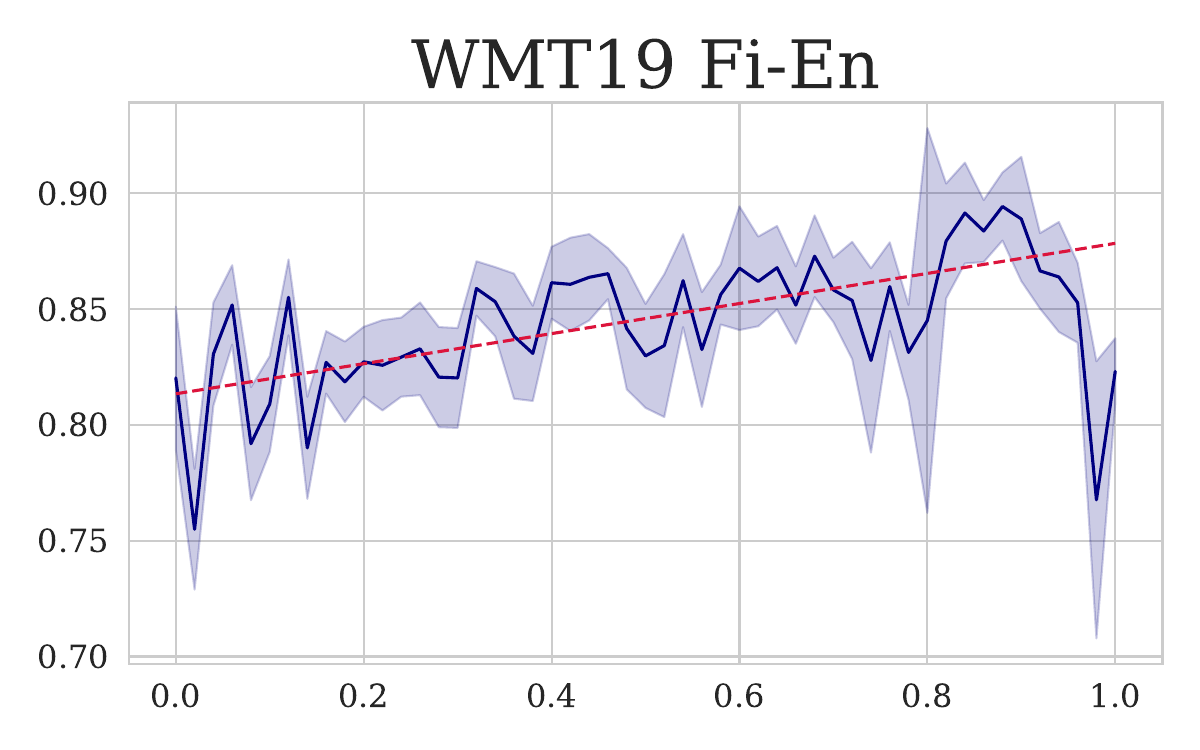}
    \end{subfigure}
    \begin{subfigure}{0.24\textwidth}
        \includegraphics[width=\linewidth]{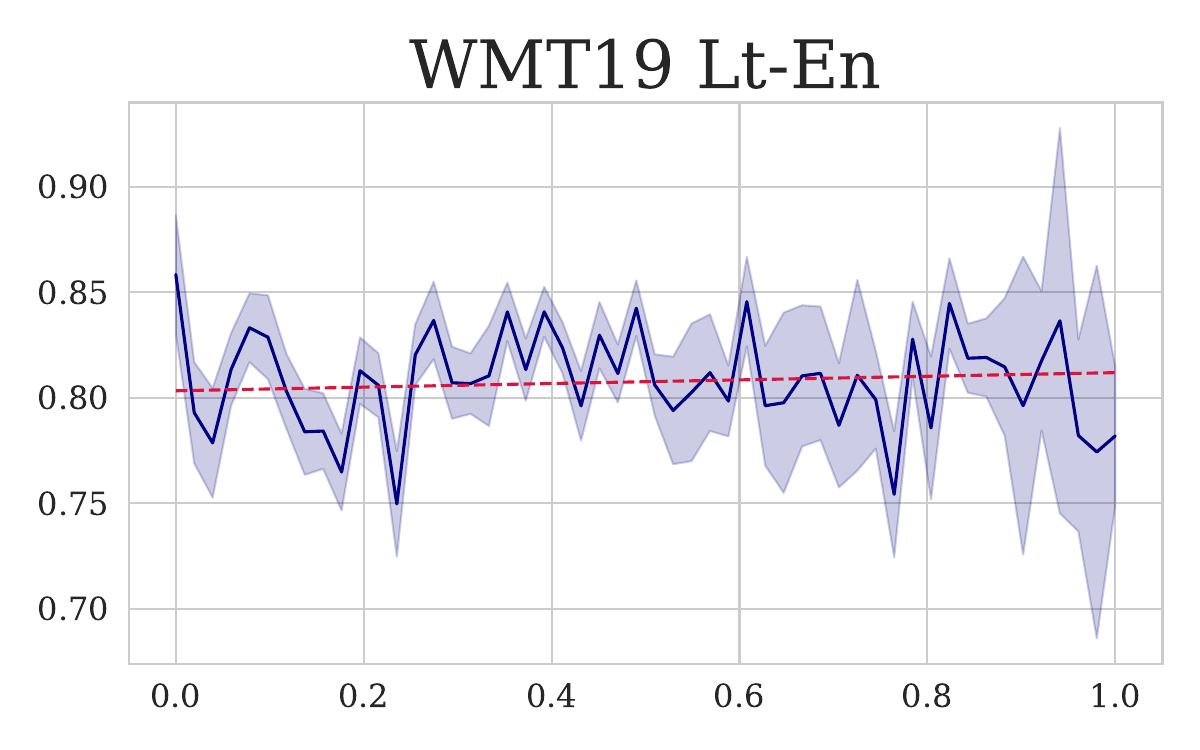}
    \end{subfigure}
    \begin{subfigure}{0.24\textwidth}
        \includegraphics[width=\linewidth]{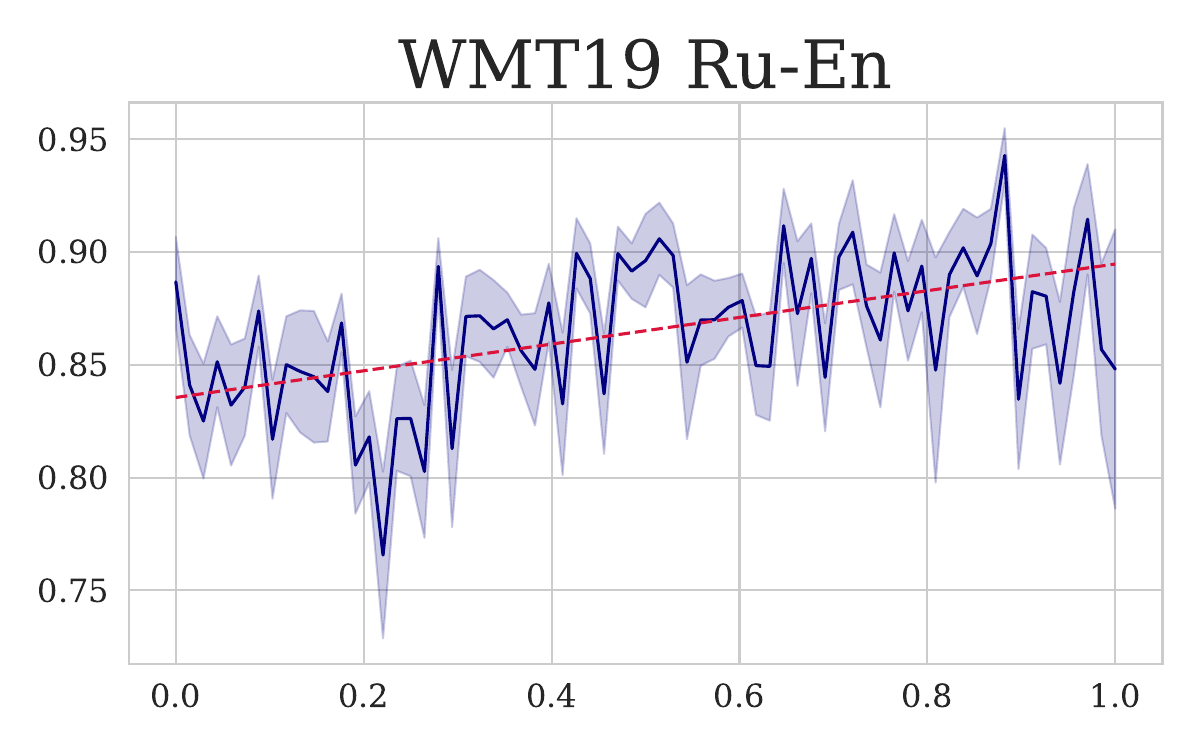}
    \end{subfigure}

    \caption{MetricX-XXL score trends with respect to normalized generated sequence length across four machine translation datasets. Each subplot shows a linear regression fit over binned MetricX-XXL scores.}
    \label{fig:metricxl_trends}
\end{figure*}

\begin{figure*}[h!]
    \centering

\vspace{-0.5em}
{\centering \textbf{\small \llama} \par}
\vspace{0.3em}

    \begin{subfigure}{0.49\textwidth}
        \includegraphics[width=\linewidth]{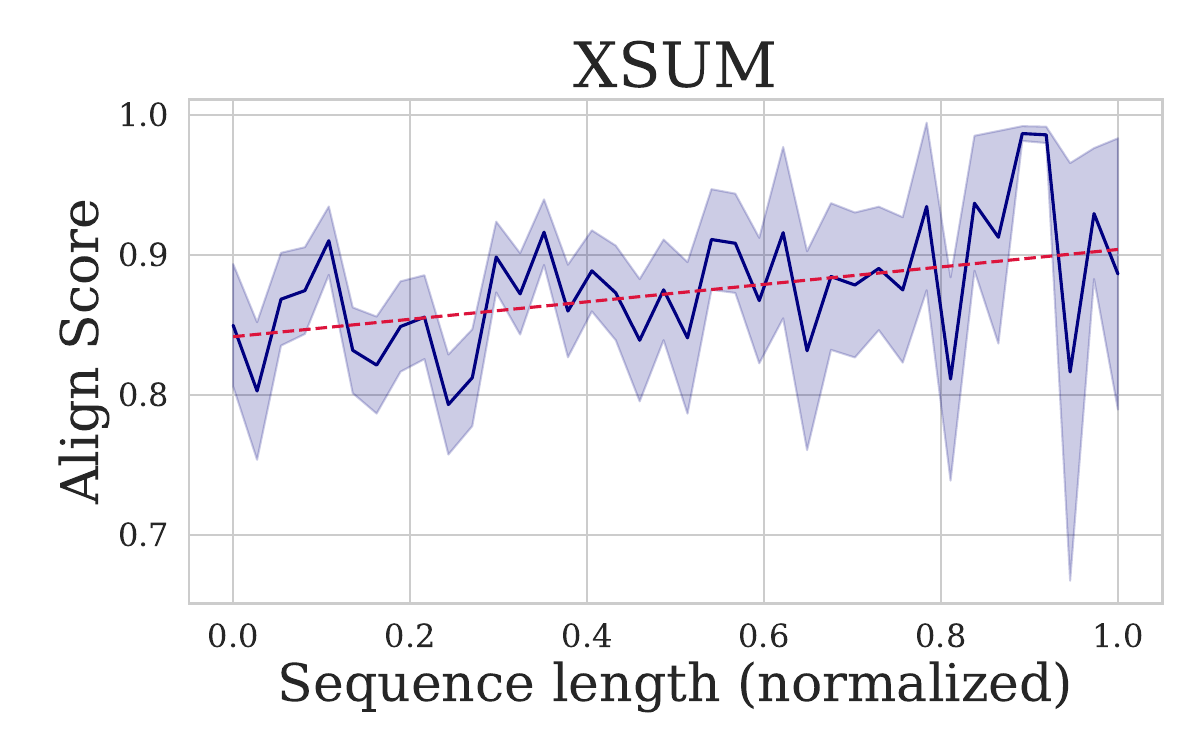}
    \end{subfigure}
    \begin{subfigure}{0.49\textwidth}
        \includegraphics[width=\linewidth]{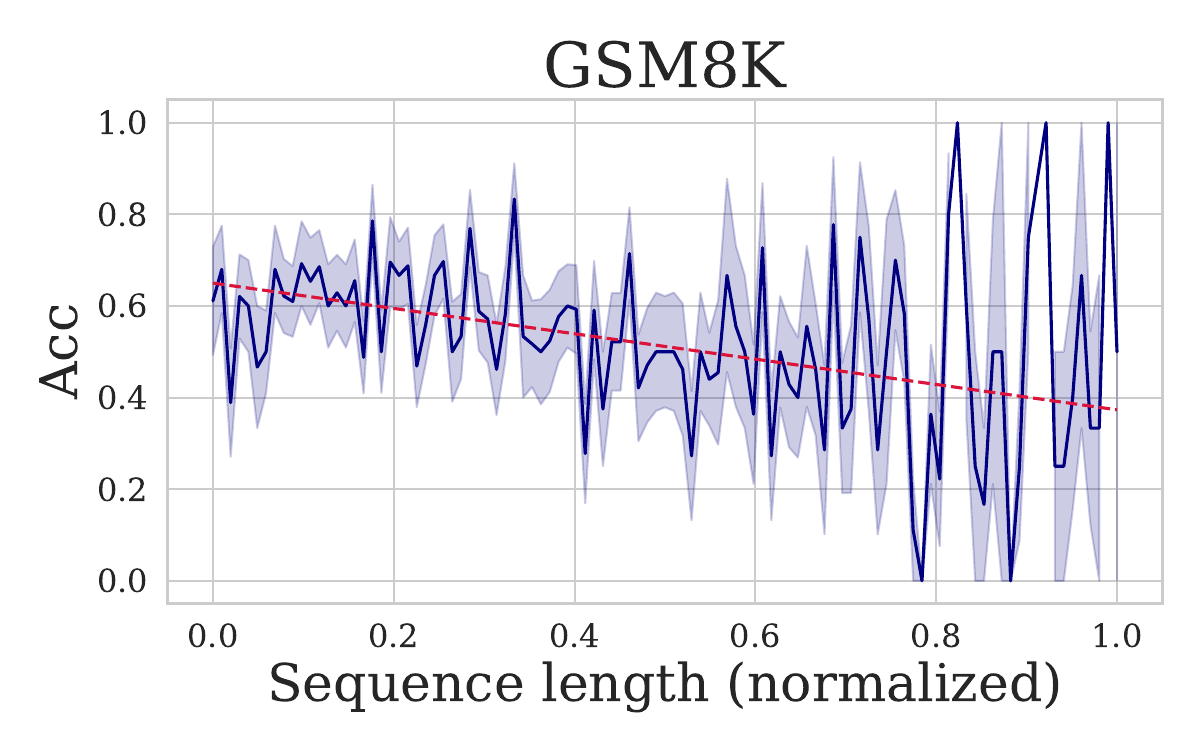}
    \end{subfigure}
    
\vspace{-0.5em}
{\centering \textbf{\small \gemma} \par}
\vspace{0.3em}

 \begin{subfigure}{0.49\textwidth}
        \includegraphics[width=\linewidth]{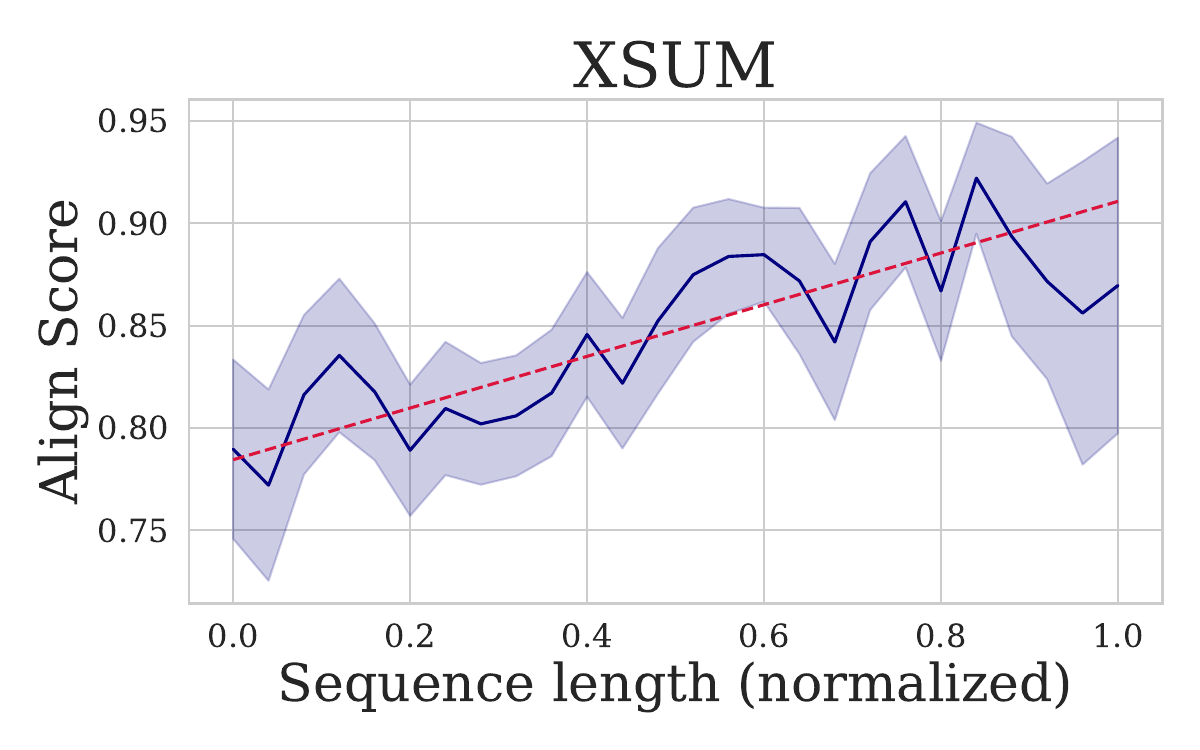}
    \end{subfigure}
    \begin{subfigure}{0.49\textwidth}
        \includegraphics[width=\linewidth]{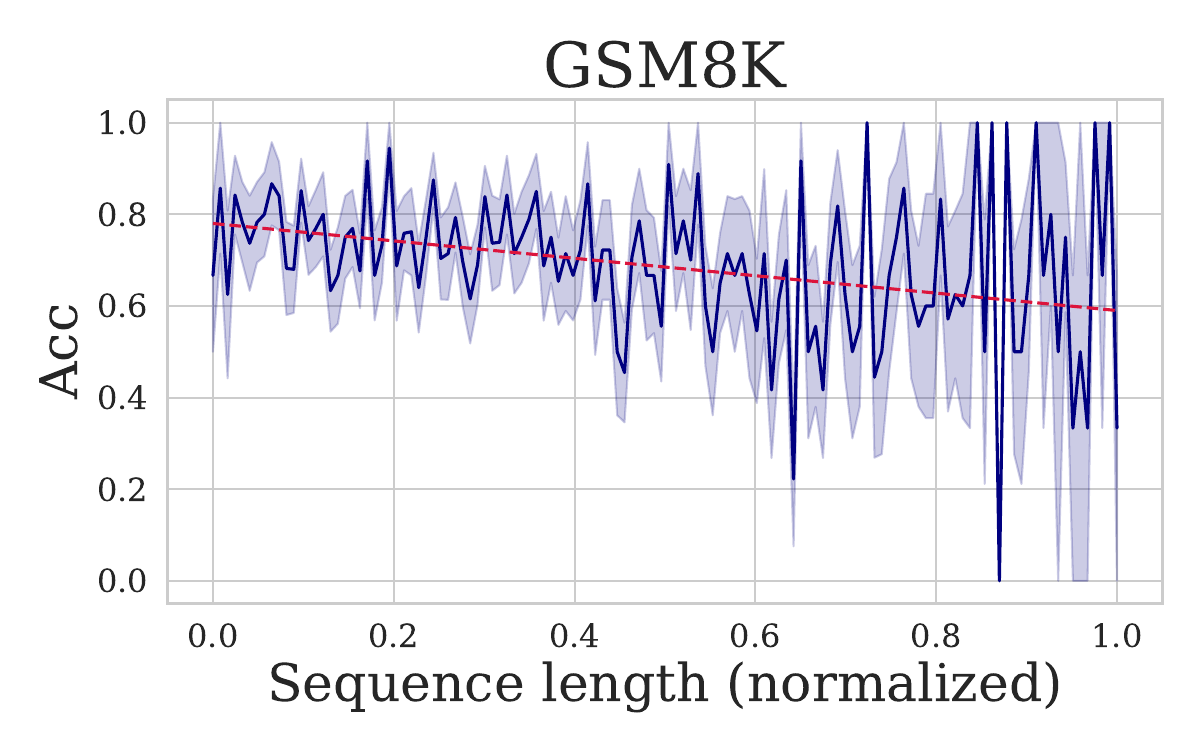}
    \end{subfigure}

    \caption{Align Score and Accuracy trends with respect to normalized generated sequence length across four machine translation datasets. Each subplot shows a linear regression fit over binned Align Score and Accuracy scores.}
    \label{fig:non_nmt_metric_trends}
\end{figure*}

  In this section we report the detailed relation between performance metrics for various tasks and length of the generated output. Figures~\ref{fig:comet_trends}, \ref{fig:xcometxxl_trends} and \ref{fig:metricxl_trends} show average normalized values of performance metrics for NMT tasks at each generated sequence length, as well as a linear OLS fit to this dependency. Figure~\ref{fig:non_nmt_metric_trends} contains similar charts for the QA and ATS tasks.

\newpage

\newpage

\newpage

\begin{table}[ht!]
\centering
\footnotesize
\begin{tabular}{l|cc|cc|cc}
\toprule
\textbf{Dataset} & \multicolumn{2}{c}{COMET} & \multicolumn{2}{c}{MetricX} & \multicolumn{2}{c}{XCOMET} \\
 & slope & p-val & slope & p-val & slope & p-val \\
 \midrule
\multicolumn{7}{c}{\textbf{\llama}} \\
\midrule
WMT14 Cs-En & -0.032 & 0.019 & 0.039 & 0.008 & -0.033 & 0.092 \\
WMT14 De-En & -0.027 & 0.176 & 0.009 & 0.651 & -0.073 & 0.004 \\
WMT14 Fr-En & -0.043 & 0.002 & 0.014 & 0.404 & -0.077 & 0.002 \\
WMT14 Ru-En & -0.100 & 0.000 & -0.088 & 0.000 & -0.029 & 0.272 \\
WMT19 De-En & -0.176 & 0.000 & -0.057 & 0.000 & -0.118 & 0.000 \\
WMT19 Fi-En & -0.016 & 0.288 & 0.042 & 0.012 & 0.067 & 0.005 \\
WMT19 Lt-En & -0.018 & 0.223 & 0.014 & 0.382 & 0.093 & 0.000 \\
WMT19 Ru-En & -0.030 & 0.011 & 0.042 & 0.001 & 0.090 & 0.000 \\
\midrule
 \multicolumn{7}{c}{\textbf{\gemma}} \\
 \midrule
WMT14 Cs-En & -0.033 & 0.018 & 0.046 & 0.001 & -0.047 & 0.018 \\
WMT14 De-En & -0.023 & 0.251 & 0.009 & 0.636 & -0.053 & 0.027 \\
WMT14 Fr-En & -0.028 & 0.048 & 0.024 & 0.139 & -0.051 & 0.035 \\
WMT14 Ru-En & -0.103 & 0.000 & -0.093 & 0.000 & -0.021 & 0.404 \\
WMT19 De-En & -0.154 & 0.000 & -0.044 & 0.000 & -0.095 & 0.000 \\
WMT19 Fi-En & -0.006 & 0.647 & 0.061 & 0.000 & 0.121 & 0.000 \\
WMT19 Lt-En & -0.022 & 0.108 & -0.000 & 0.998 & 0.110 & 0.000 \\
WMT19 Ru-En & -0.034 & 0.004 & 0.051 & 0.000 & 0.120 & 0.000 \\
\midrule
\multicolumn{7}{c}{\textbf{\eurollm}} \\
\midrule
WMT14 Cs-En & -0.047 & 0.001 & 0.033 & 0.010 & -0.011 & 0.572 \\
WMT14 De-En & -0.016 & 0.413 & 0.005 & 0.779 & -0.035 & 0.157 \\
WMT14 Fr-En & -0.029 & 0.042 & 0.017 & 0.294 & -0.034 & 0.152 \\
WMT14 Ru-En & -0.101 & 0.000 & -0.102 & 0.000 & -0.031 & 0.250 \\
WMT19 De-En & -0.156 & 0.000 & -0.050 & 0.000 & -0.096 & 0.000 \\
WMT19 Fi-En & 0.007 & 0.648 & 0.065 & 0.000 & 0.170 & 0.000 \\
WMT19 Lt-En & -0.027 & 0.053 & 0.009 & 0.523 & 0.156 & 0.000 \\
WMT19 Ru-En & -0.034 & 0.003 & 0.059 & 0.000 & 0.135 & 0.000 \\
\bottomrule
\end{tabular}
\caption{Regression slopes and p-values measuring the correlation between output length and three machine translation quality metrics (Comet, MetricX-XXL, Xcomet-XXL) across different translation datasets and models. }
\end{table}

\newpage

\begin{table}[ht!]
\centering
\footnotesize
\begin{tabular}{l|cc}
\toprule
\textbf{Dataset} & \multicolumn{2}{c}{AlignScore} \\
 & slope & p-val \\
 \midrule
 \multicolumn{3}{c}{\textbf{\llama}} \\
 \midrule
XSum & 0.062 & 0.042 \\
\midrule
\multicolumn{3}{c}{\textbf{\gemma}} \\
\midrule
XSum & 0.126 & 0.000 \\
\bottomrule
\end{tabular}
\caption{Regression slopes and p-values measuring the correlation between output length and summarization quality metric (Align Score).}
\end{table}

\begin{table}[ht!]
\centering
\footnotesize
\begin{tabular}{l|cc}
\toprule
\textbf{Dataset} & \multicolumn{2}{c}{Accuracy} \\
 & slope & p-val \\
 \midrule
 \multicolumn{3}{c}{\textbf{\llama}} \\
 \midrule
 GSM8k & -0.277 & 0.000 \\
\midrule
\multicolumn{3}{c}{\textbf{\gemma}} \\
\midrule
GSM8k & -0.190 & 0.000 \\

\bottomrule
\end{tabular}
\caption{Regression slopes and p-values measuring the correlation between output length and QA quality metric (Accuracy).}
\end{table}

\newpage

\subsection{UQ Values vs Generation Length}
\label{sec:length_effects_ue}
\begin{figure*}[h!]
    \centering
    \vspace{-0.5em}
    {\centering \textbf{\small WMT14 De-En} \par}
    \vspace{0.2em}
    \begin{subfigure}{0.15\textwidth}
        \includegraphics[width=\linewidth]{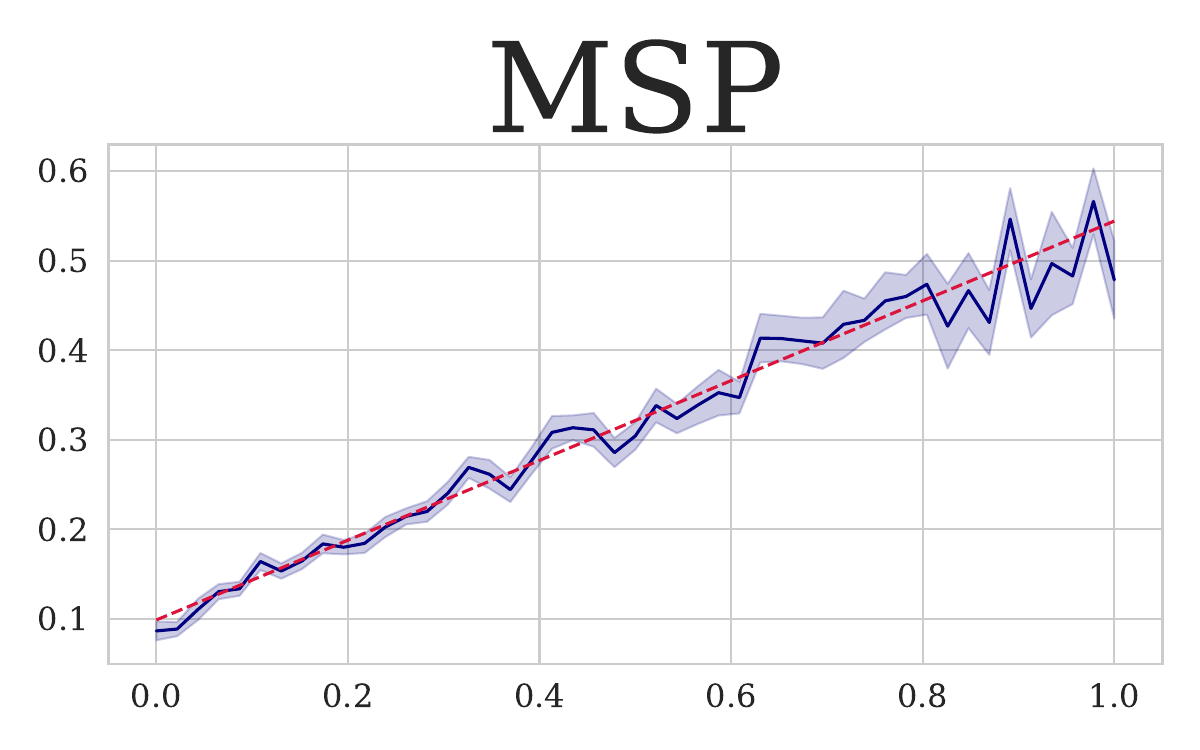}
    \end{subfigure}
    \begin{subfigure}{0.15\textwidth}
        \includegraphics[width=\linewidth]{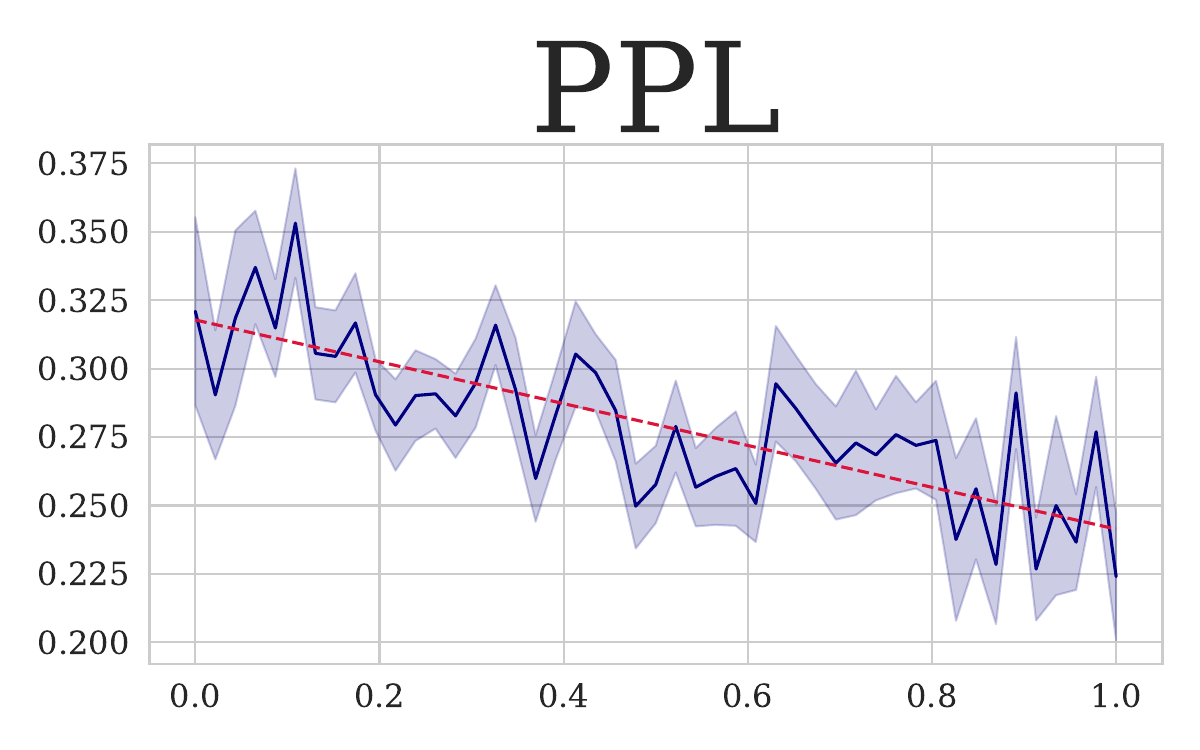}
    \end{subfigure}
    \begin{subfigure}{0.15\textwidth}
        \includegraphics[width=\linewidth]{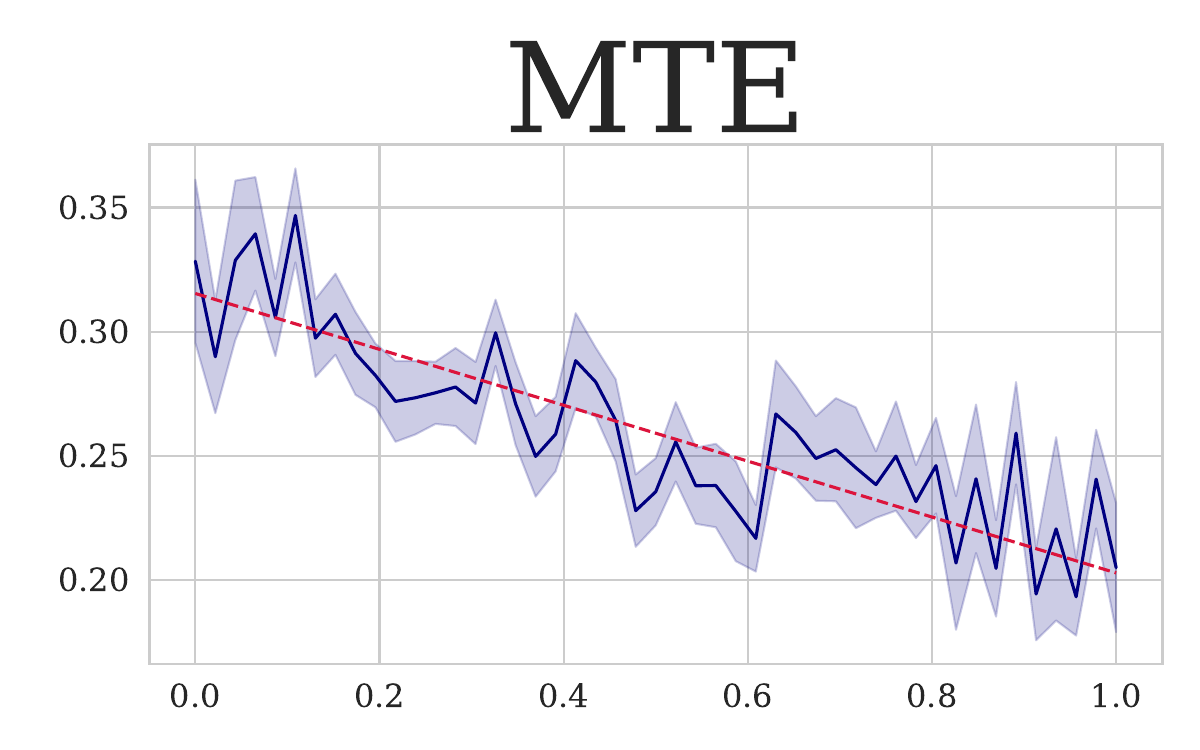}
    \end{subfigure}
    \begin{subfigure}{0.15\textwidth}
        \includegraphics[width=\linewidth]{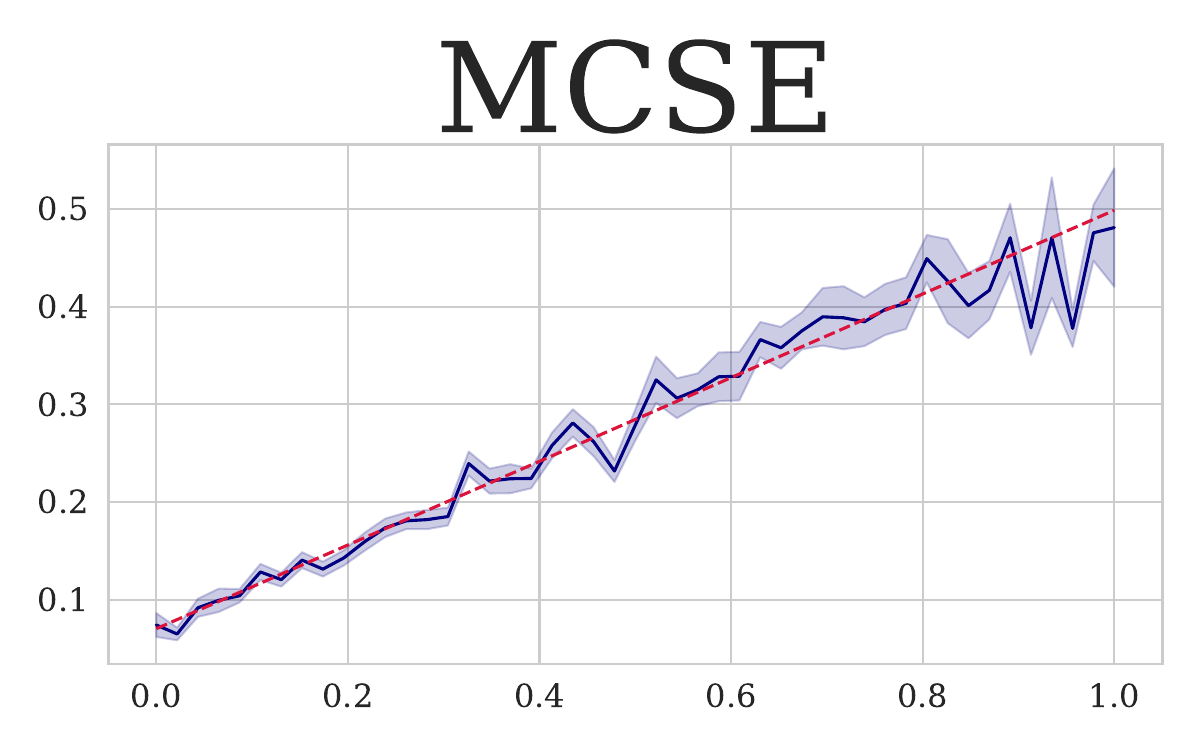}
    \end{subfigure}
    \begin{subfigure}{0.15\textwidth}
        \includegraphics[width=\linewidth]{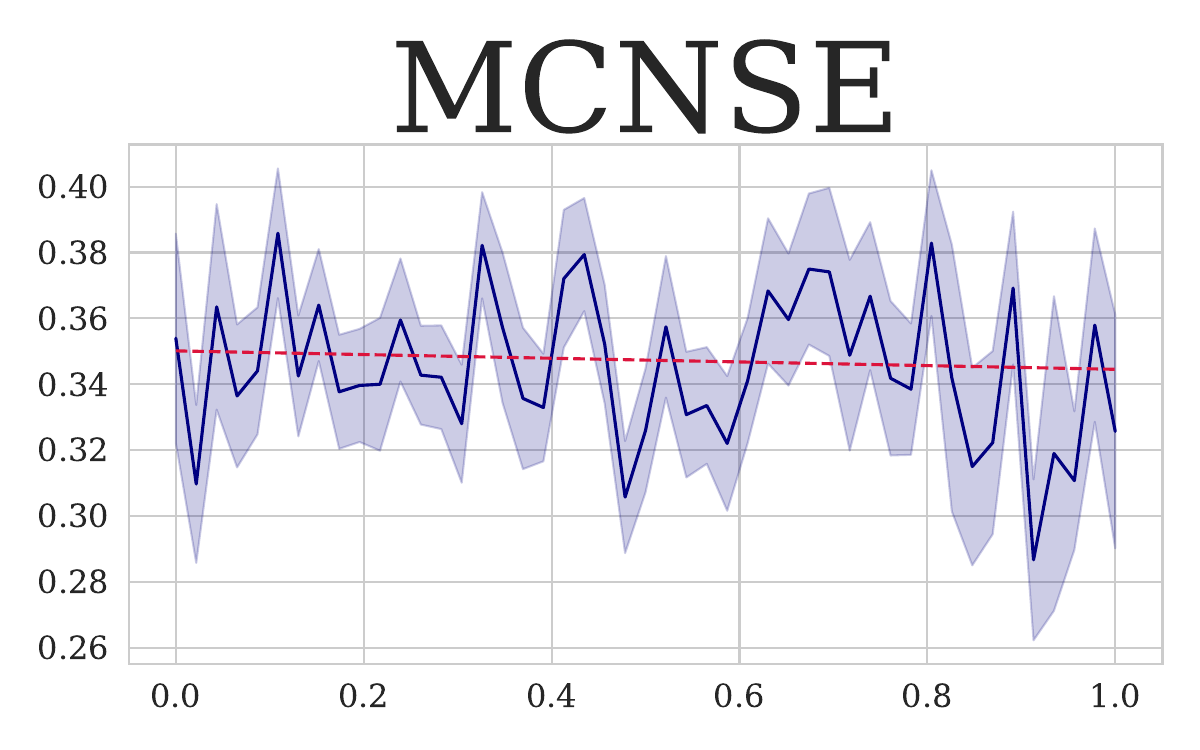}
    \end{subfigure}
    \begin{subfigure}{0.15\textwidth}
        \includegraphics[width=\linewidth]{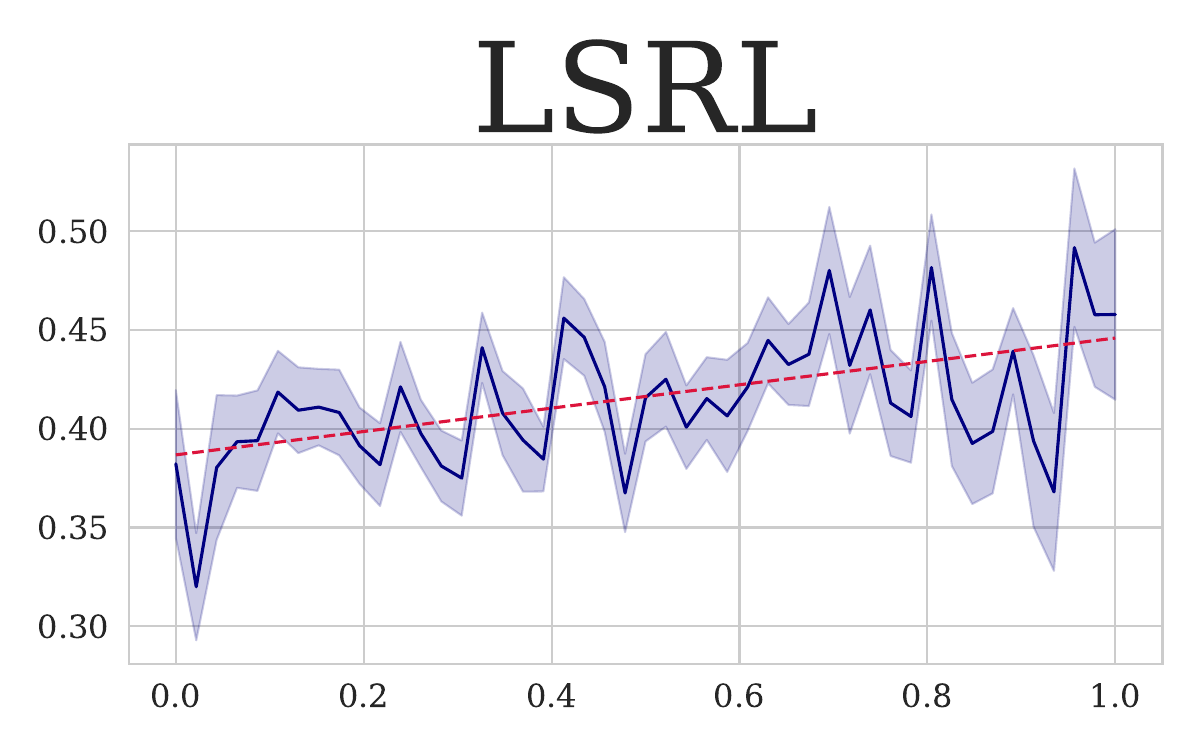}
    \end{subfigure}
    \vspace{0.3em}
    {\centering \textbf{\small WMT14 Fr-En} \par}
    \vspace{0.2em}
    \begin{subfigure}{0.15\textwidth}
        \includegraphics[width=\linewidth]{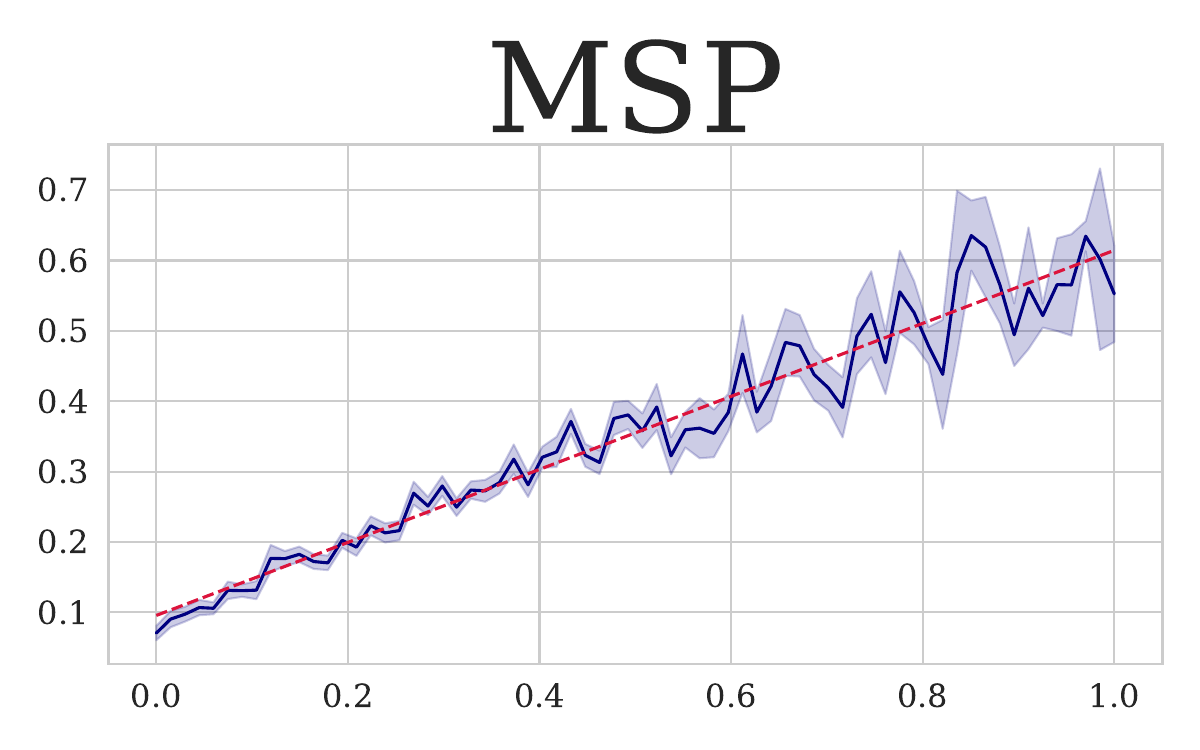}
    \end{subfigure}
    \begin{subfigure}{0.15\textwidth}
        \includegraphics[width=\linewidth]{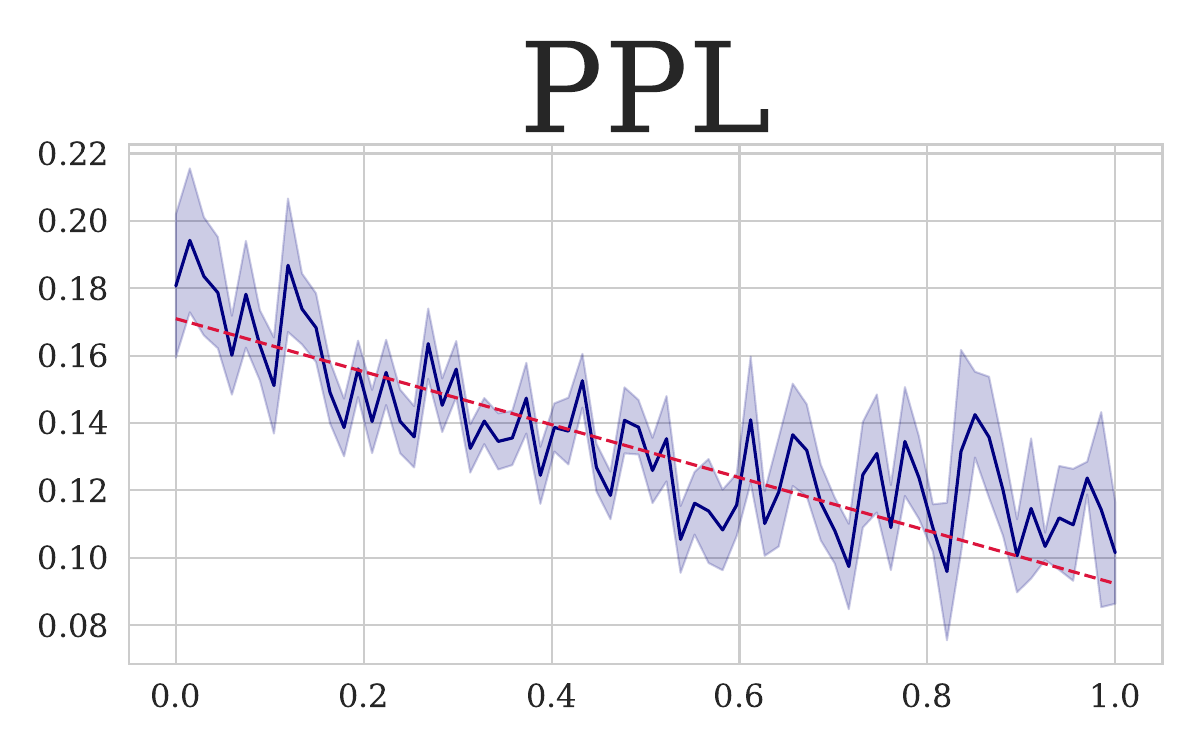}
    \end{subfigure}
    \begin{subfigure}{0.15\textwidth}
        \includegraphics[width=\linewidth]{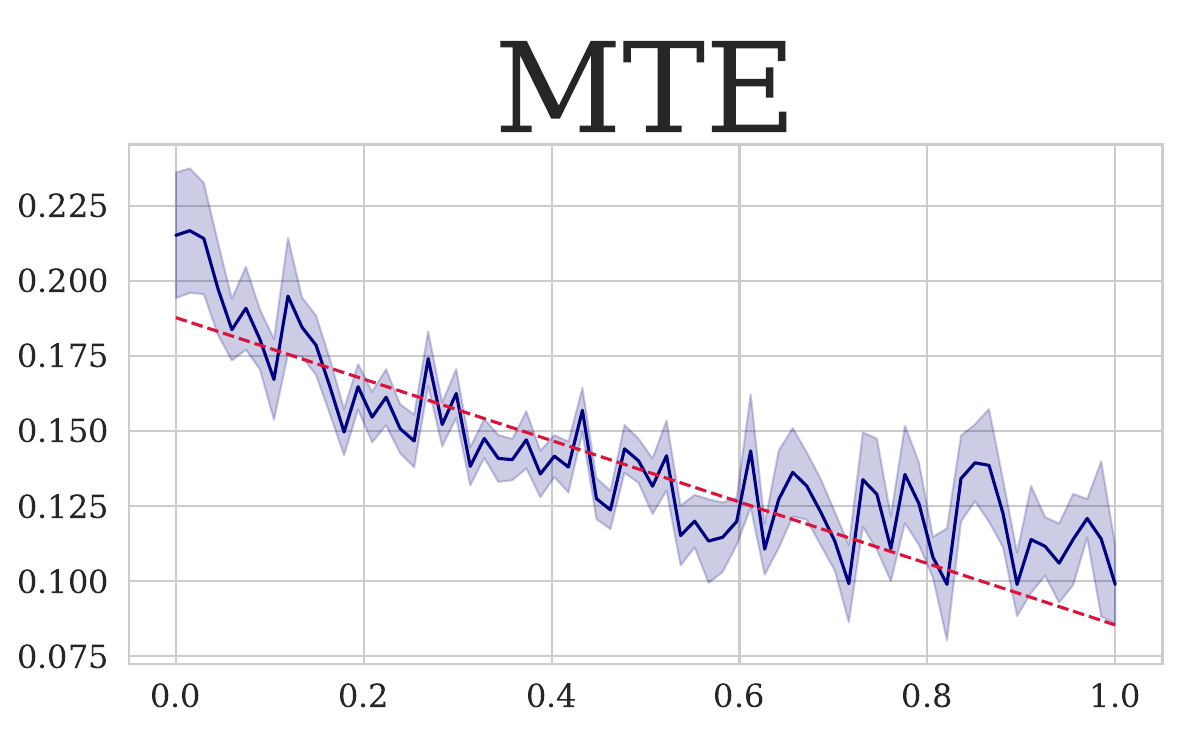}
    \end{subfigure}
    \begin{subfigure}{0.15\textwidth}
        \includegraphics[width=\linewidth]{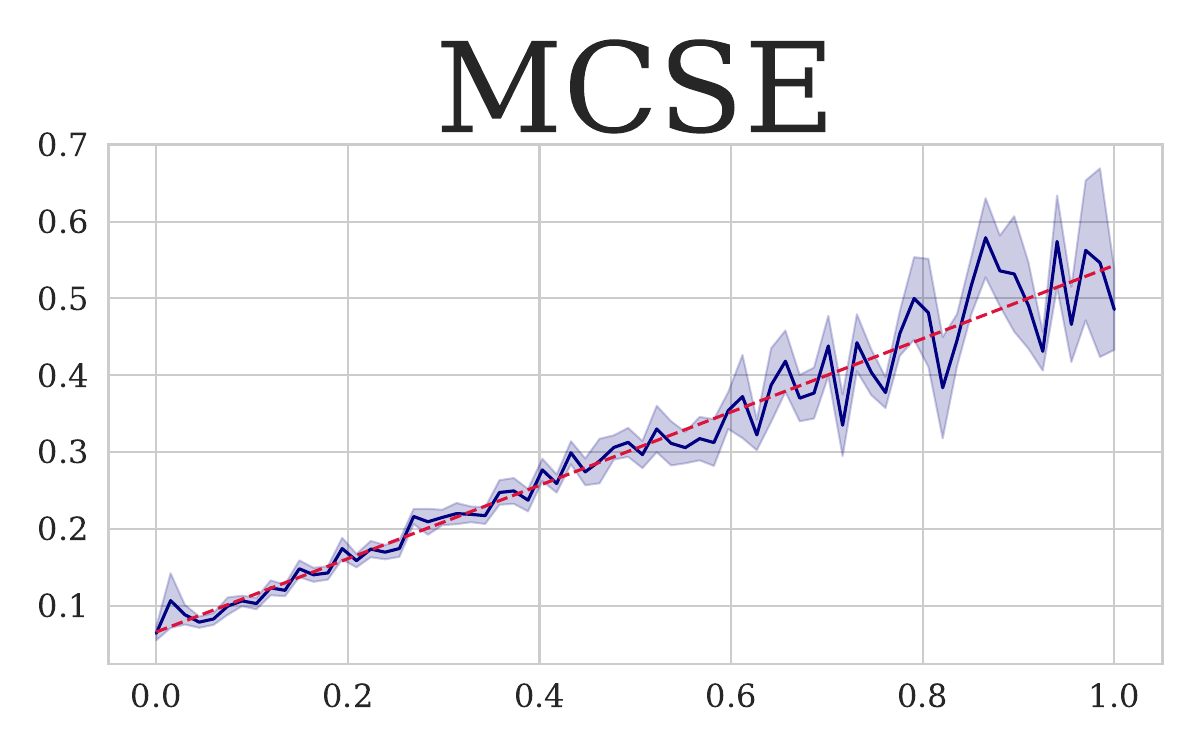}
    \end{subfigure}
    \begin{subfigure}{0.15\textwidth}
        \includegraphics[width=\linewidth]{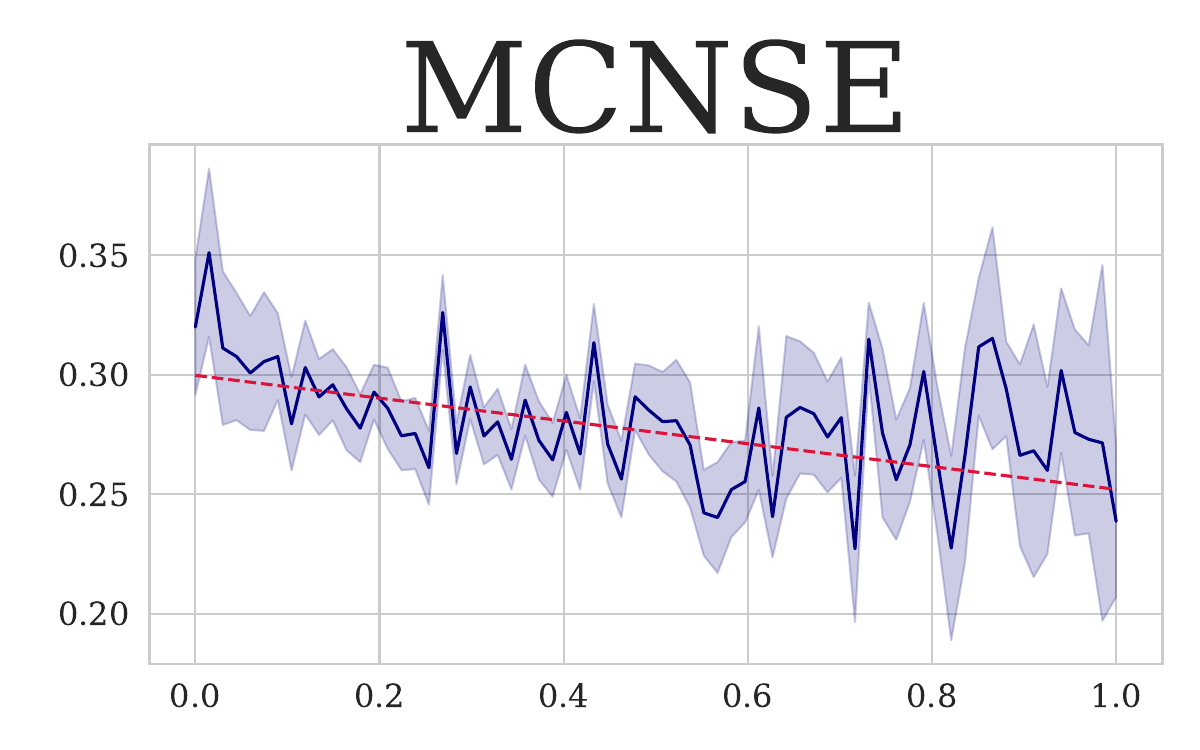}
    \end{subfigure}
    \begin{subfigure}{0.15\textwidth}
        \includegraphics[width=\linewidth]{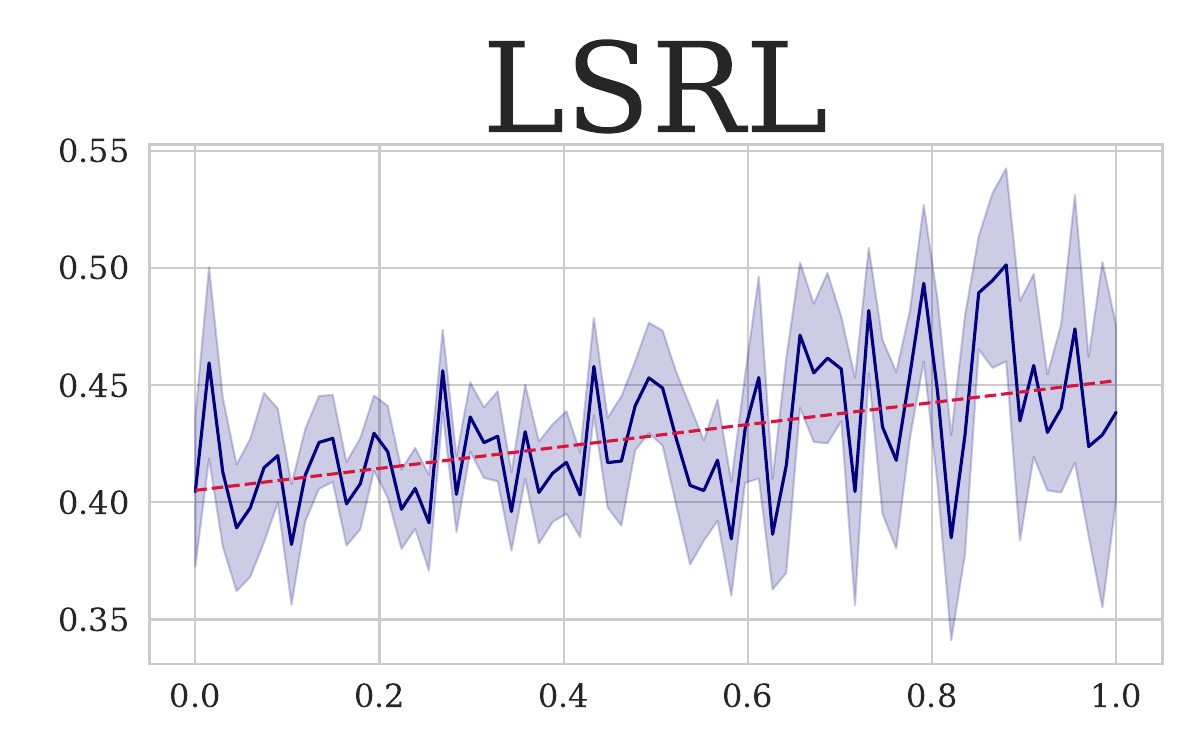}
    \end{subfigure}
    \vspace{0.3em}
    {\centering \textbf{\small WMT14 Cs-En} \par}
    \vspace{0.2em}
    \begin{subfigure}{0.15\textwidth}
        \includegraphics[width=\linewidth]{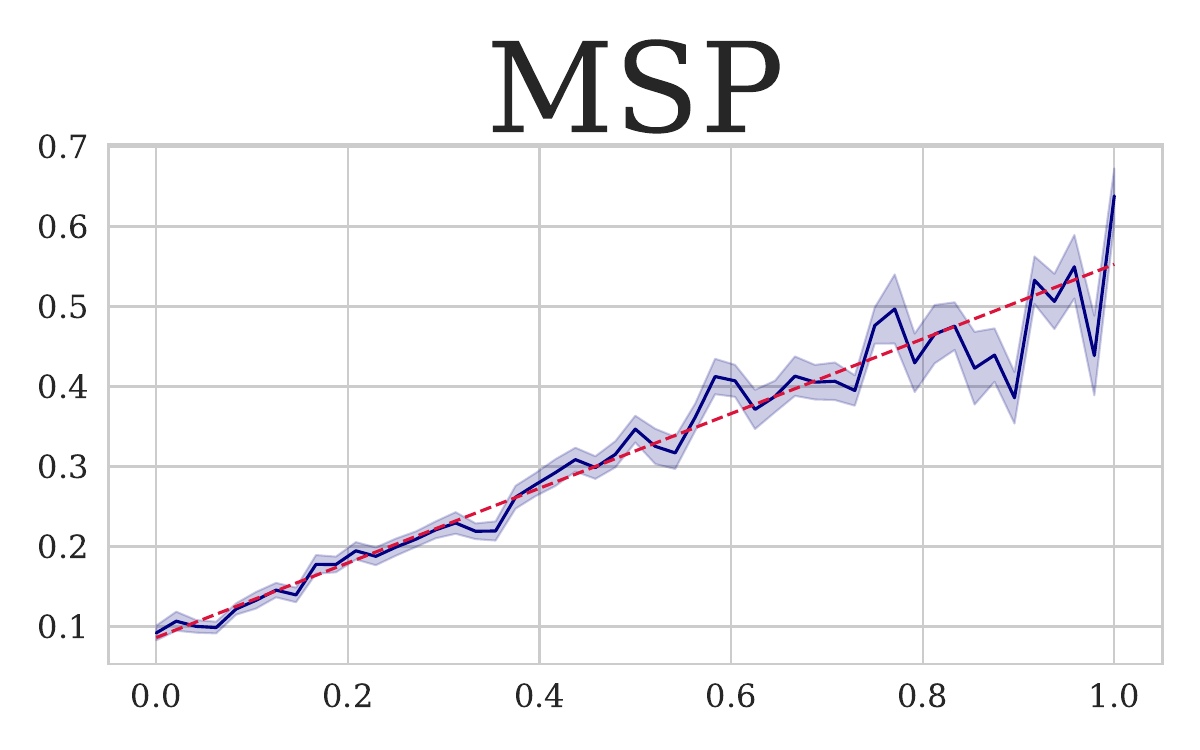}
    \end{subfigure}
    \begin{subfigure}{0.15\textwidth}
        \includegraphics[width=\linewidth]{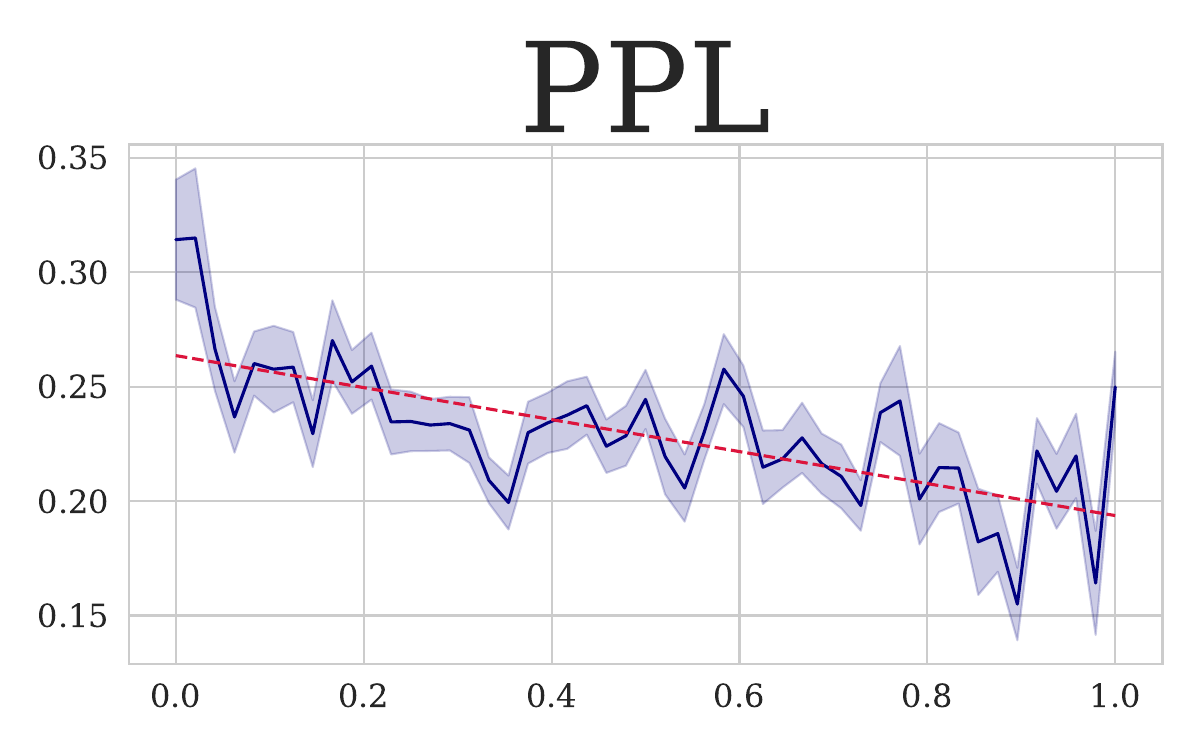}
    \end{subfigure}
    \begin{subfigure}{0.15\textwidth}
        \includegraphics[width=\linewidth]{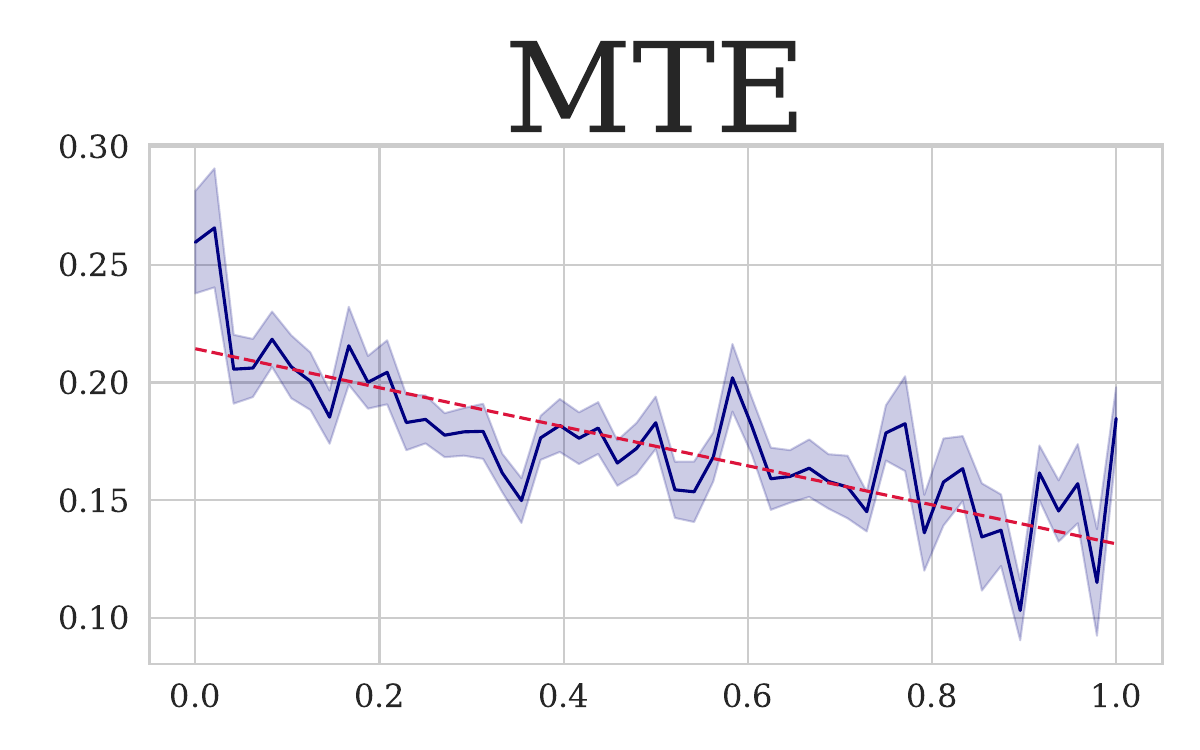}
    \end{subfigure}
    \begin{subfigure}{0.15\textwidth}
        \includegraphics[width=\linewidth]{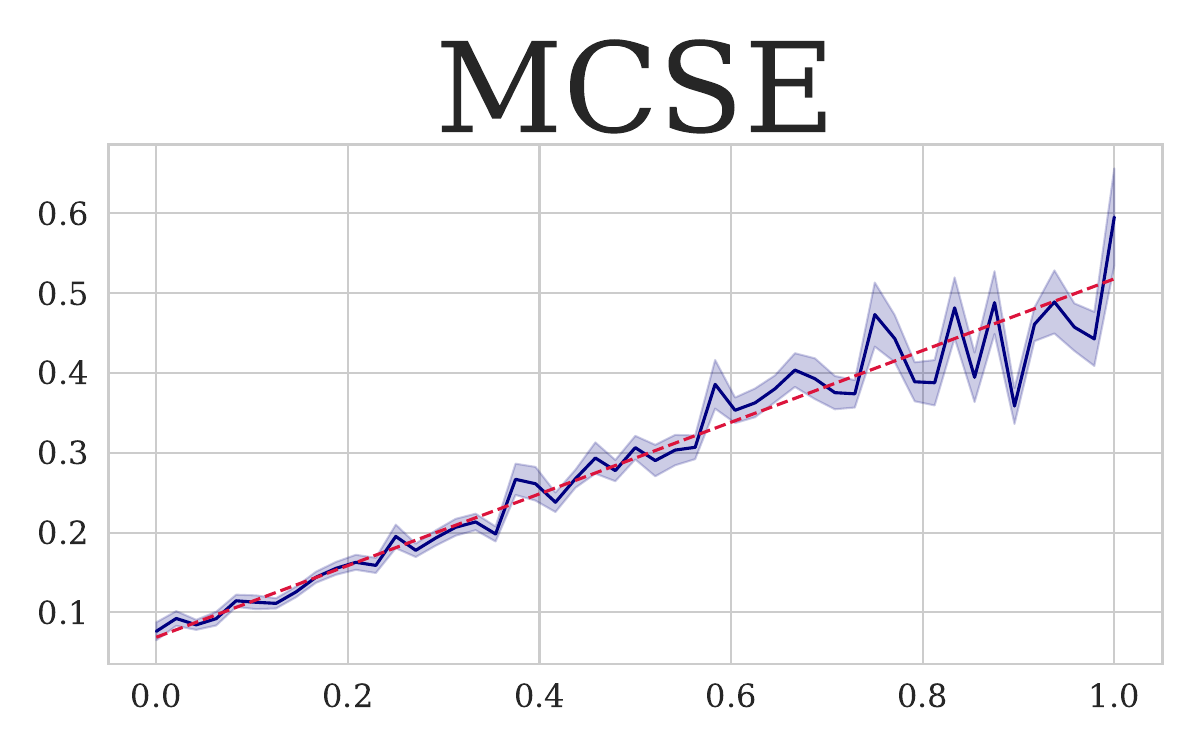}
    \end{subfigure}
    \begin{subfigure}{0.15\textwidth}
        \includegraphics[width=\linewidth]{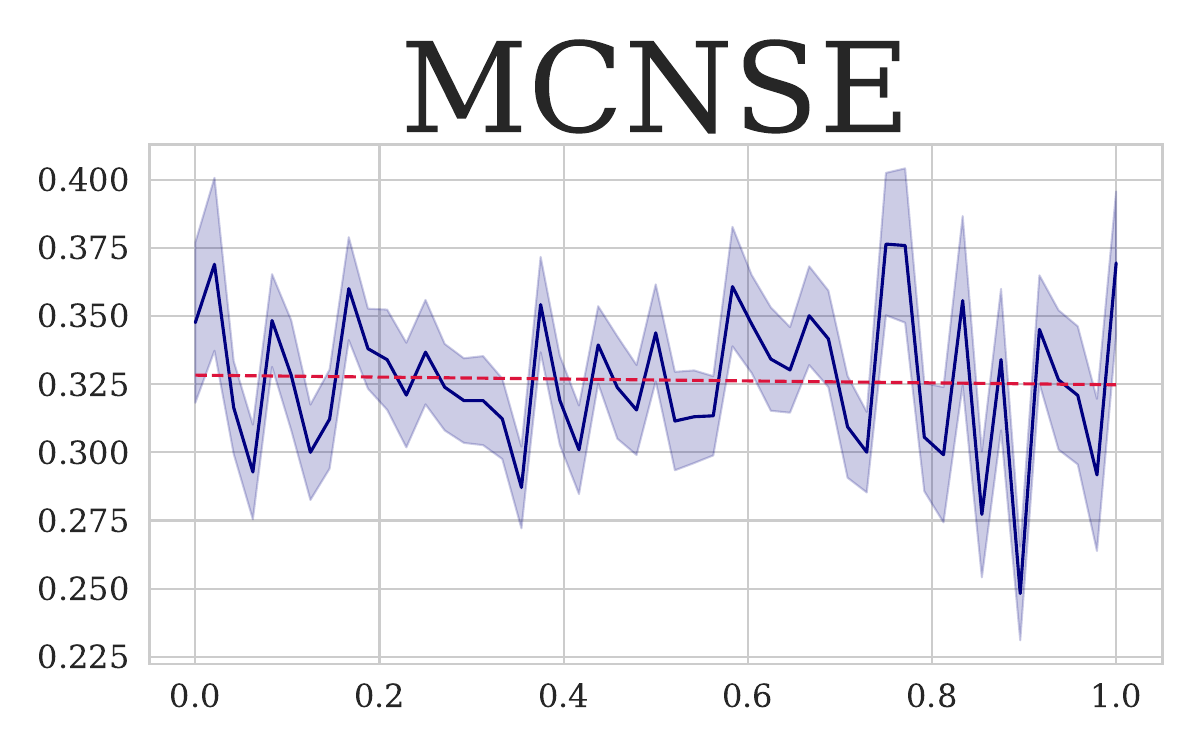}
    \end{subfigure}
    \begin{subfigure}{0.15\textwidth}
        \includegraphics[width=\linewidth]{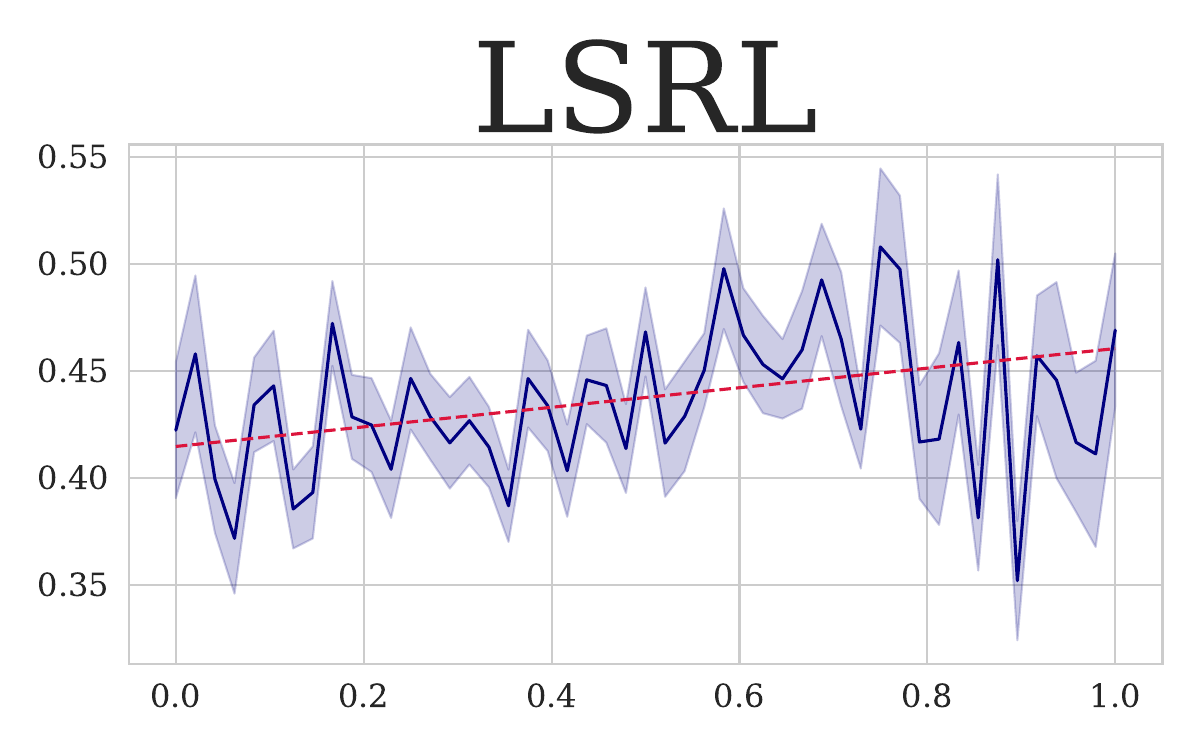}
    \end{subfigure}
    \vspace{0.3em}
    {\centering \textbf{\small WMT14 Ru-En} \par}
    \vspace{0.2em}
    \begin{subfigure}{0.15\textwidth}
        \includegraphics[width=\linewidth]{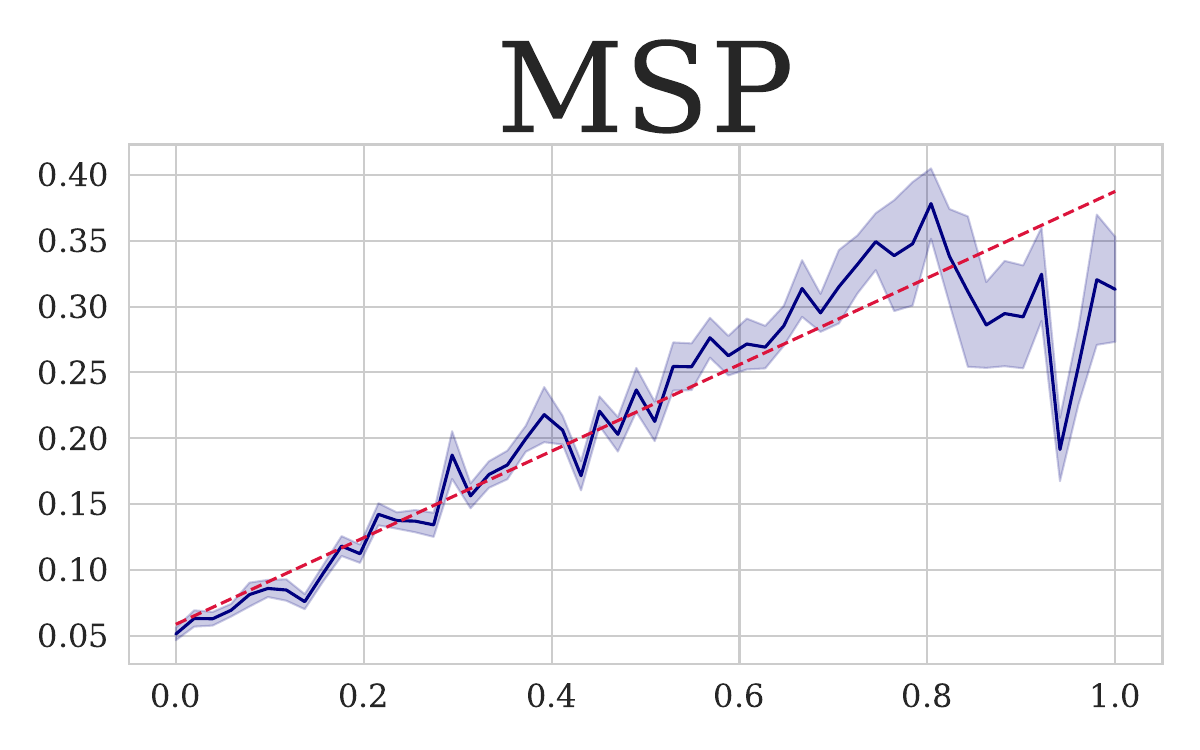}
    \end{subfigure}
    \begin{subfigure}{0.15\textwidth}
        \includegraphics[width=\linewidth]{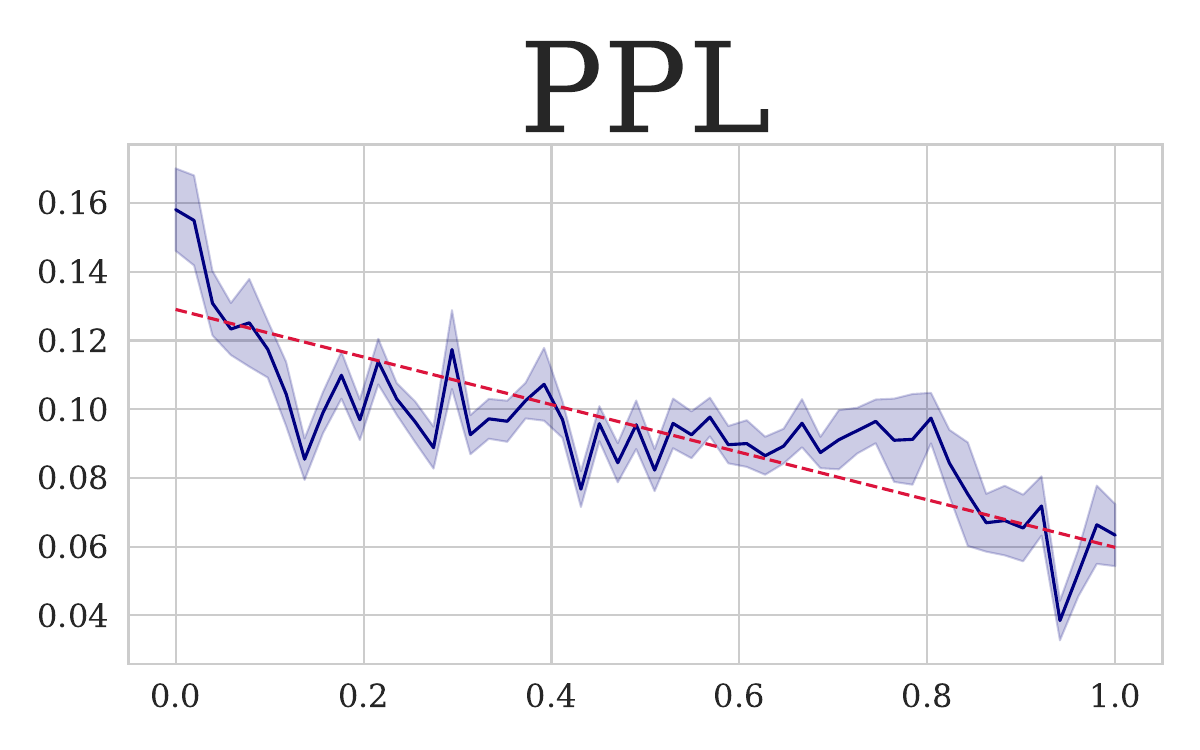}
    \end{subfigure}
    \begin{subfigure}{0.15\textwidth}
        \includegraphics[width=\linewidth]{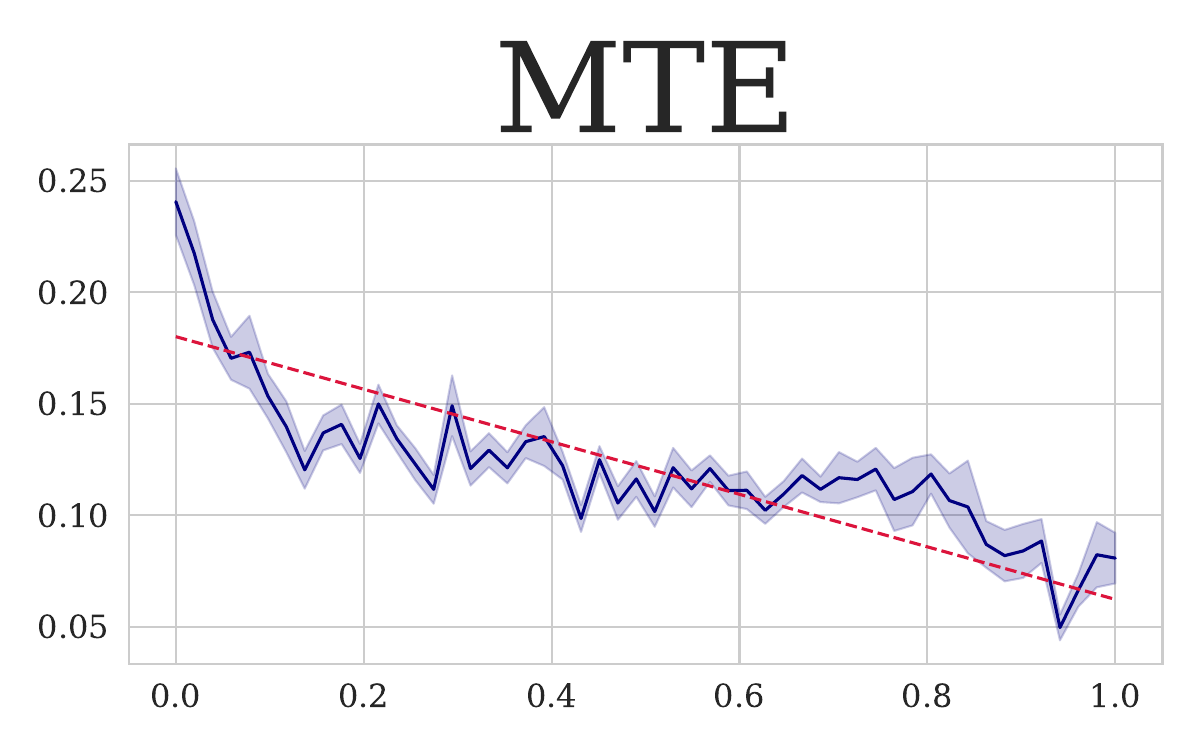}
    \end{subfigure}
    \begin{subfigure}{0.15\textwidth}
        \includegraphics[width=\linewidth]{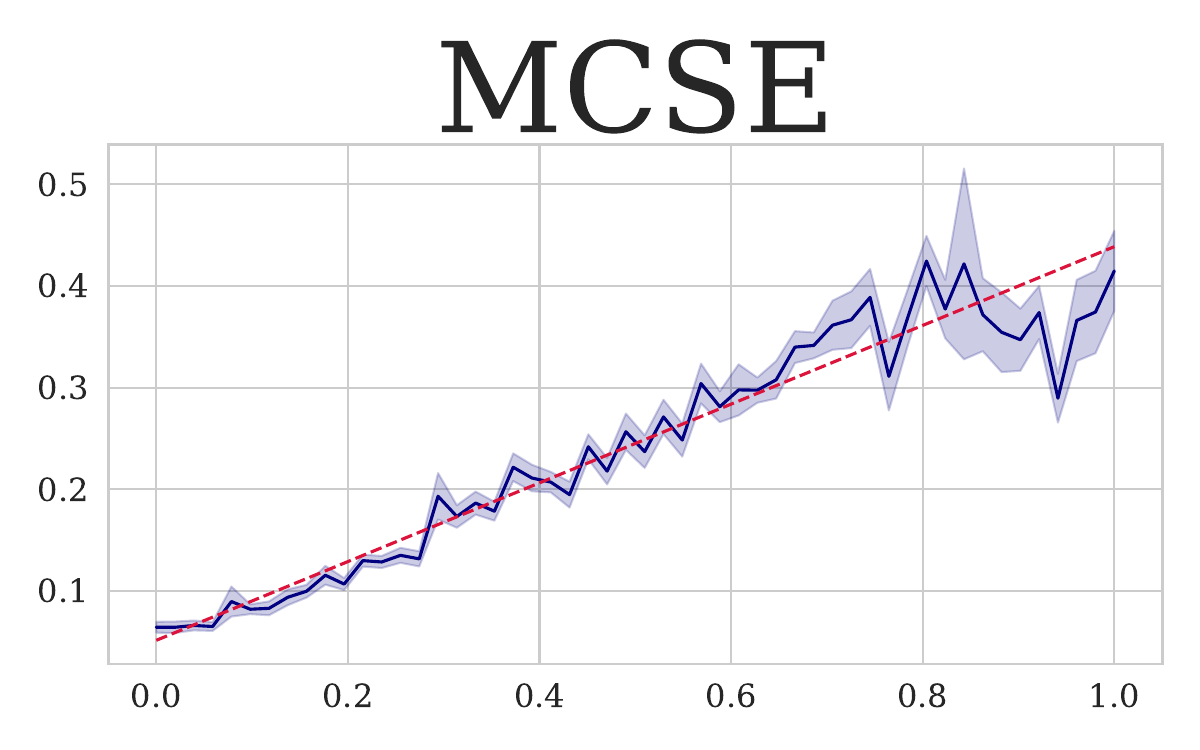}
    \end{subfigure}
    \begin{subfigure}{0.15\textwidth}
        \includegraphics[width=\linewidth]{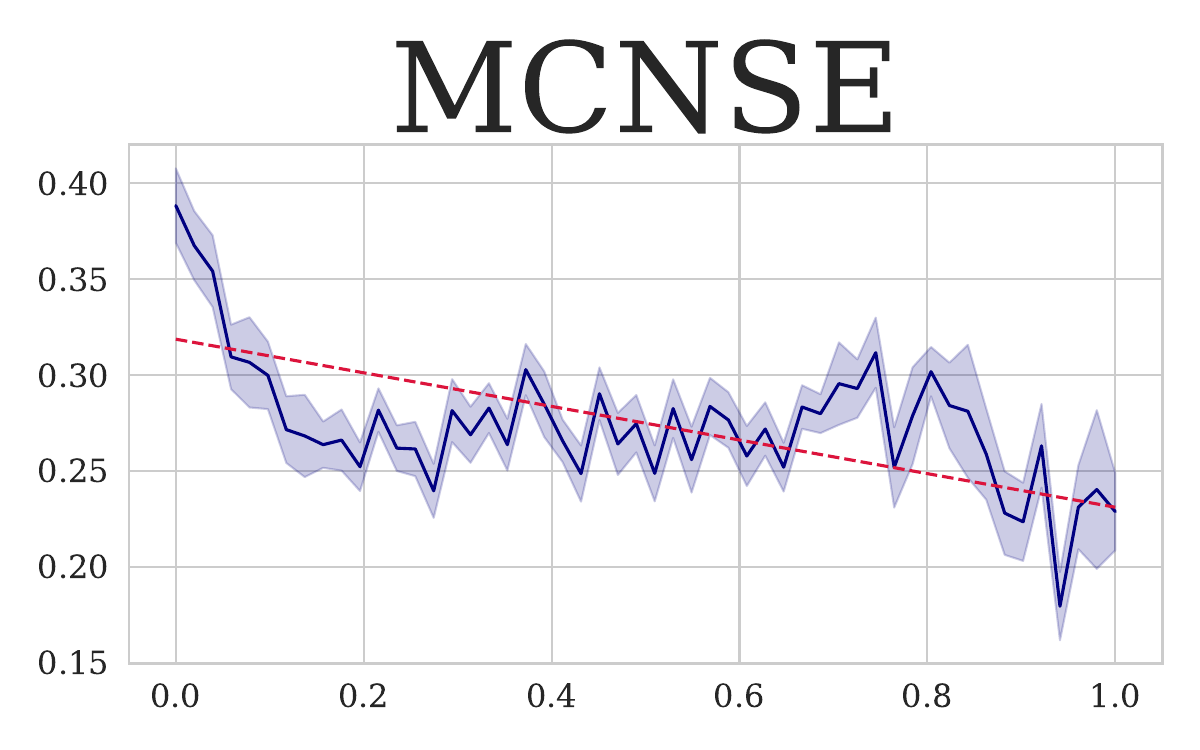}
    \end{subfigure}
    \begin{subfigure}{0.15\textwidth}
        \includegraphics[width=\linewidth]{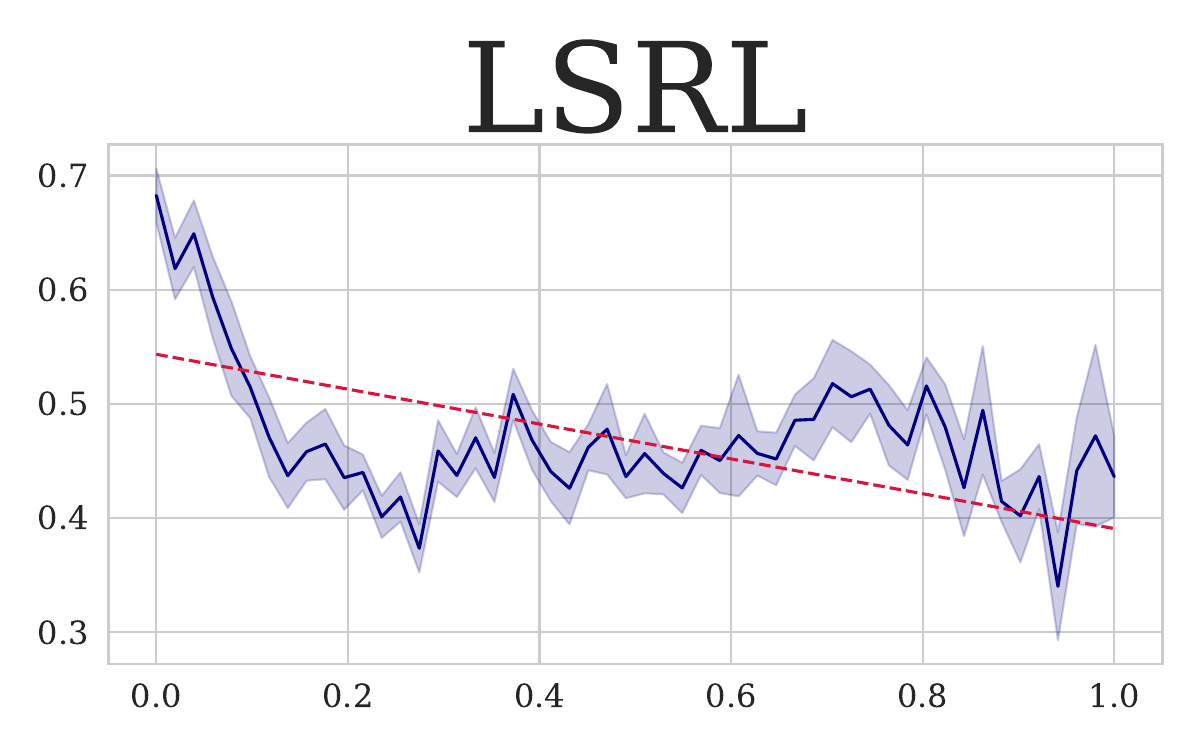}
    \end{subfigure}
    \vspace{0.3em}
    {\centering \textbf{\small WMT19 Ru-En} \par}
    \vspace{0.2em}
    \begin{subfigure}{0.15\textwidth}
        \includegraphics[width=\linewidth]{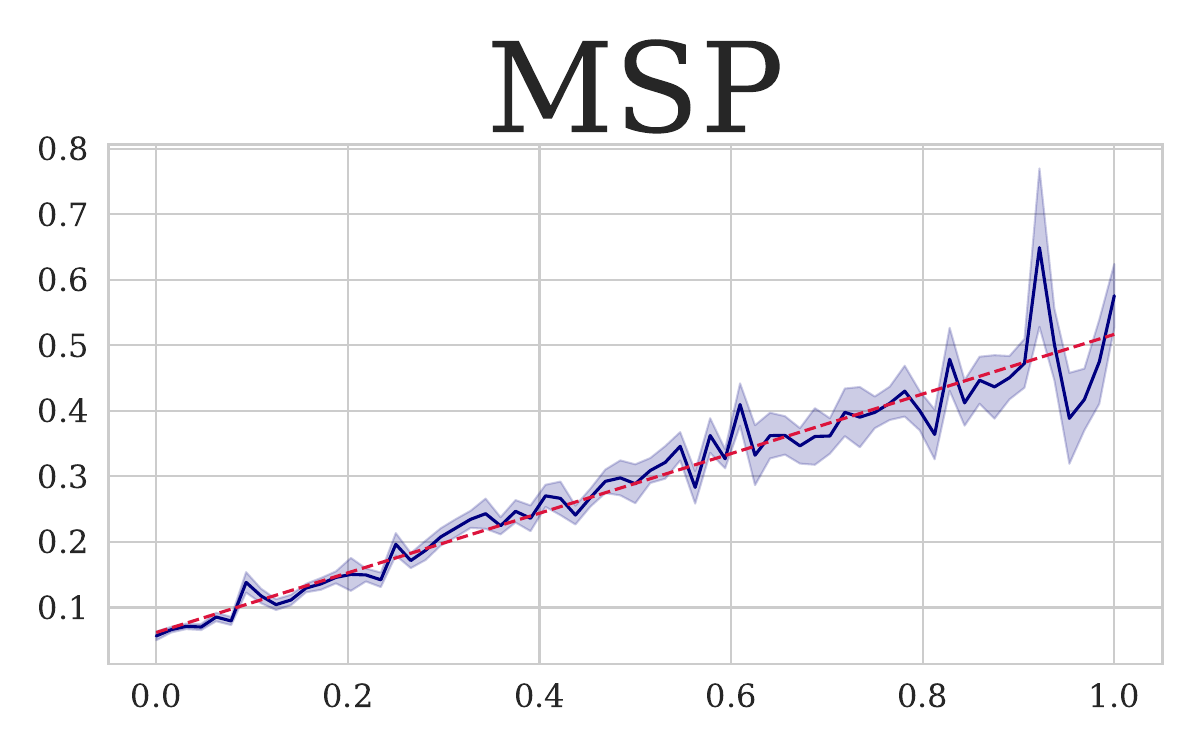}
    \end{subfigure}
    \begin{subfigure}{0.15\textwidth}
        \includegraphics[width=\linewidth]{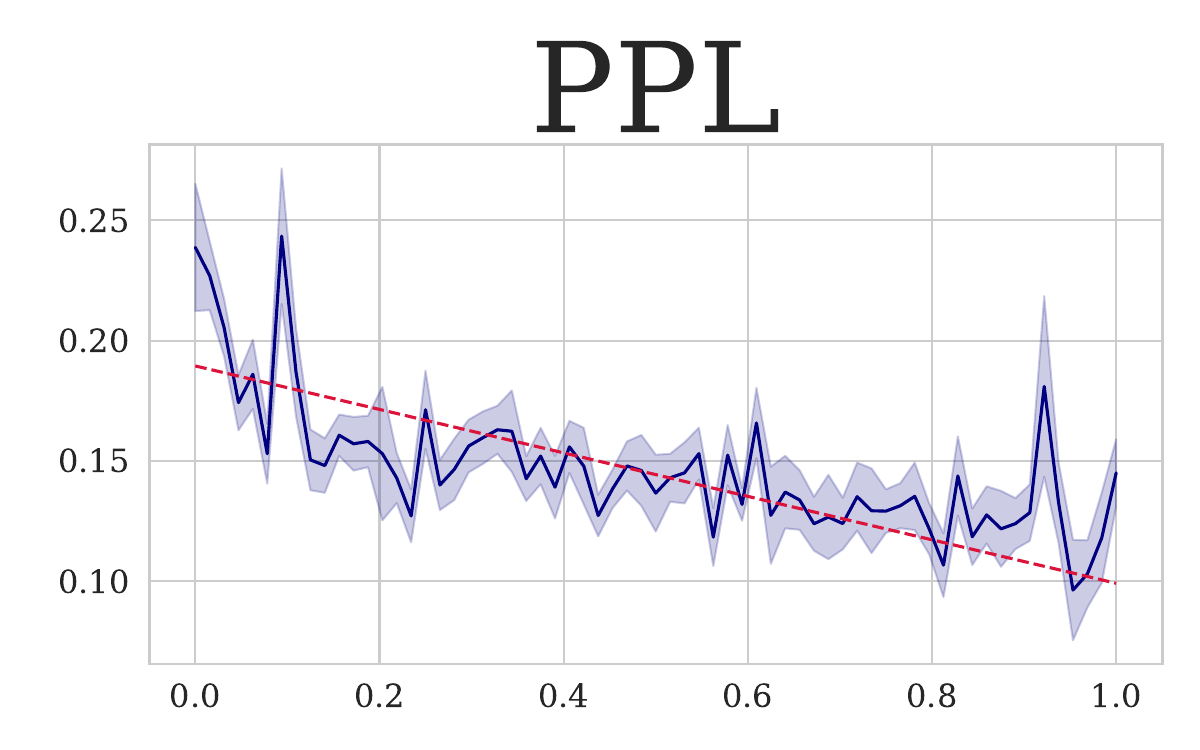}
    \end{subfigure}
    \begin{subfigure}{0.15\textwidth}
        \includegraphics[width=\linewidth]{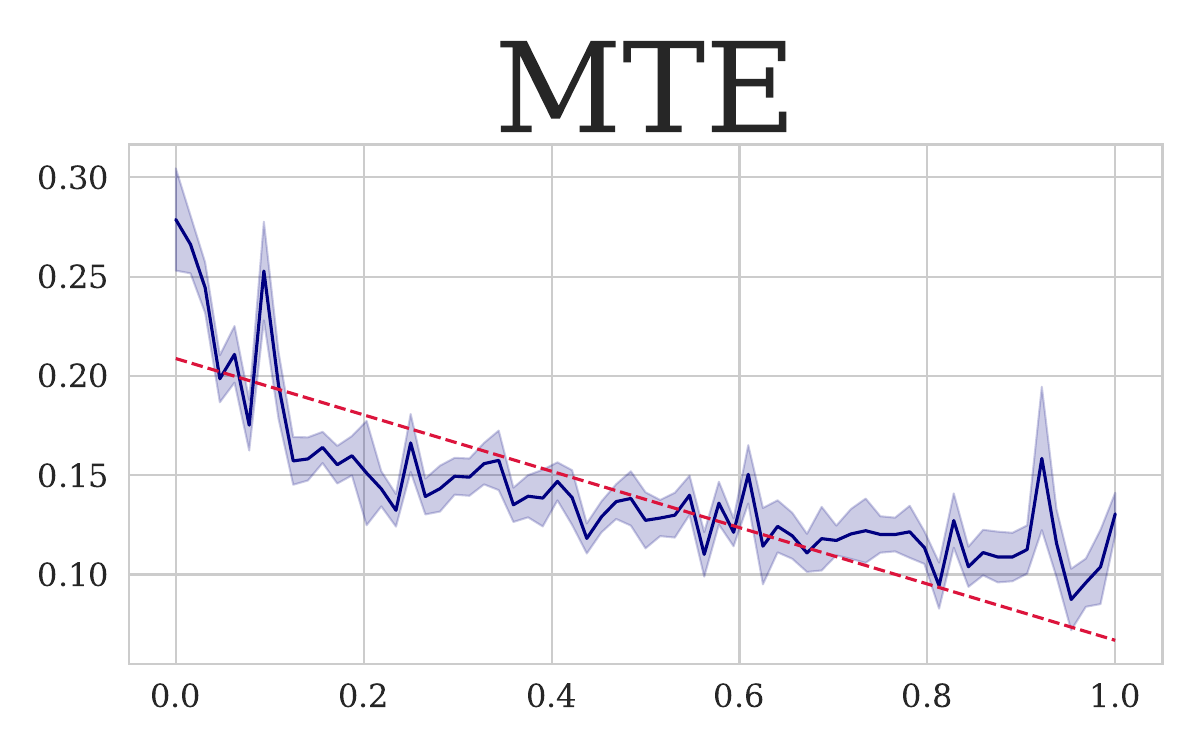}
    \end{subfigure}
    \begin{subfigure}{0.15\textwidth}
        \includegraphics[width=\linewidth]{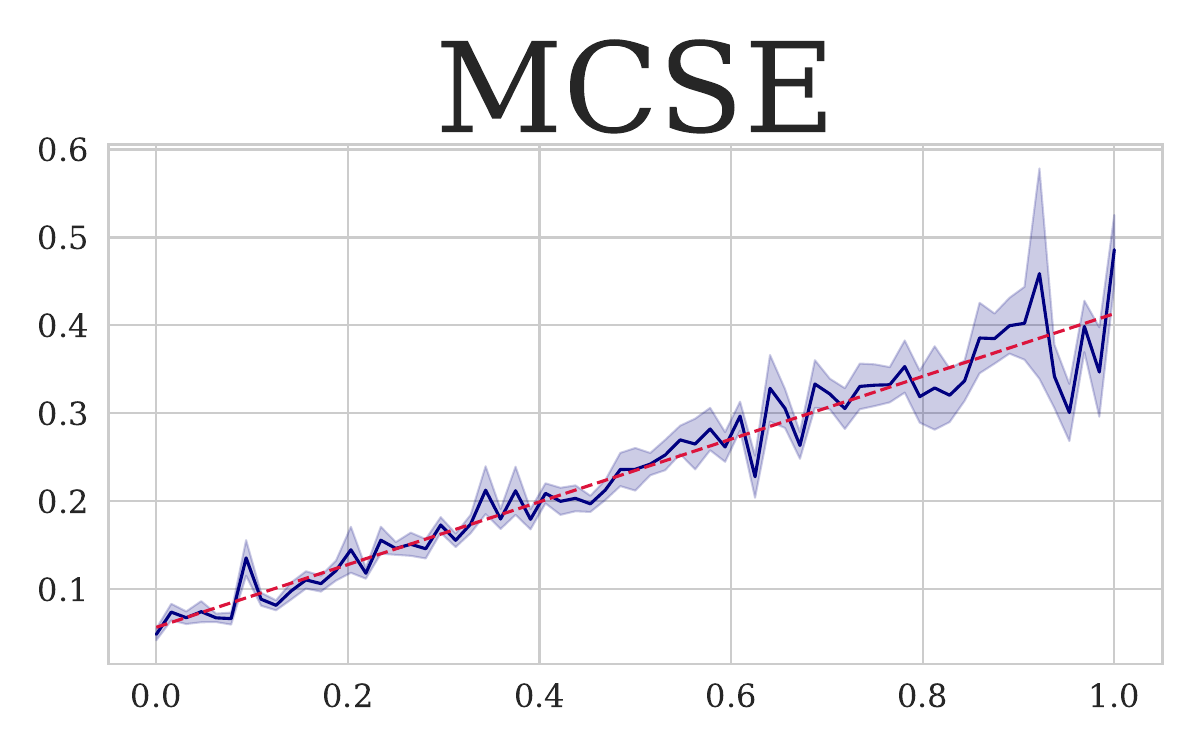}
    \end{subfigure}
    \begin{subfigure}{0.15\textwidth}
        \includegraphics[width=\linewidth]{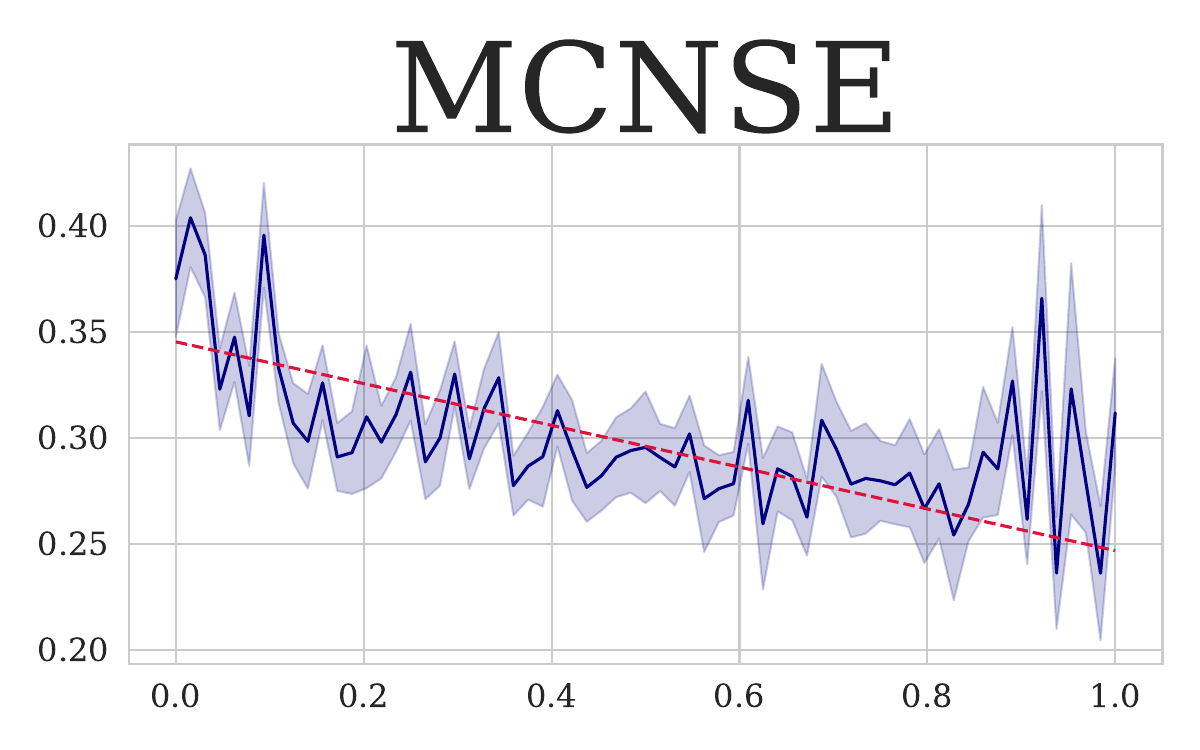}
    \end{subfigure}
    \begin{subfigure}{0.15\textwidth}
        \includegraphics[width=\linewidth]{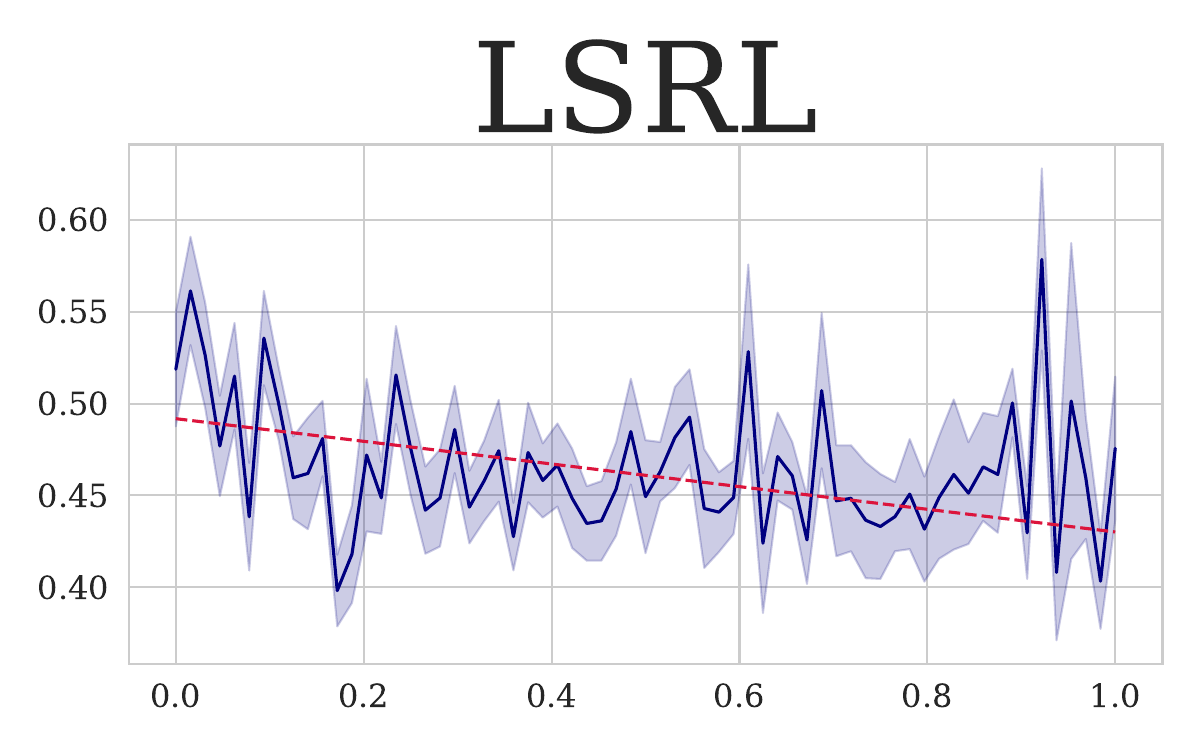}
    \end{subfigure}
    
    \vspace{0.3em}
    {\centering \textbf{\small WMT19 Fi-En} \par}
    \vspace{0.2em}
    \begin{subfigure}{0.15\textwidth}
        \includegraphics[width=\linewidth]{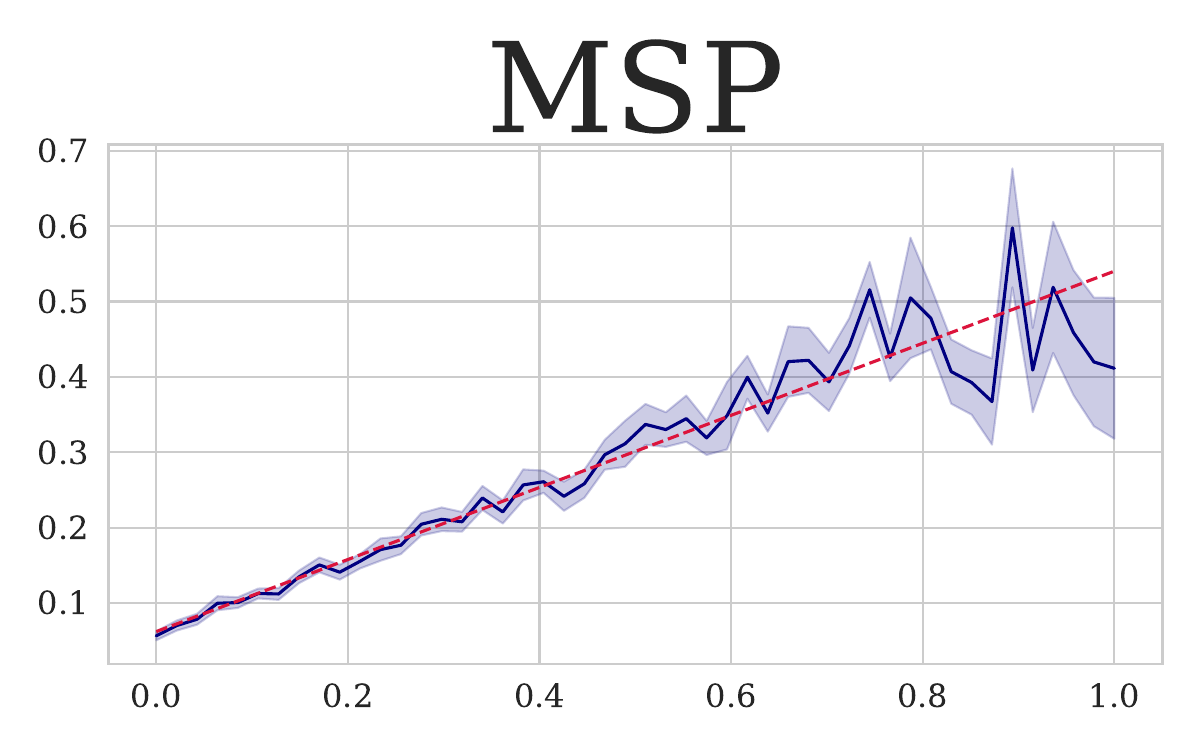}
    \end{subfigure}
    \begin{subfigure}{0.15\textwidth}
        \includegraphics[width=\linewidth]{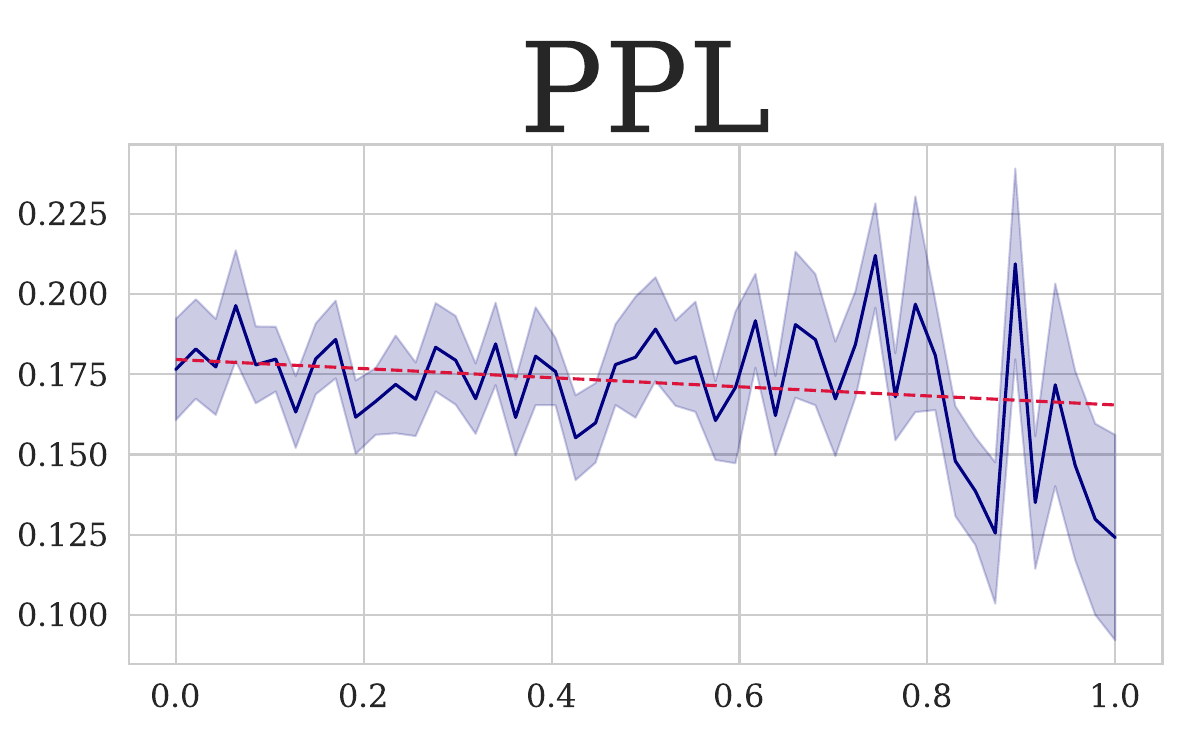}
    \end{subfigure}
    \begin{subfigure}{0.15\textwidth}
        \includegraphics[width=\linewidth]{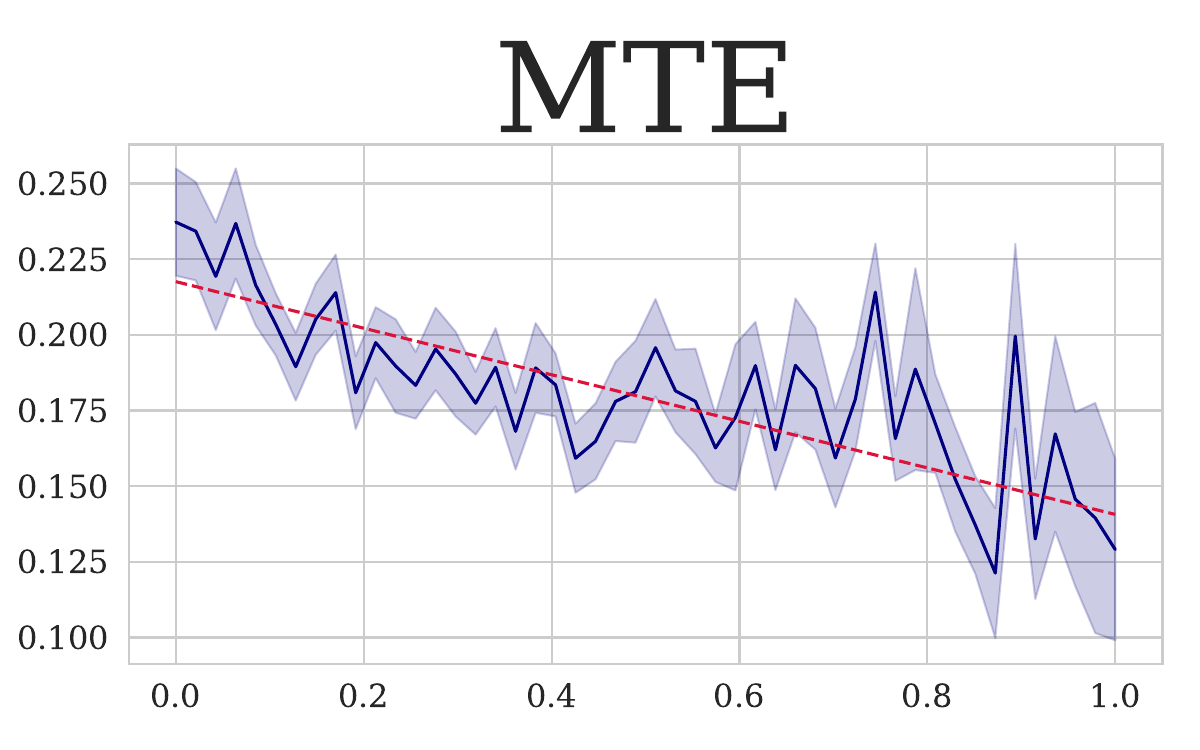}
    \end{subfigure}
    \begin{subfigure}{0.15\textwidth}
        \includegraphics[width=\linewidth]{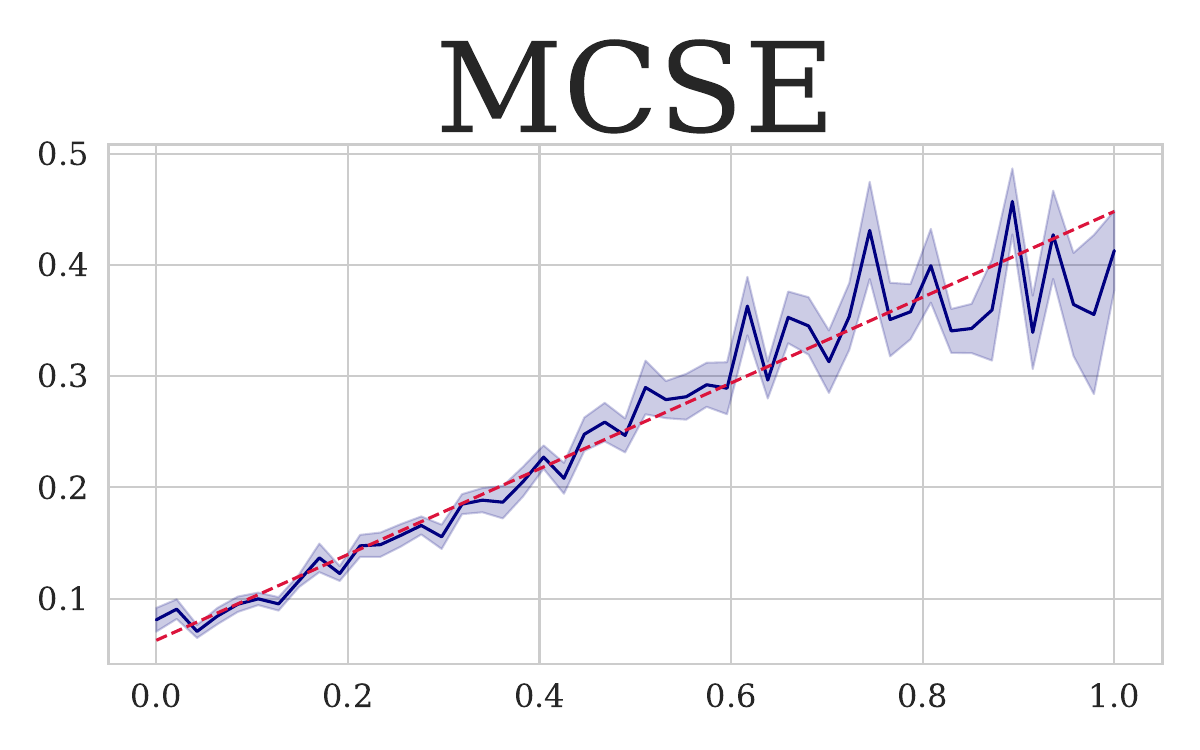}
    \end{subfigure}
    \begin{subfigure}{0.15\textwidth}
        \includegraphics[width=\linewidth]{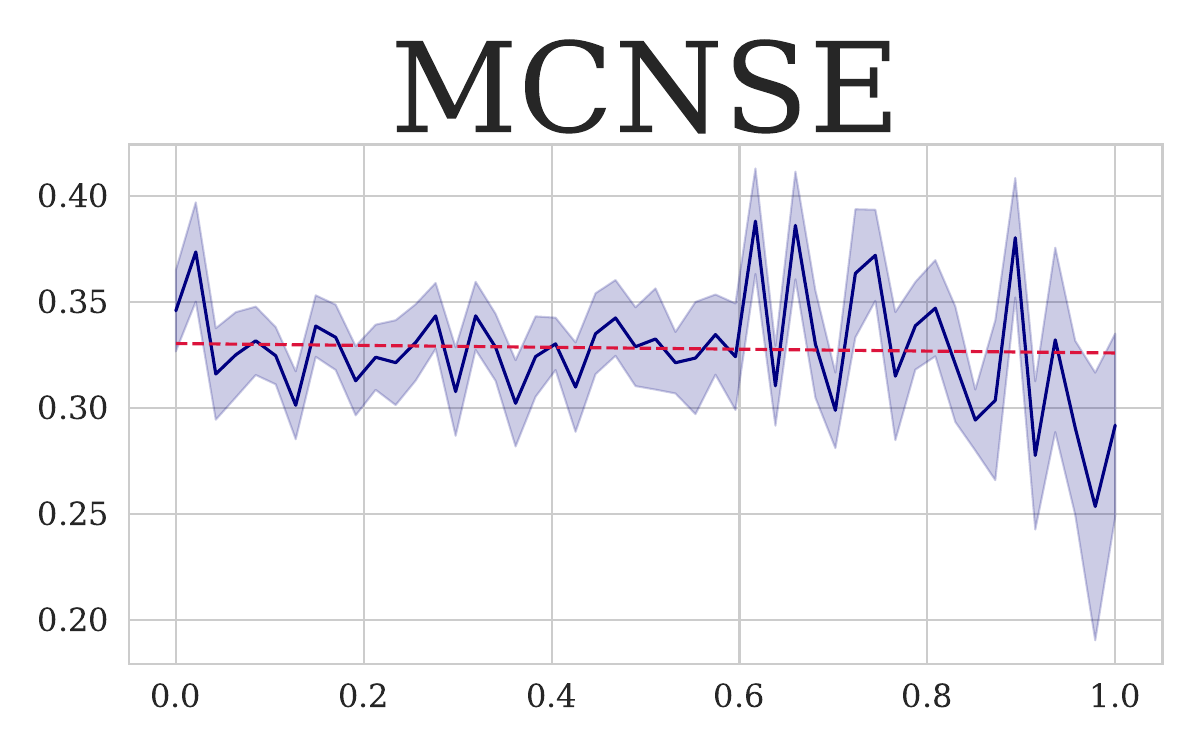}
    \end{subfigure}
    \begin{subfigure}{0.15\textwidth}
        \includegraphics[width=\linewidth]{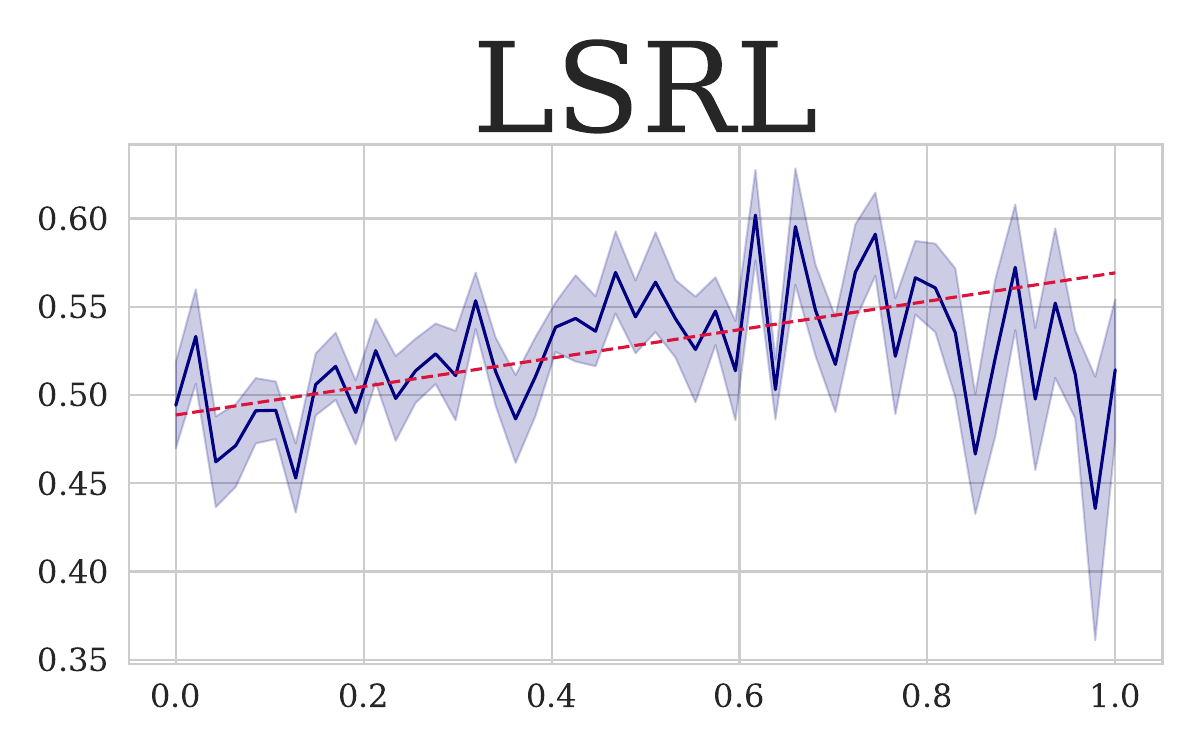}
    \end{subfigure}
    \vspace{0.3em}
    {\centering \textbf{\small WMT19 De-En} \par}
    \vspace{0.2em}
    \begin{subfigure}{0.15\textwidth}
        \includegraphics[width=\linewidth]{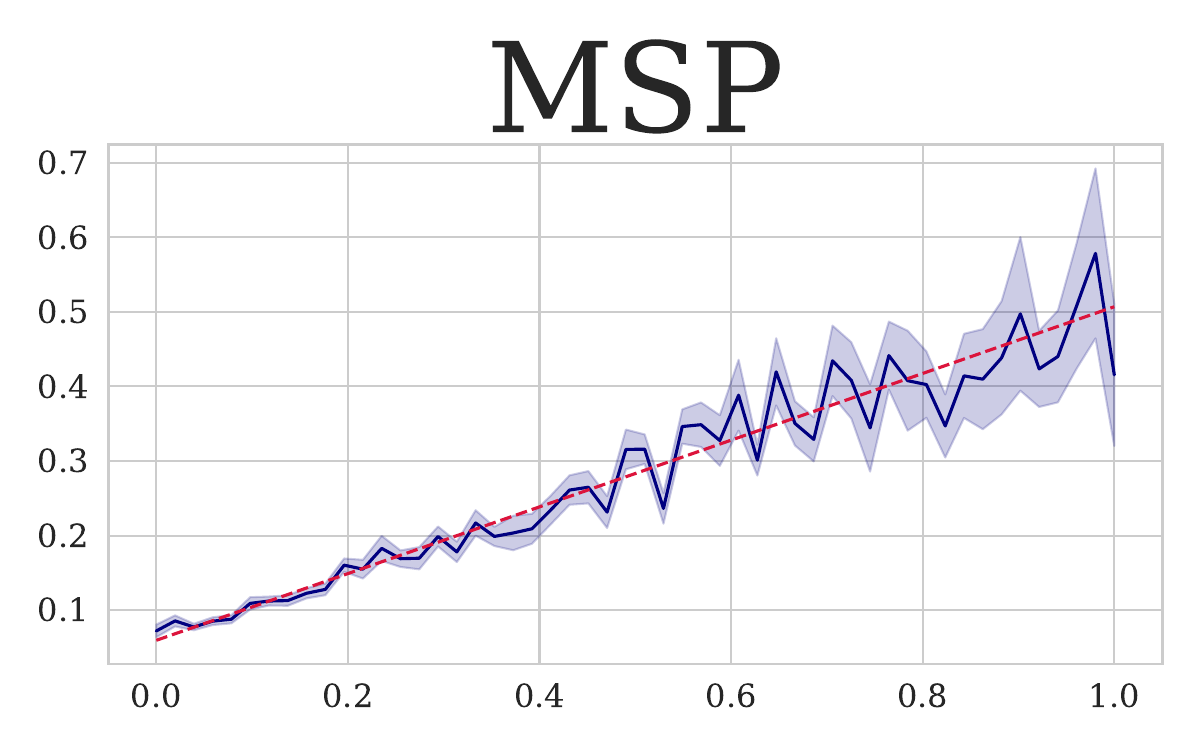}
    \end{subfigure}
    \begin{subfigure}{0.15\textwidth}
        \includegraphics[width=\linewidth]{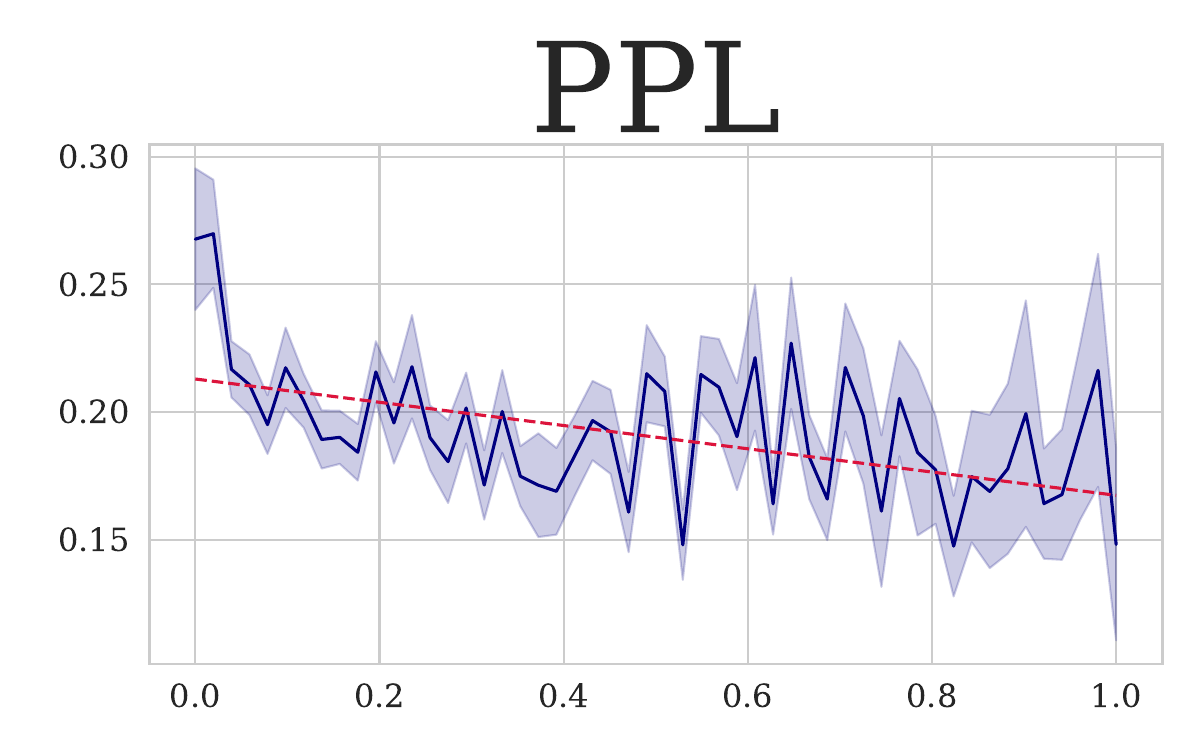}
    \end{subfigure}
    \begin{subfigure}{0.15\textwidth}
        \includegraphics[width=\linewidth]{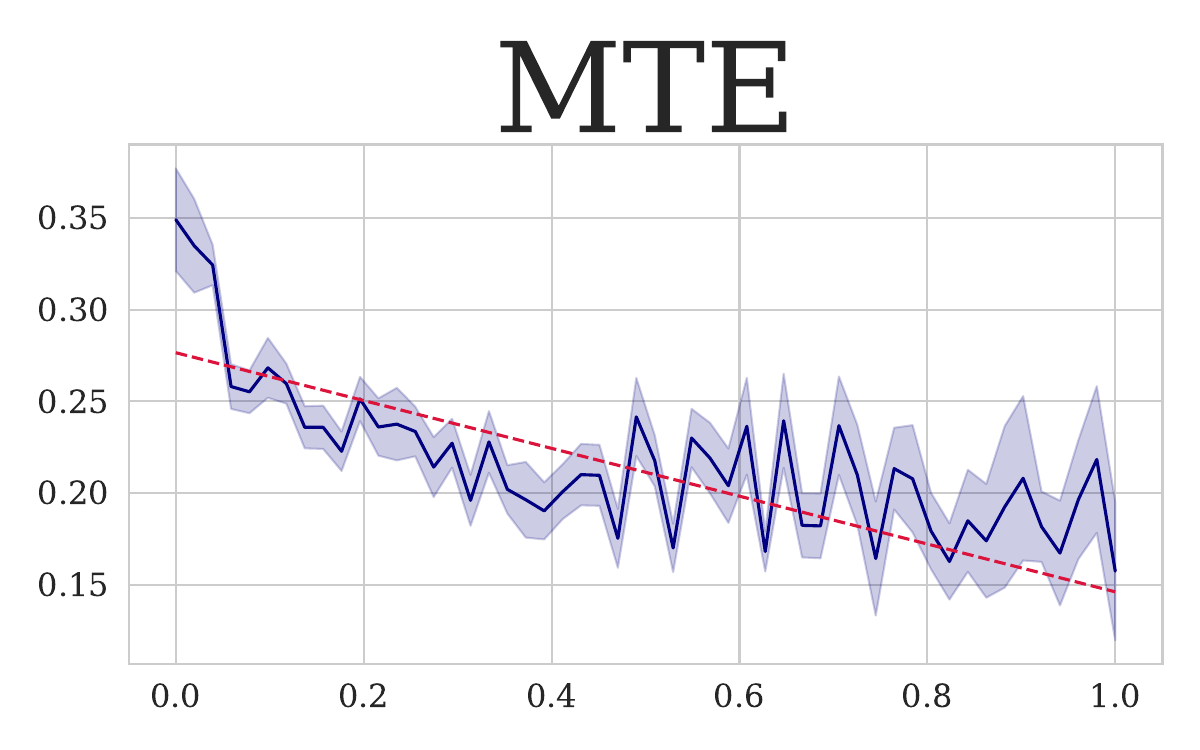}
    \end{subfigure}
    \begin{subfigure}{0.15\textwidth}
        \includegraphics[width=\linewidth]{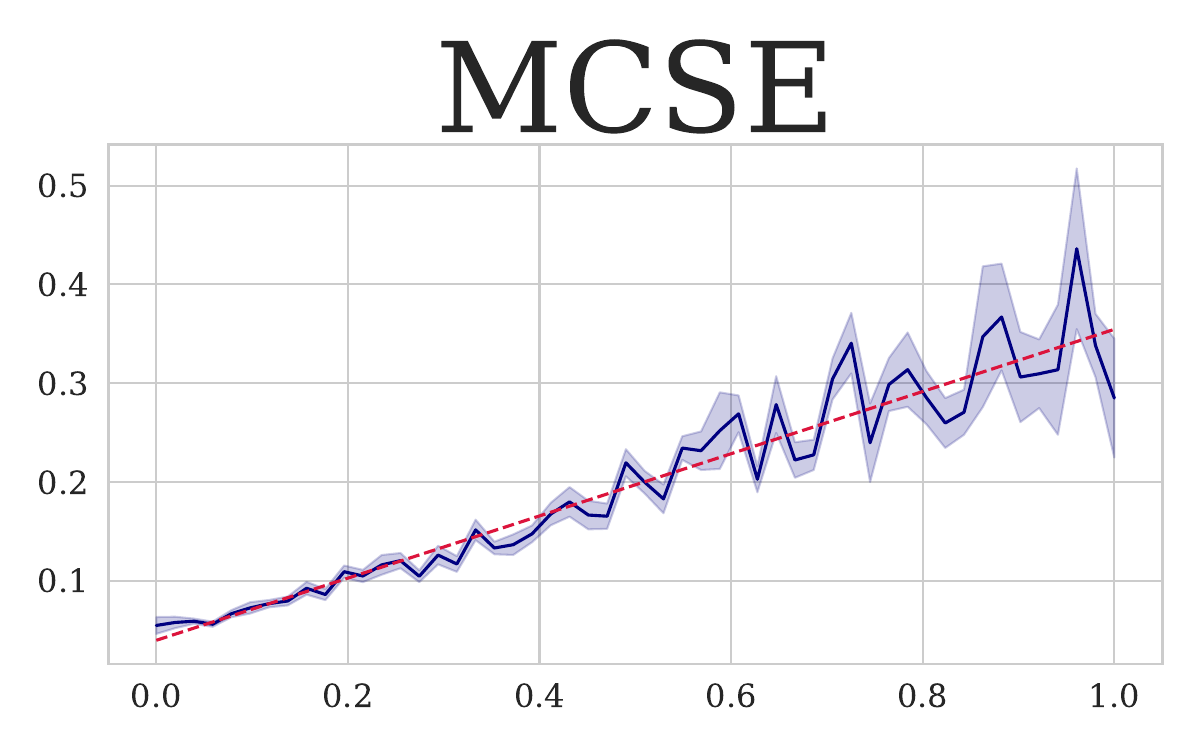}
    \end{subfigure}
    \begin{subfigure}{0.15\textwidth}
        \includegraphics[width=\linewidth]{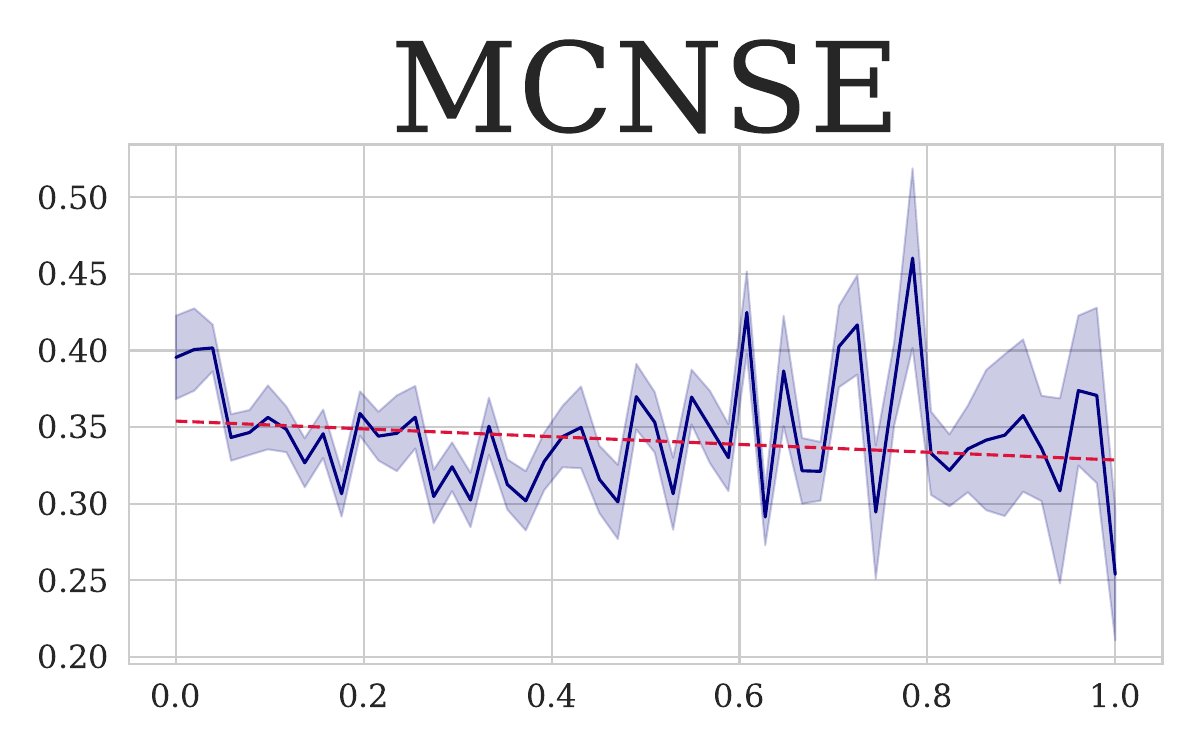}
    \end{subfigure}
    \begin{subfigure}{0.15\textwidth}
        \includegraphics[width=\linewidth]{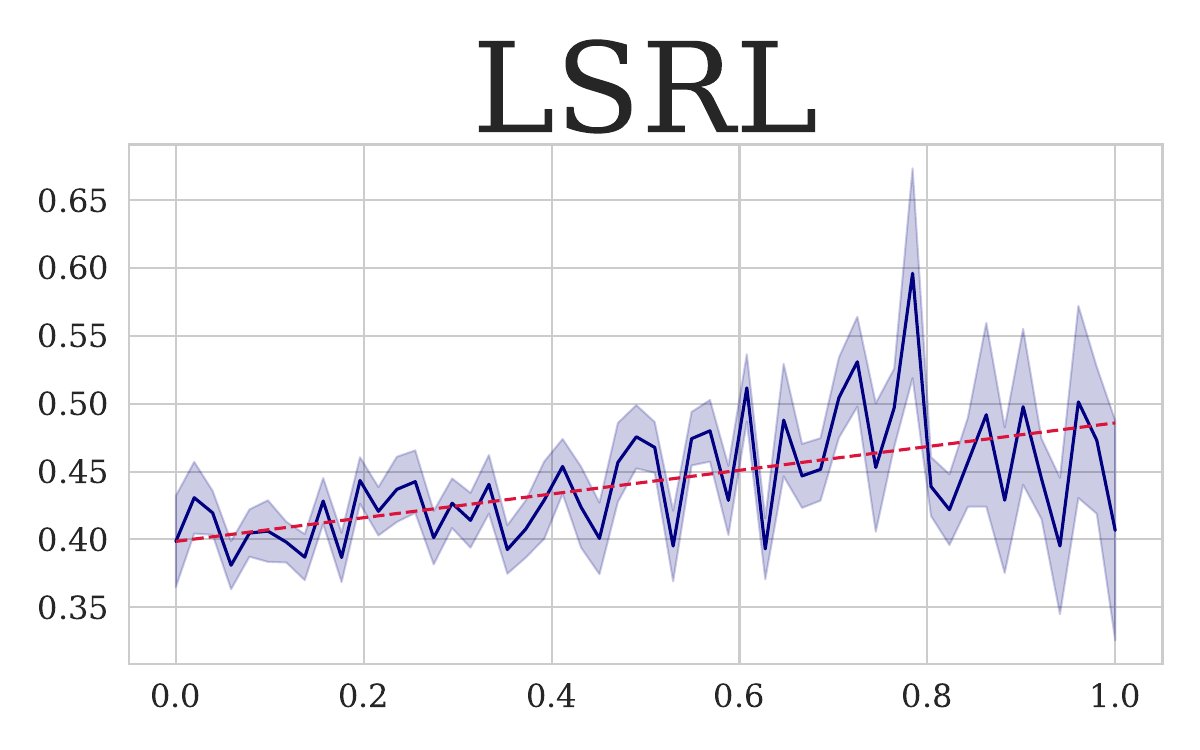}
    \end{subfigure}
    \vspace{0.3em}
    {\centering \textbf{\small WMT19 Lt-En} \par}
    \vspace{0.2em}
    \begin{subfigure}{0.15\textwidth}
        \includegraphics[width=\linewidth]{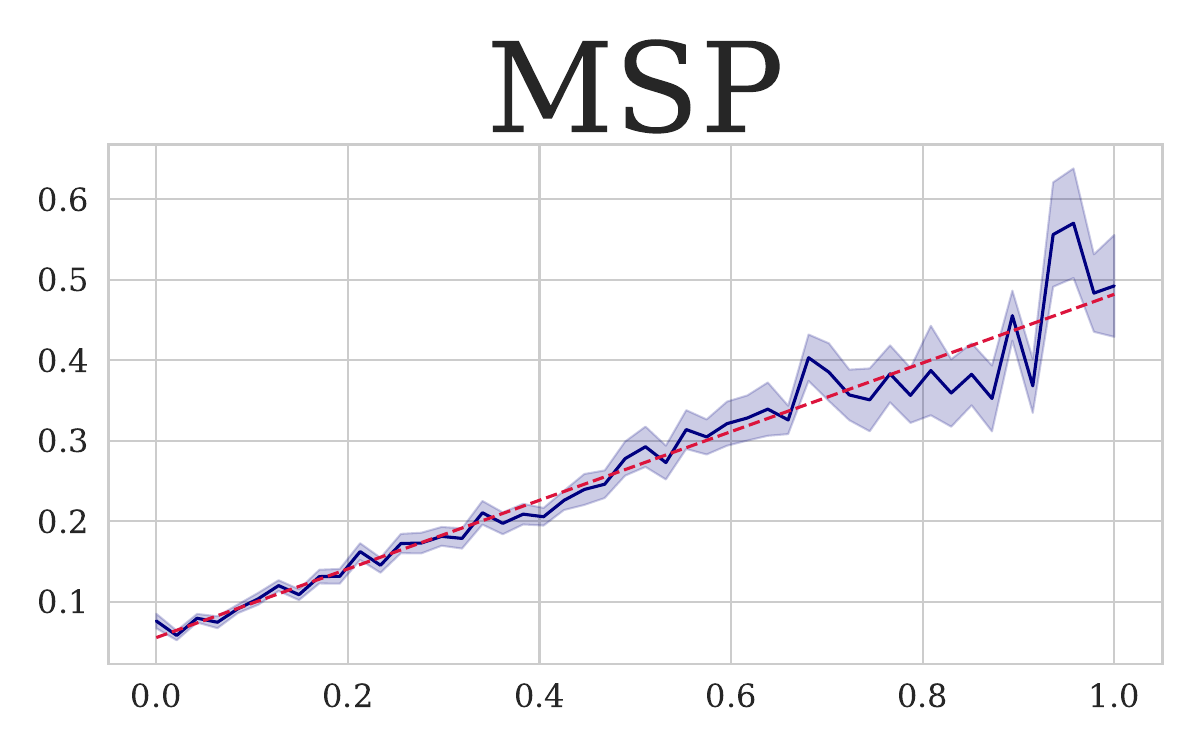}
    \end{subfigure}
    \begin{subfigure}{0.15\textwidth}
        \includegraphics[width=\linewidth]{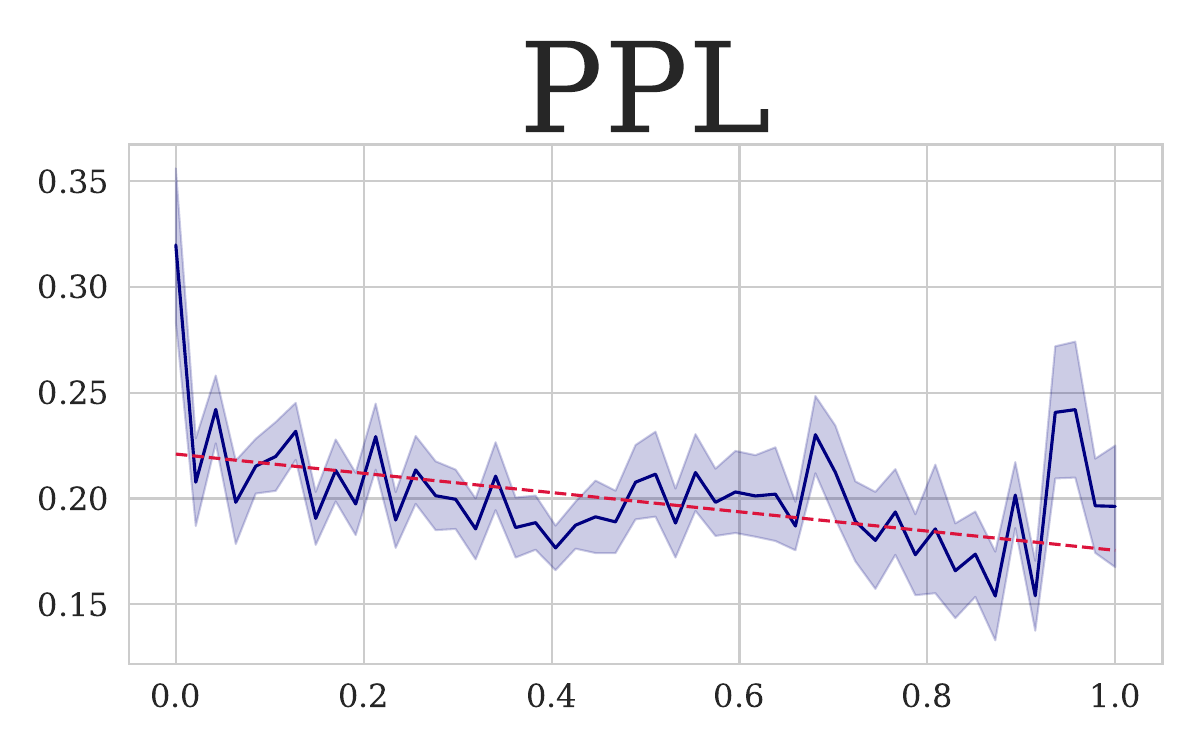}
    \end{subfigure}
    \begin{subfigure}{0.15\textwidth}
        \includegraphics[width=\linewidth]{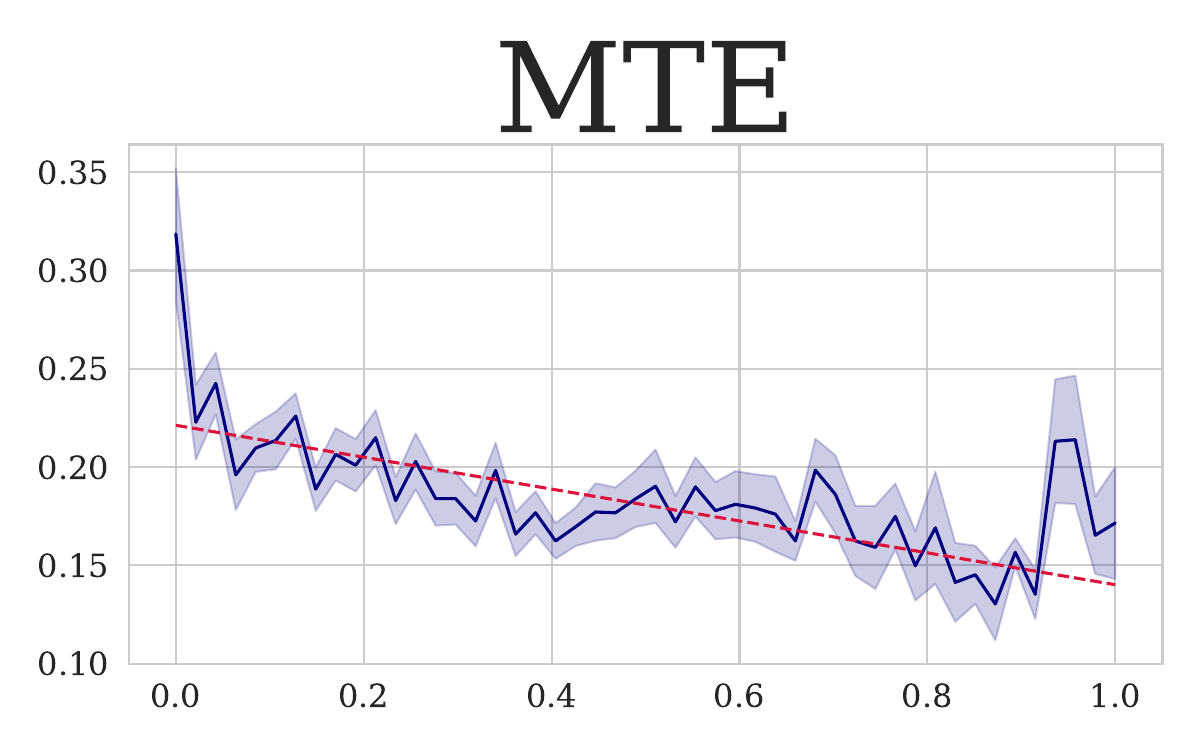}
    \end{subfigure}
    \begin{subfigure}{0.15\textwidth}
        \includegraphics[width=\linewidth]{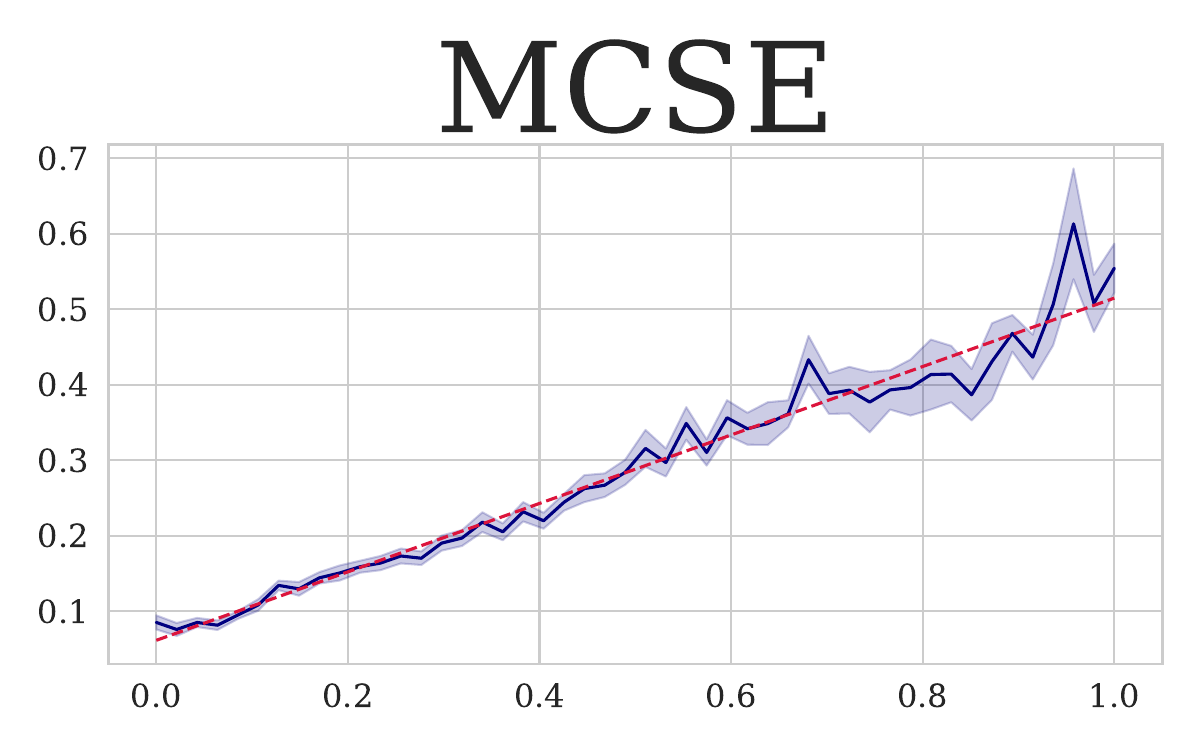}
    \end{subfigure}
    \begin{subfigure}{0.15\textwidth}
        \includegraphics[width=\linewidth]{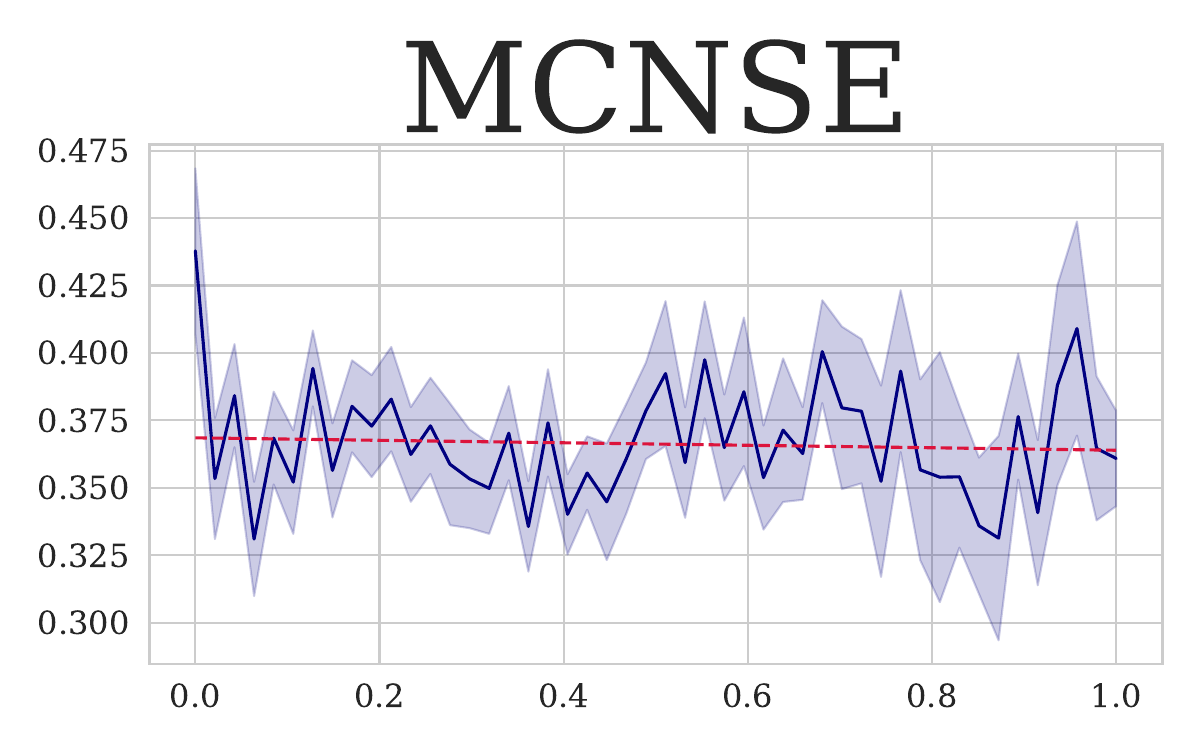}
    \end{subfigure}
    \begin{subfigure}{0.15\textwidth}
        \includegraphics[width=\linewidth]{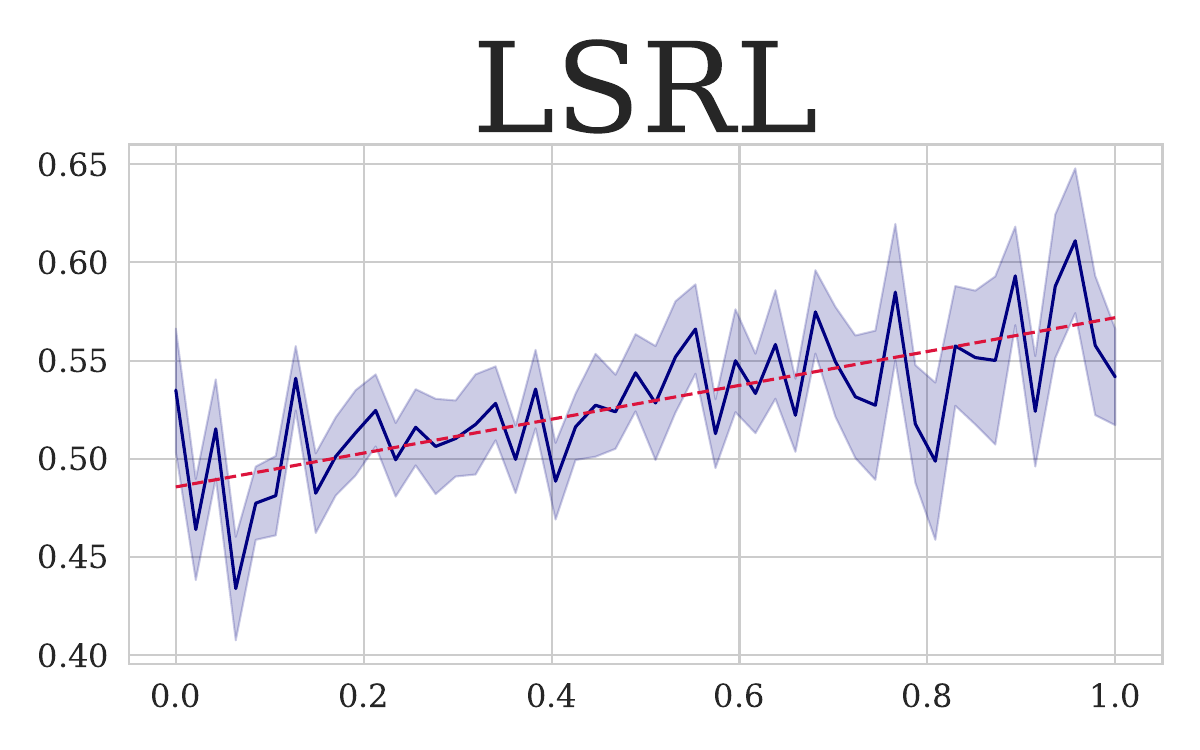}
    \end{subfigure}
    \vspace{0.3em}

 {\centering \textbf{\small XSUM} \par}
    \vspace{0.2em}
    \begin{subfigure}{0.15\textwidth}
        \includegraphics[width=\linewidth]{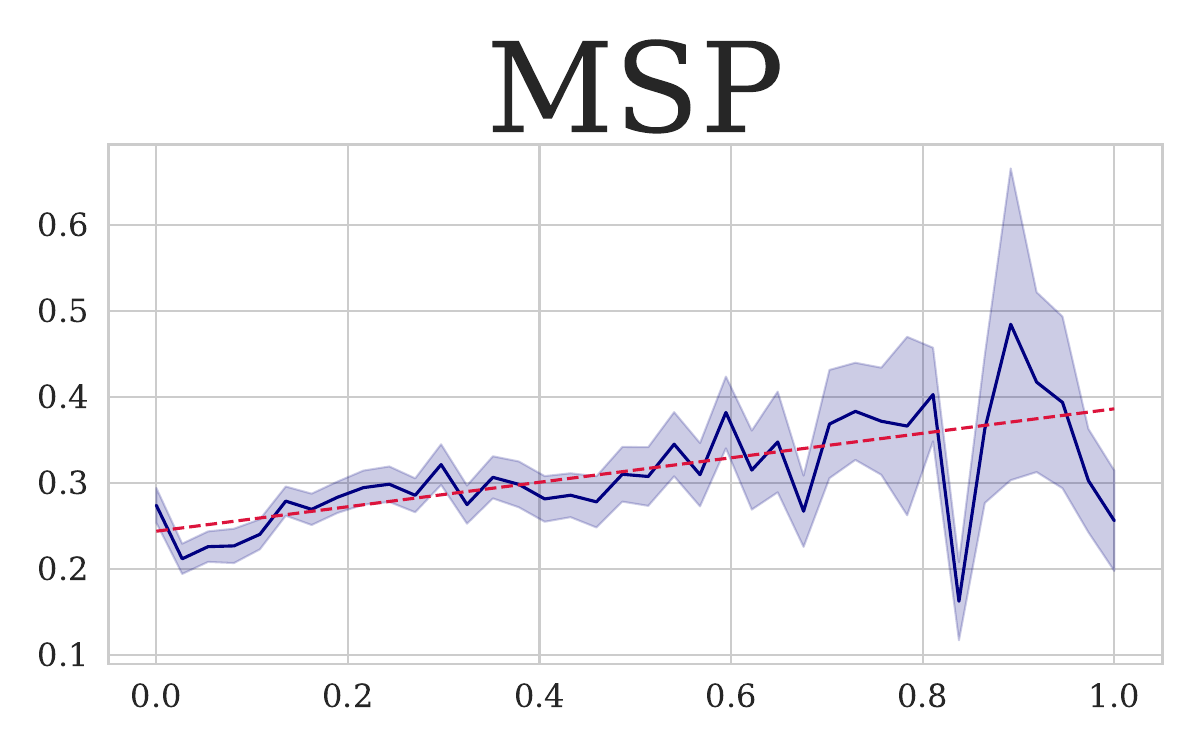}
    \end{subfigure}
    \begin{subfigure}{0.15\textwidth}
        \includegraphics[width=\linewidth]{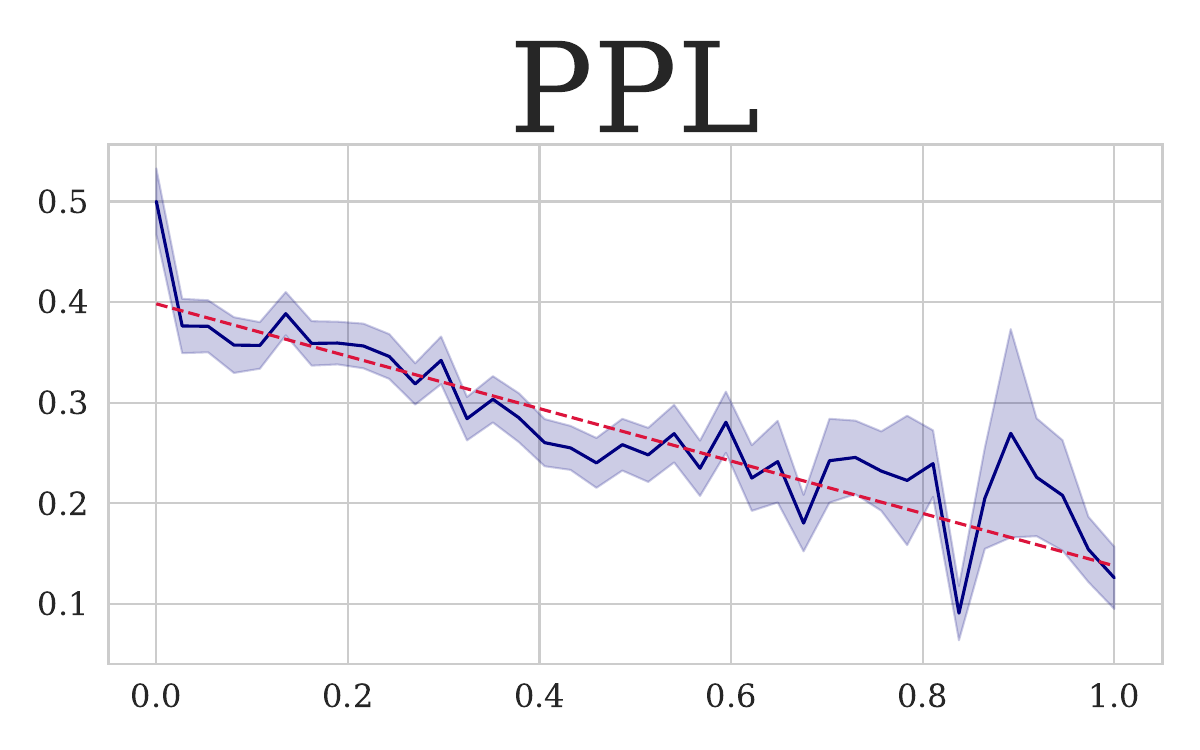}
    \end{subfigure}
    \begin{subfigure}{0.15\textwidth}
        \includegraphics[width=\linewidth]{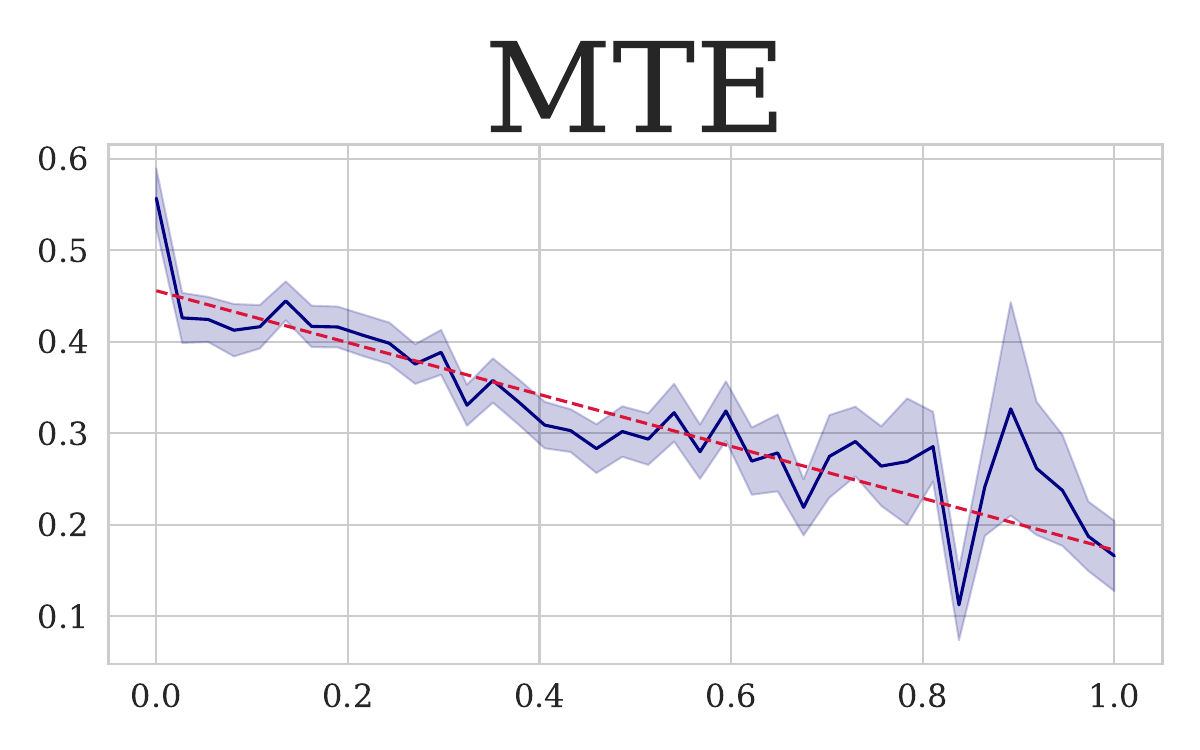}
    \end{subfigure}
    \begin{subfigure}{0.15\textwidth}
        \includegraphics[width=\linewidth]{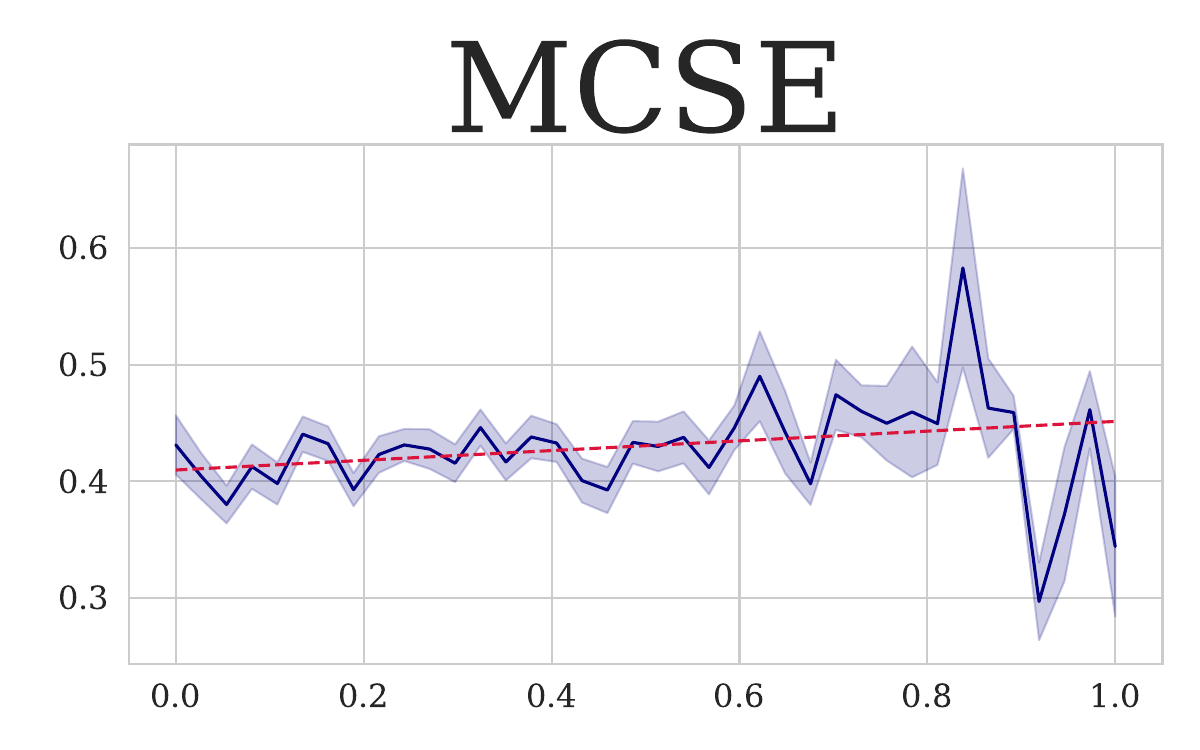}
    \end{subfigure}
    \begin{subfigure}{0.15\textwidth}
        \includegraphics[width=\linewidth]{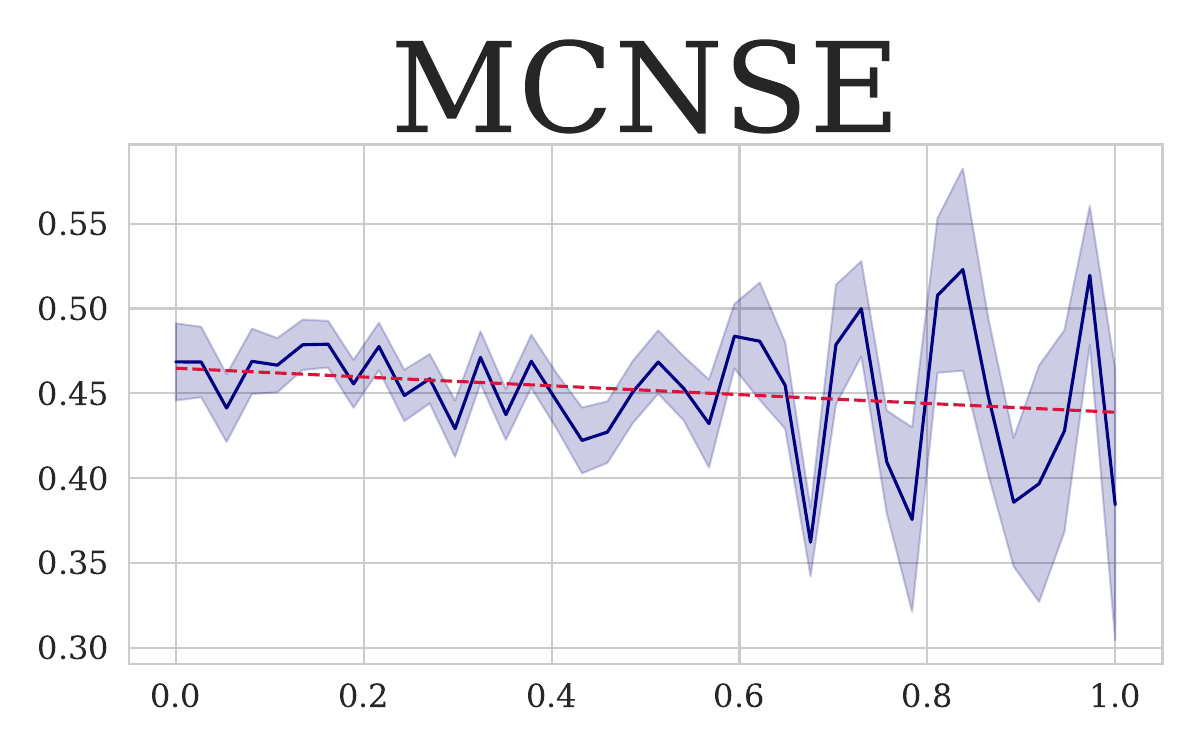}
    \end{subfigure}
    \begin{subfigure}{0.15\textwidth}
        \includegraphics[width=\linewidth]{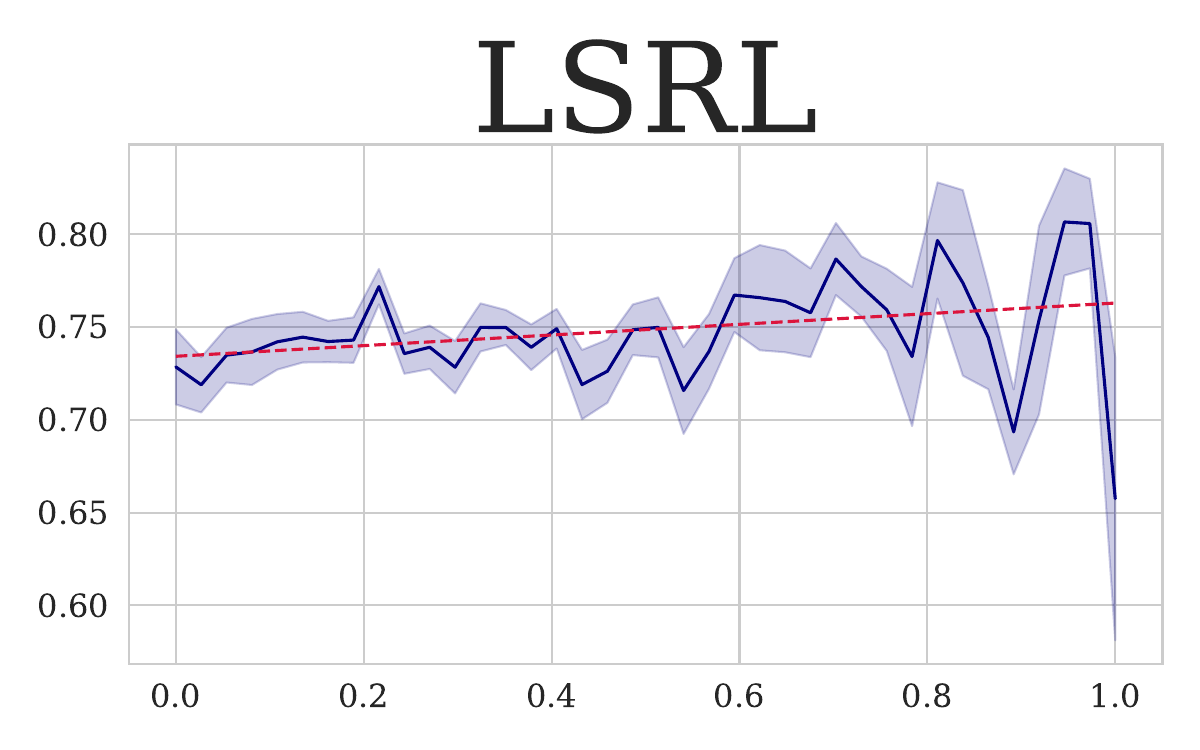}
    \end{subfigure}
    \vspace{0.3em}

     {\centering \textbf{\small GSM8K} \par}
    \vspace{0.2em}
    \begin{subfigure}{0.15\textwidth}
        \includegraphics[width=\linewidth]{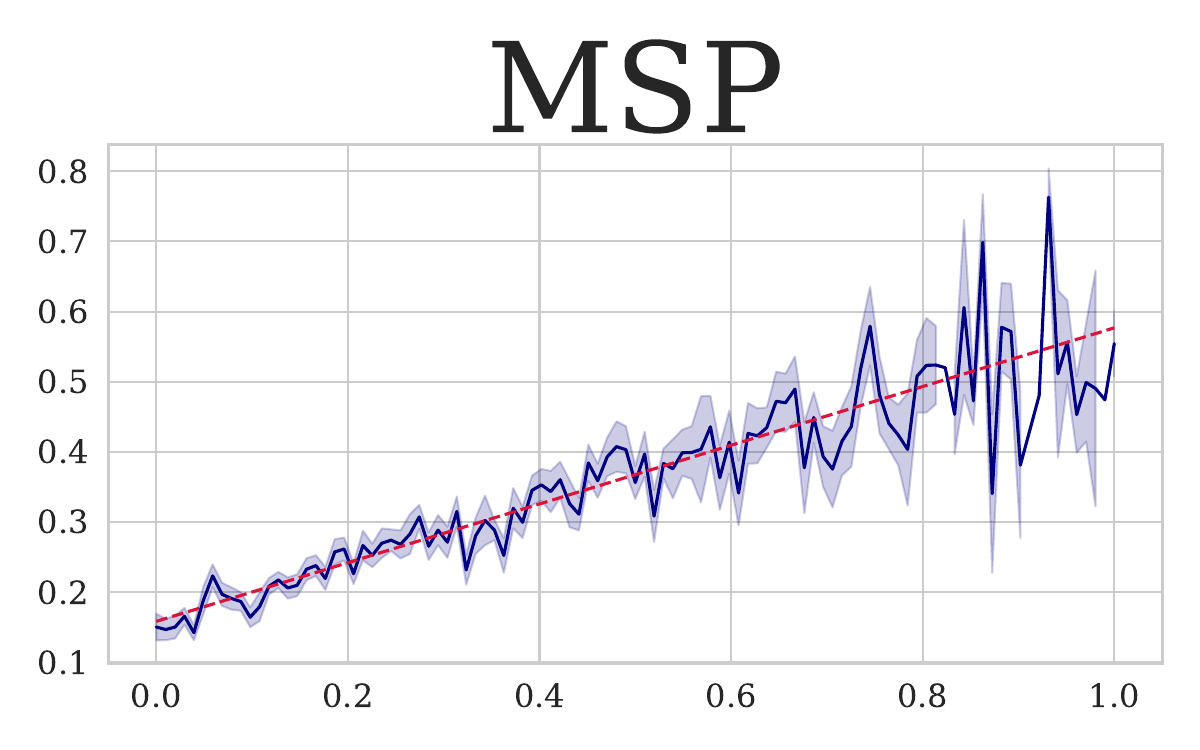}
    \end{subfigure}
    \begin{subfigure}{0.15\textwidth}
        \includegraphics[width=\linewidth]{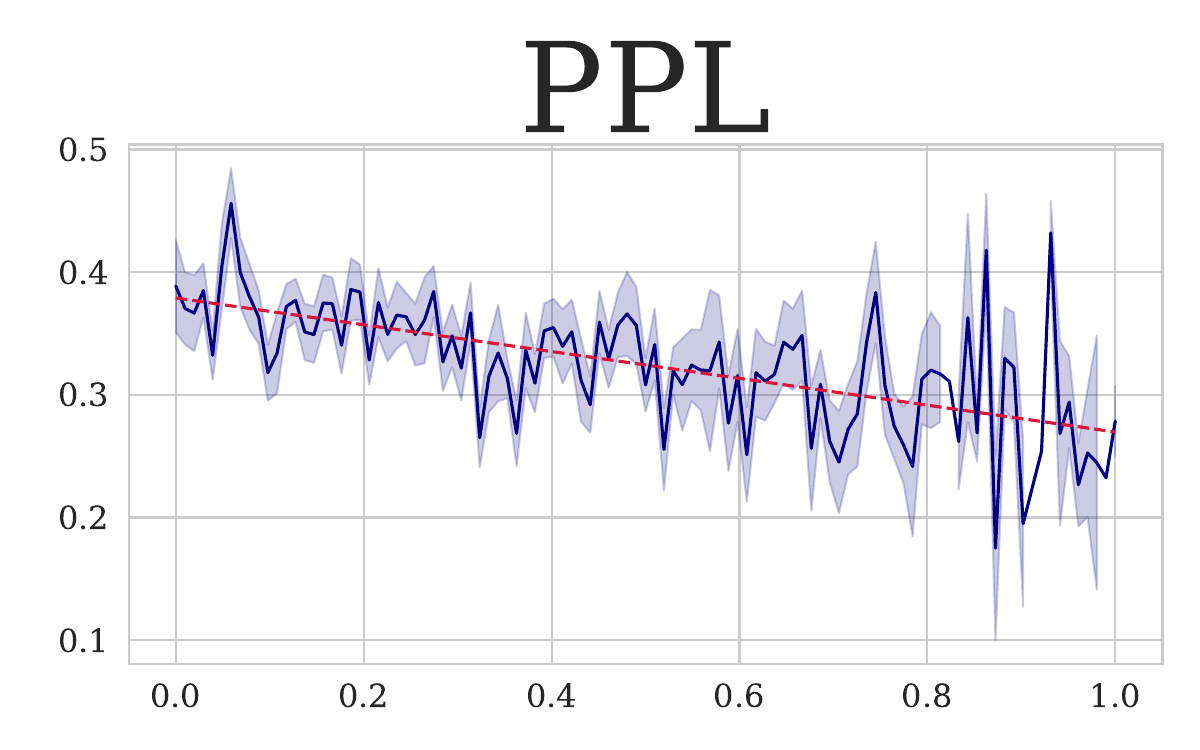}
    \end{subfigure}
    \begin{subfigure}{0.15\textwidth}
        \includegraphics[width=\linewidth]{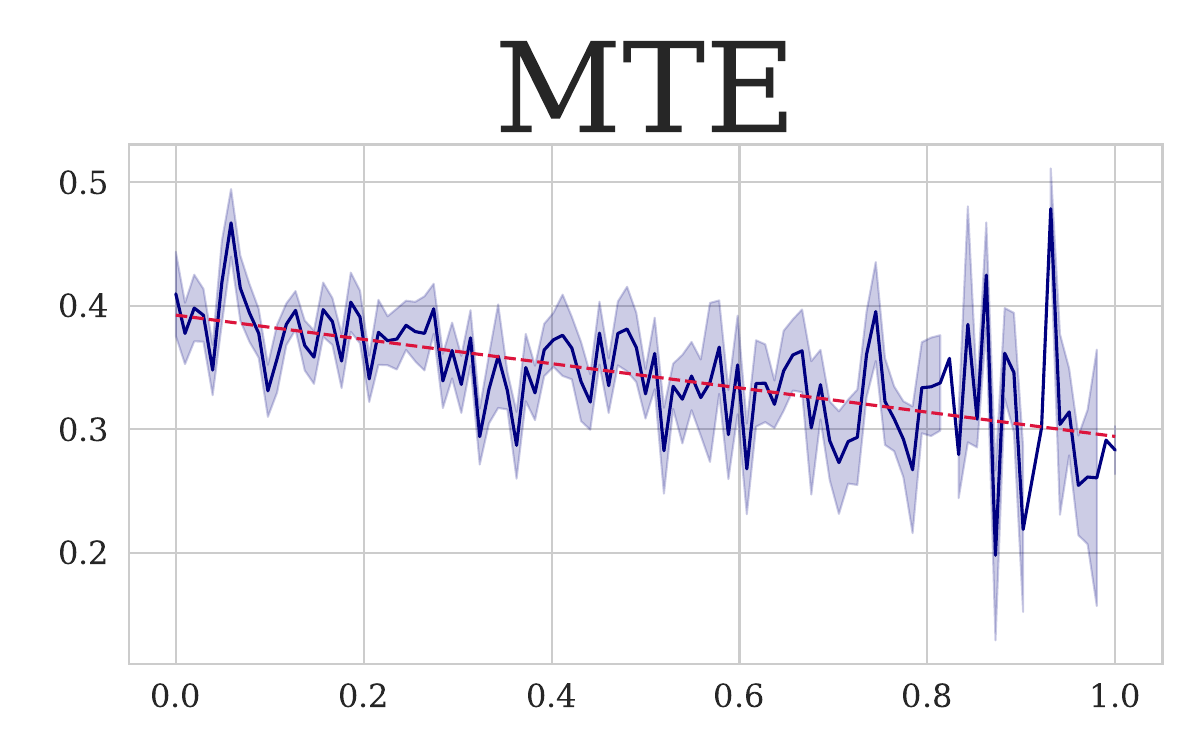}
    \end{subfigure}
    \begin{subfigure}{0.15\textwidth}
        \includegraphics[width=\linewidth]{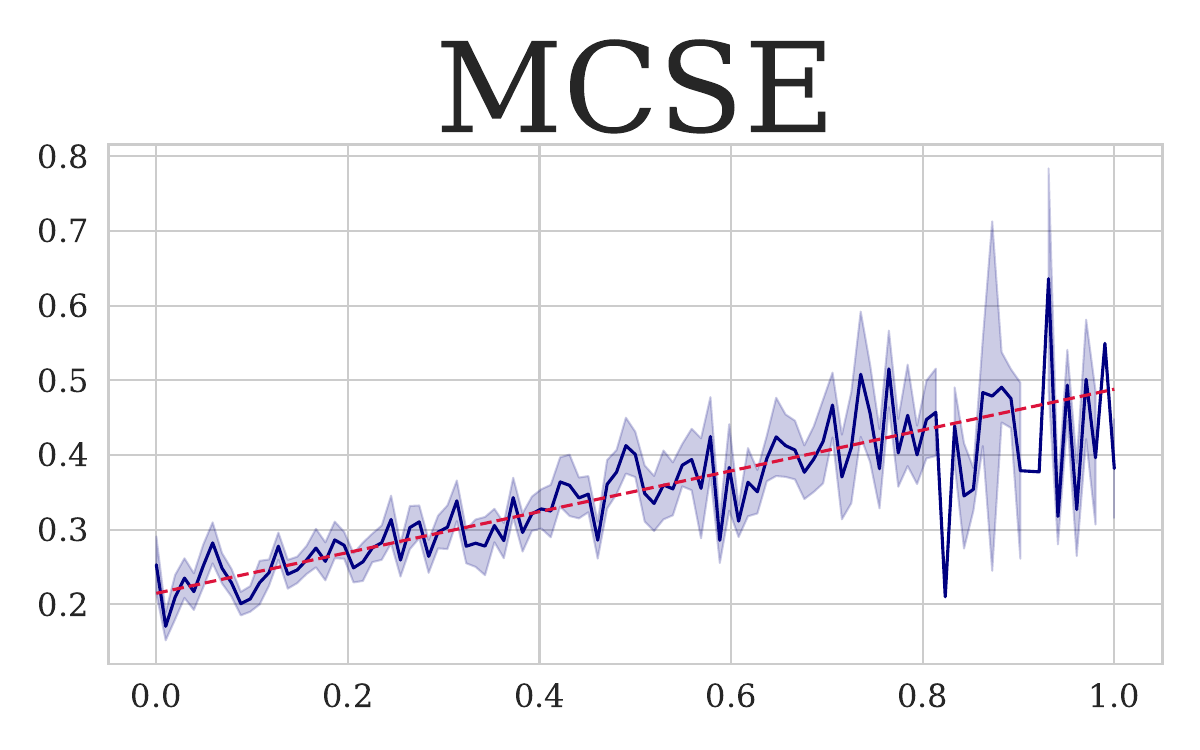}
    \end{subfigure}
    \begin{subfigure}{0.15\textwidth}
        \includegraphics[width=\linewidth]{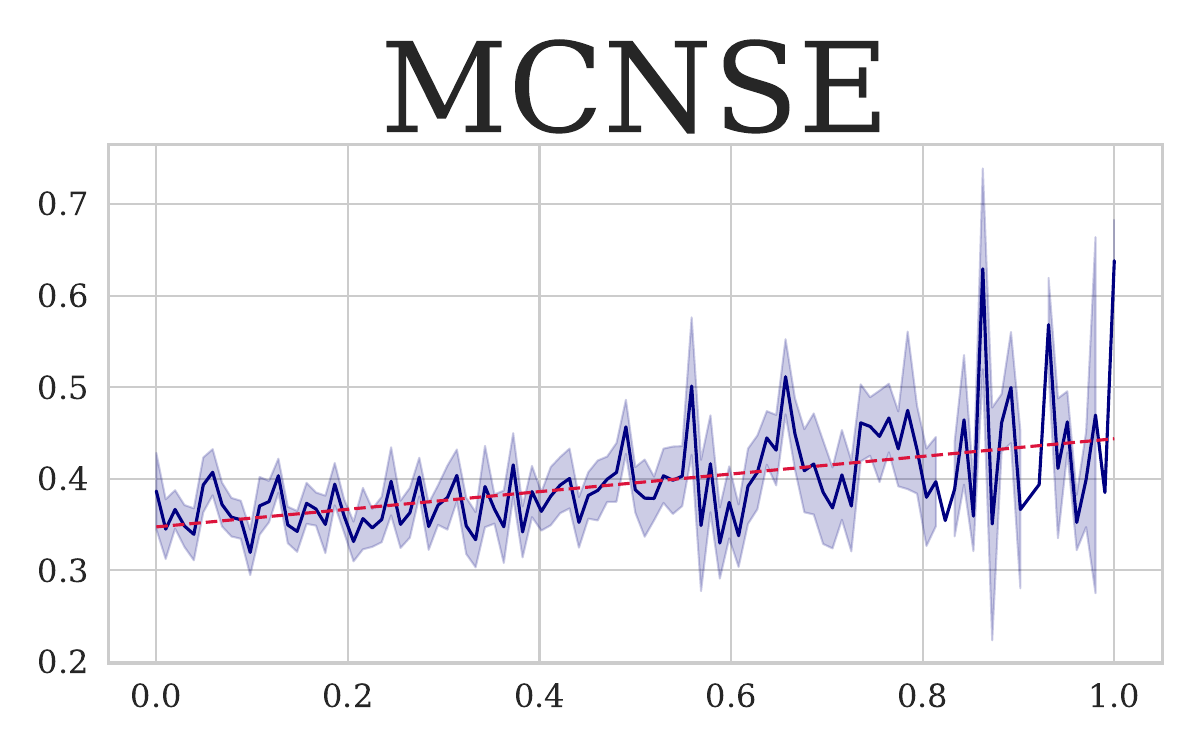}
    \end{subfigure}
    \begin{subfigure}{0.15\textwidth}
        \includegraphics[width=\linewidth]{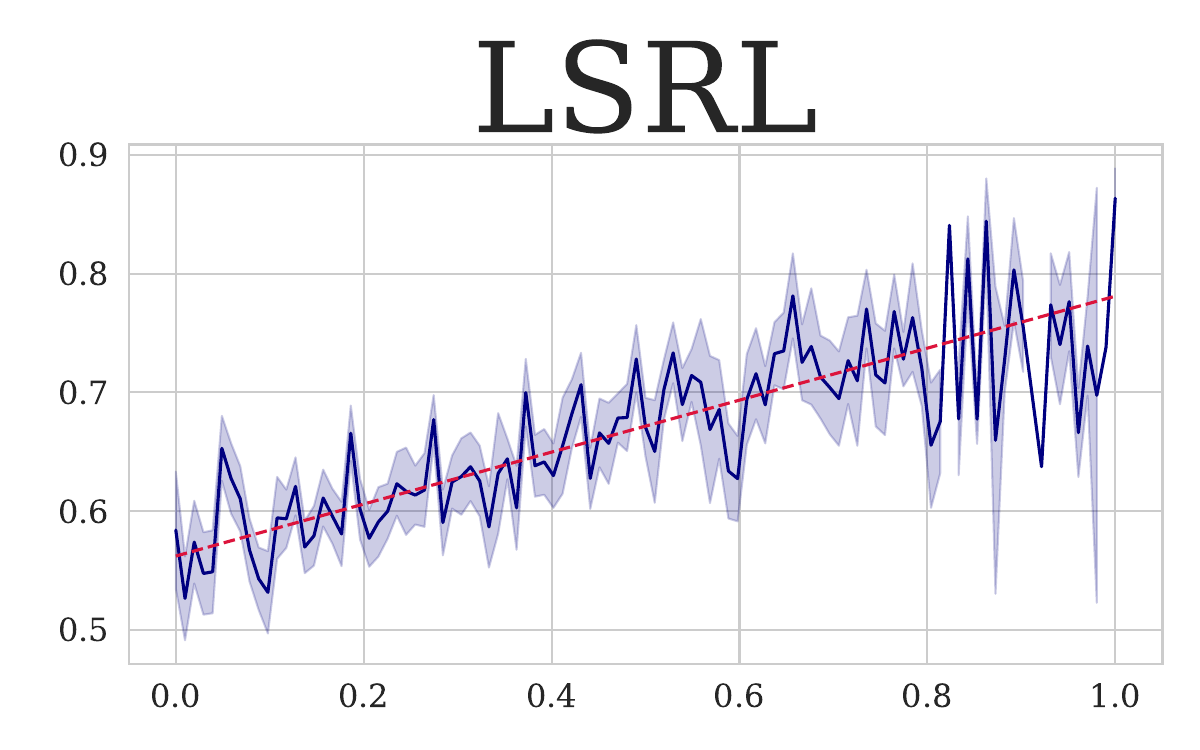}
    \end{subfigure}
    \vspace{0.3em}
    
        \caption{Uncertainty metric trends for model \textbf{LLAMA} across all datasets.}
    \label{fig:ue_metrics_llama}
\end{figure*}

\begin{figure*}[h!]
    \centering
    \vspace{-0.5em}
    {\centering \textbf{\small WMT14 De-En} \par}
    \vspace{0.2em}
    \begin{subfigure}{0.15\textwidth}
        \includegraphics[width=\linewidth]{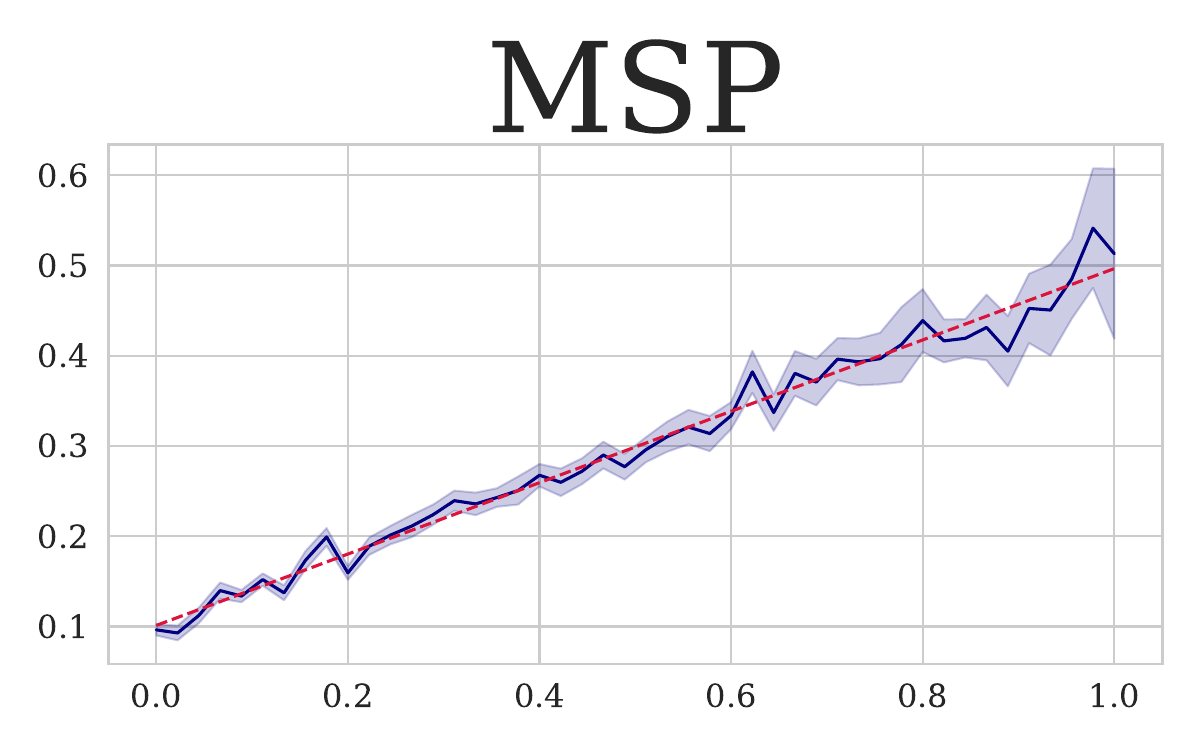}
    \end{subfigure}
    \begin{subfigure}{0.15\textwidth}
        \includegraphics[width=\linewidth]{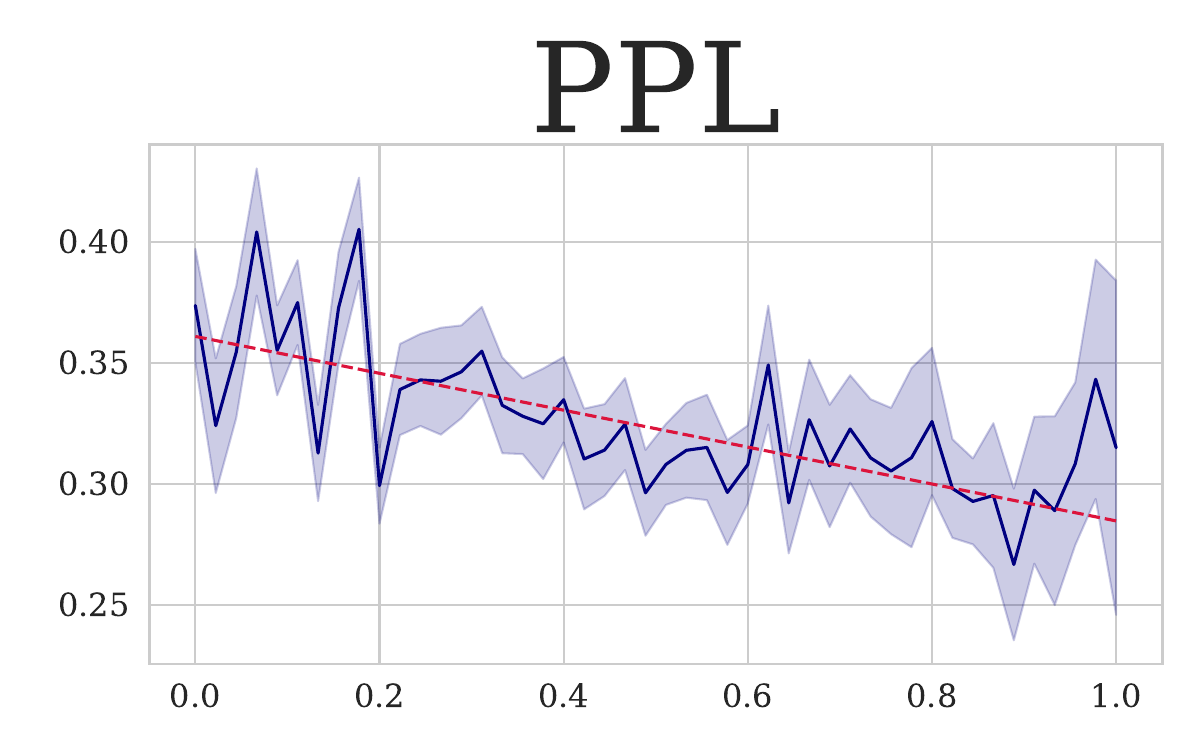}
    \end{subfigure}
    \begin{subfigure}{0.15\textwidth}
        \includegraphics[width=\linewidth]{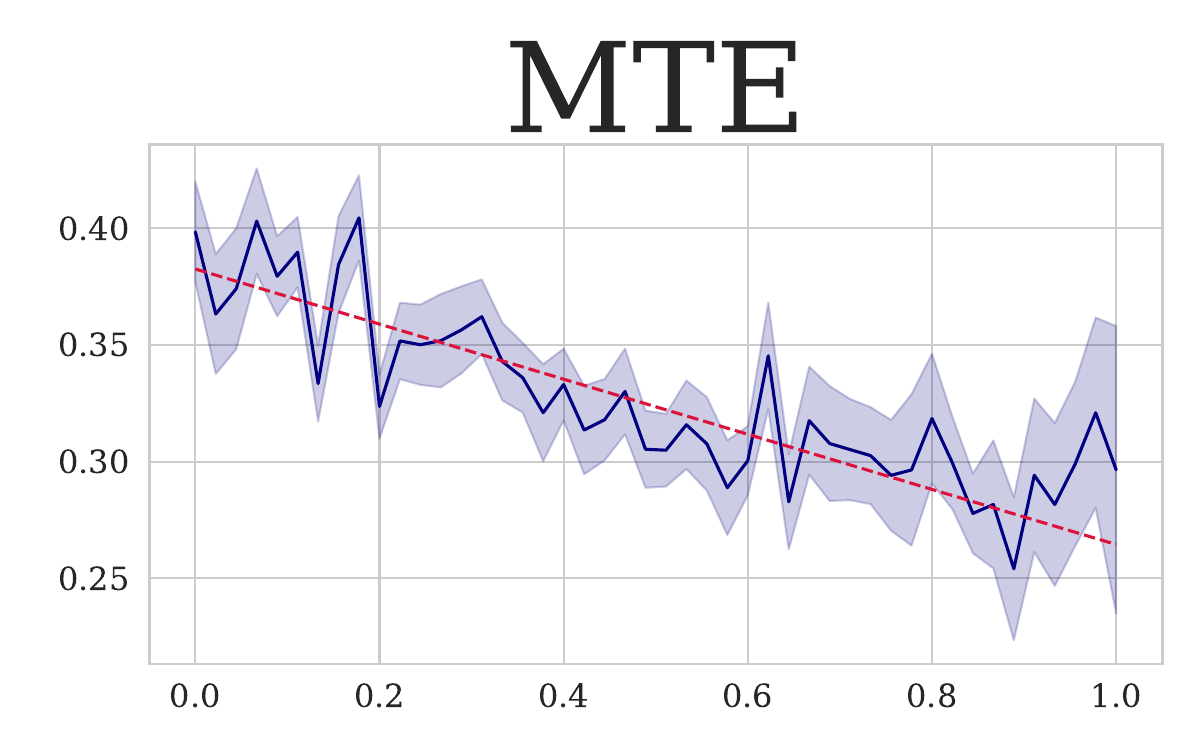}
    \end{subfigure}
    \begin{subfigure}{0.15\textwidth}
        \includegraphics[width=\linewidth]{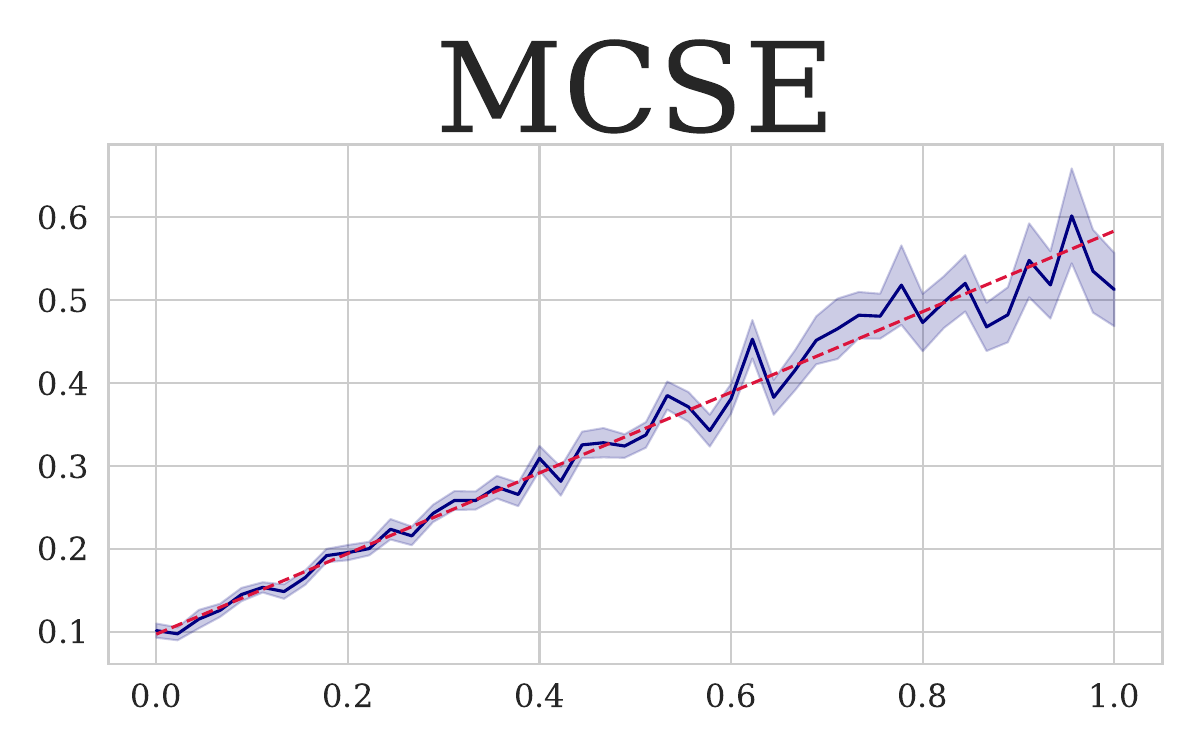}
    \end{subfigure}
    \begin{subfigure}{0.15\textwidth}
        \includegraphics[width=\linewidth]{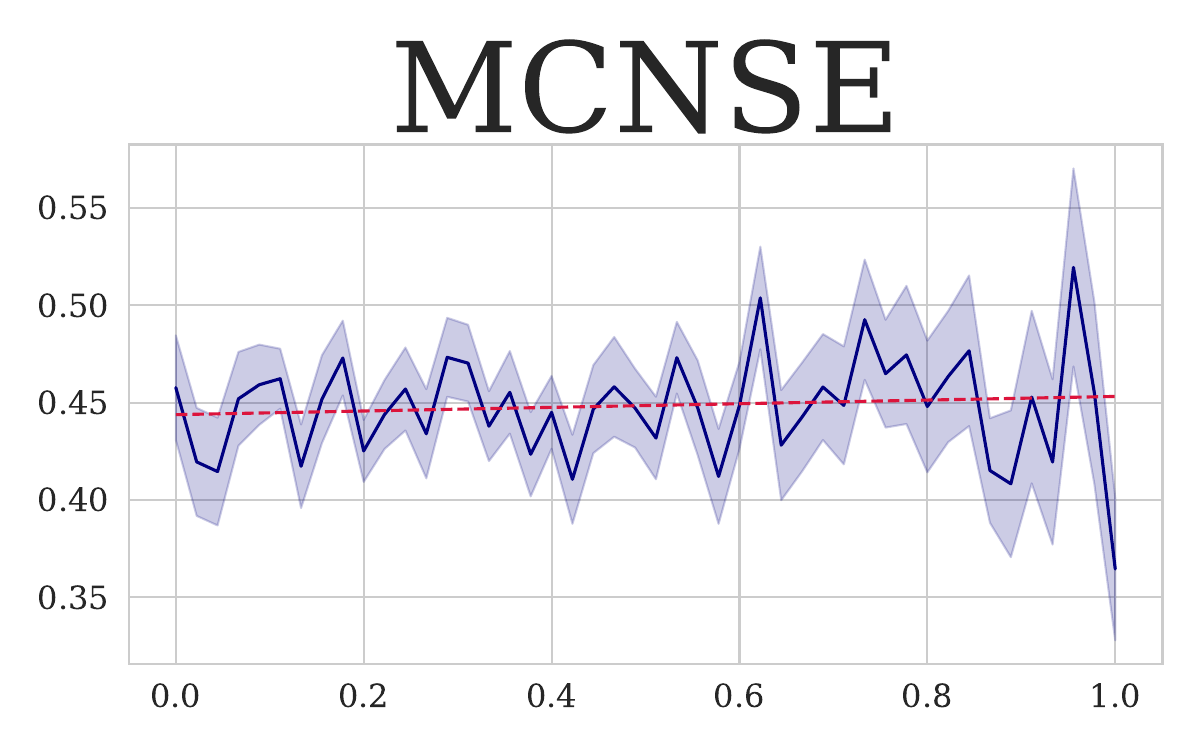}
    \end{subfigure}
    \begin{subfigure}{0.15\textwidth}
        \includegraphics[width=\linewidth]{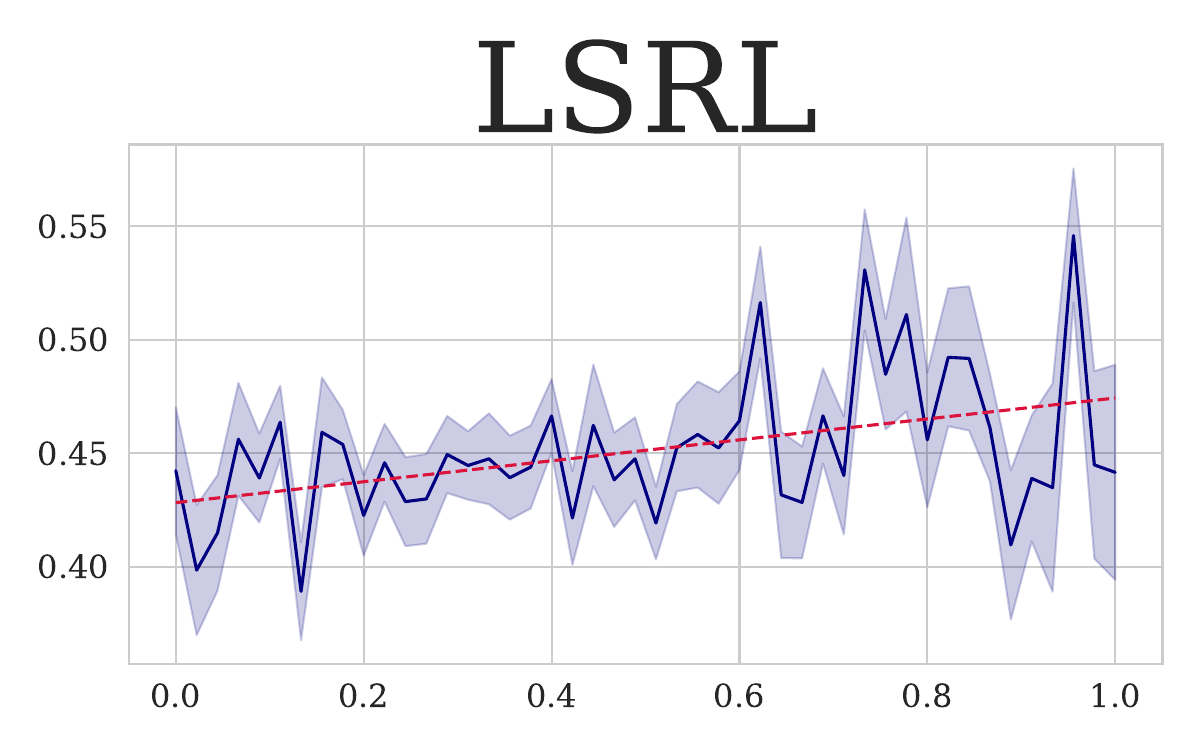}
    \end{subfigure}
    \vspace{0.3em}
    {\centering \textbf{\small WMT14 Fr-En} \par}
    \vspace{0.2em}
    \begin{subfigure}{0.15\textwidth}
        \includegraphics[width=\linewidth]{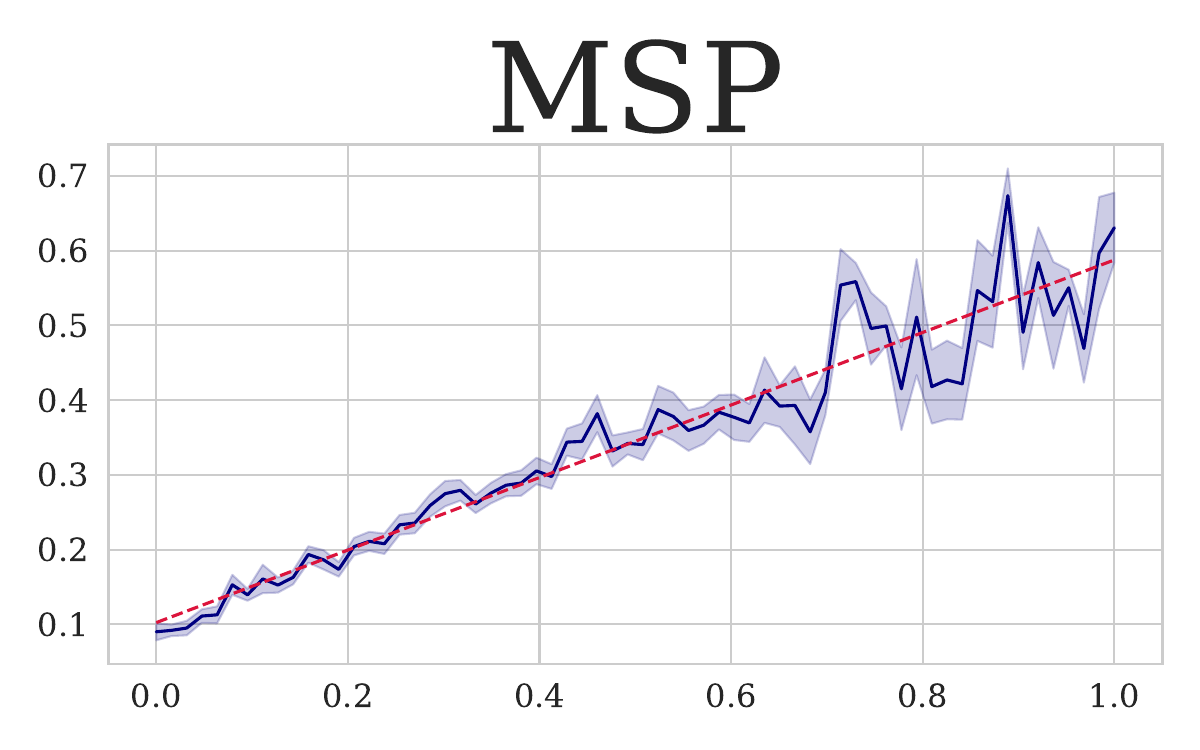}
    \end{subfigure}
    \begin{subfigure}{0.15\textwidth}
        \includegraphics[width=\linewidth]{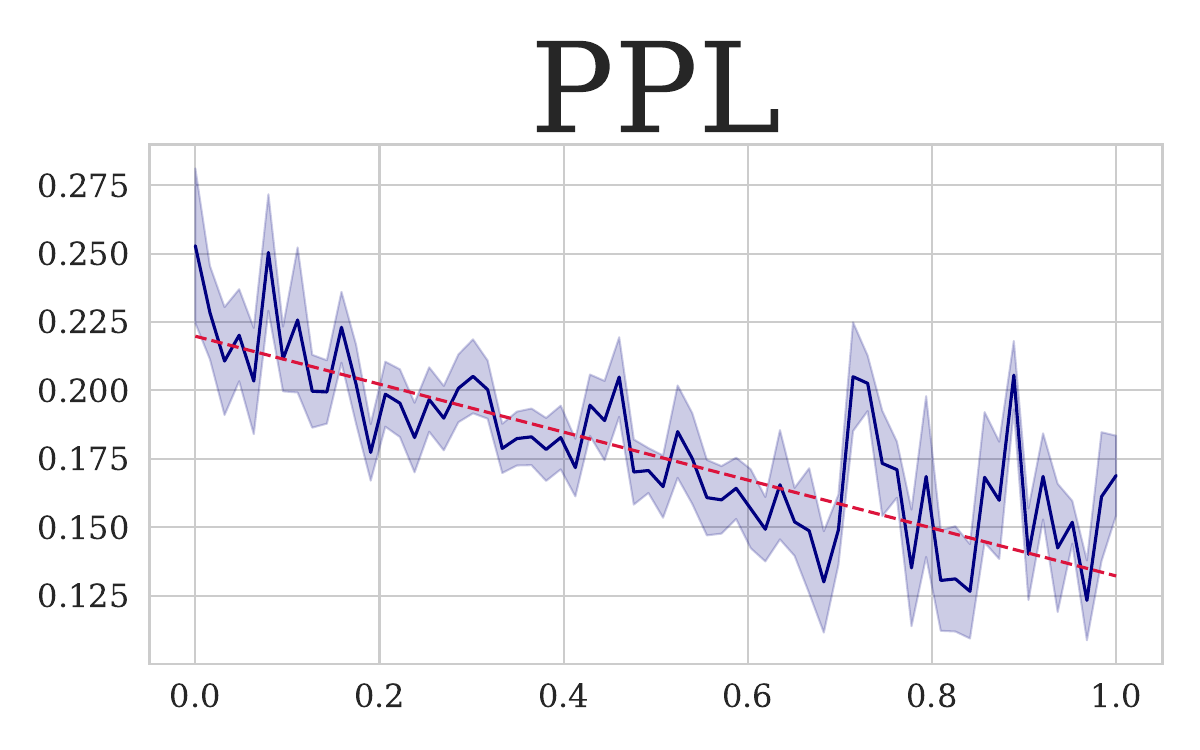}
    \end{subfigure}
    \begin{subfigure}{0.15\textwidth}
        \includegraphics[width=\linewidth]{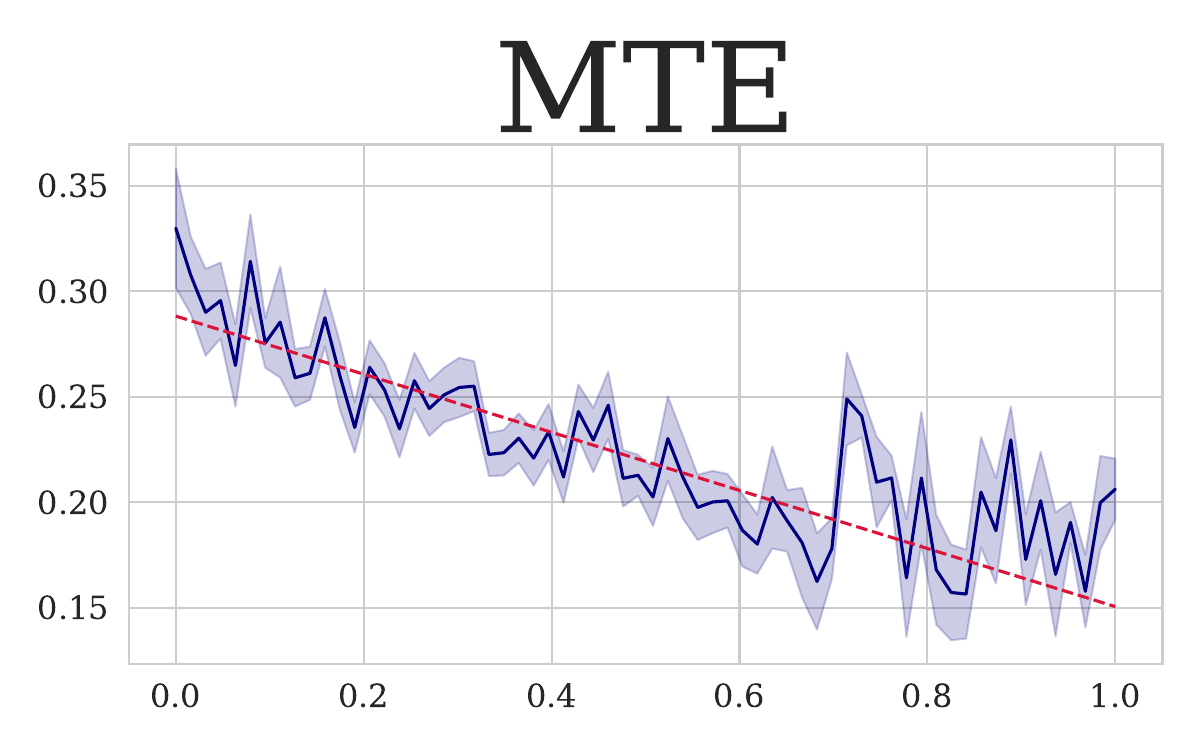}
    \end{subfigure}
    \begin{subfigure}{0.15\textwidth}
        \includegraphics[width=\linewidth]{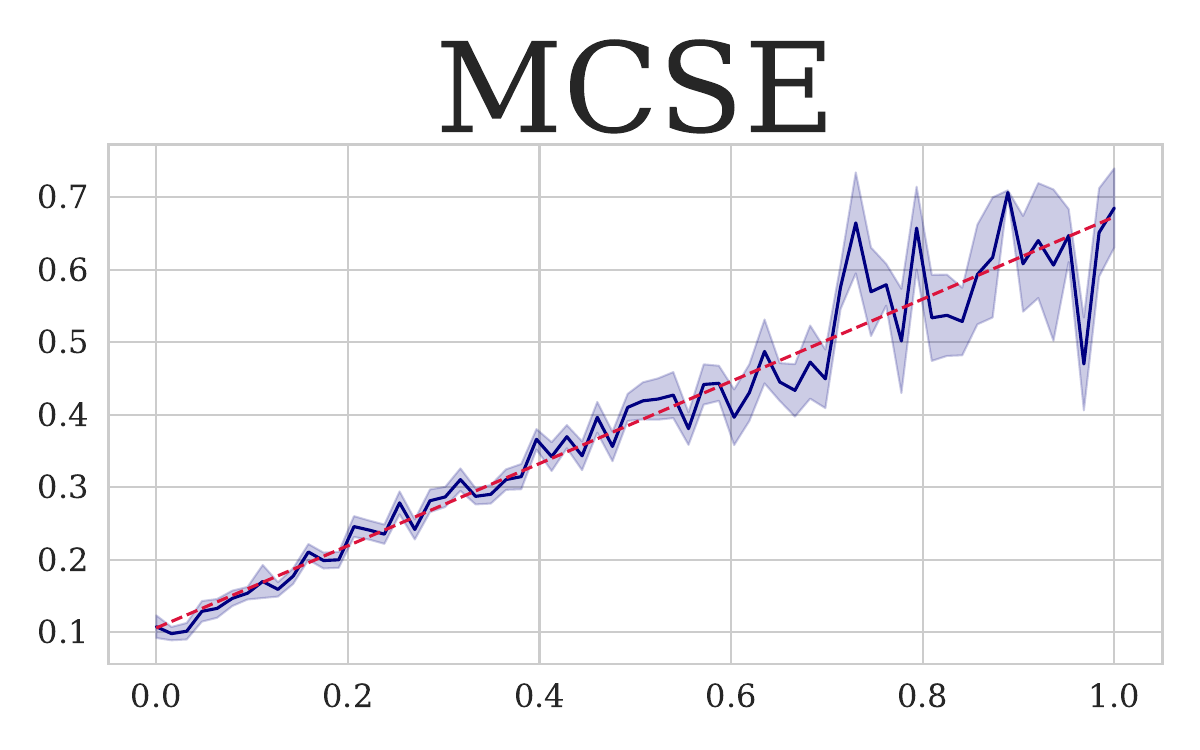}
    \end{subfigure}
    \begin{subfigure}{0.15\textwidth}
        \includegraphics[width=\linewidth]{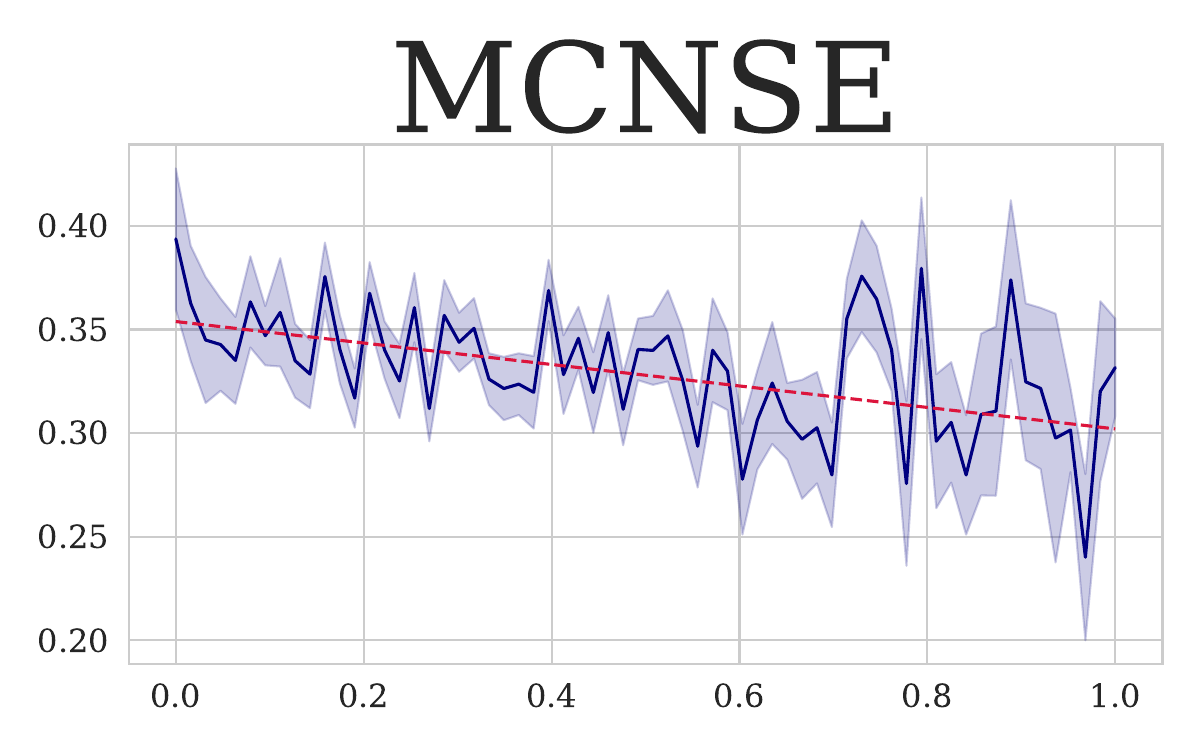}
    \end{subfigure}
    \begin{subfigure}{0.15\textwidth}
        \includegraphics[width=\linewidth]{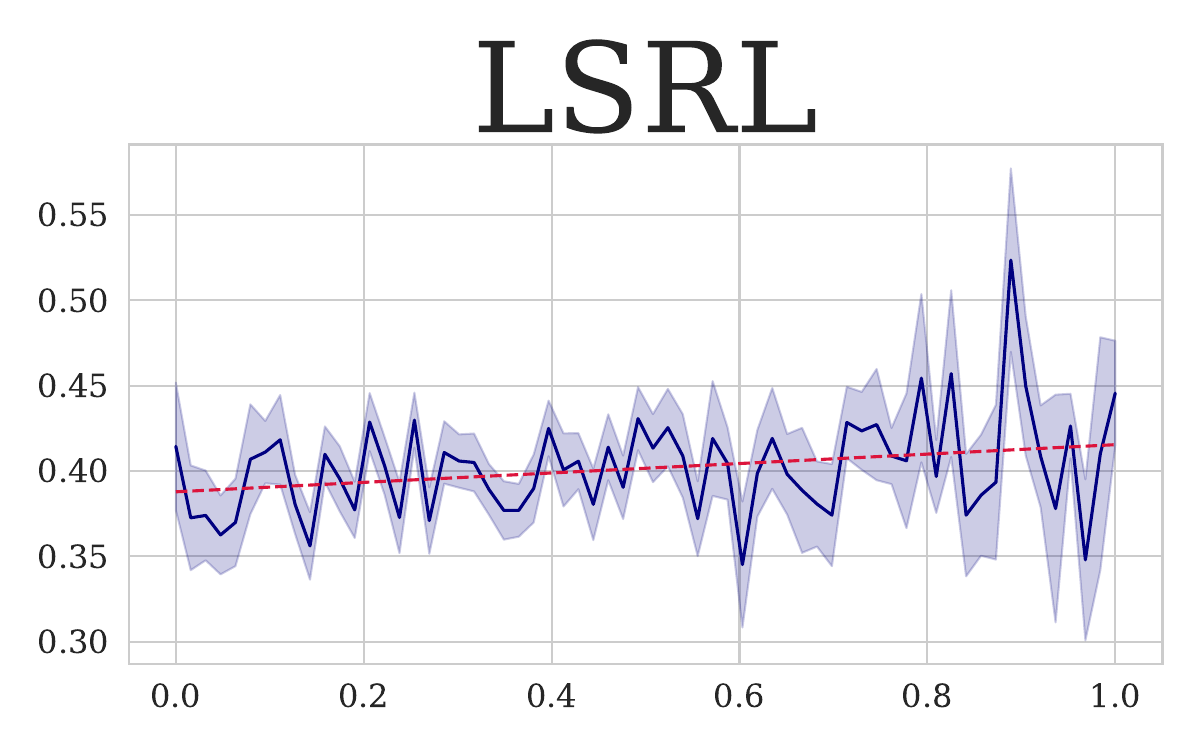}
    \end{subfigure}
    \vspace{0.3em}
    {\centering \textbf{\small WMT14 Cs-En} \par}
    \vspace{0.2em}
    \begin{subfigure}{0.15\textwidth}
        \includegraphics[width=\linewidth]{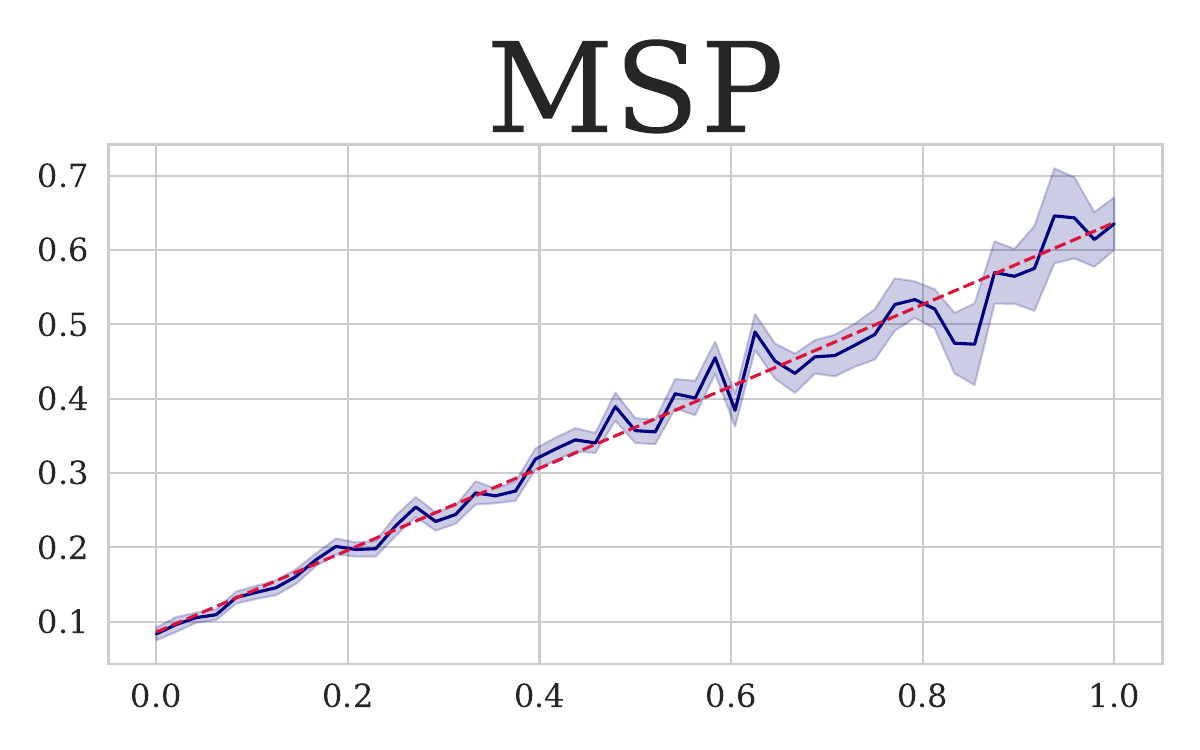}
    \end{subfigure}
    \begin{subfigure}{0.15\textwidth}
        \includegraphics[width=\linewidth]{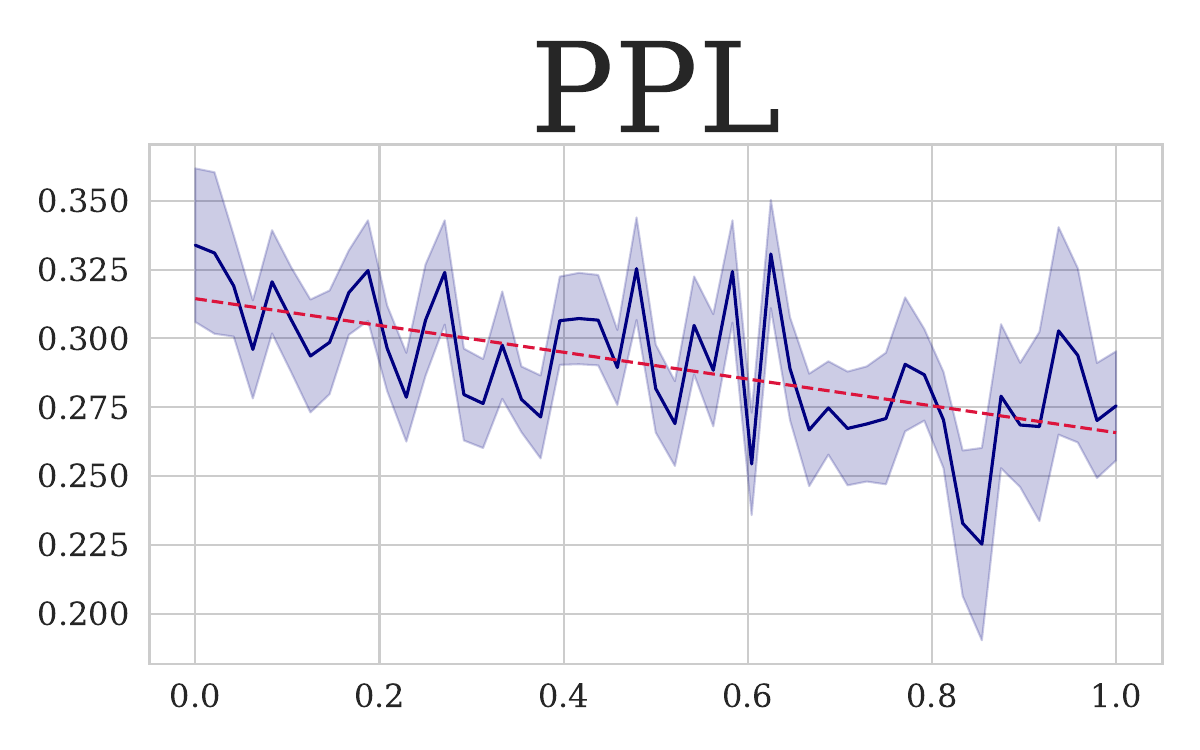}
    \end{subfigure}
    \begin{subfigure}{0.15\textwidth}
        \includegraphics[width=\linewidth]{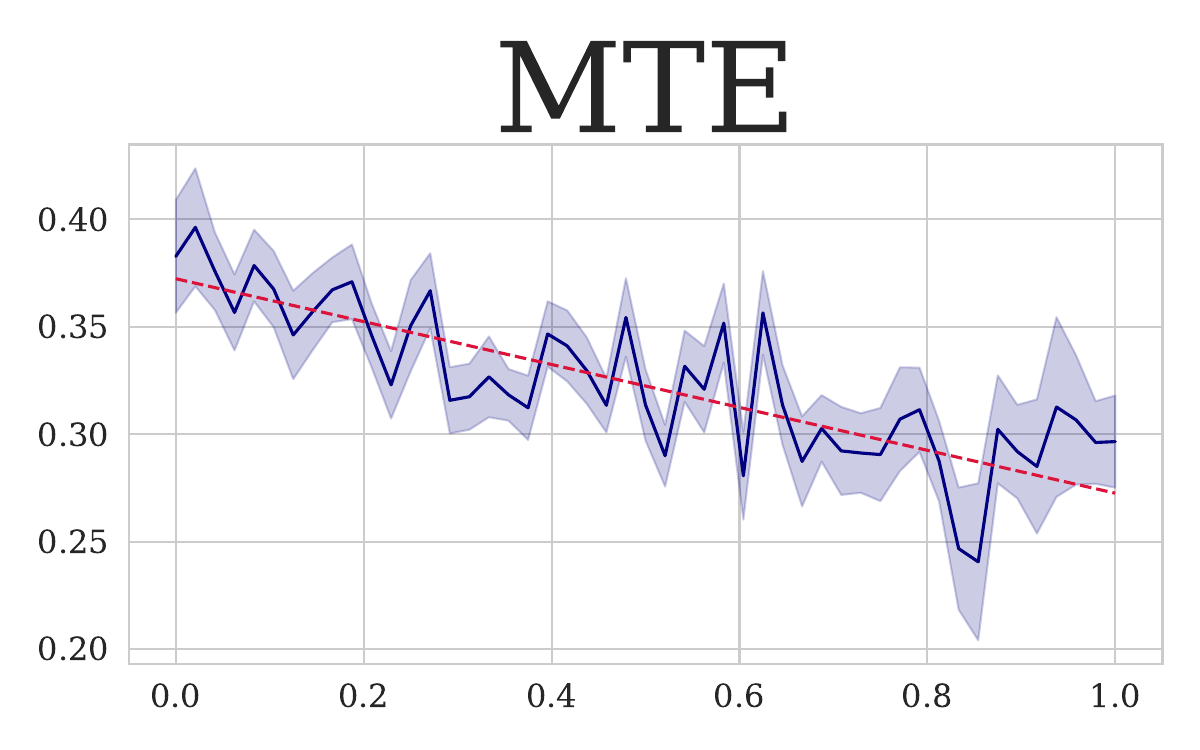}
    \end{subfigure}
    \begin{subfigure}{0.15\textwidth}
        \includegraphics[width=\linewidth]{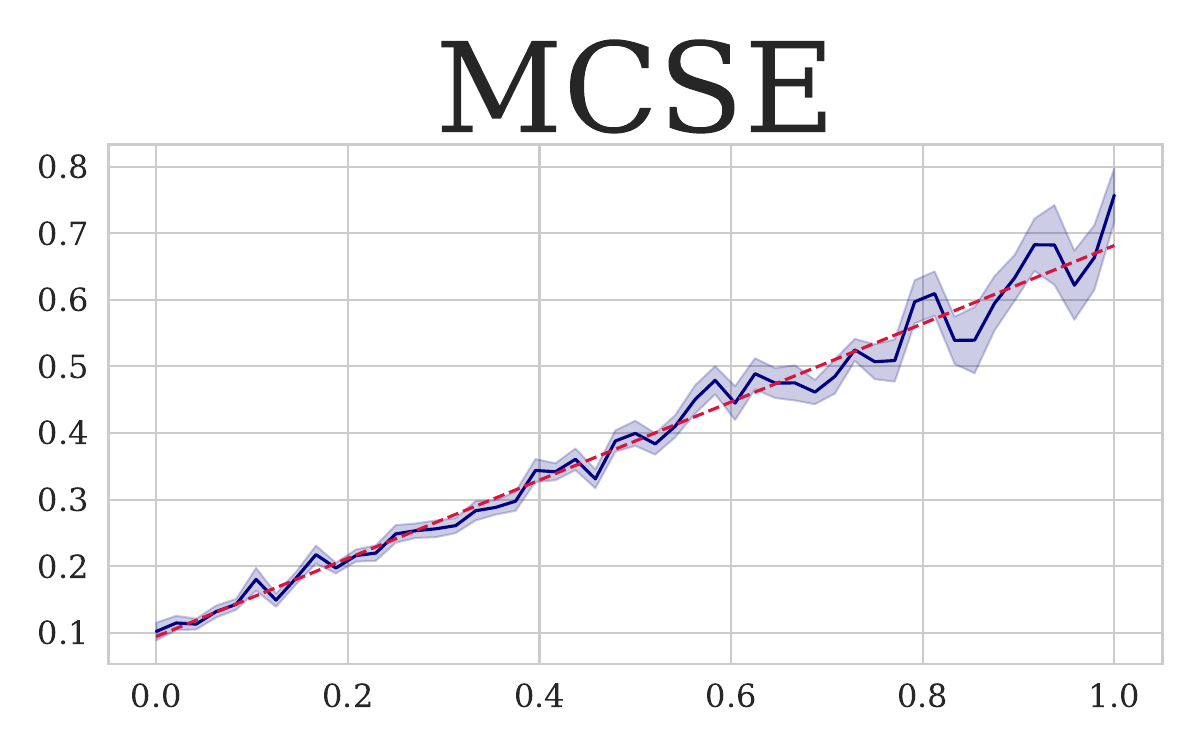}
    \end{subfigure}
    \begin{subfigure}{0.15\textwidth}
        \includegraphics[width=\linewidth]{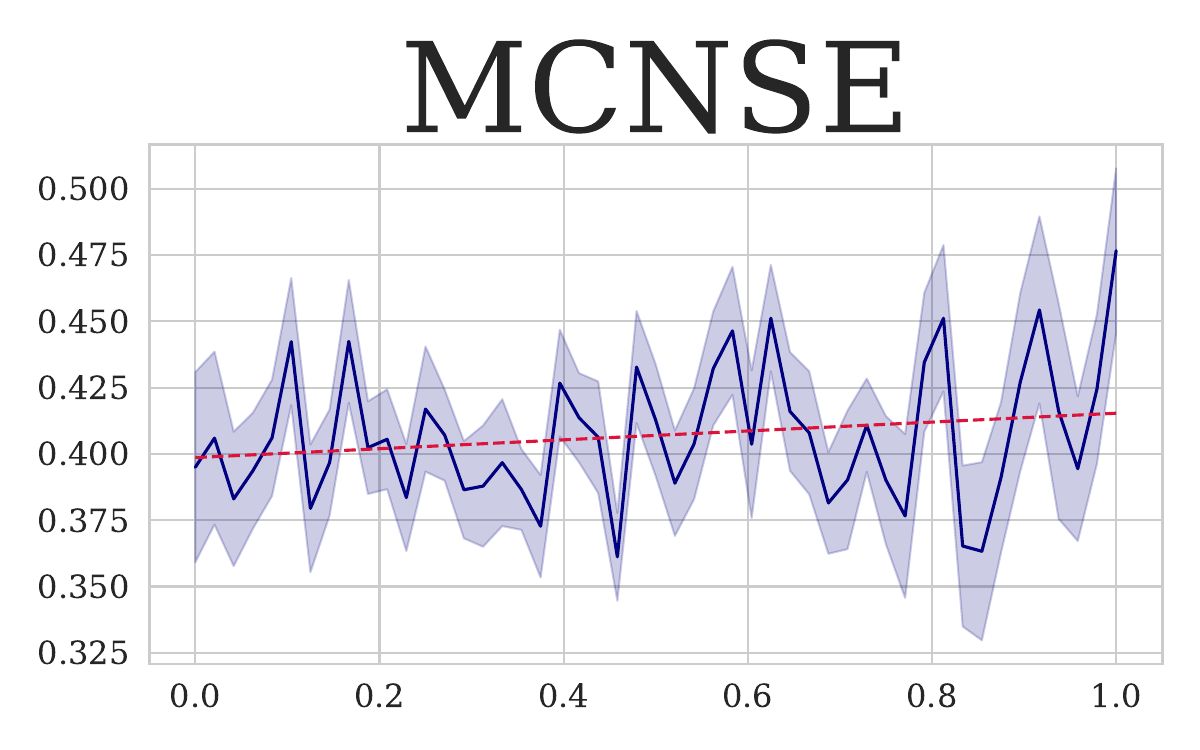}
    \end{subfigure}
    \begin{subfigure}{0.15\textwidth}
        \includegraphics[width=\linewidth]{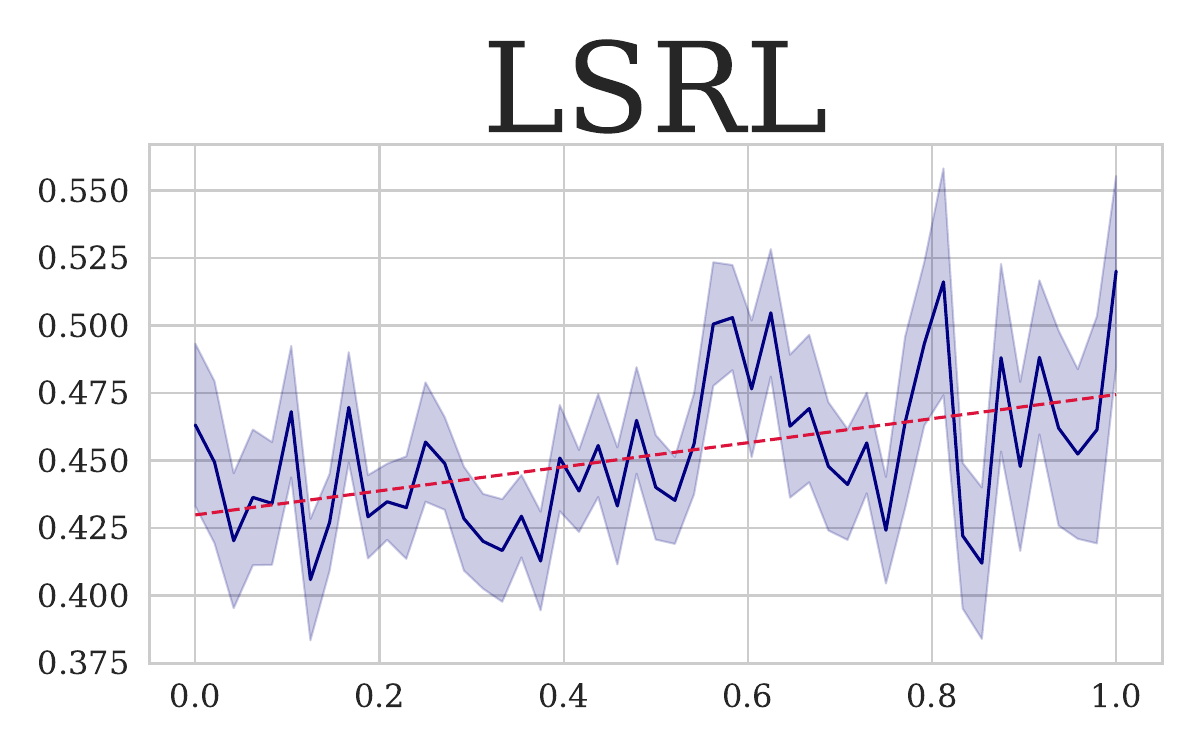}
    \end{subfigure}
    \vspace{0.3em}
    {\centering \textbf{\small WMT14 Ru-En} \par}
    \vspace{0.2em}
    \begin{subfigure}{0.15\textwidth}
        \includegraphics[width=\linewidth]{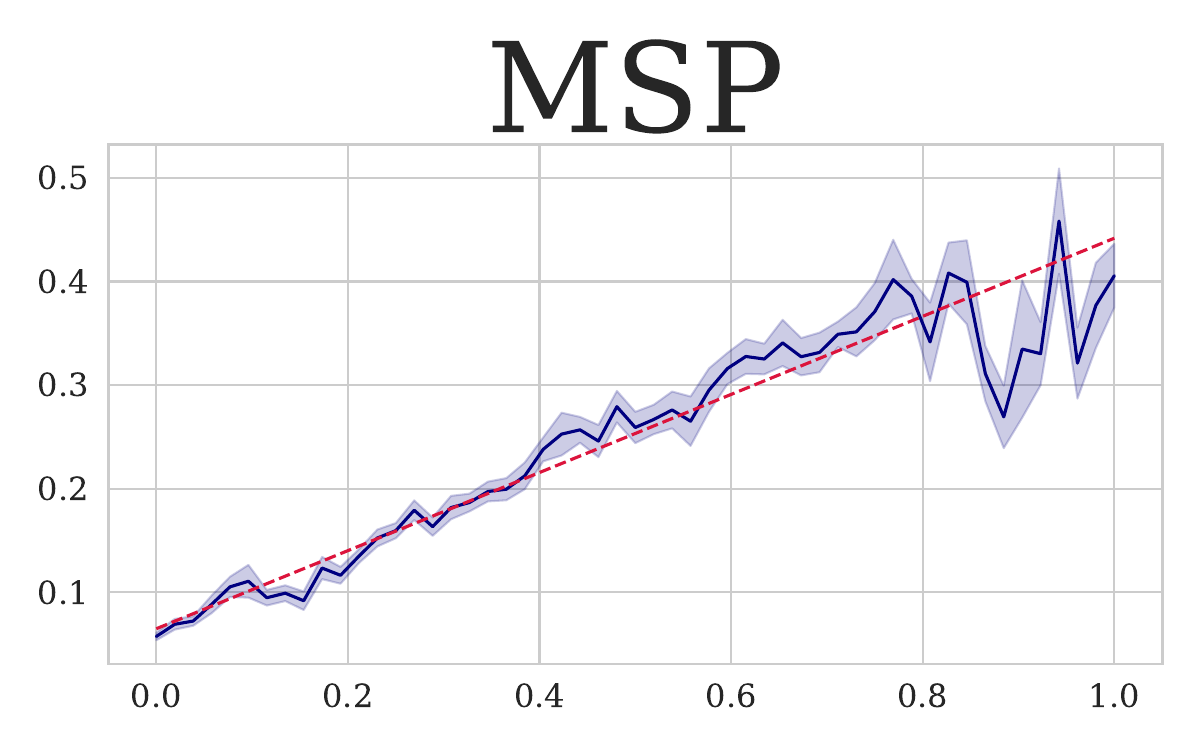}
    \end{subfigure}
    \begin{subfigure}{0.15\textwidth}
        \includegraphics[width=\linewidth]{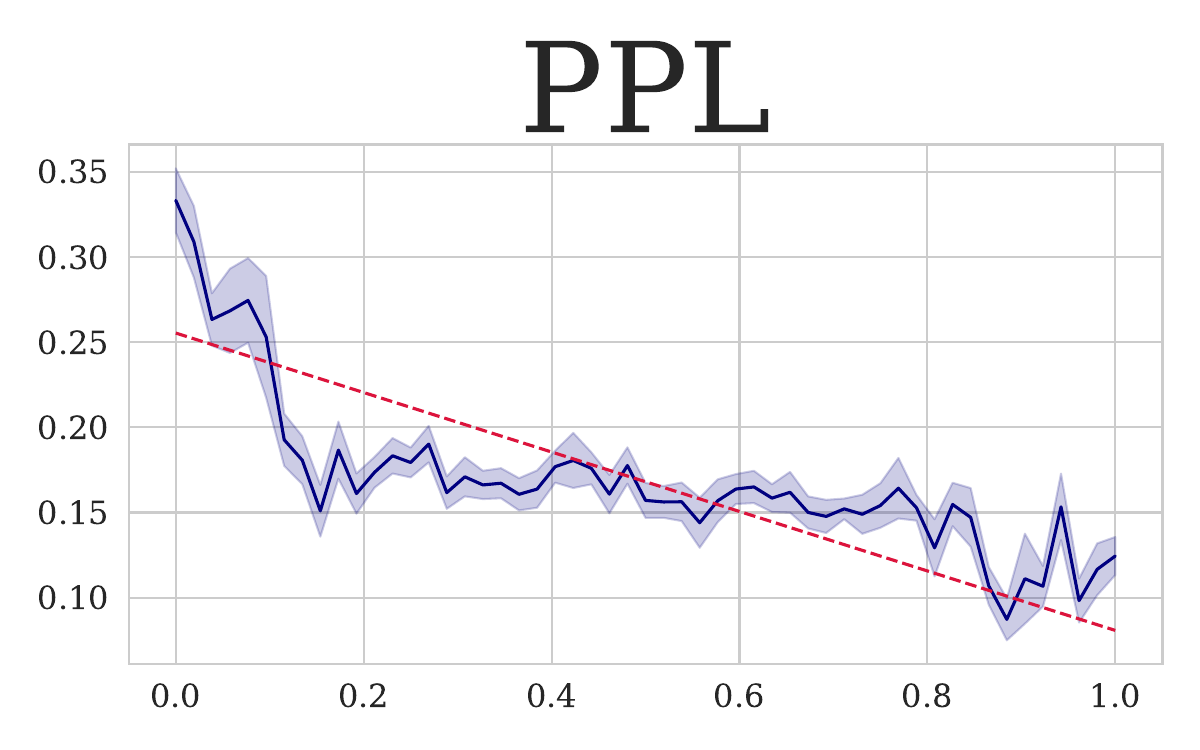}
    \end{subfigure}
    \begin{subfigure}{0.15\textwidth}
        \includegraphics[width=\linewidth]{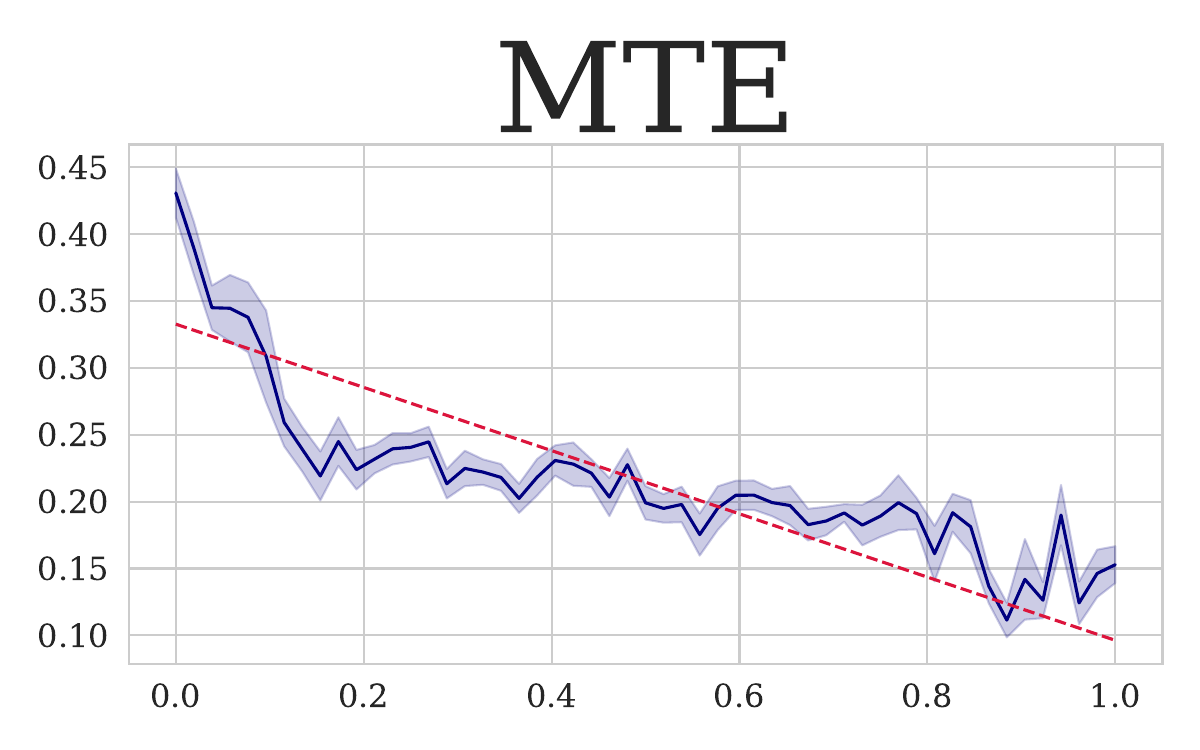}
    \end{subfigure}
    \begin{subfigure}{0.15\textwidth}
        \includegraphics[width=\linewidth]{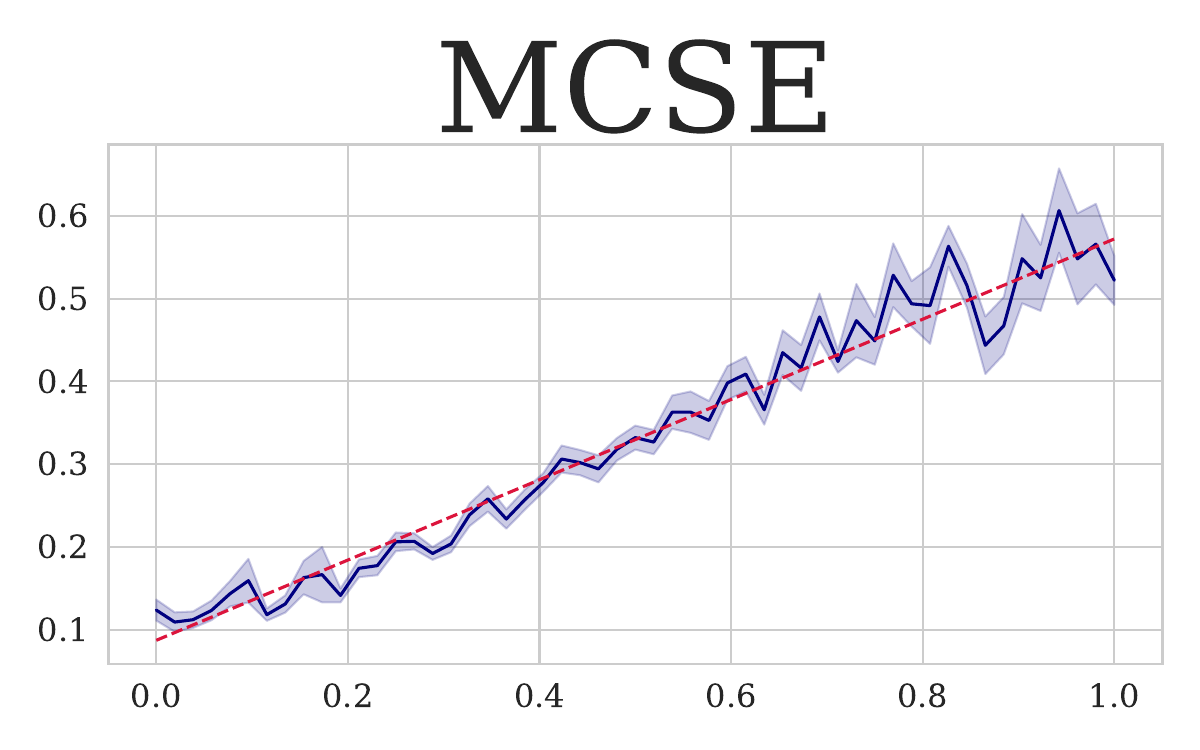}
    \end{subfigure}
    \begin{subfigure}{0.15\textwidth}
        \includegraphics[width=\linewidth]{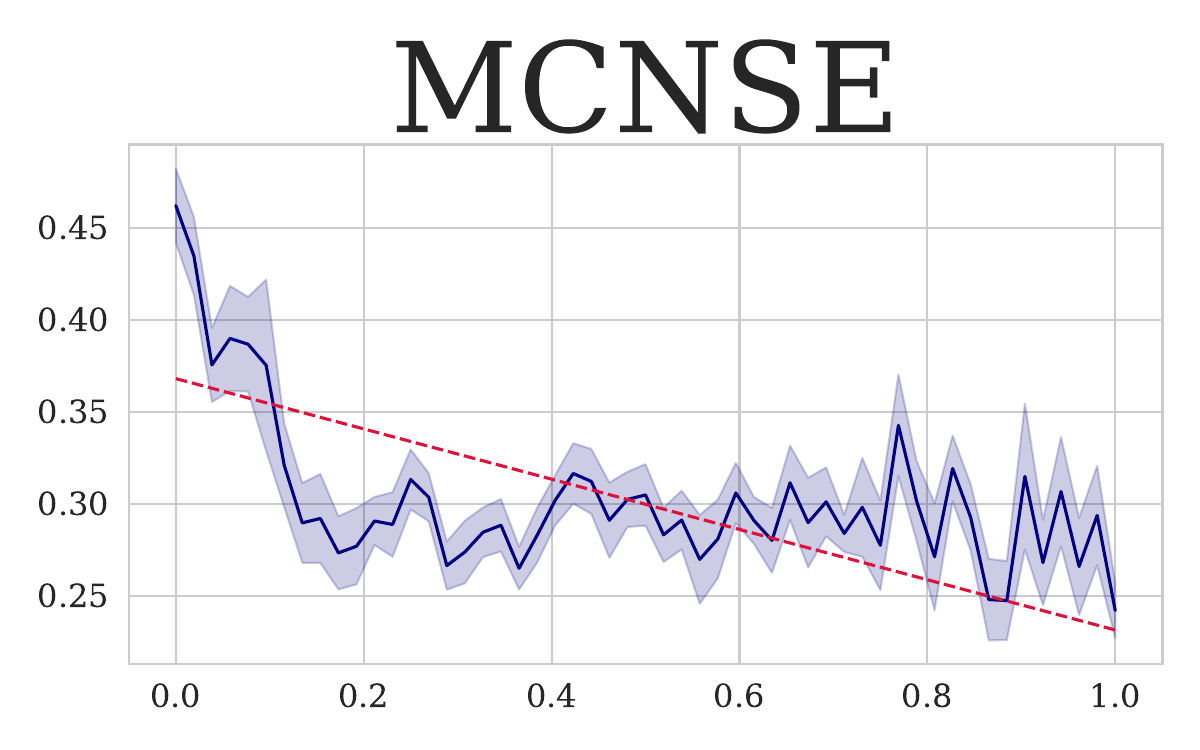}
    \end{subfigure}
    \begin{subfigure}{0.15\textwidth}
        \includegraphics[width=\linewidth]{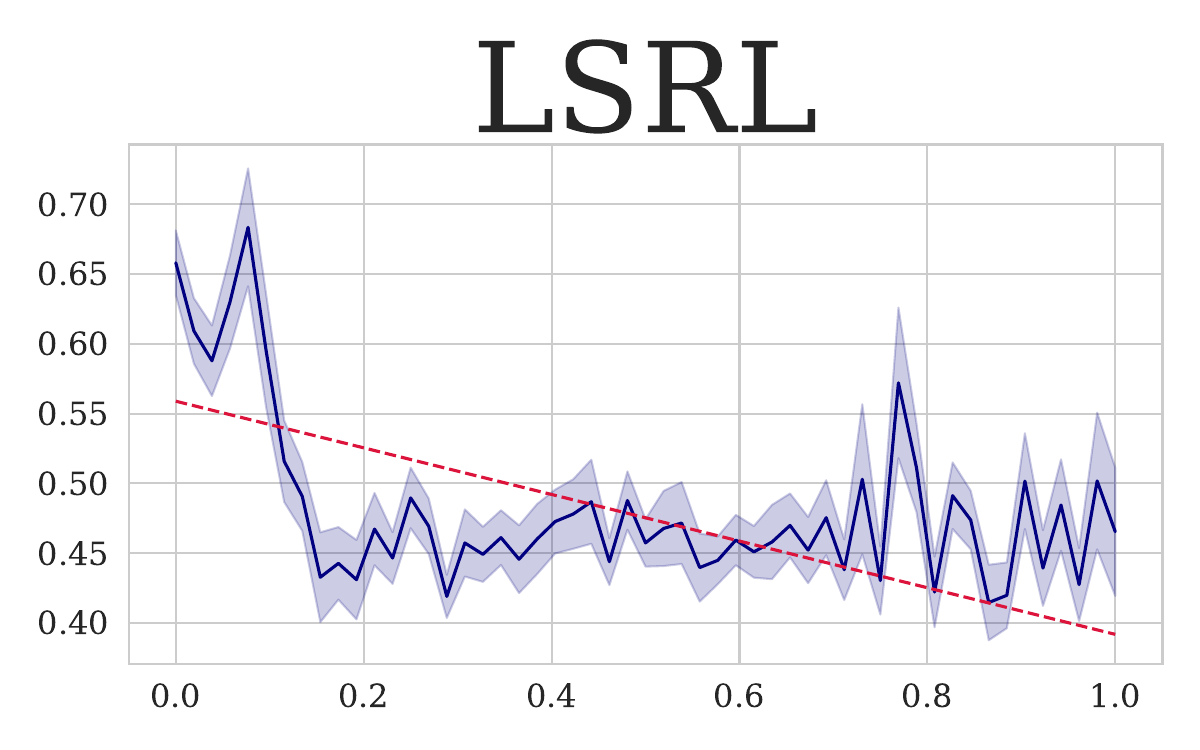}
    \end{subfigure}
    \vspace{0.3em}
    {\centering \textbf{\small WMT19 Ru-En} \par}
    \vspace{0.2em}
    \begin{subfigure}{0.15\textwidth}
        \includegraphics[width=\linewidth]{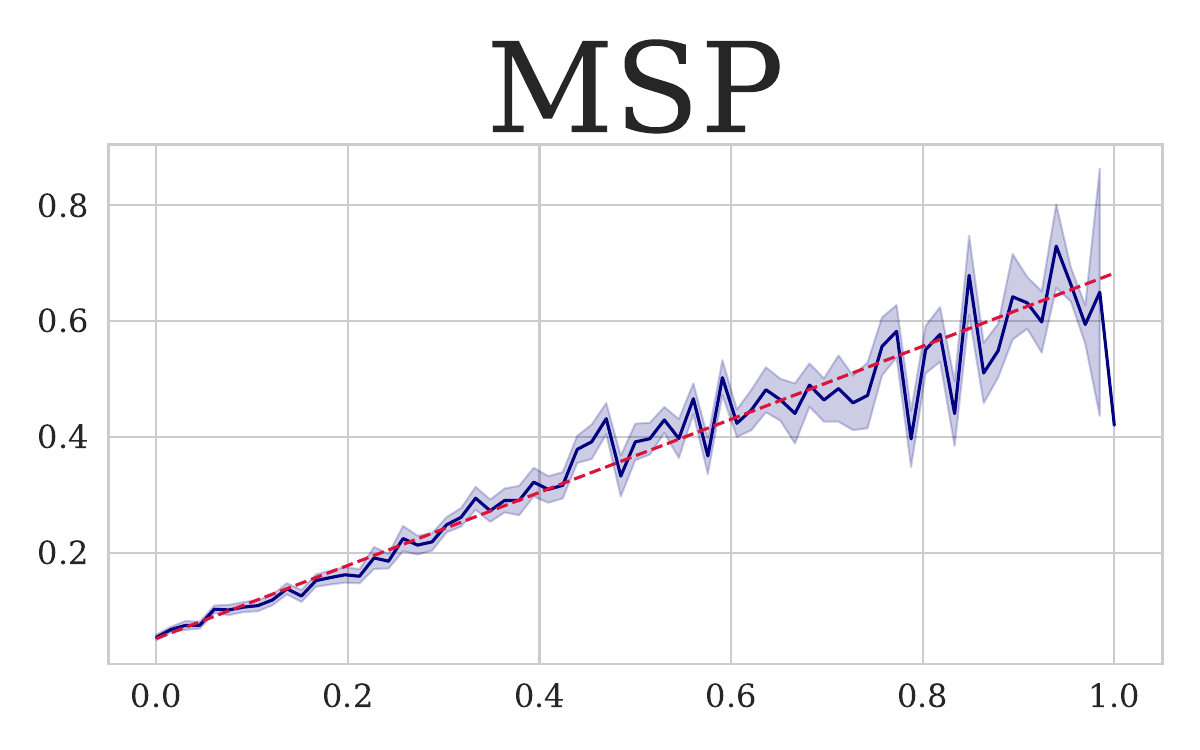}
    \end{subfigure}
    \begin{subfigure}{0.15\textwidth}
        \includegraphics[width=\linewidth]{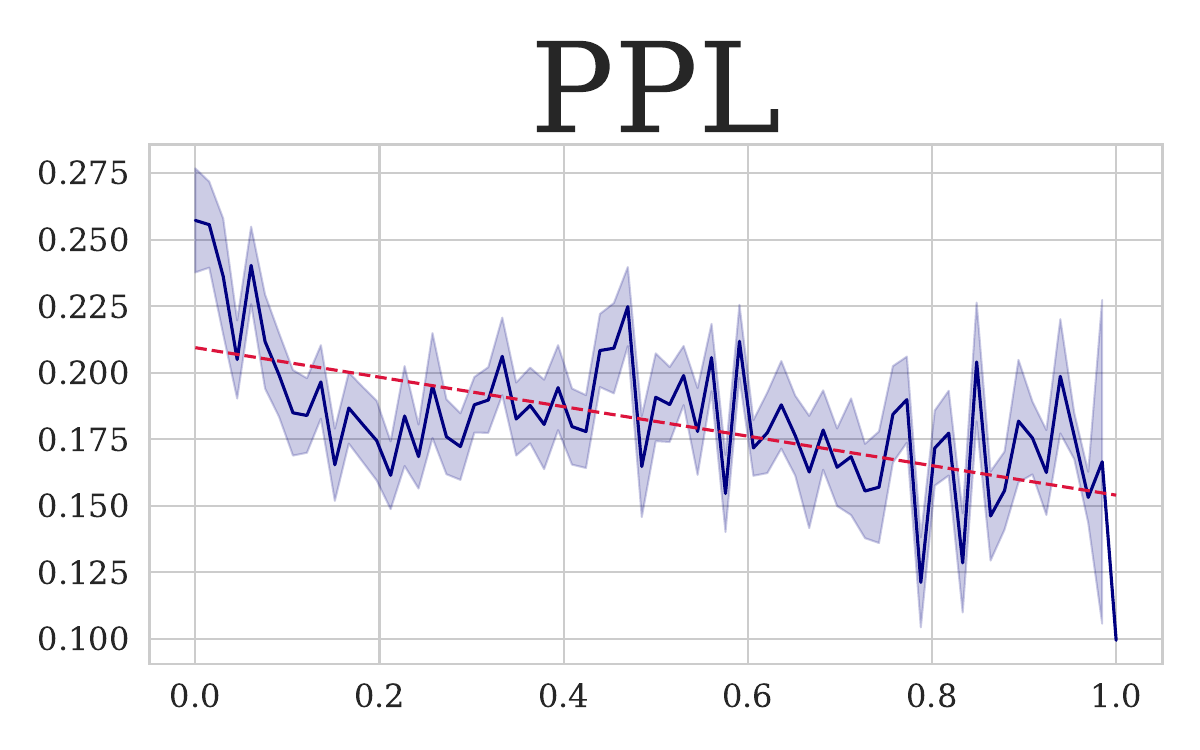}
    \end{subfigure}
    \begin{subfigure}{0.15\textwidth}
        \includegraphics[width=\linewidth]{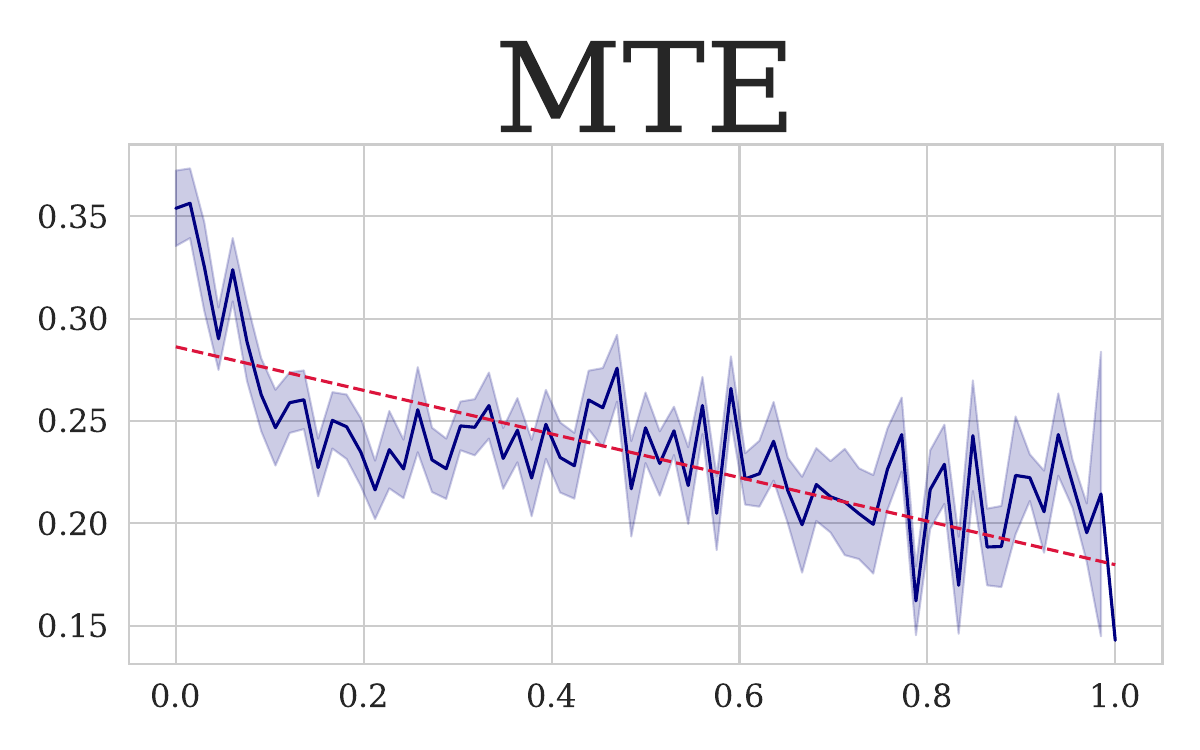}
    \end{subfigure}
    \begin{subfigure}{0.15\textwidth}
        \includegraphics[width=\linewidth]{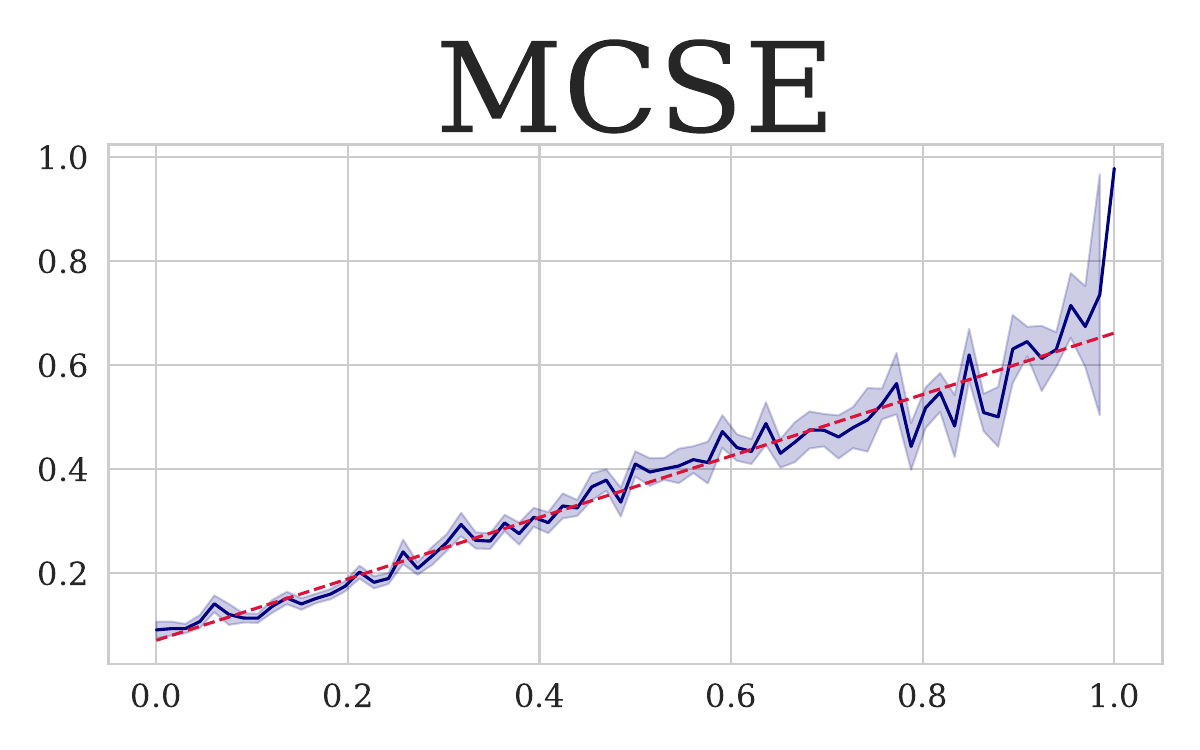}
    \end{subfigure}
    \begin{subfigure}{0.15\textwidth}
        \includegraphics[width=\linewidth]{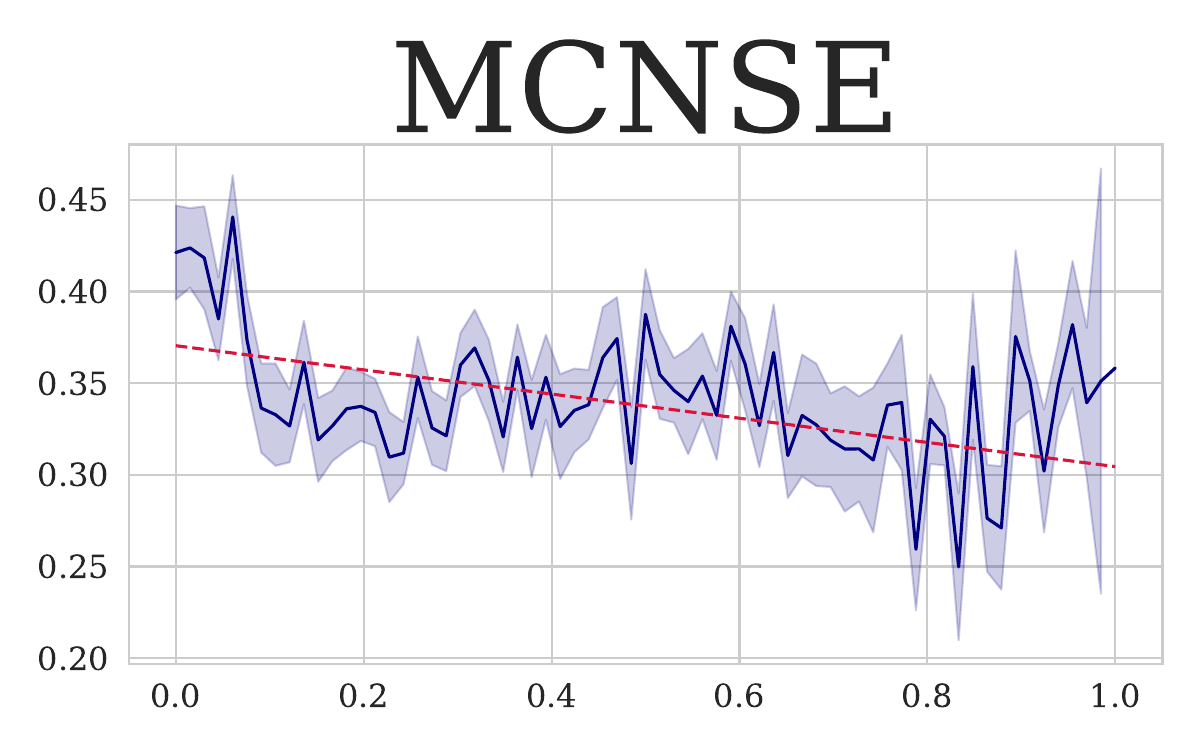}
    \end{subfigure}
    \begin{subfigure}{0.15\textwidth}
        \includegraphics[width=\linewidth]{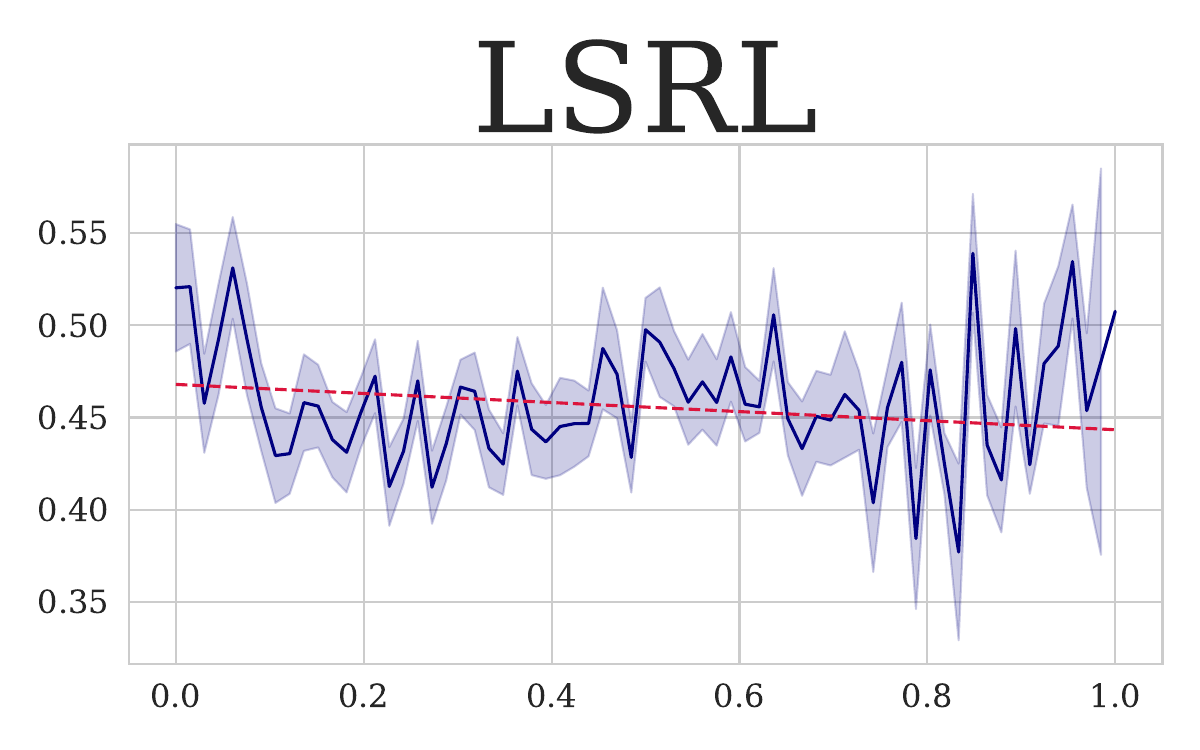}
    \end{subfigure}
    \vspace{0.3em}
    {\centering \textbf{\small WMT19 Fi-En} \par}
    \vspace{0.2em}
    \begin{subfigure}{0.15\textwidth}
        \includegraphics[width=\linewidth]{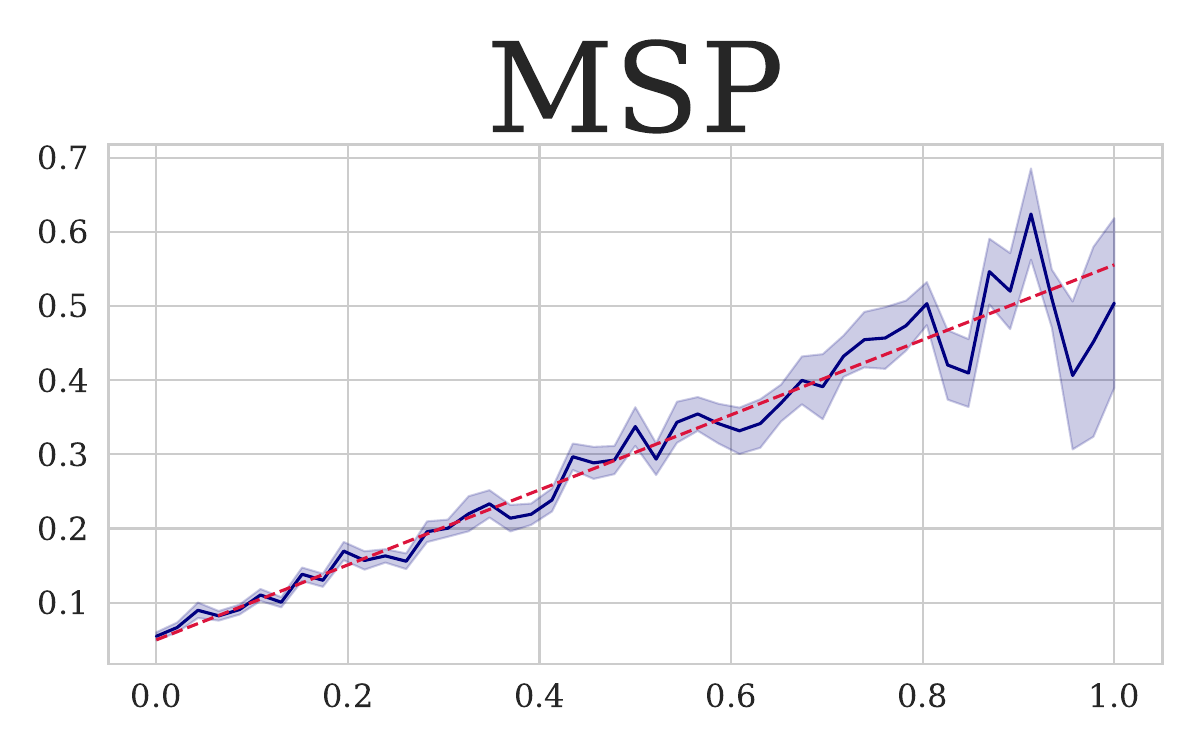}
    \end{subfigure}
    \begin{subfigure}{0.15\textwidth}
        \includegraphics[width=\linewidth]{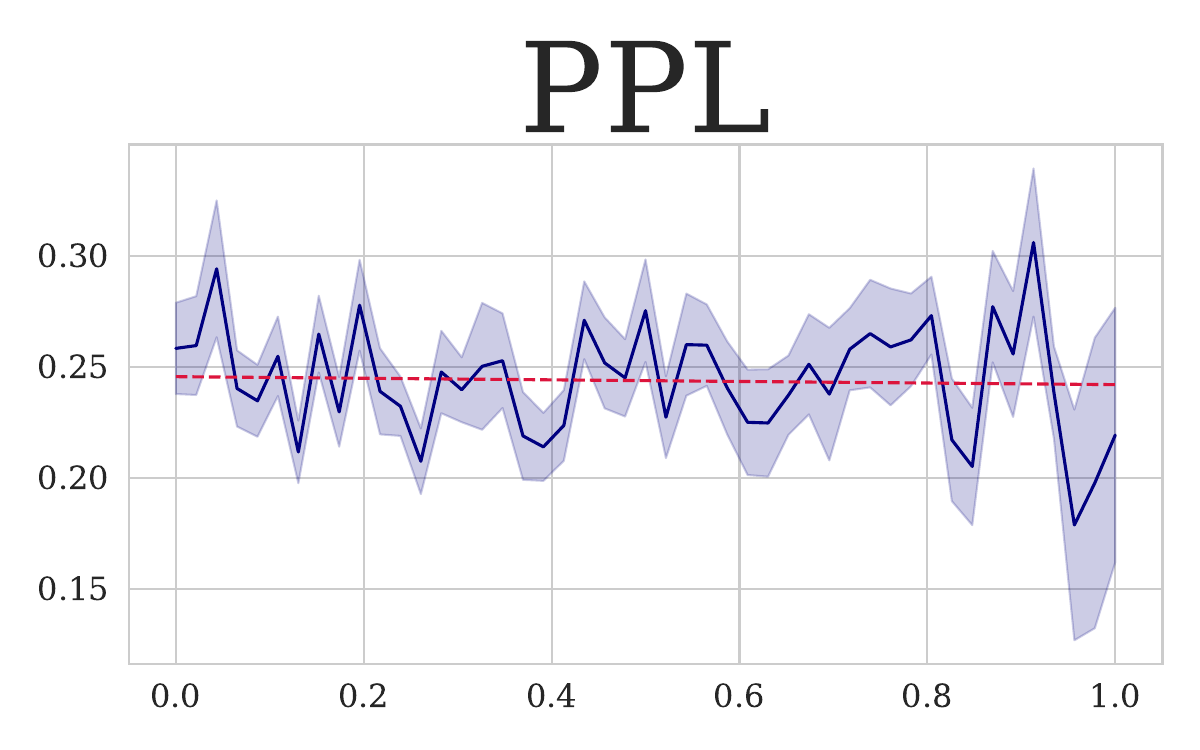}
    \end{subfigure}
    \begin{subfigure}{0.15\textwidth}
        \includegraphics[width=\linewidth]{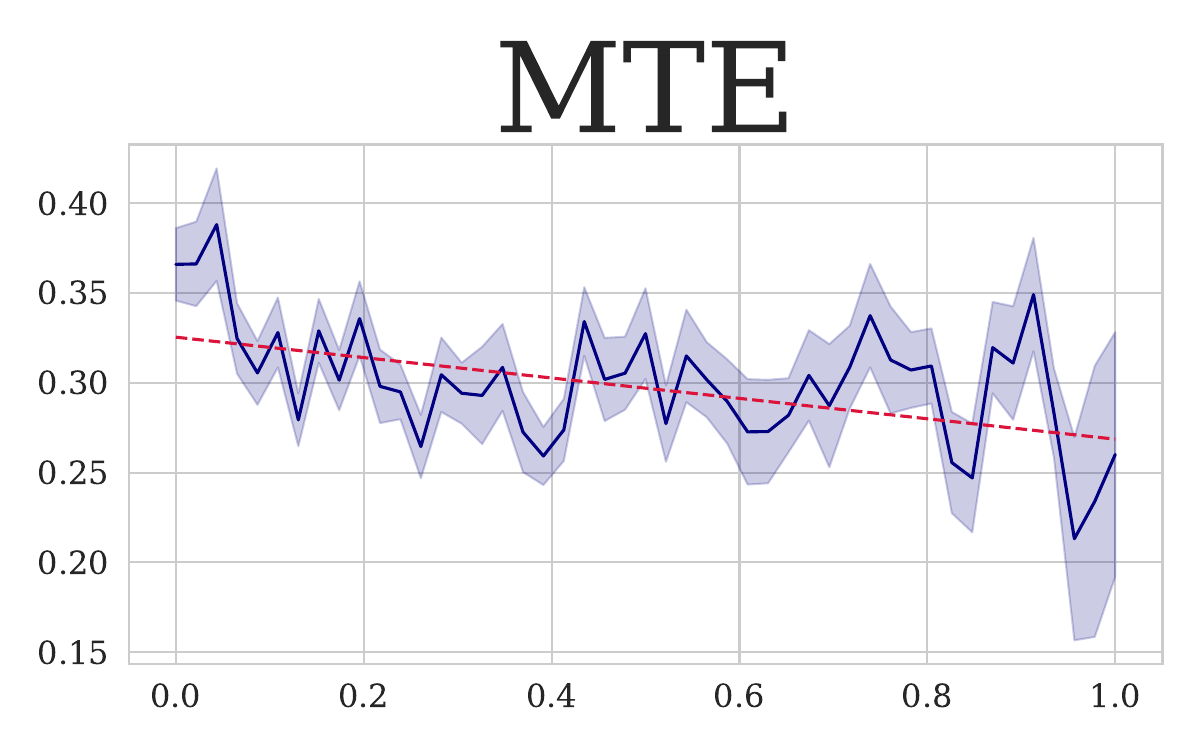}
    \end{subfigure}
    \begin{subfigure}{0.15\textwidth}
        \includegraphics[width=\linewidth]{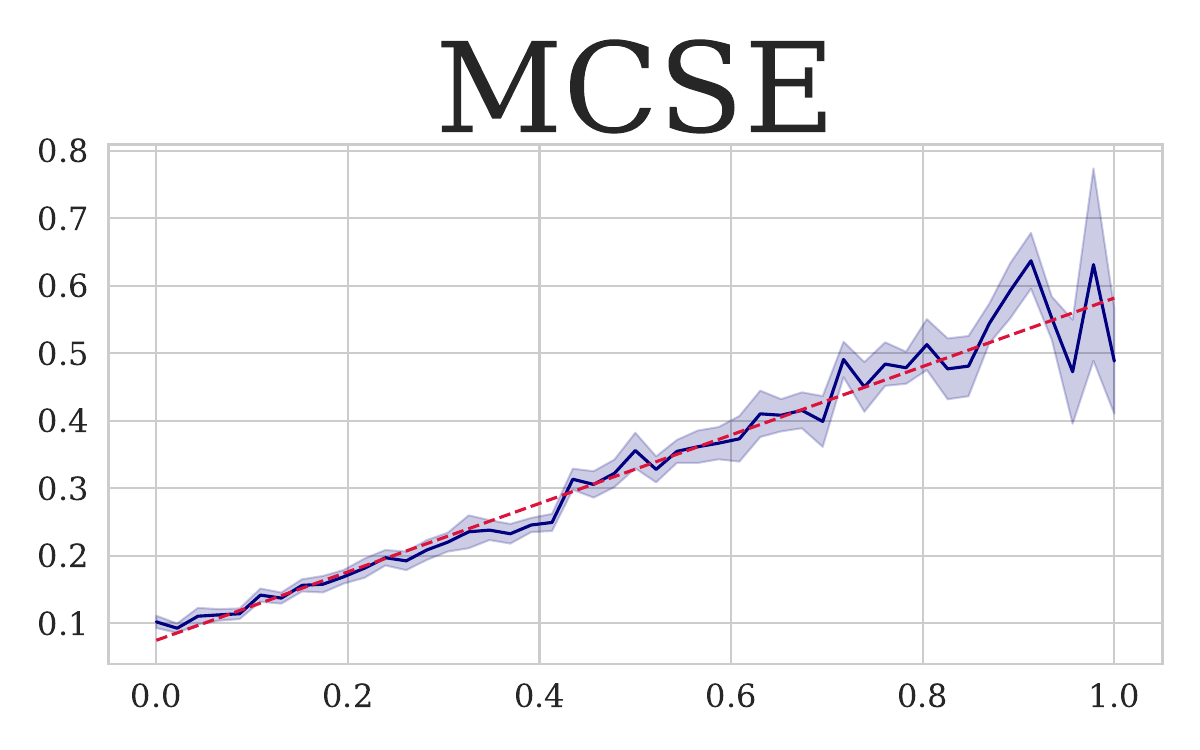}
    \end{subfigure}
    \begin{subfigure}{0.15\textwidth}
        \includegraphics[width=\linewidth]{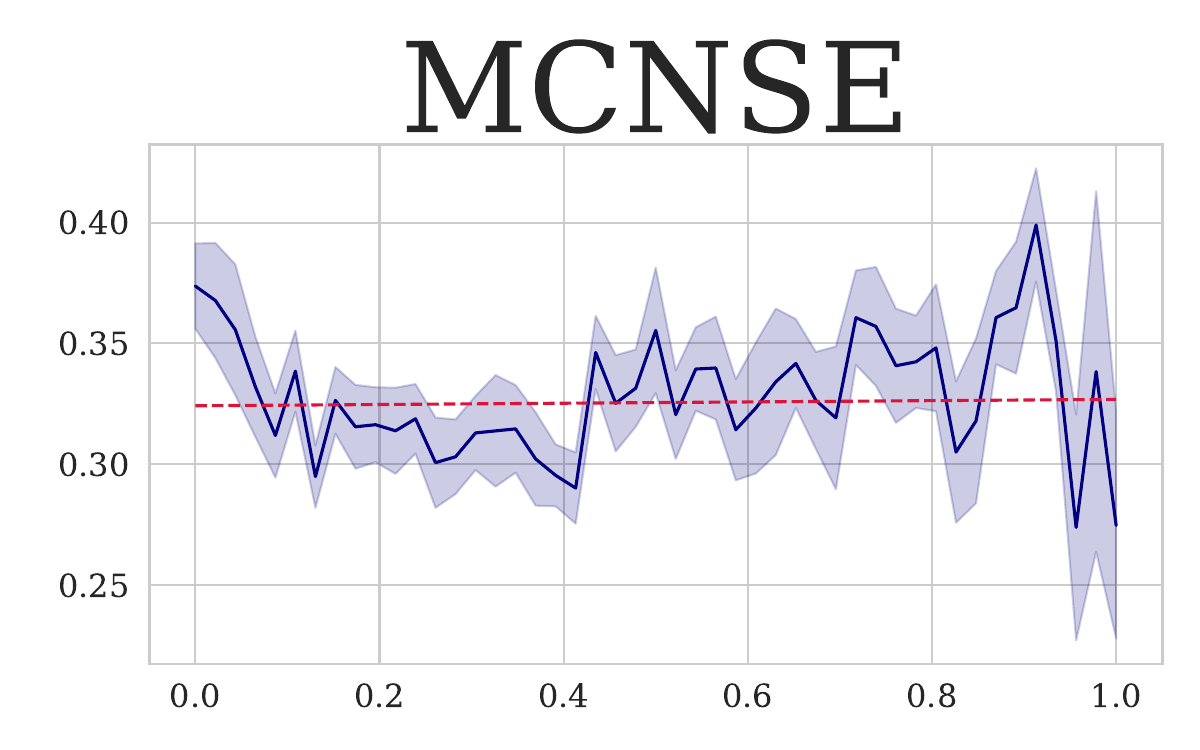}
    \end{subfigure}
    \begin{subfigure}{0.15\textwidth}
        \includegraphics[width=\linewidth]{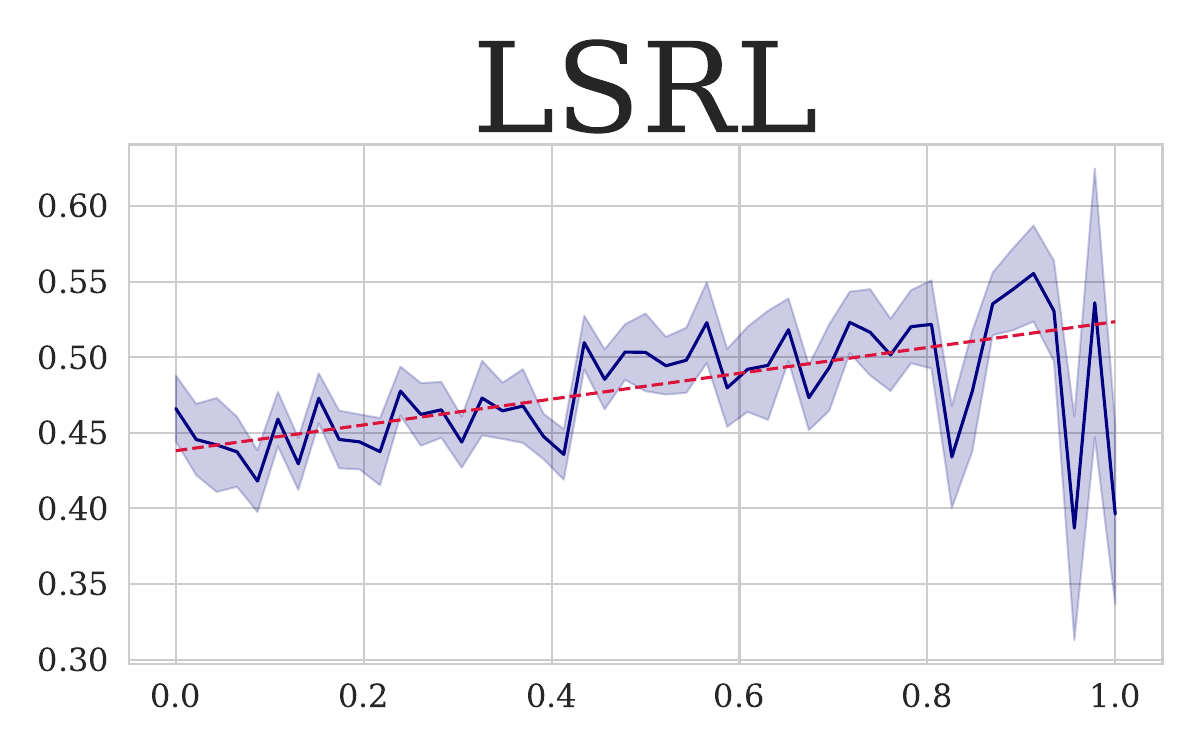}
    \end{subfigure}
    \vspace{0.3em}
    {\centering \textbf{\small WMT19 De-En} \par}
    \vspace{0.2em}
    \begin{subfigure}{0.15\textwidth}
        \includegraphics[width=\linewidth]{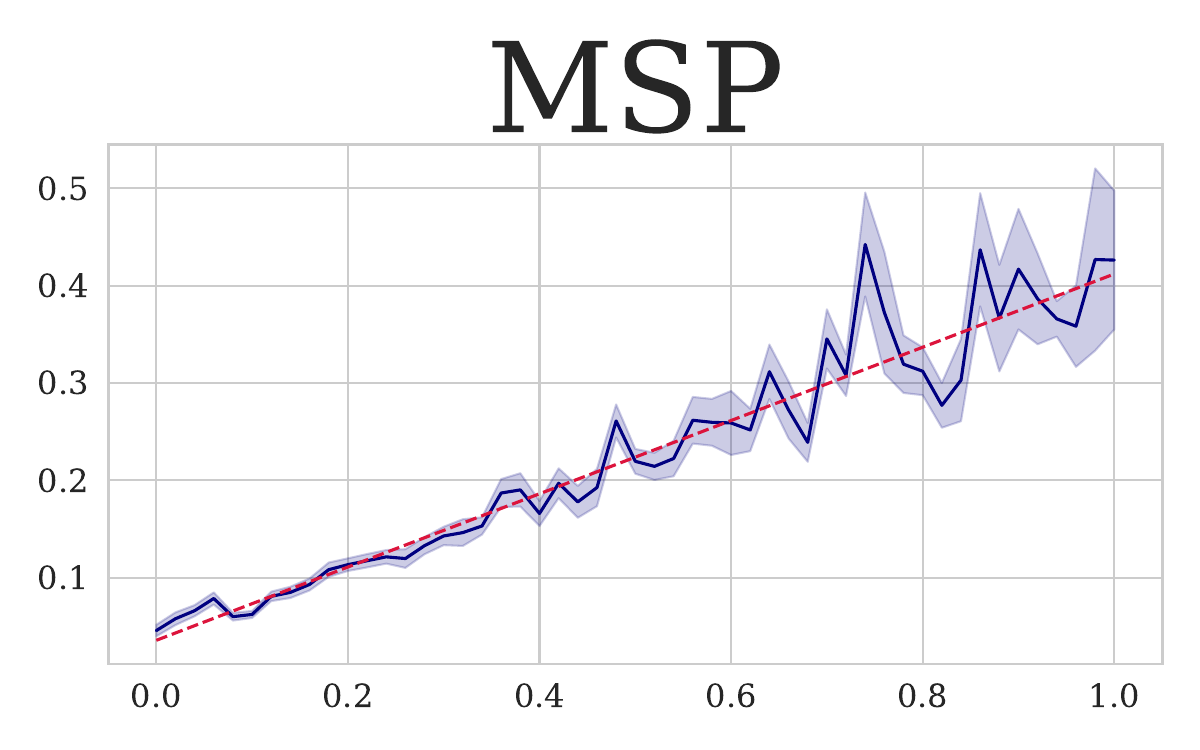}
    \end{subfigure}
    \begin{subfigure}{0.15\textwidth}
        \includegraphics[width=\linewidth]{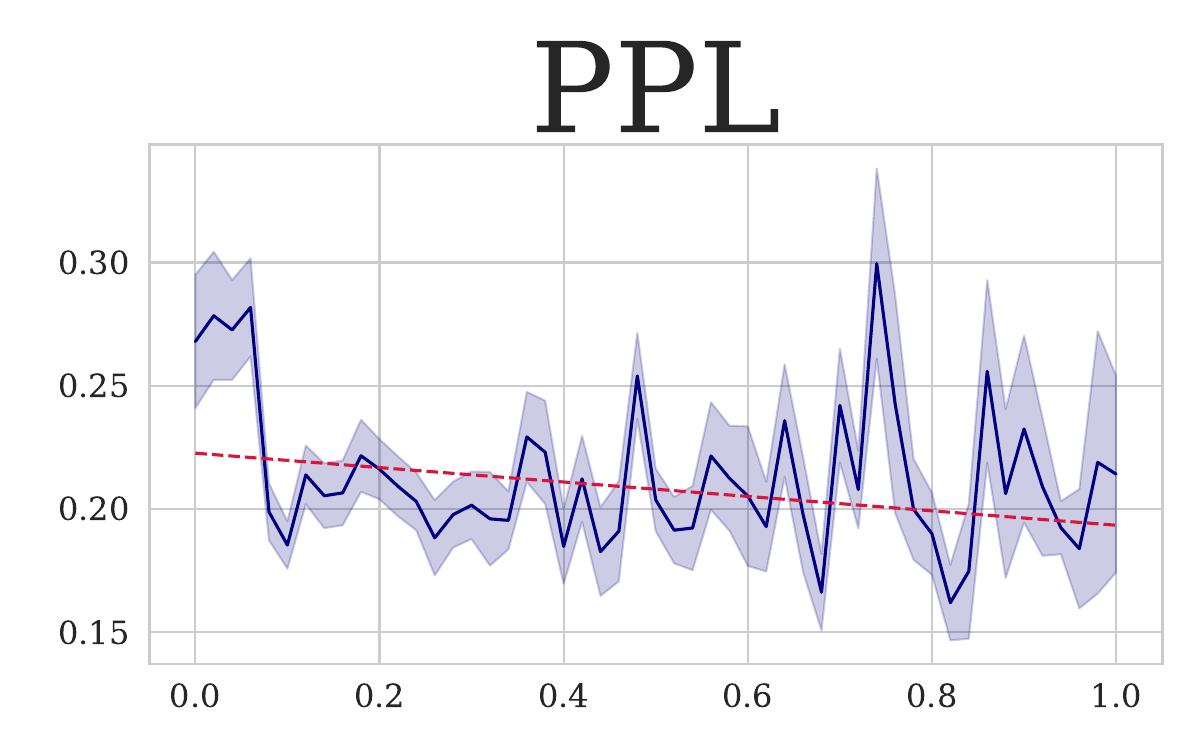}
    \end{subfigure}
    \begin{subfigure}{0.15\textwidth}
        \includegraphics[width=\linewidth]{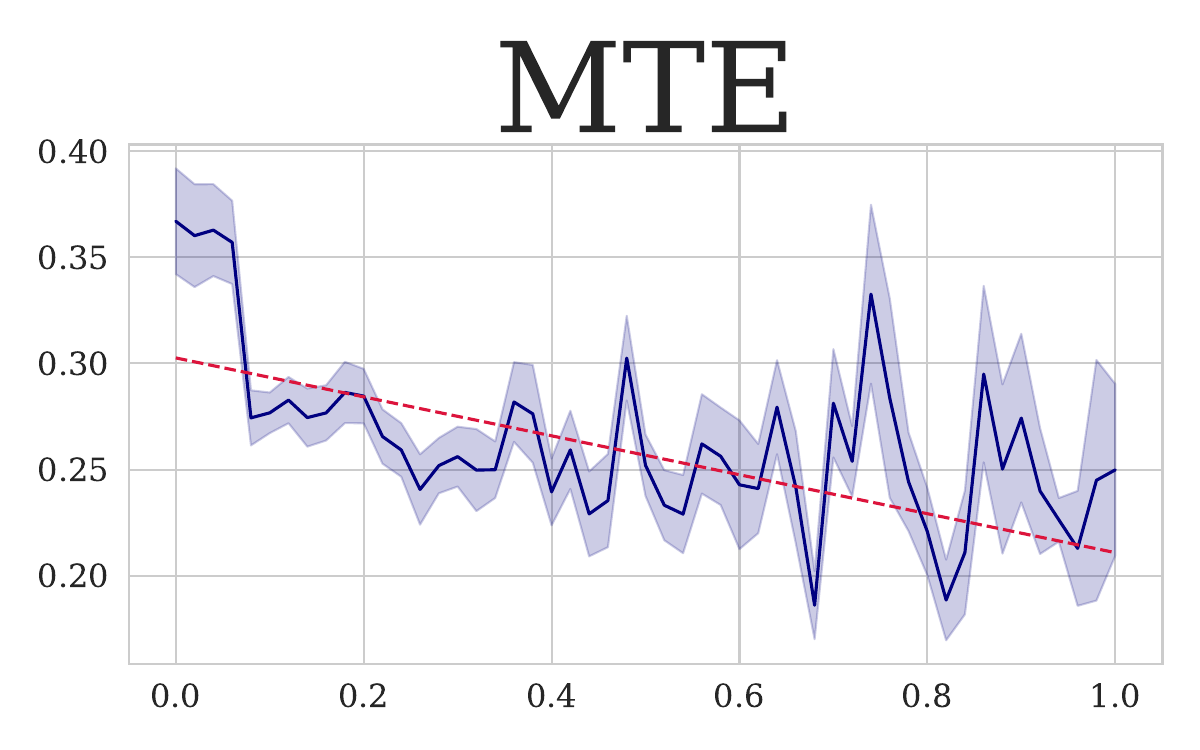}
    \end{subfigure}
    \begin{subfigure}{0.15\textwidth}
        \includegraphics[width=\linewidth]{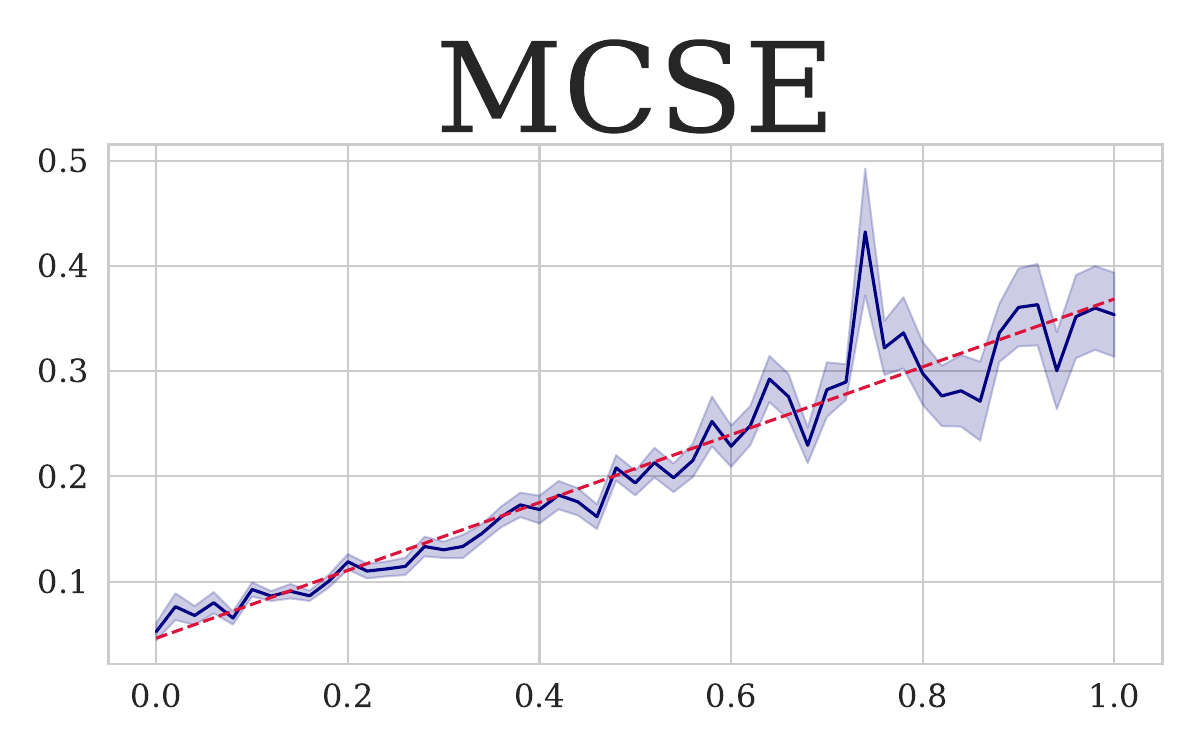}
    \end{subfigure}
    \begin{subfigure}{0.15\textwidth}
        \includegraphics[width=\linewidth]{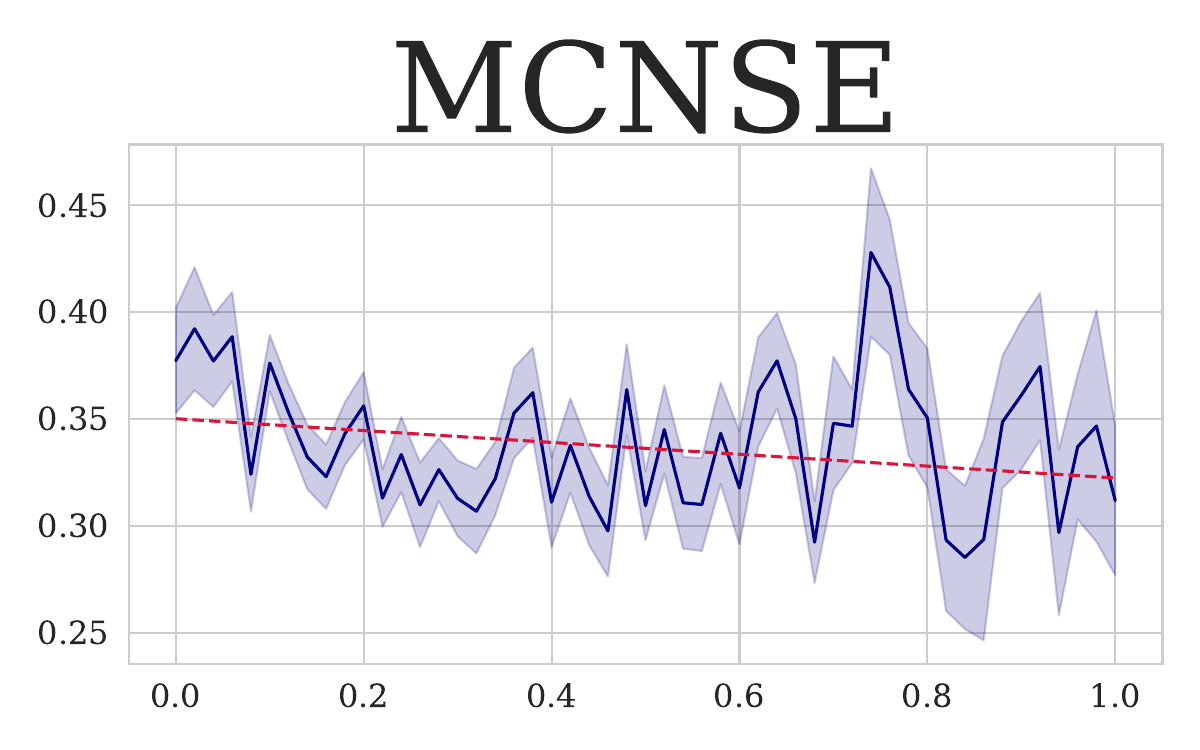}
    \end{subfigure}
    \begin{subfigure}{0.15\textwidth}
        \includegraphics[width=\linewidth]{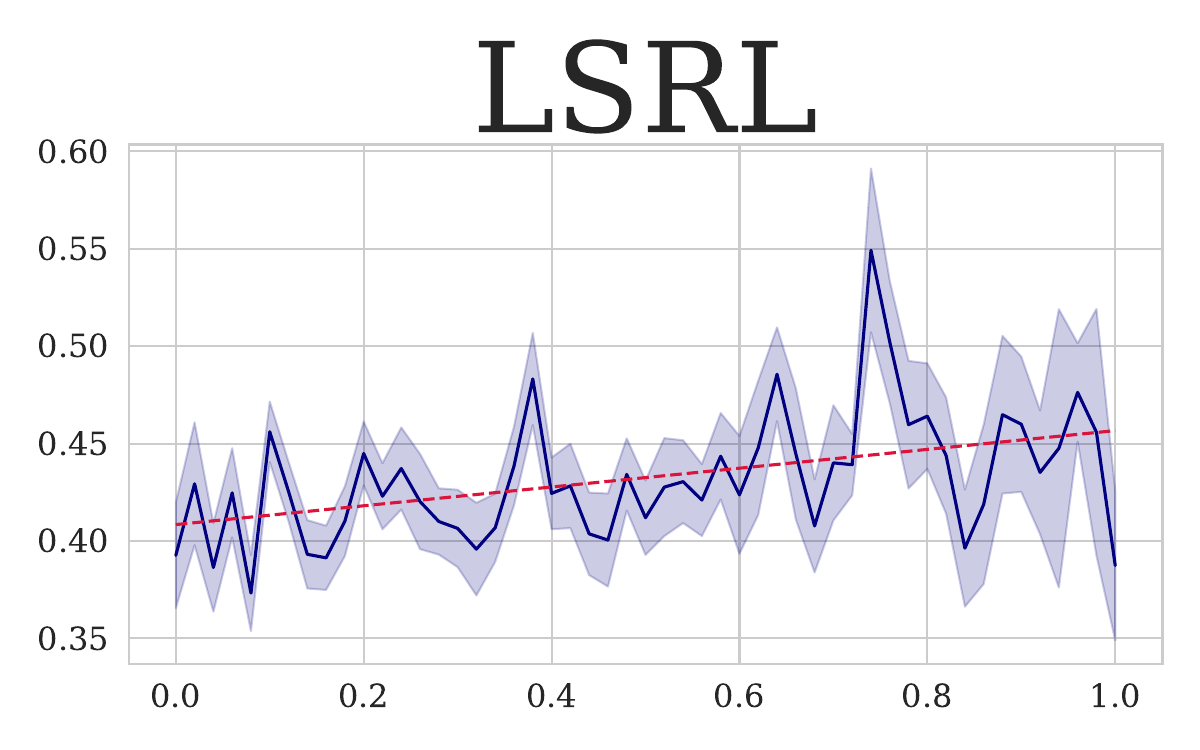}
    \end{subfigure}
    \vspace{0.3em}
    {\centering \textbf{\small WMT19 Lt-En} \par}
    \vspace{0.2em}
    \begin{subfigure}{0.15\textwidth}
        \includegraphics[width=\linewidth]{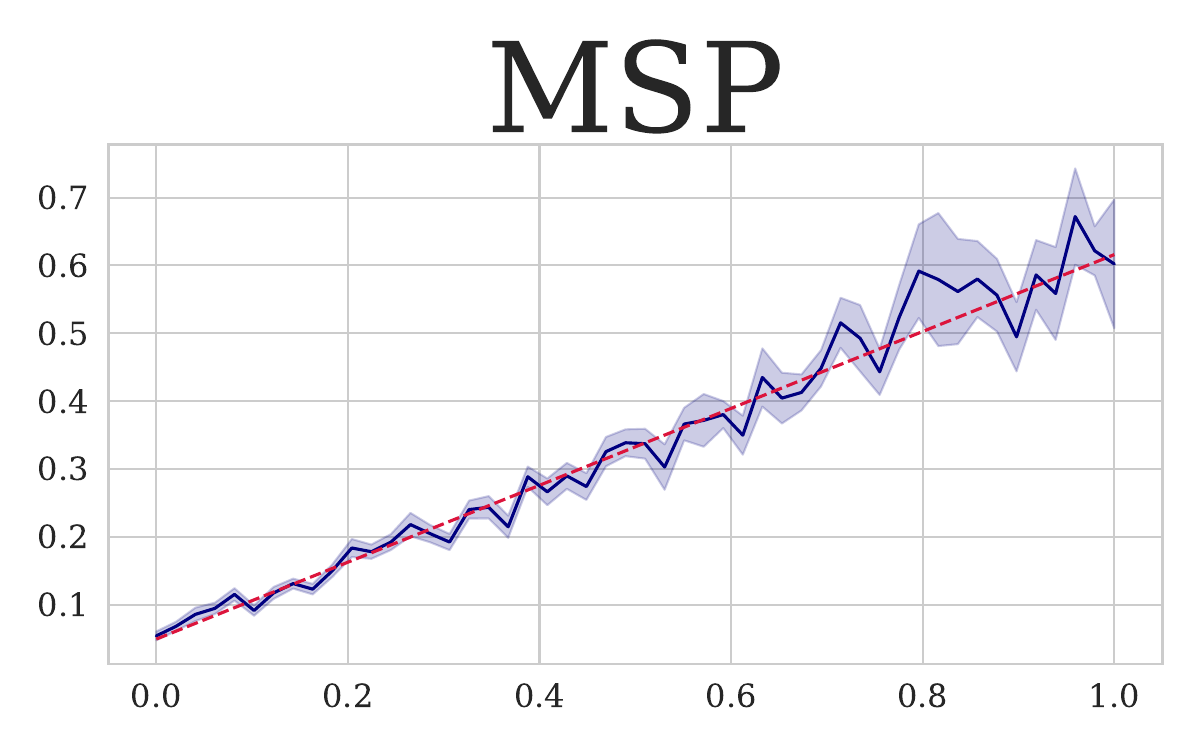}
    \end{subfigure}
    \begin{subfigure}{0.15\textwidth}
        \includegraphics[width=\linewidth]{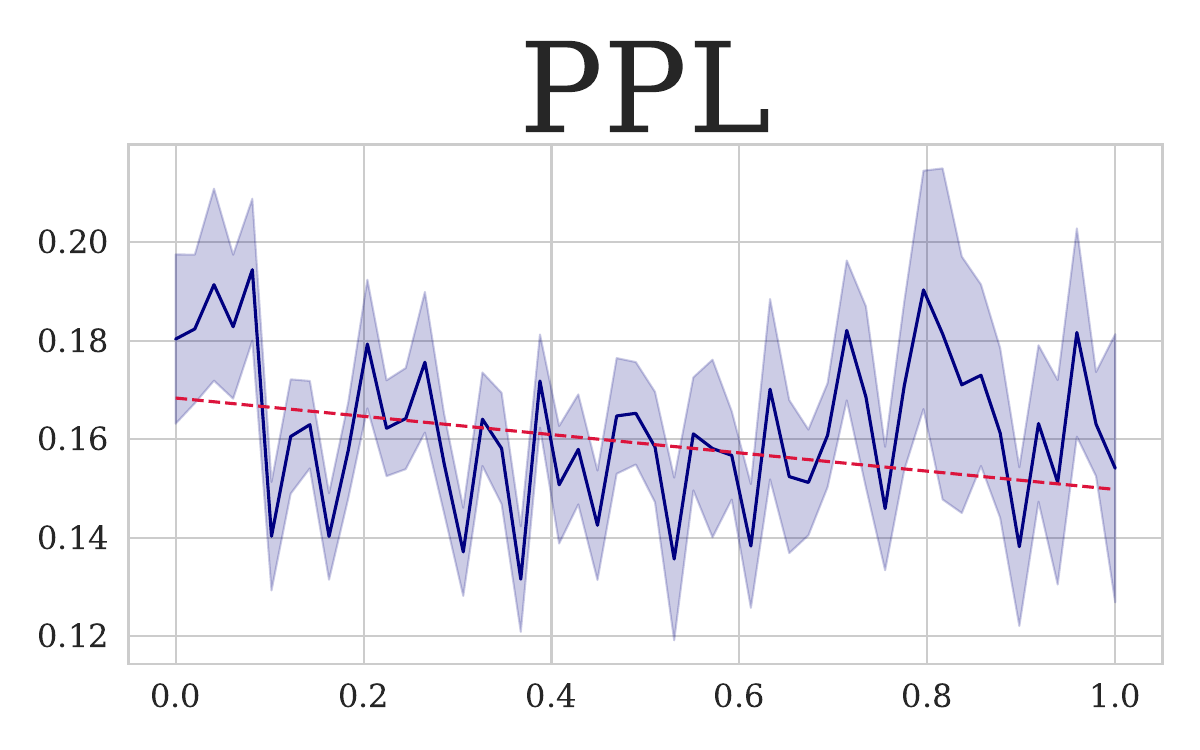}
    \end{subfigure}
    \begin{subfigure}{0.15\textwidth}
        \includegraphics[width=\linewidth]{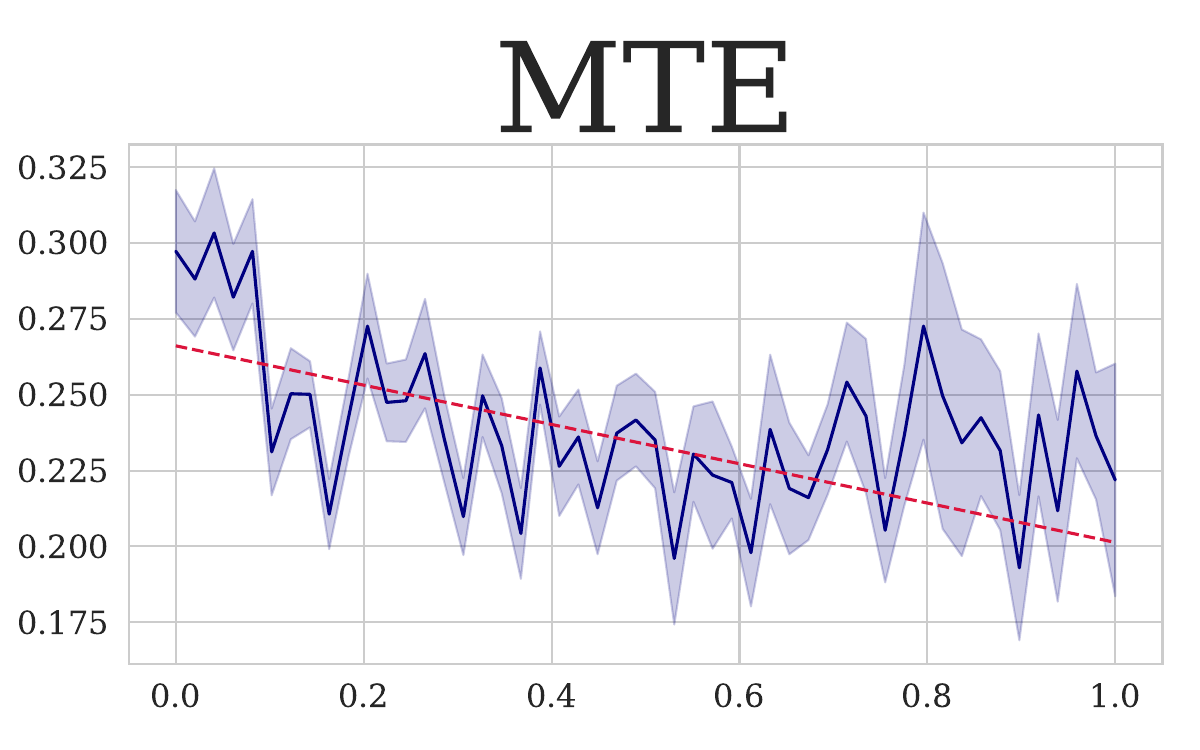}
    \end{subfigure}
    \begin{subfigure}{0.15\textwidth}
        \includegraphics[width=\linewidth]{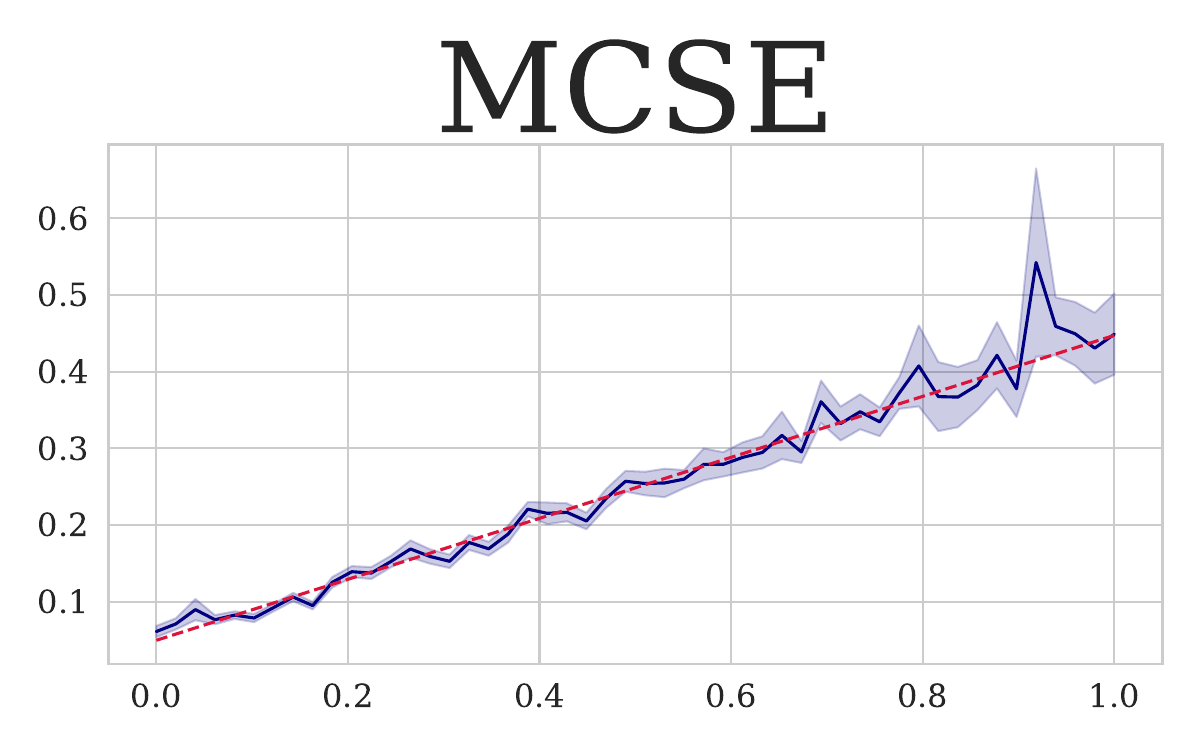}
    \end{subfigure}
    \begin{subfigure}{0.15\textwidth}
        \includegraphics[width=\linewidth]{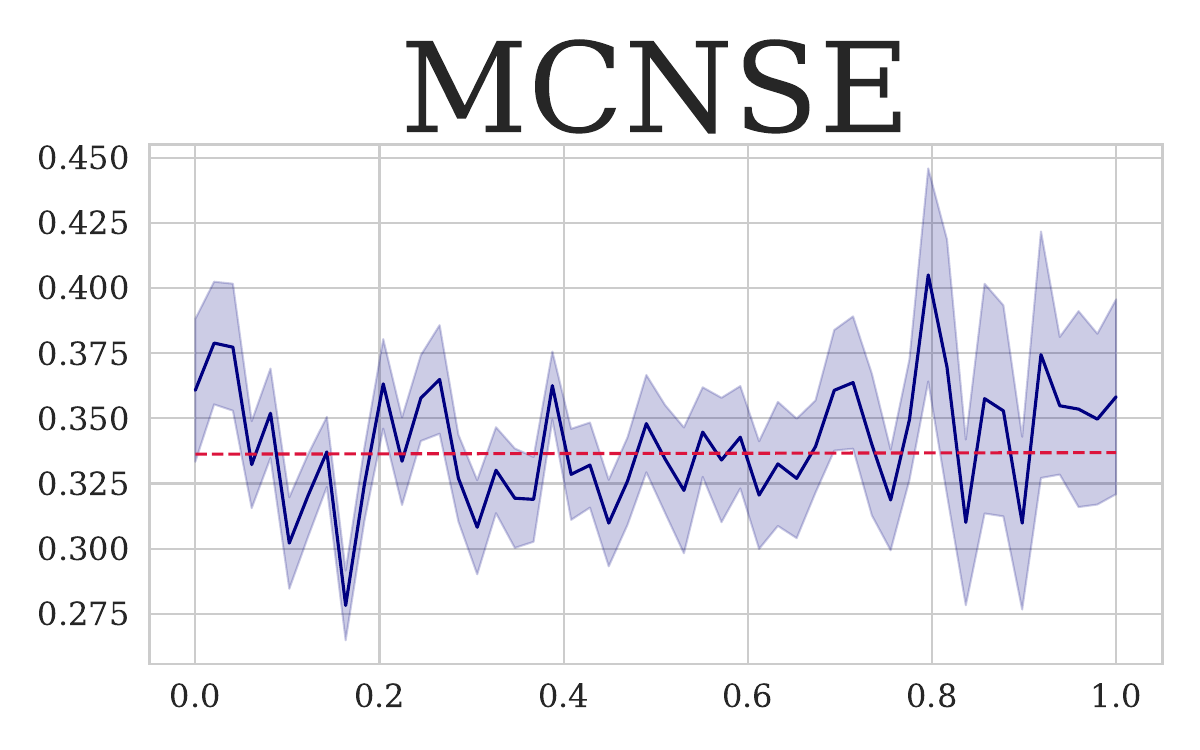}
    \end{subfigure}
    \begin{subfigure}{0.15\textwidth}
        \includegraphics[width=\linewidth]{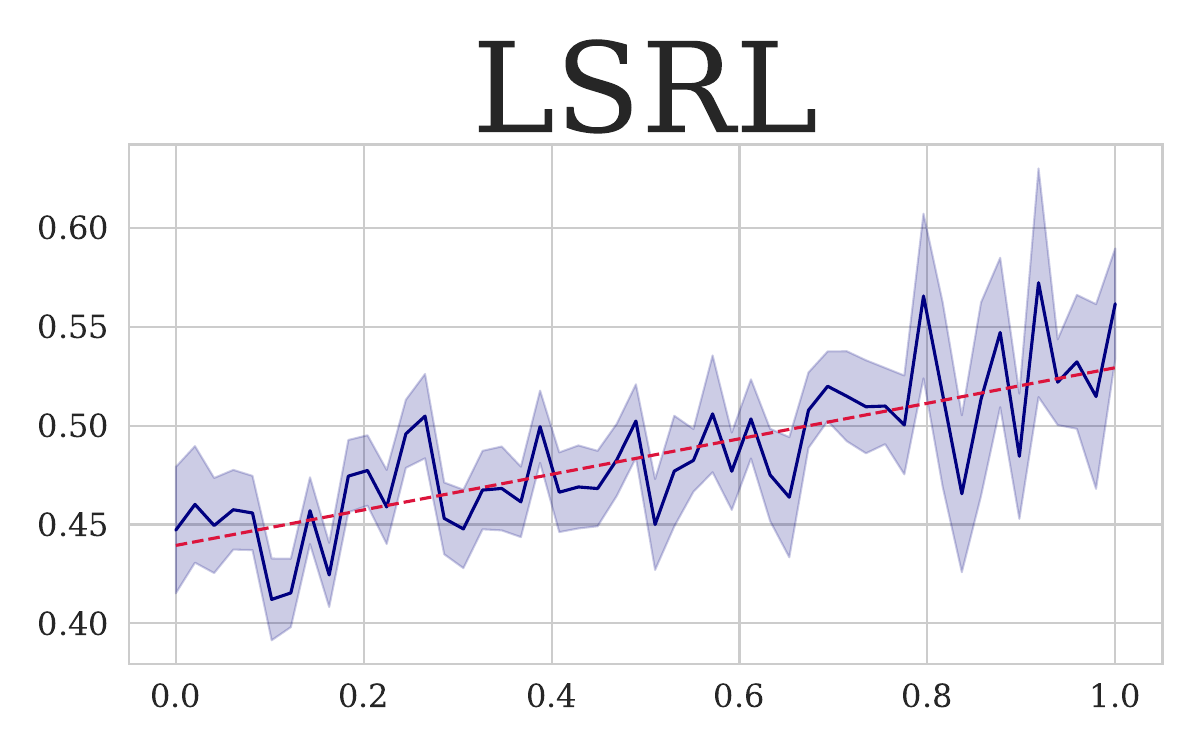}
    \end{subfigure}
    \vspace{0.3em}

 {\centering \textbf{\small XSUM} \par}
    \vspace{0.2em}
    \begin{subfigure}{0.15\textwidth}
        \includegraphics[width=\linewidth]{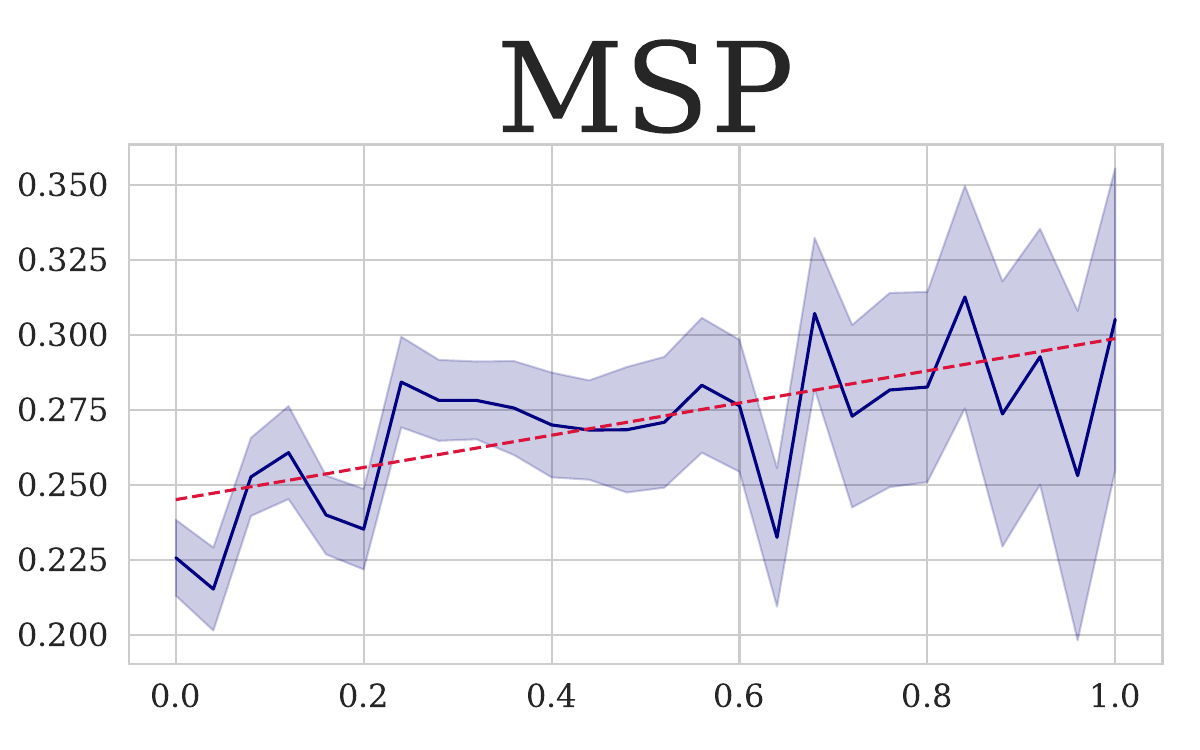}
    \end{subfigure}
    \begin{subfigure}{0.15\textwidth}
        \includegraphics[width=\linewidth]{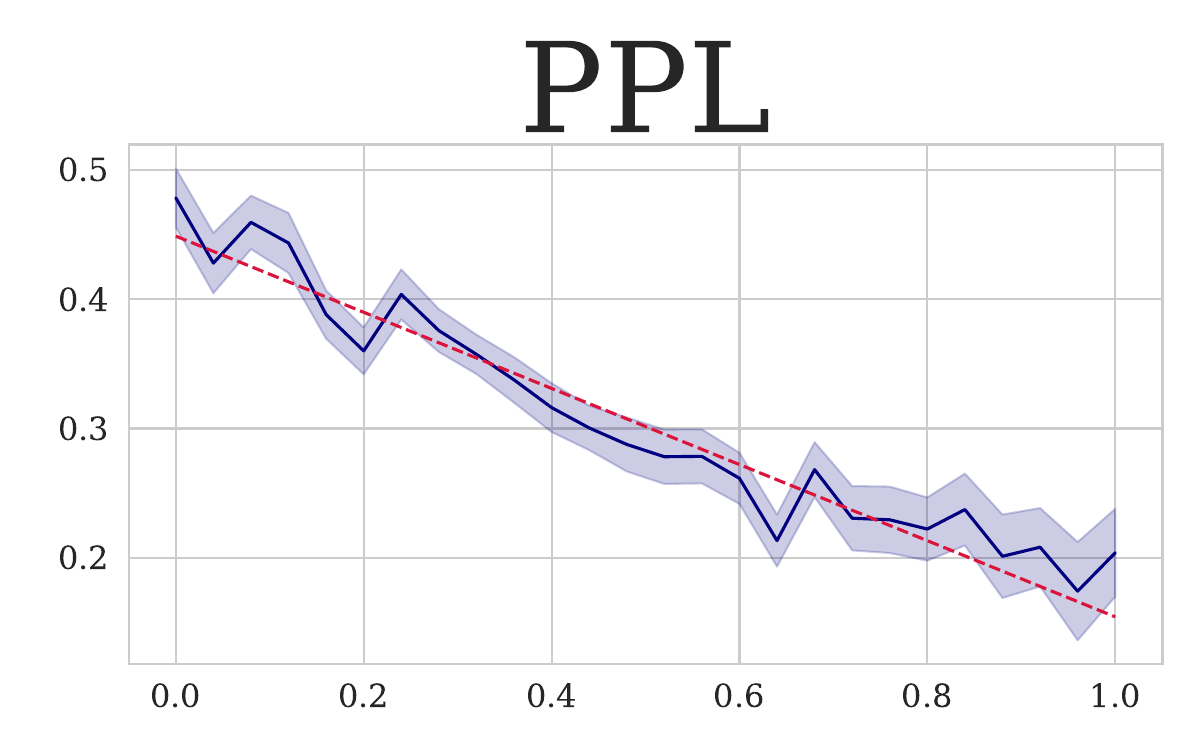}
    \end{subfigure}
    \begin{subfigure}{0.15\textwidth}
        \includegraphics[width=\linewidth]{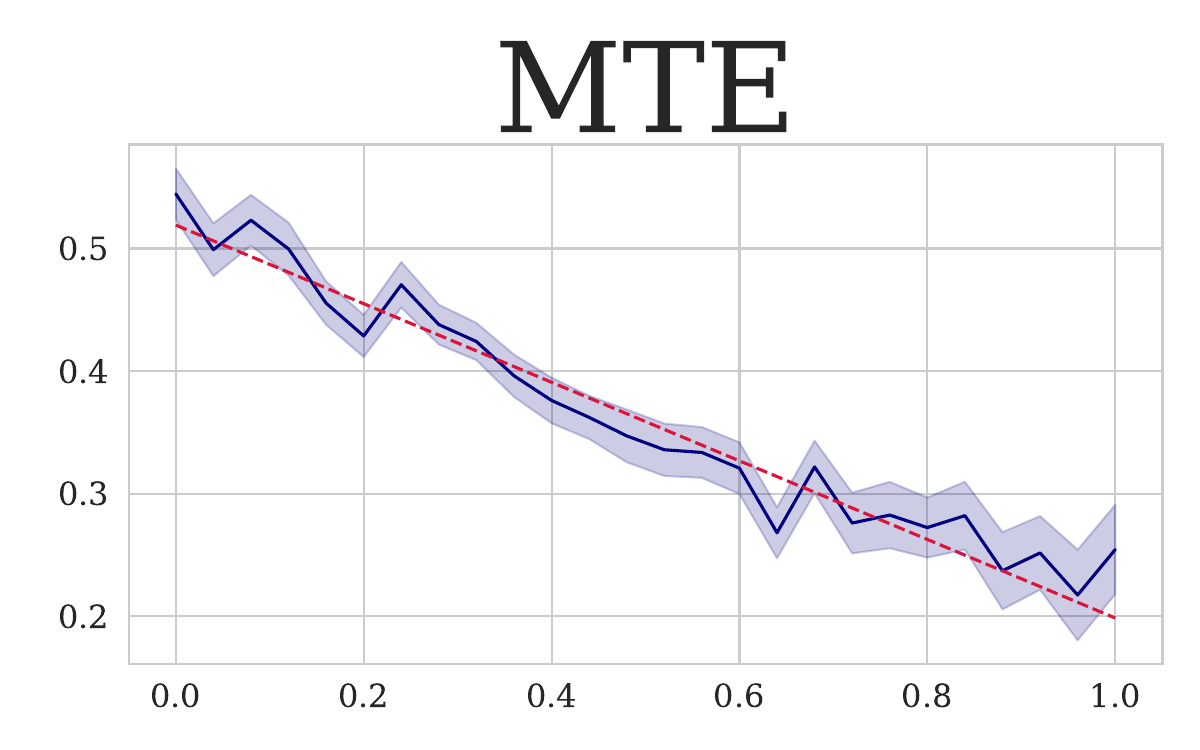}
    \end{subfigure}
    \begin{subfigure}{0.15\textwidth}
        \includegraphics[width=\linewidth]{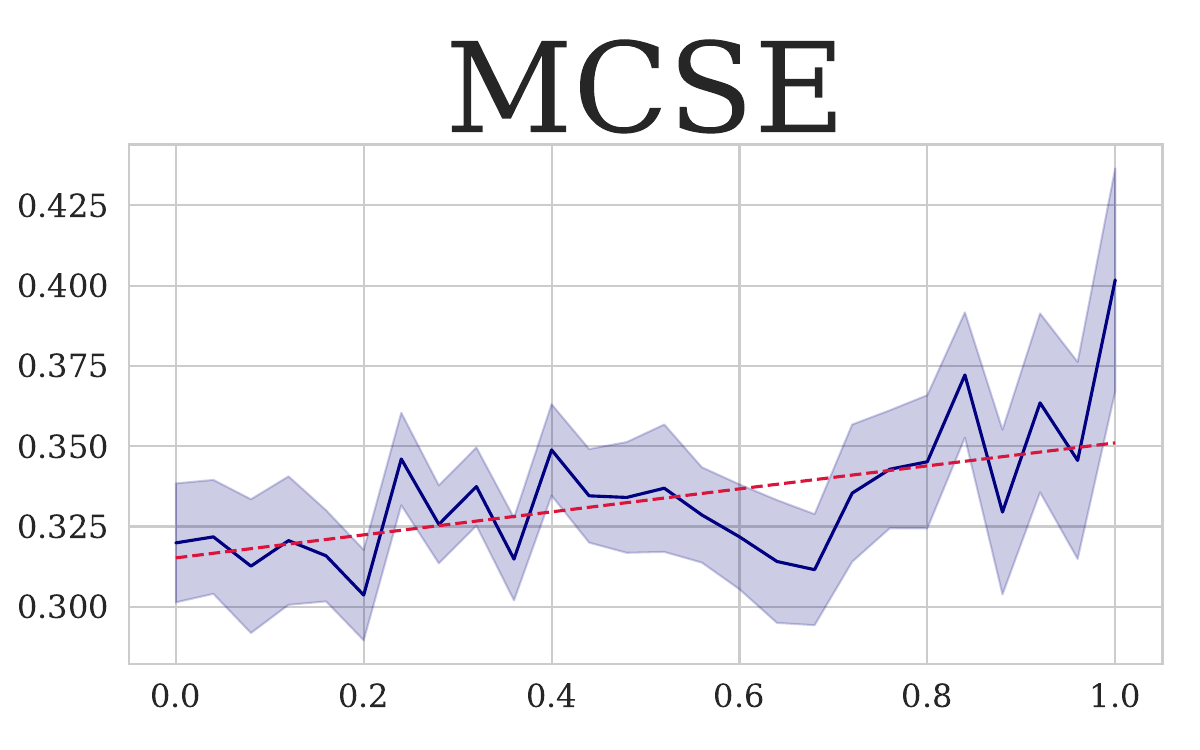}
    \end{subfigure}
    \begin{subfigure}{0.15\textwidth}
        \includegraphics[width=\linewidth]{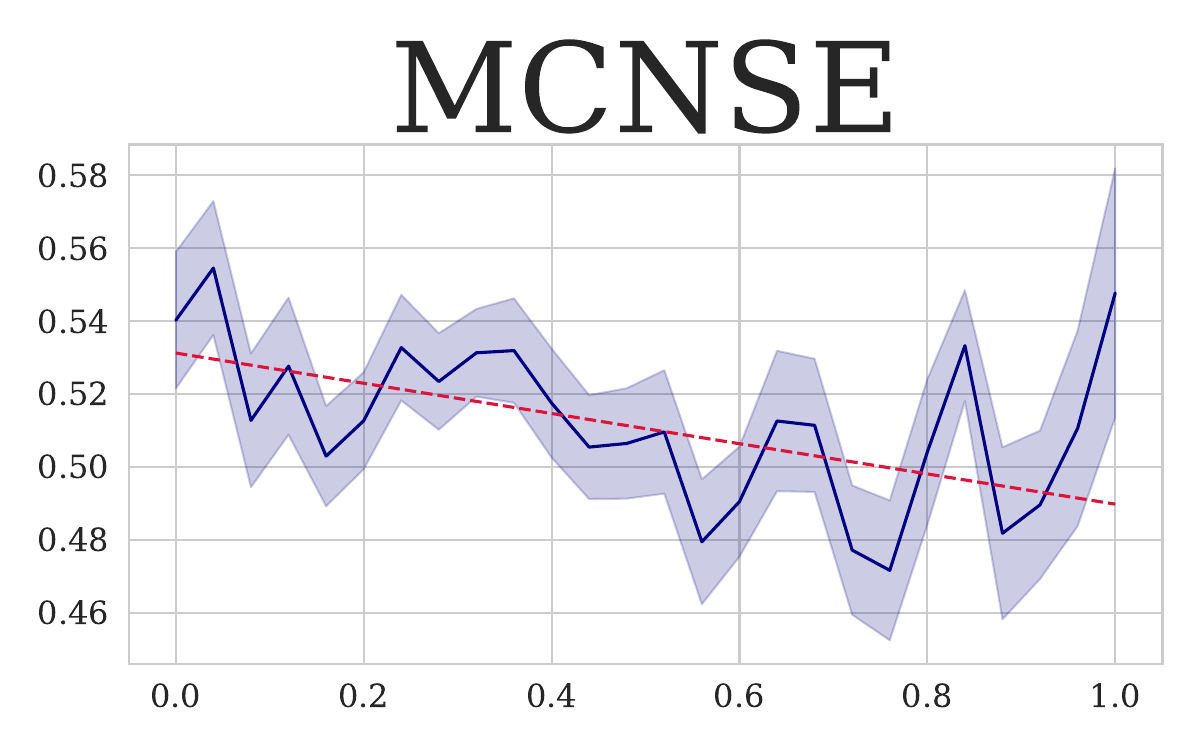}
    \end{subfigure}
    \begin{subfigure}{0.15\textwidth}
        \includegraphics[width=\linewidth]{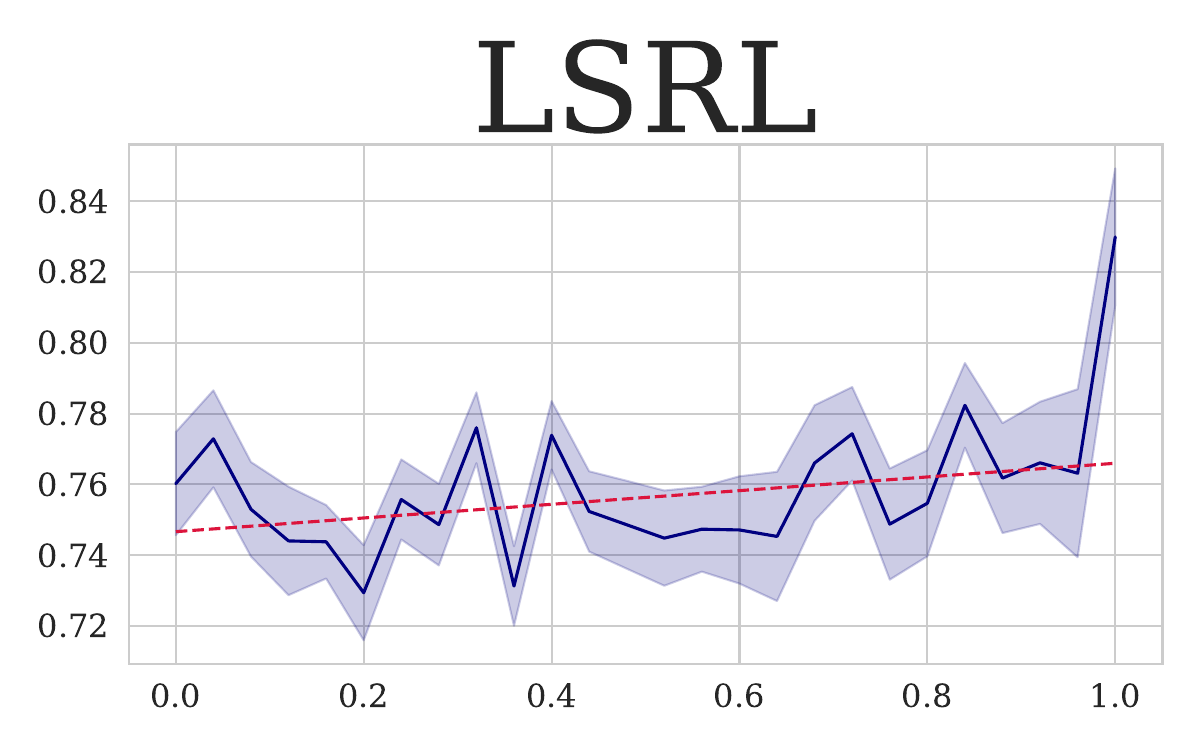}
    \end{subfigure}
    \vspace{0.3em}

     {\centering \textbf{\small GSM8K} \par}
    \vspace{0.2em}
    \begin{subfigure}{0.15\textwidth}
        \includegraphics[width=\linewidth]{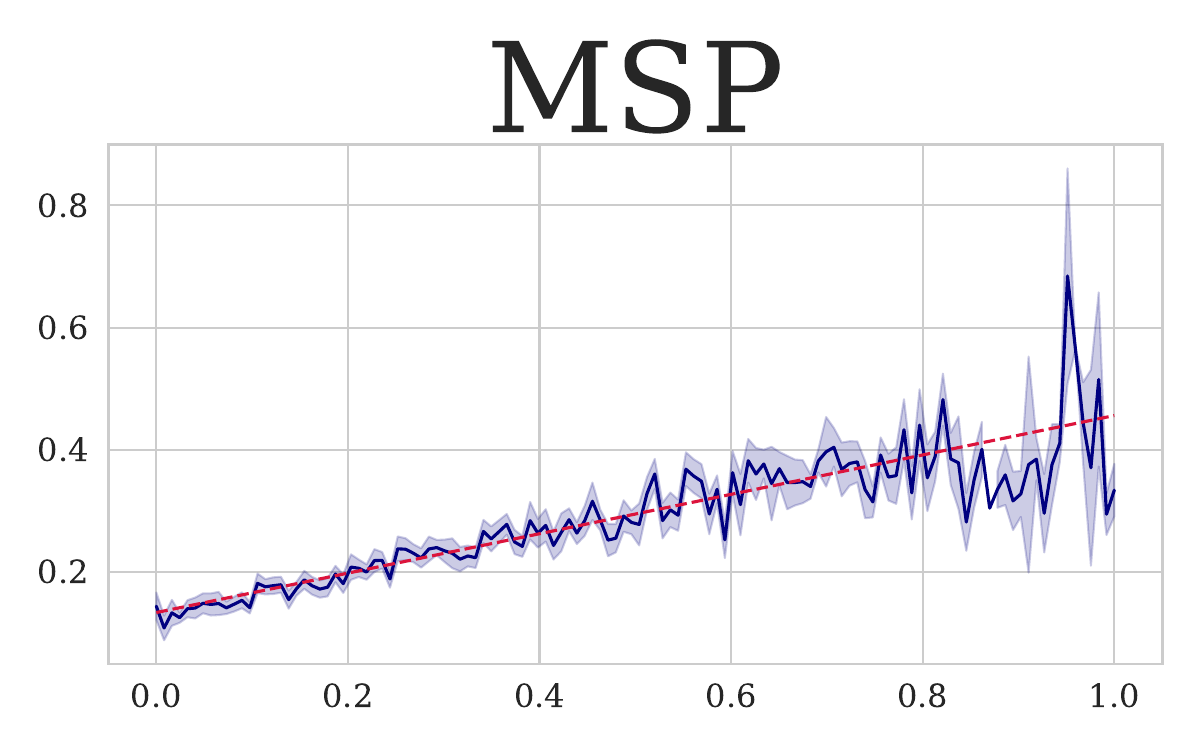}
    \end{subfigure}
    \begin{subfigure}{0.15\textwidth}
        \includegraphics[width=\linewidth]{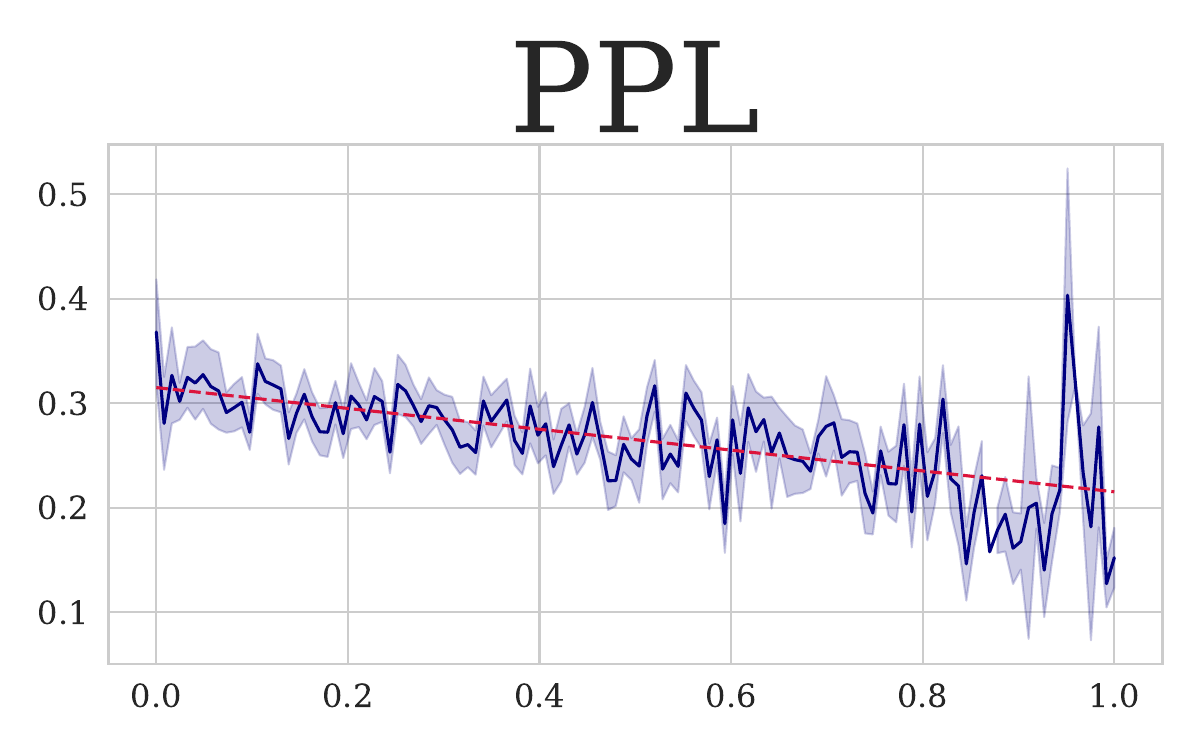}
    \end{subfigure}
    \begin{subfigure}{0.15\textwidth}
        \includegraphics[width=\linewidth]{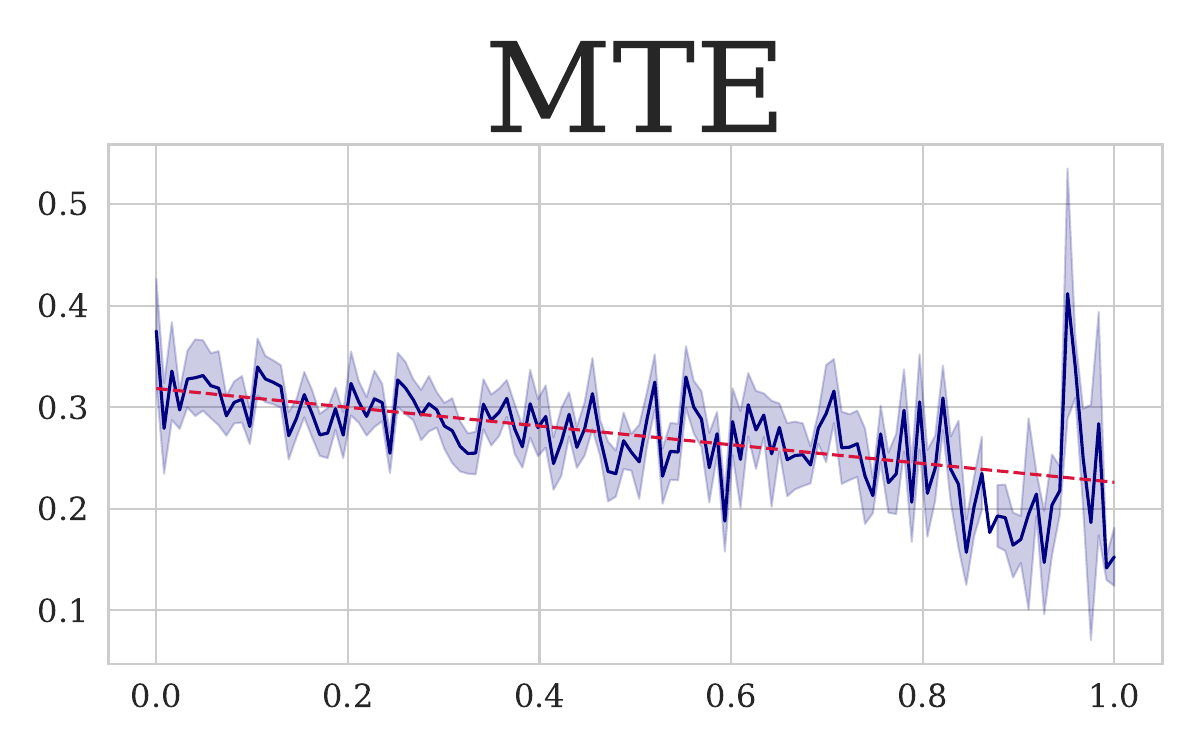}
    \end{subfigure}
    \begin{subfigure}{0.15\textwidth}
        \includegraphics[width=\linewidth]{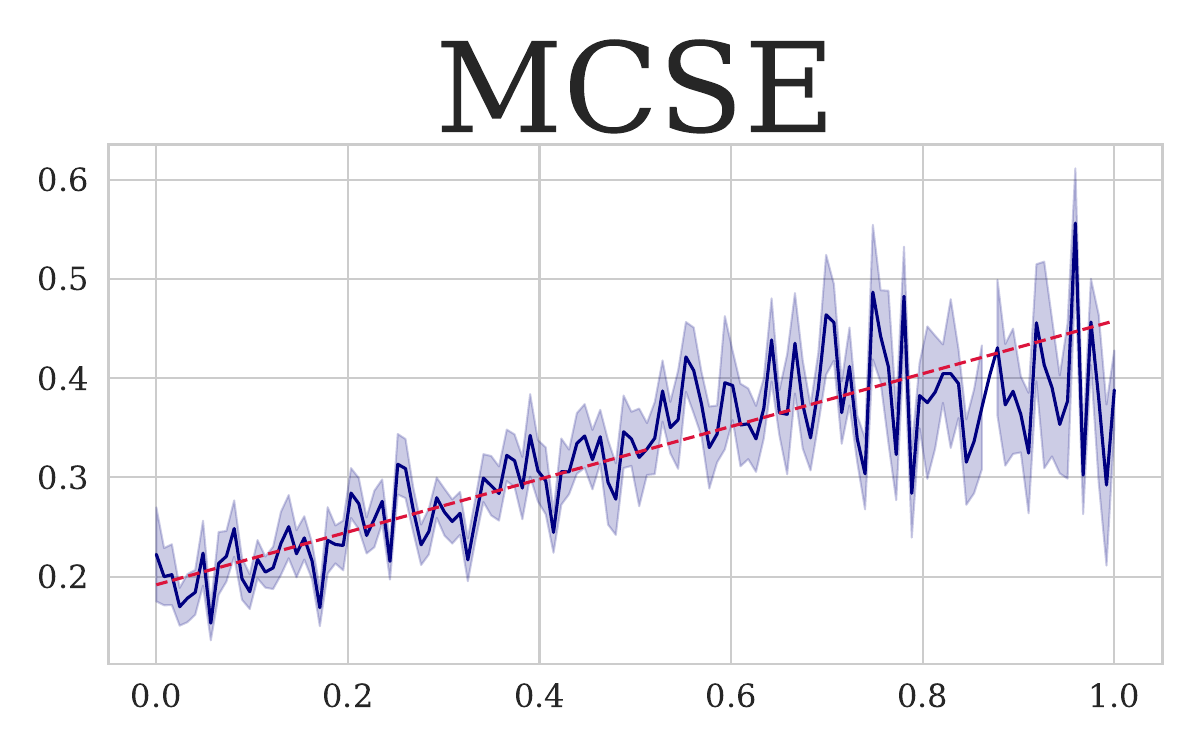}
    \end{subfigure}
    \begin{subfigure}{0.15\textwidth}
        \includegraphics[width=\linewidth]{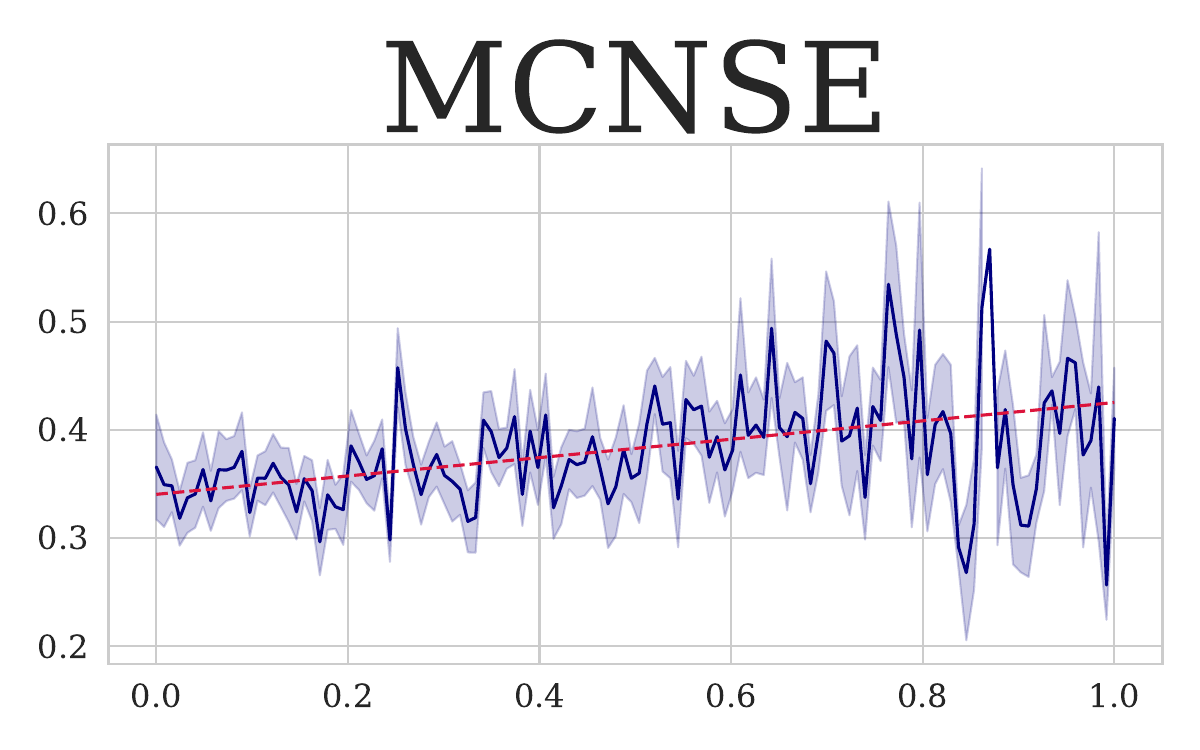}
    \end{subfigure}
    \begin{subfigure}{0.15\textwidth}
        \includegraphics[width=\linewidth]{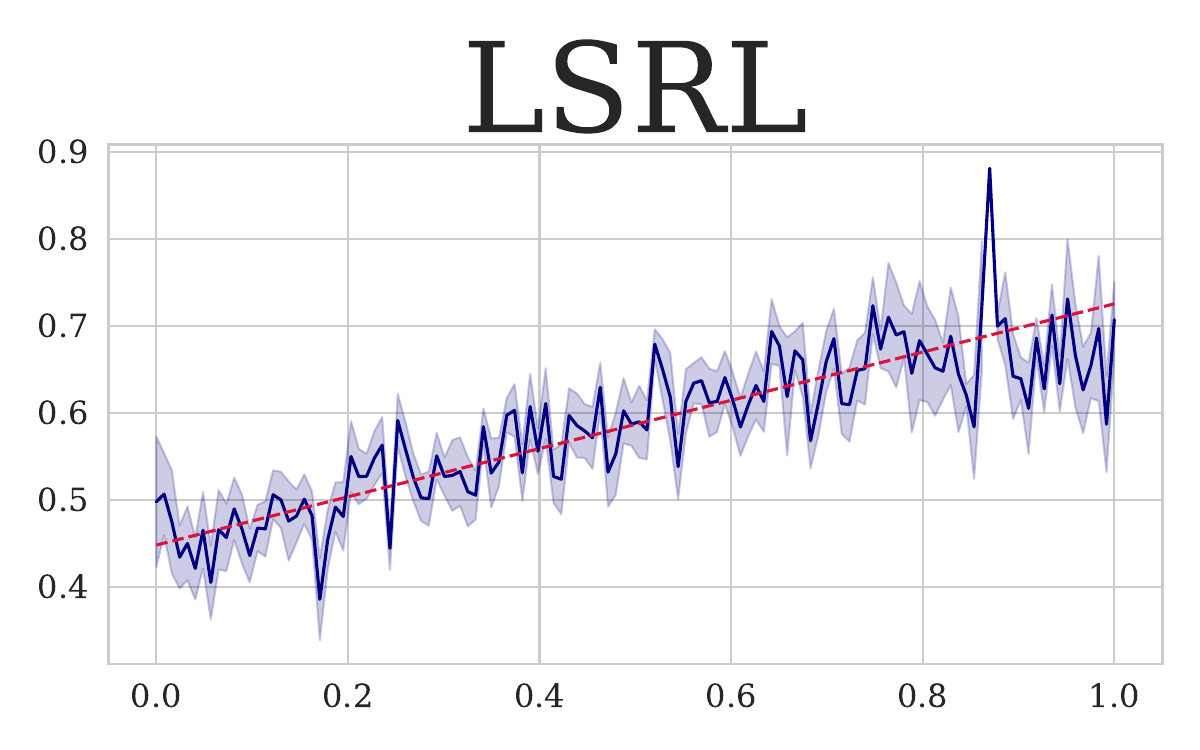}
    \end{subfigure}
    \vspace{0.3em}

        \caption{Uncertainty metric trends for model \textbf{GEMMA} across all datasets.}
    \label{fig:ue_metrics_gemma}
\end{figure*}

\begin{figure*}[h!]
    \centering
    \vspace{-0.5em}
    {\centering \textbf{\small WMT14 De-En} \par}
    \vspace{0.2em}
    \begin{subfigure}{0.15\textwidth}
        \includegraphics[width=\linewidth]{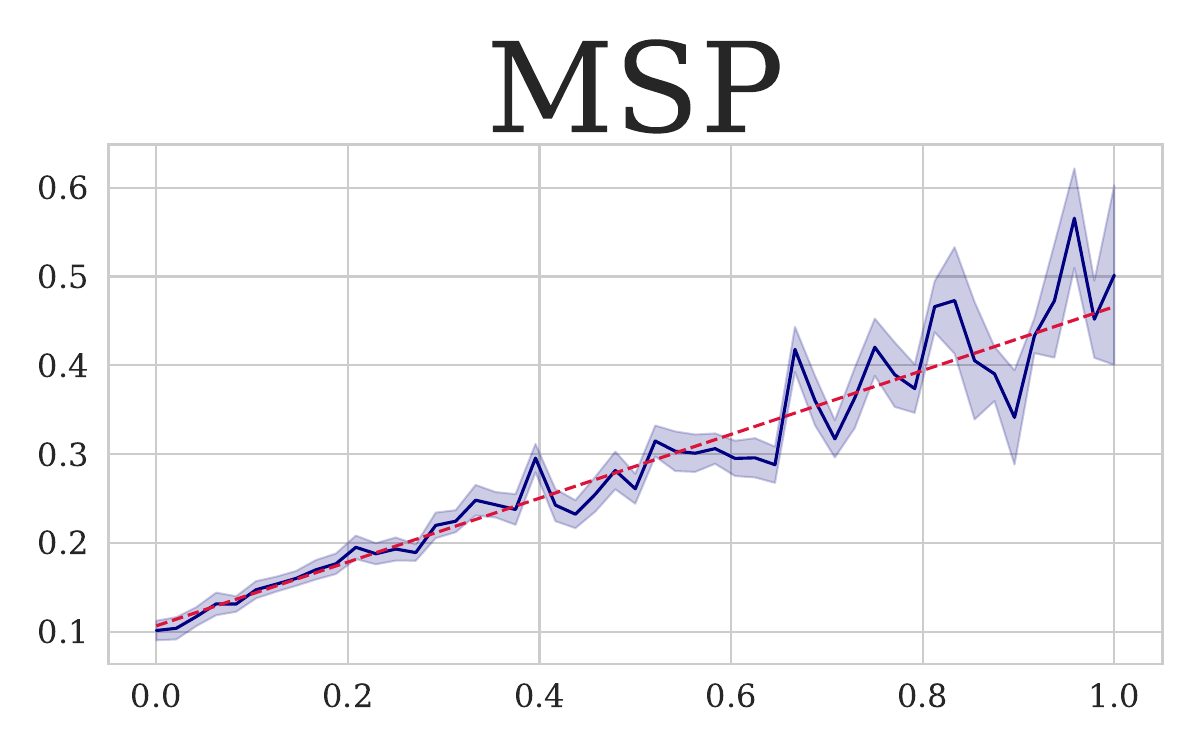}
    \end{subfigure}
    \begin{subfigure}{0.15\textwidth}
        \includegraphics[width=\linewidth]{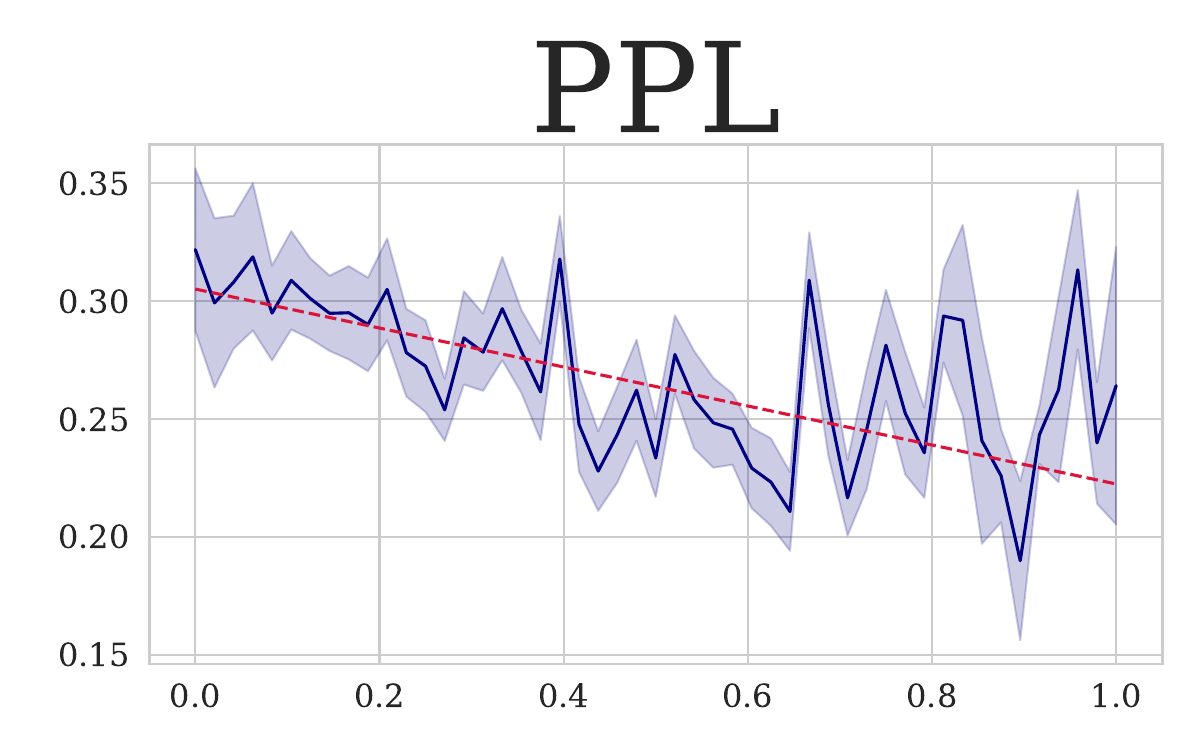}
    \end{subfigure}
    \begin{subfigure}{0.15\textwidth}
        \includegraphics[width=\linewidth]{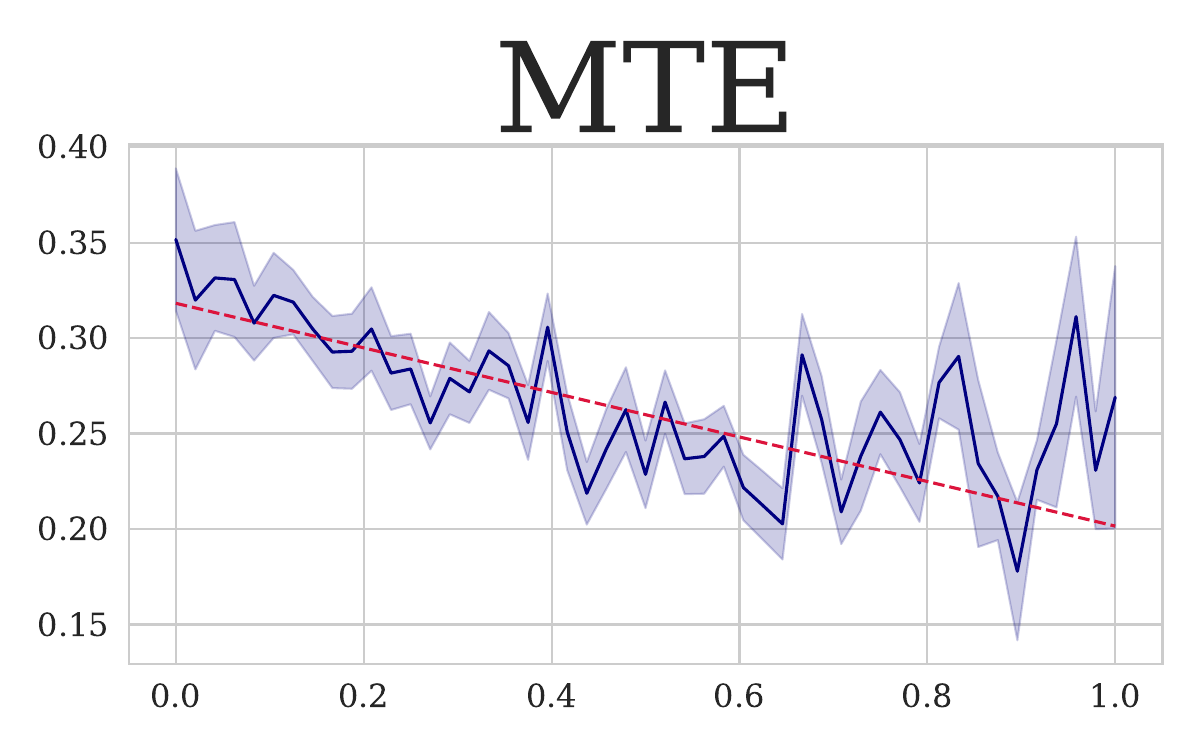}
    \end{subfigure}
    \begin{subfigure}{0.15\textwidth}
        \includegraphics[width=\linewidth]{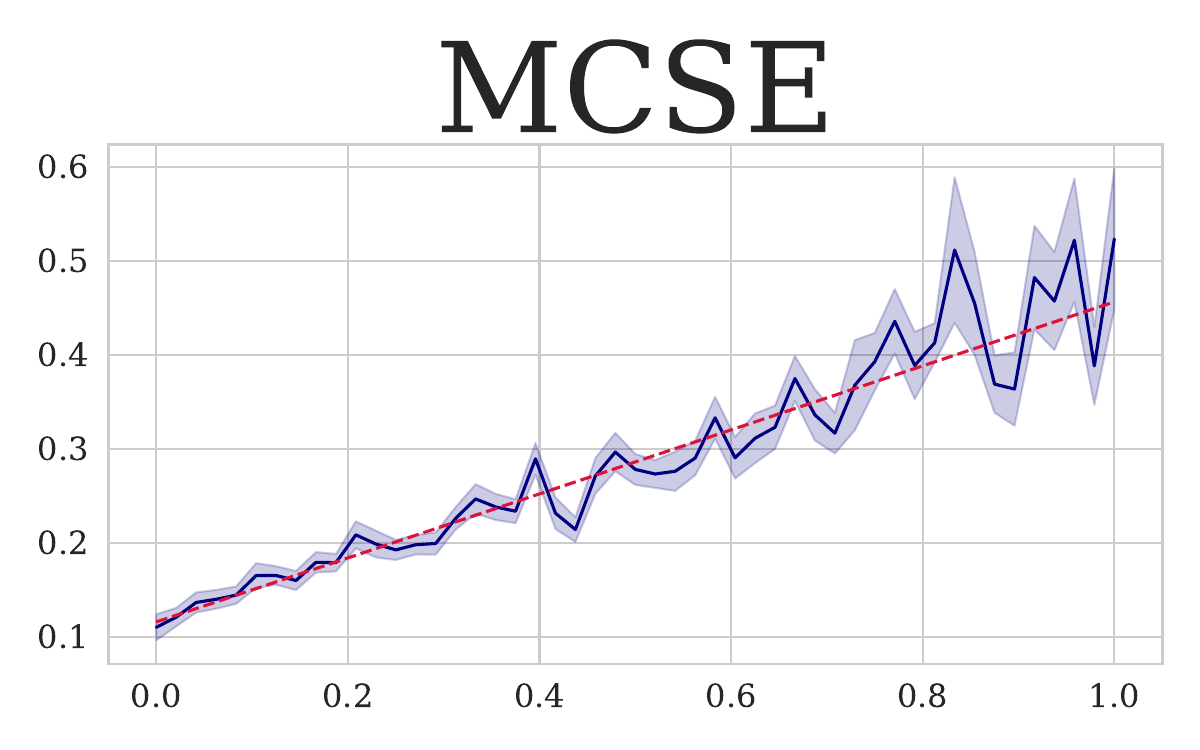}
    \end{subfigure}
    \begin{subfigure}{0.15\textwidth}
        \includegraphics[width=\linewidth]{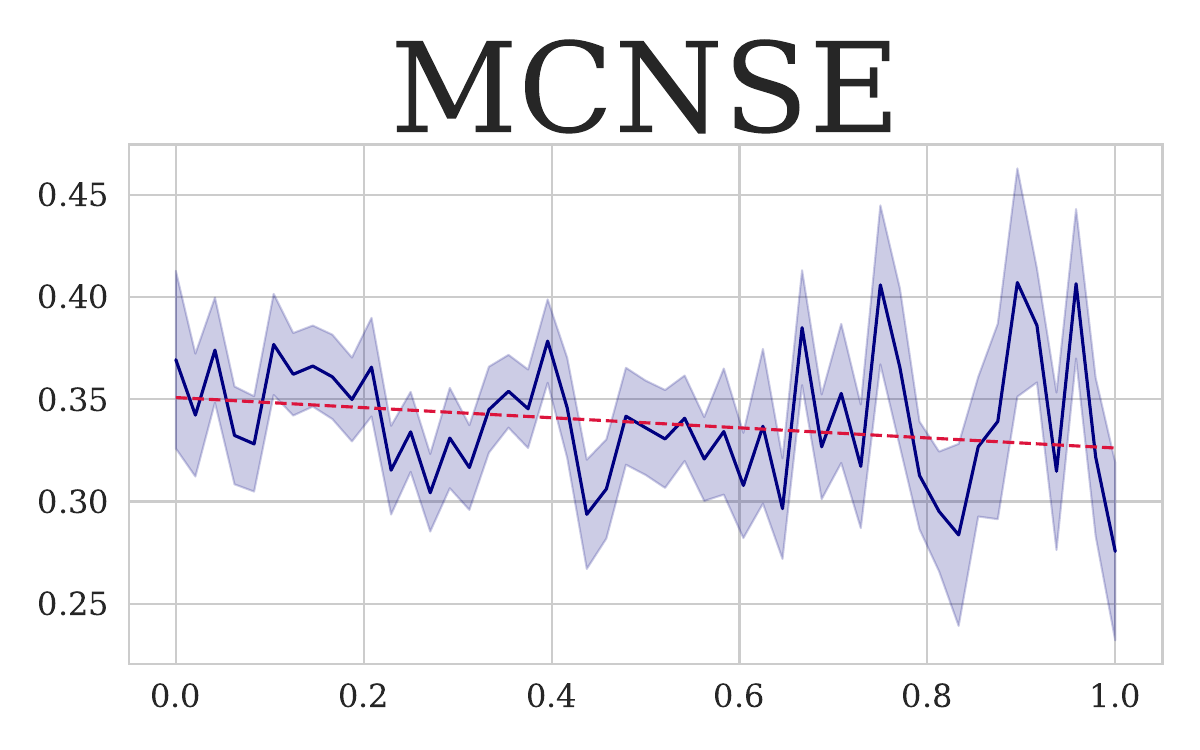}
    \end{subfigure}
    \begin{subfigure}{0.15\textwidth}
        \includegraphics[width=\linewidth]{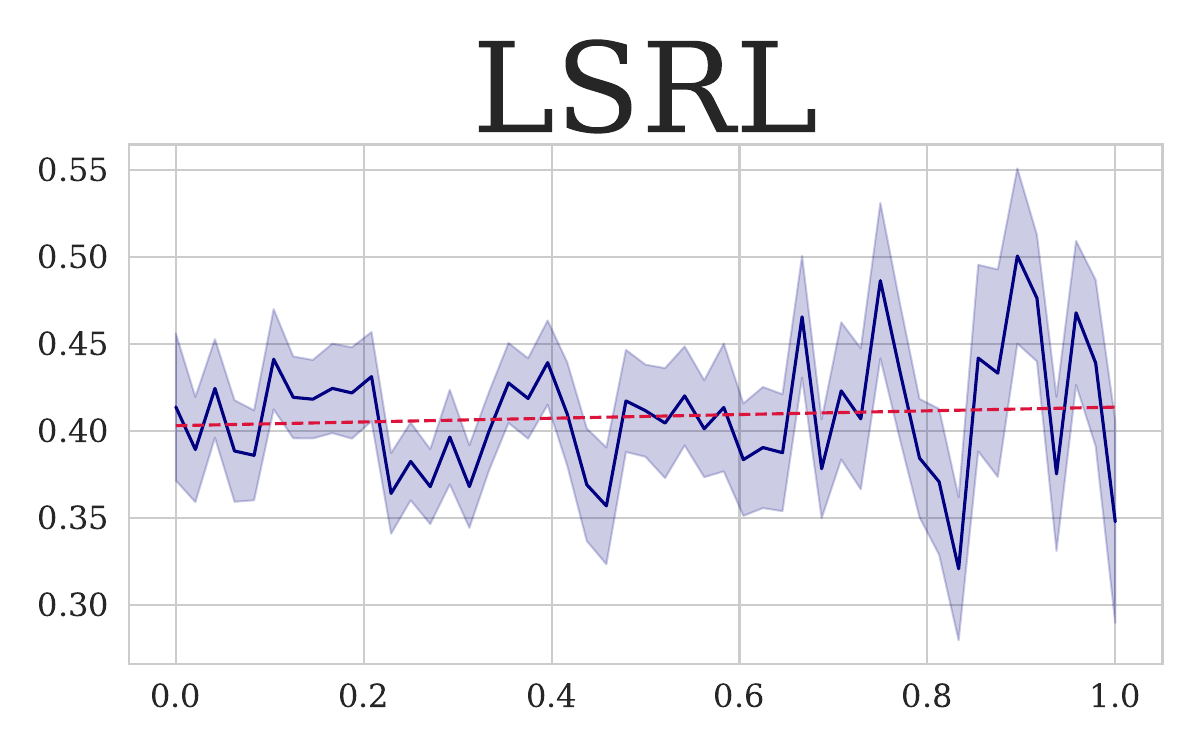}
    \end{subfigure}
    \vspace{0.3em}
    {\centering \textbf{\small WMT14 Fr-En} \par}
    \vspace{0.2em}
    \begin{subfigure}{0.15\textwidth}
        \includegraphics[width=\linewidth]{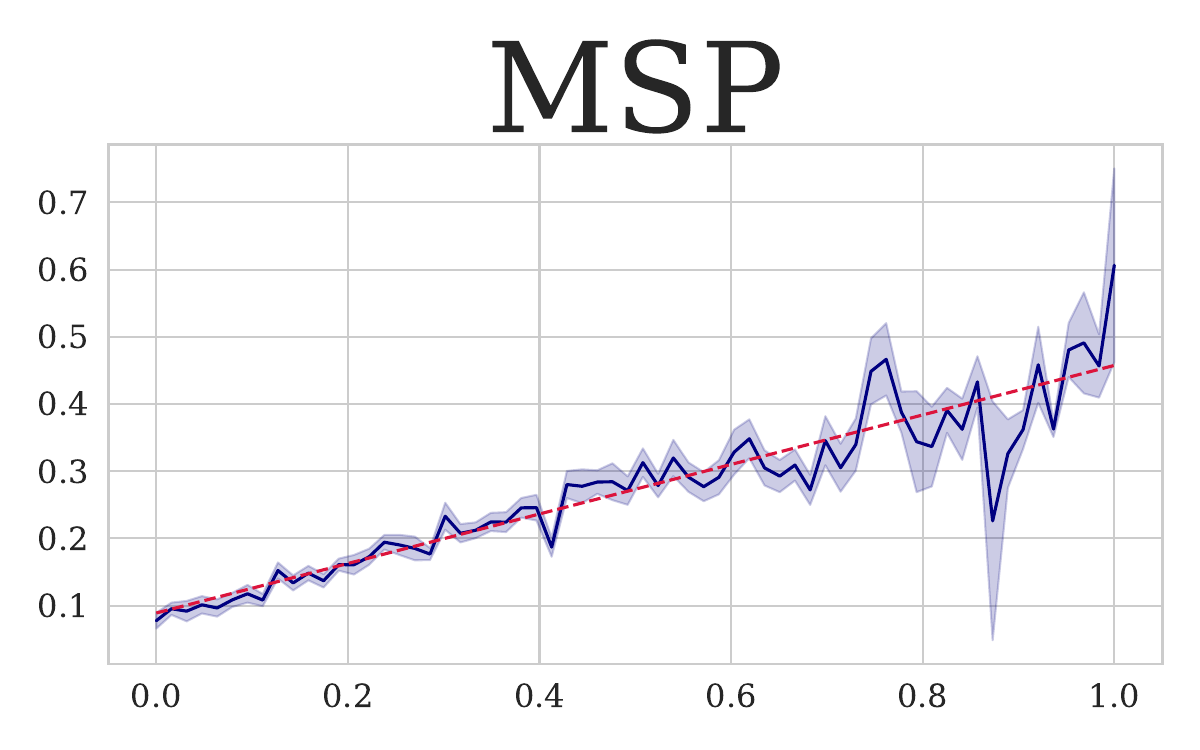}
    \end{subfigure}
    \begin{subfigure}{0.15\textwidth}
        \includegraphics[width=\linewidth]{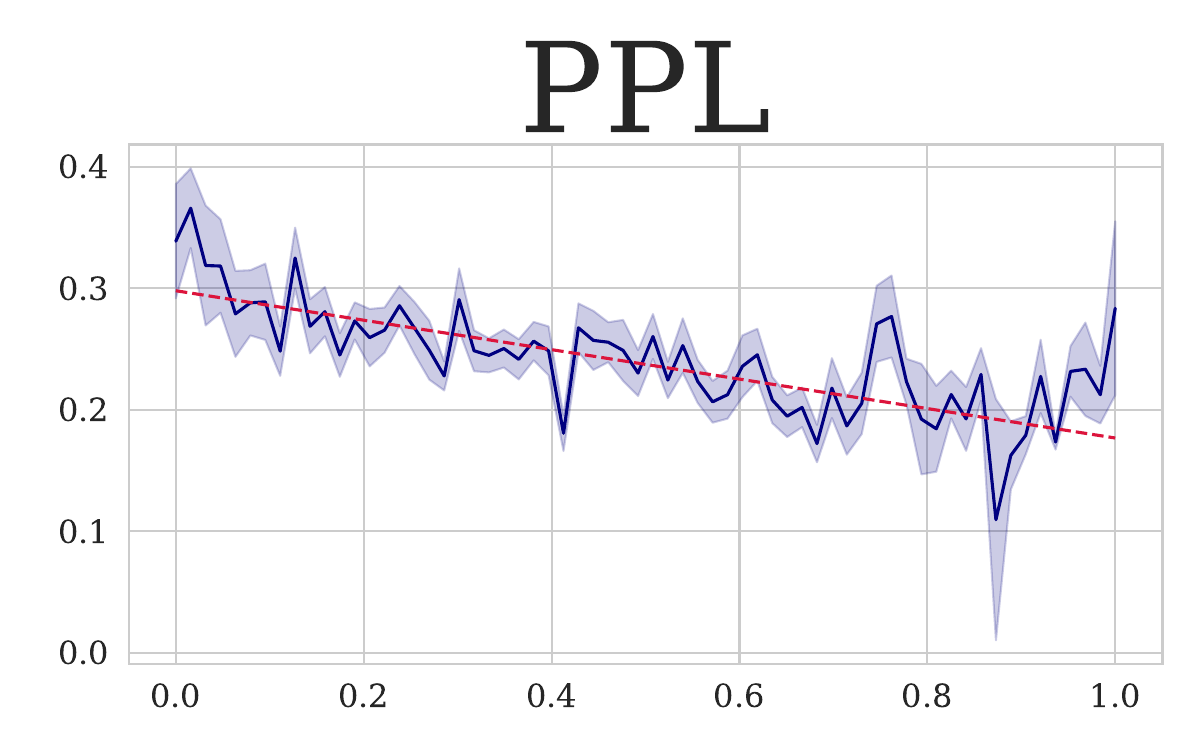}
    \end{subfigure}
    \begin{subfigure}{0.15\textwidth}
        \includegraphics[width=\linewidth]{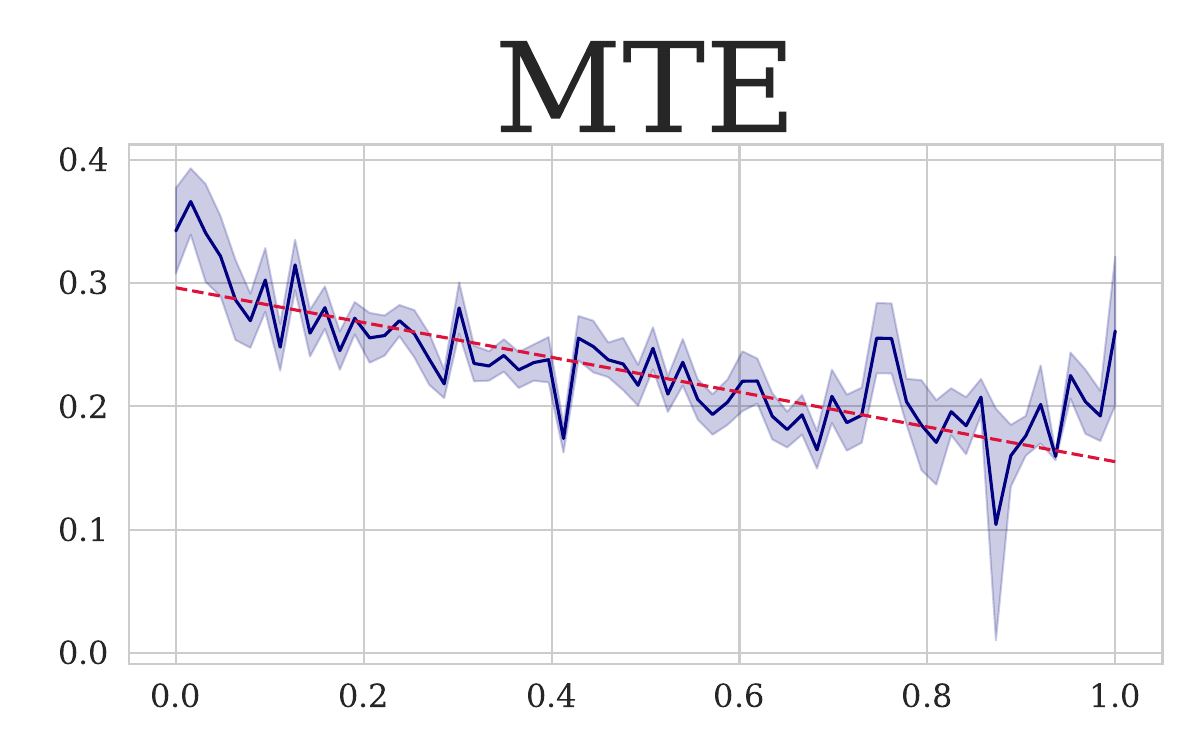}
    \end{subfigure}
    \begin{subfigure}{0.15\textwidth}
        \includegraphics[width=\linewidth]{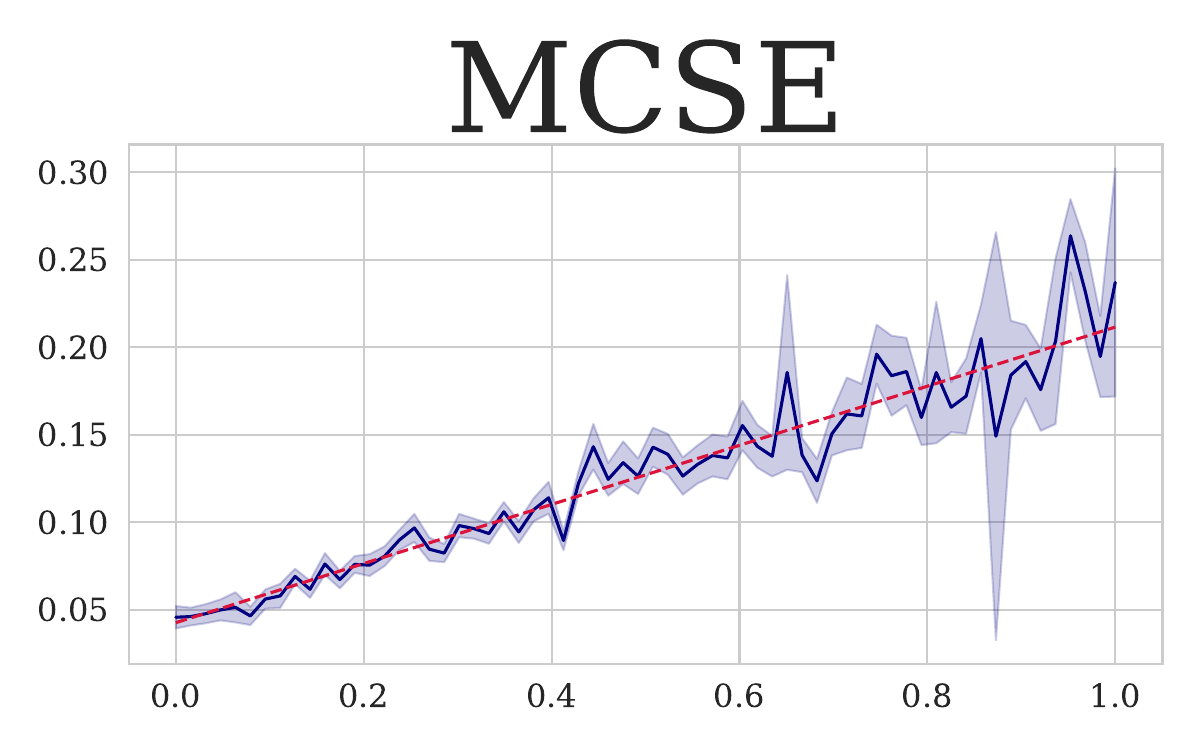}
    \end{subfigure}
    \begin{subfigure}{0.15\textwidth}
        \includegraphics[width=\linewidth]{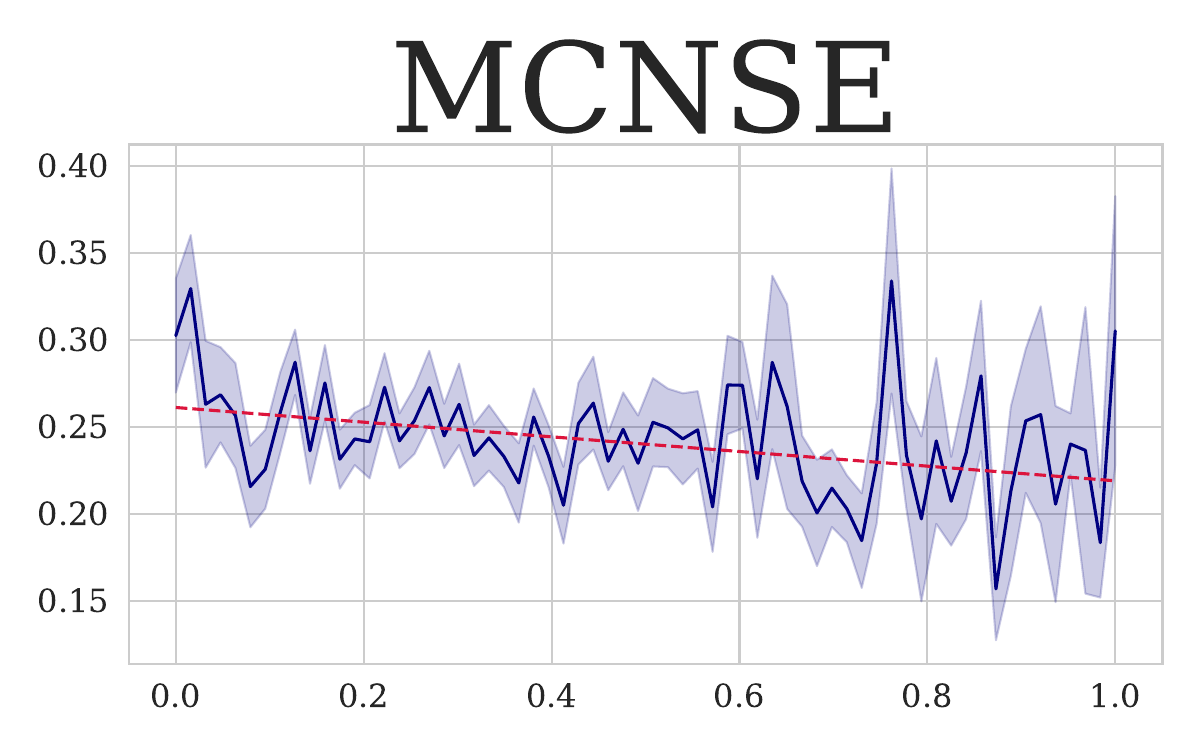}
    \end{subfigure}
    \begin{subfigure}{0.15\textwidth}
        \includegraphics[width=\linewidth]{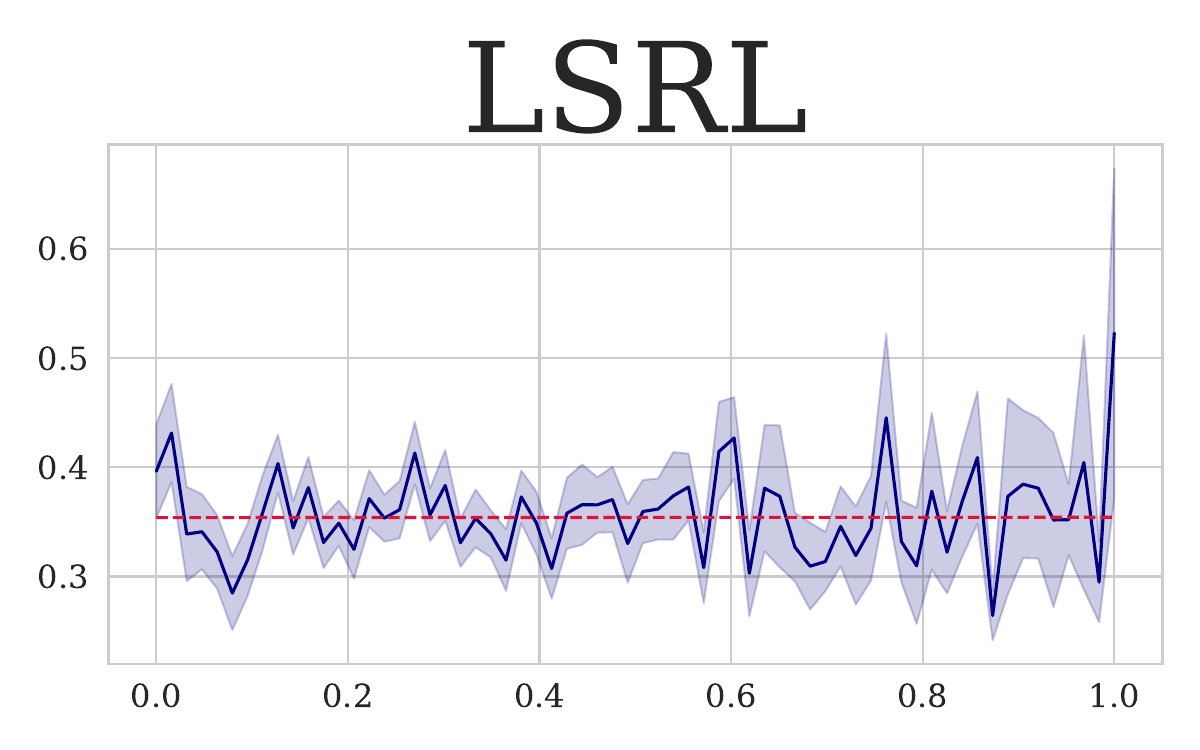}
    \end{subfigure}
    \vspace{0.3em}
    {\centering \textbf{\small WMT14 Cs-En} \par}
    \vspace{0.2em}
    \begin{subfigure}{0.15\textwidth}
        \includegraphics[width=\linewidth]{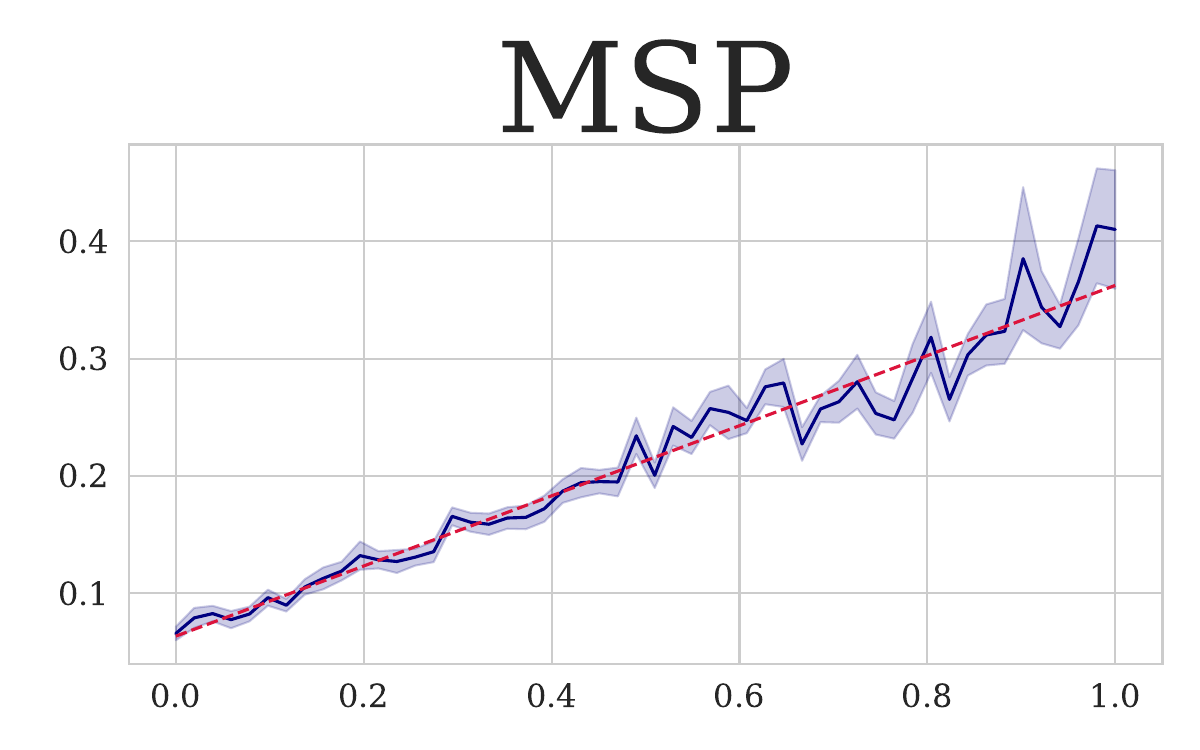}
    \end{subfigure}
    \begin{subfigure}{0.15\textwidth}
        \includegraphics[width=\linewidth]{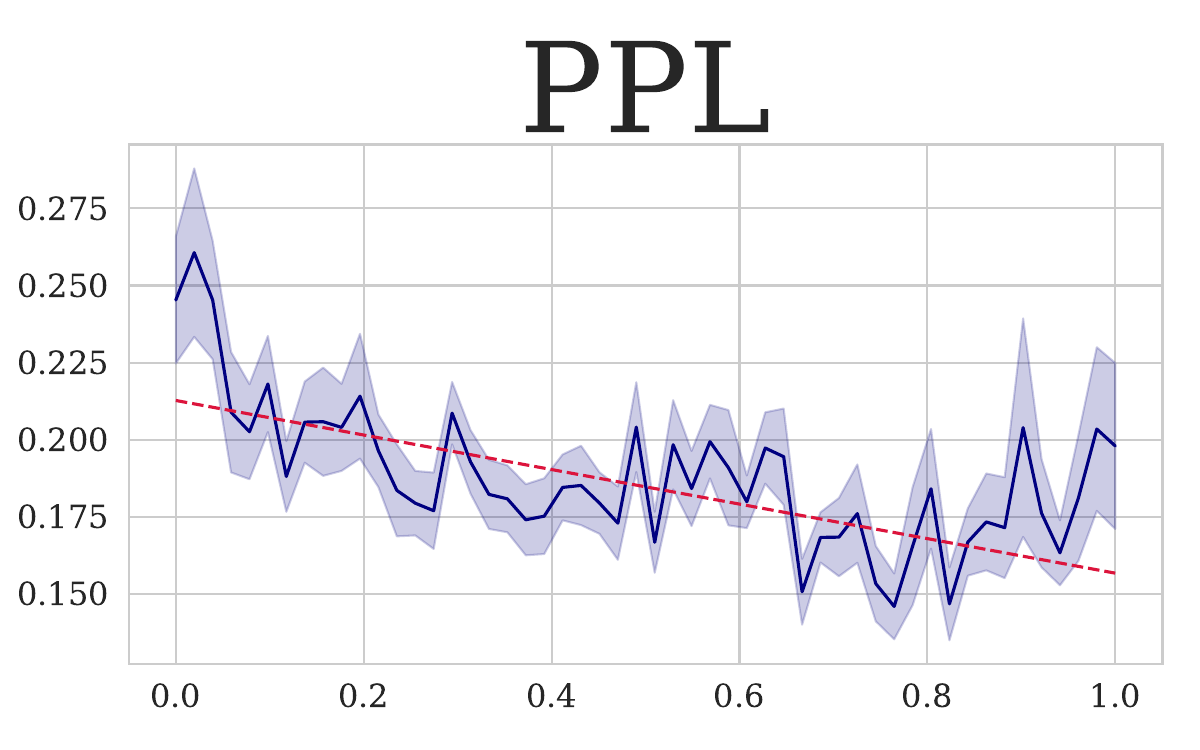}
    \end{subfigure}
    \begin{subfigure}{0.15\textwidth}
        \includegraphics[width=\linewidth]{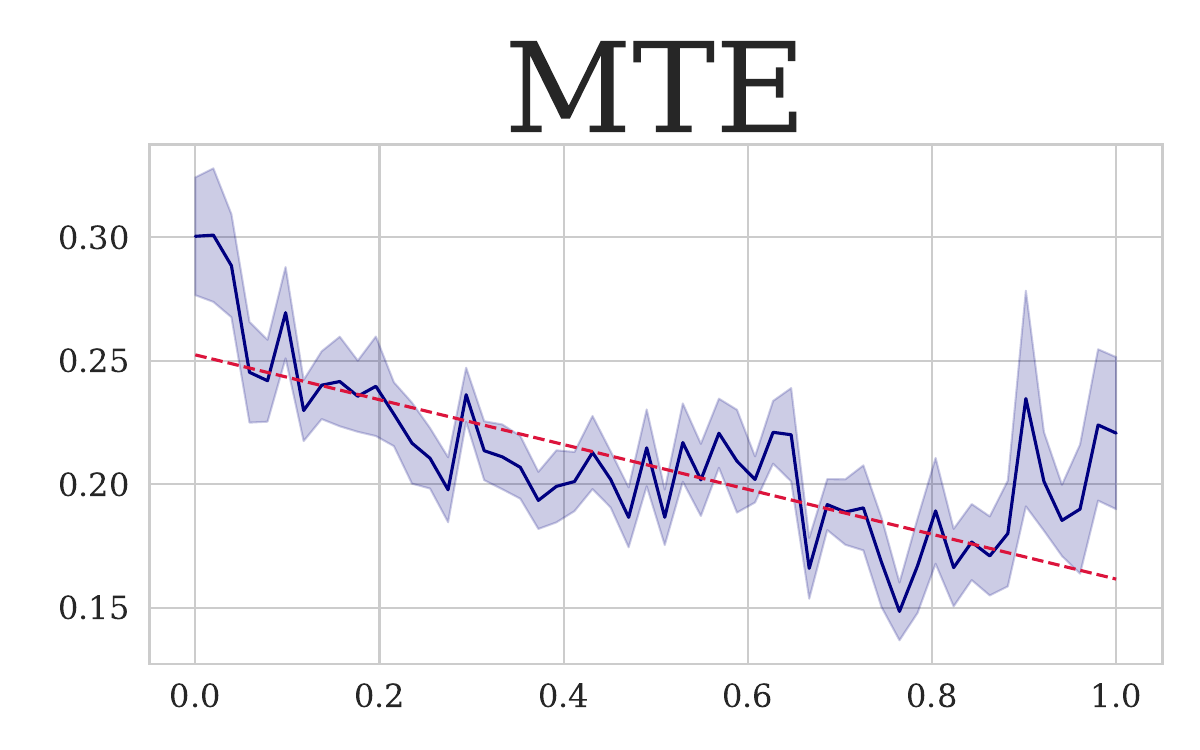}
    \end{subfigure}
    \begin{subfigure}{0.15\textwidth}
        \includegraphics[width=\linewidth]{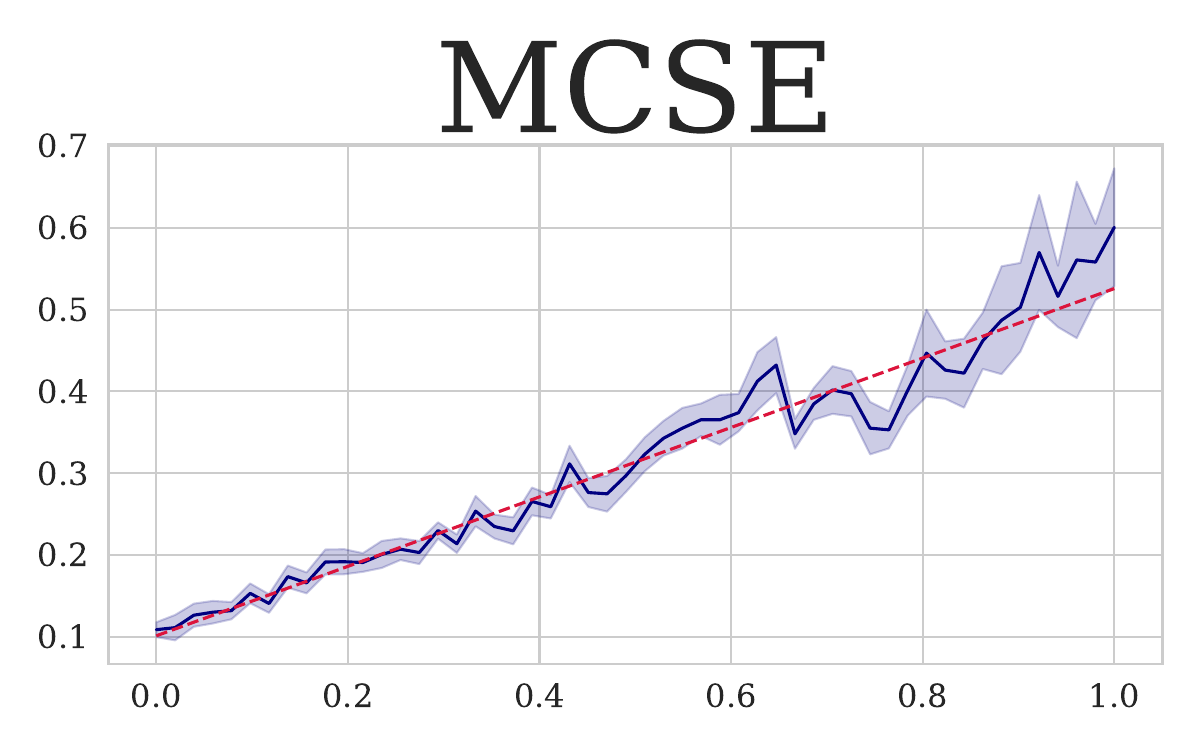}
    \end{subfigure}
    \begin{subfigure}{0.15\textwidth}
        \includegraphics[width=\linewidth]{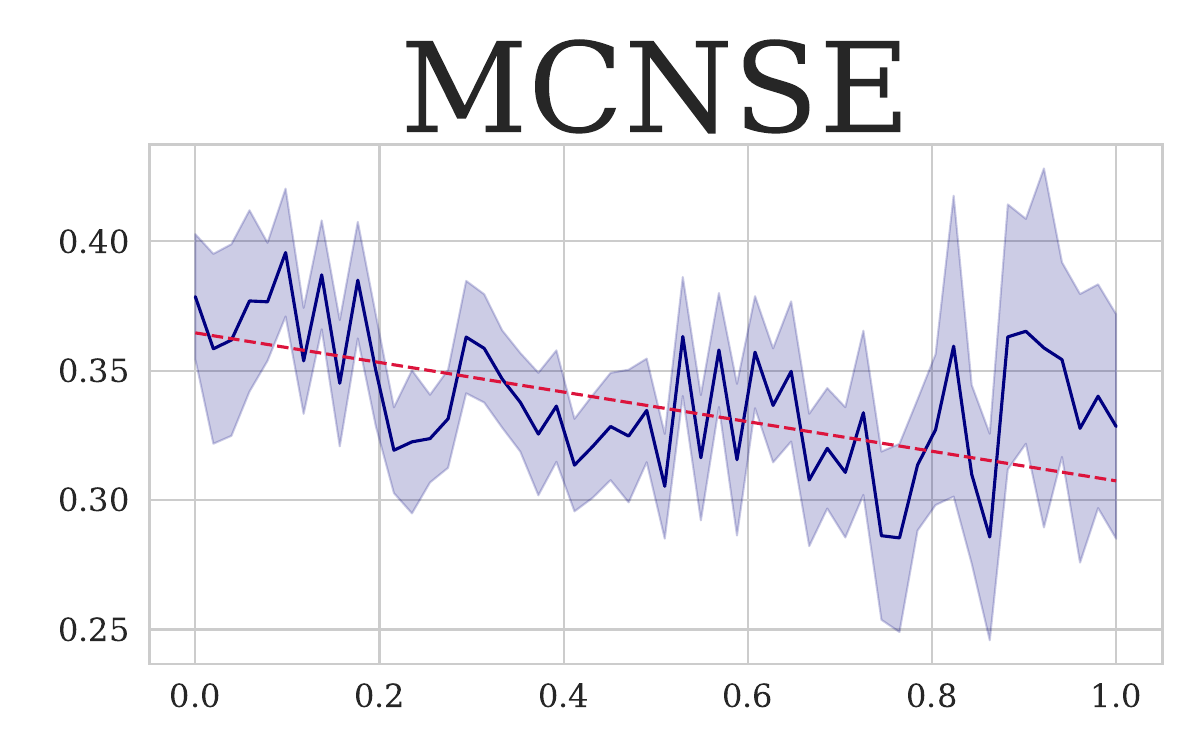}
    \end{subfigure}
    \begin{subfigure}{0.15\textwidth}
        \includegraphics[width=\linewidth]{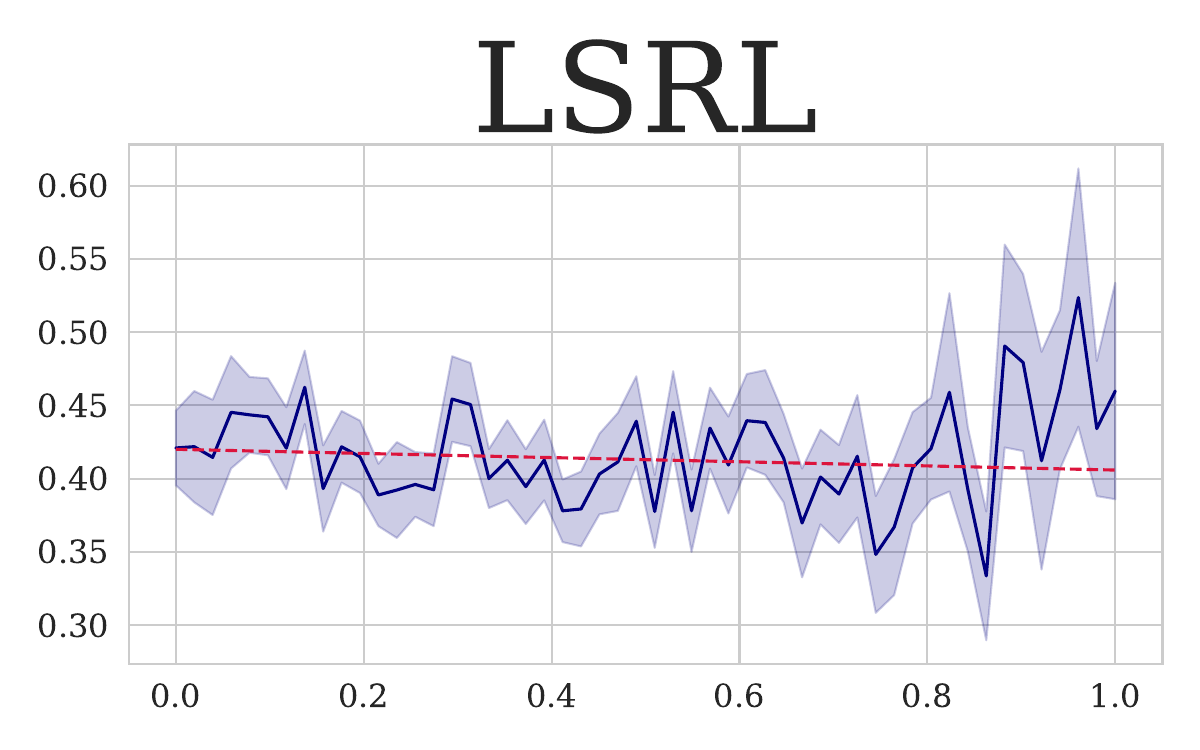}
    \end{subfigure}
    \vspace{0.3em}
    {\centering \textbf{\small WMT14 Ru-En} \par}
    \vspace{0.2em}
    \begin{subfigure}{0.15\textwidth}
        \includegraphics[width=\linewidth]{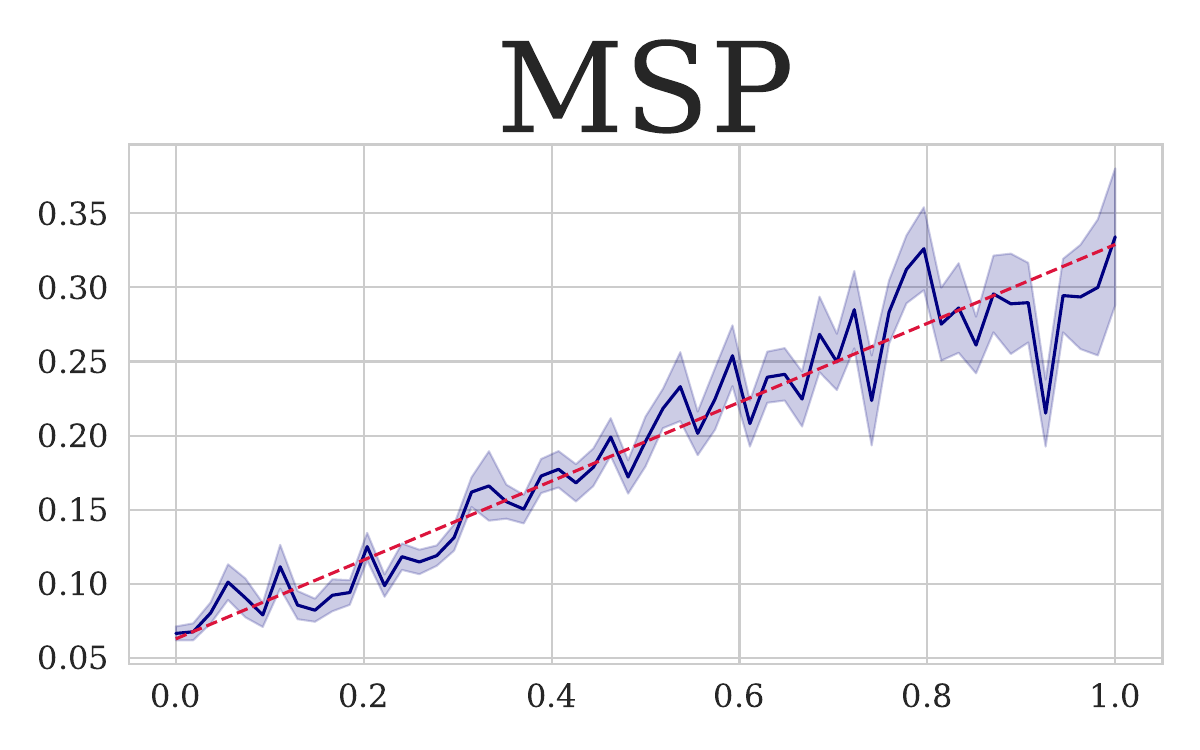}
    \end{subfigure}
    \begin{subfigure}{0.15\textwidth}
        \includegraphics[width=\linewidth]{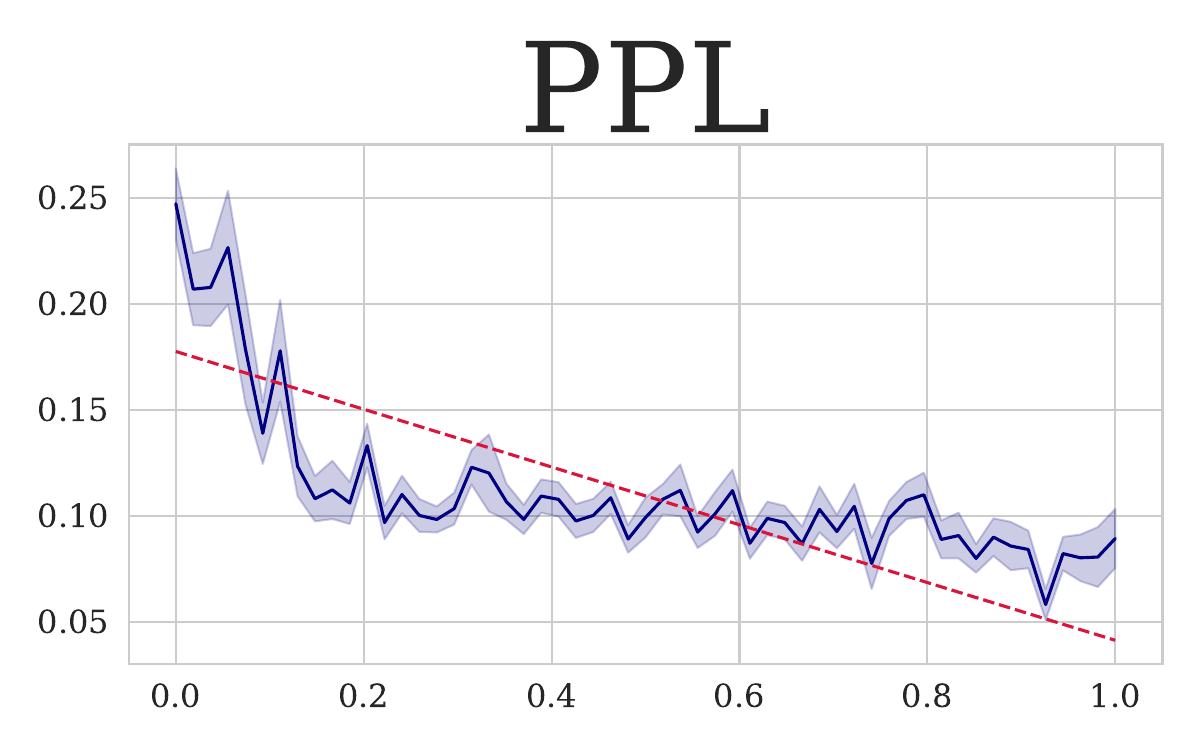}
    \end{subfigure}
    \begin{subfigure}{0.15\textwidth}
        \includegraphics[width=\linewidth]{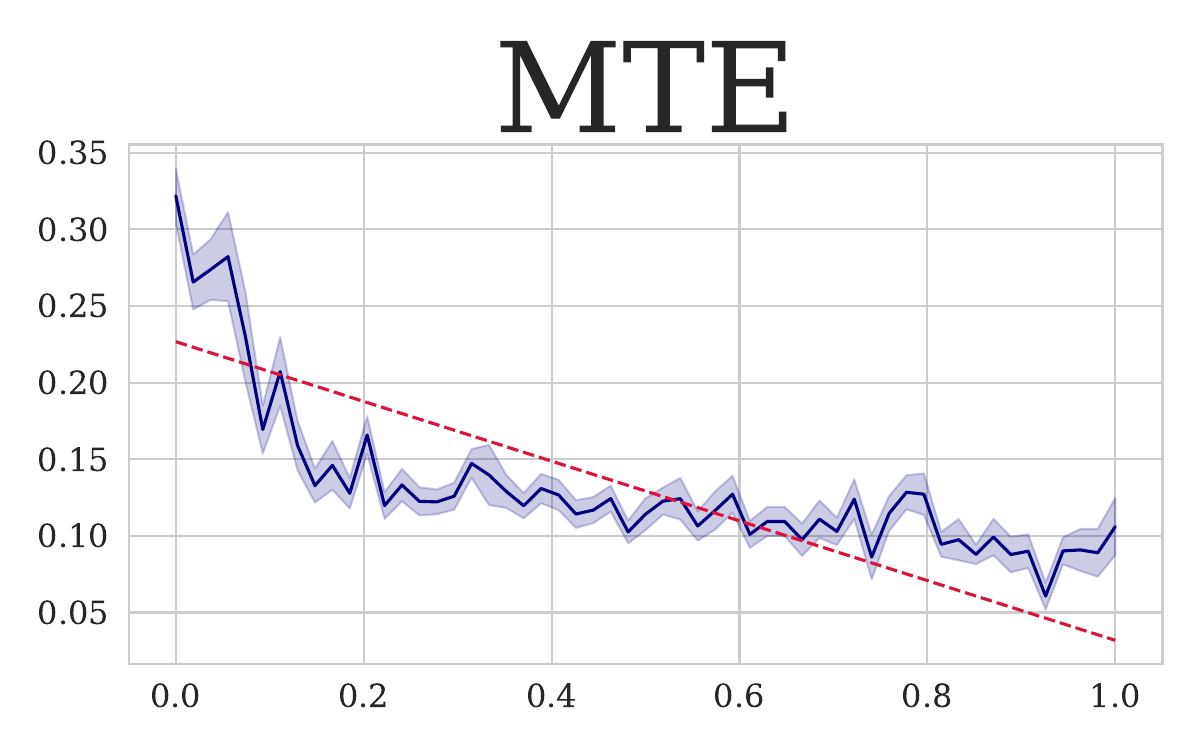}
    \end{subfigure}
    \begin{subfigure}{0.15\textwidth}
        \includegraphics[width=\linewidth]{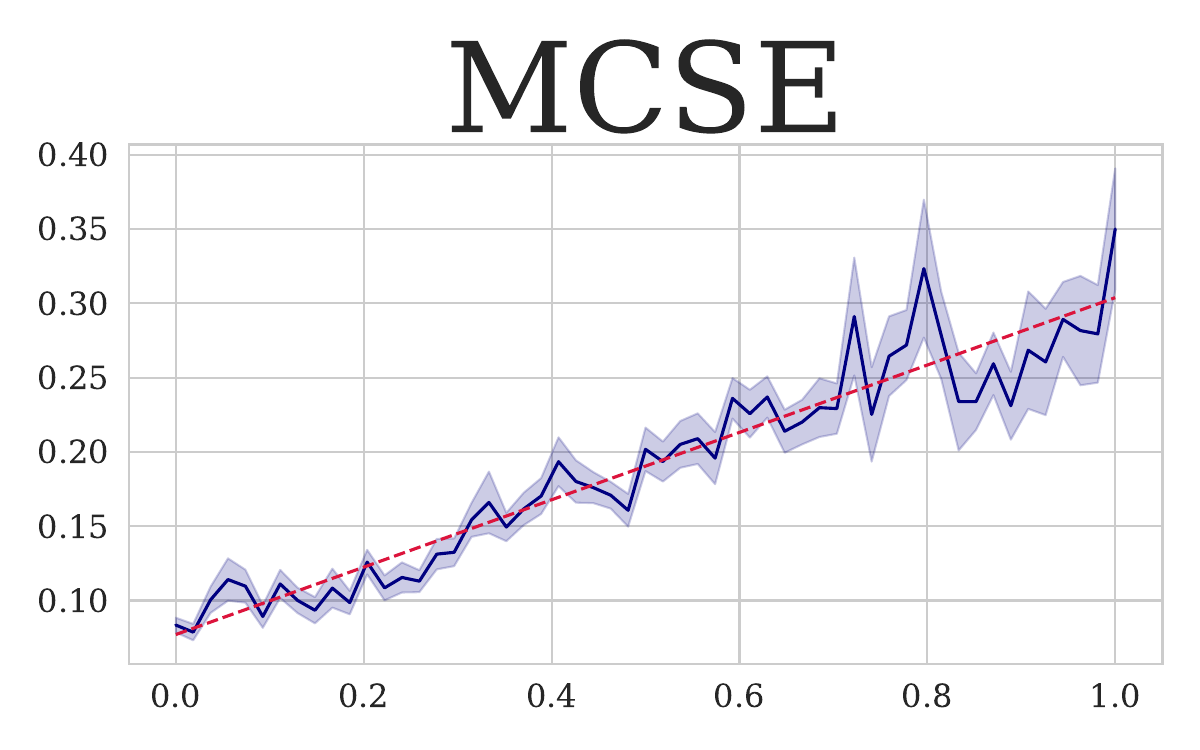}
    \end{subfigure}
    \begin{subfigure}{0.15\textwidth}
        \includegraphics[width=\linewidth]{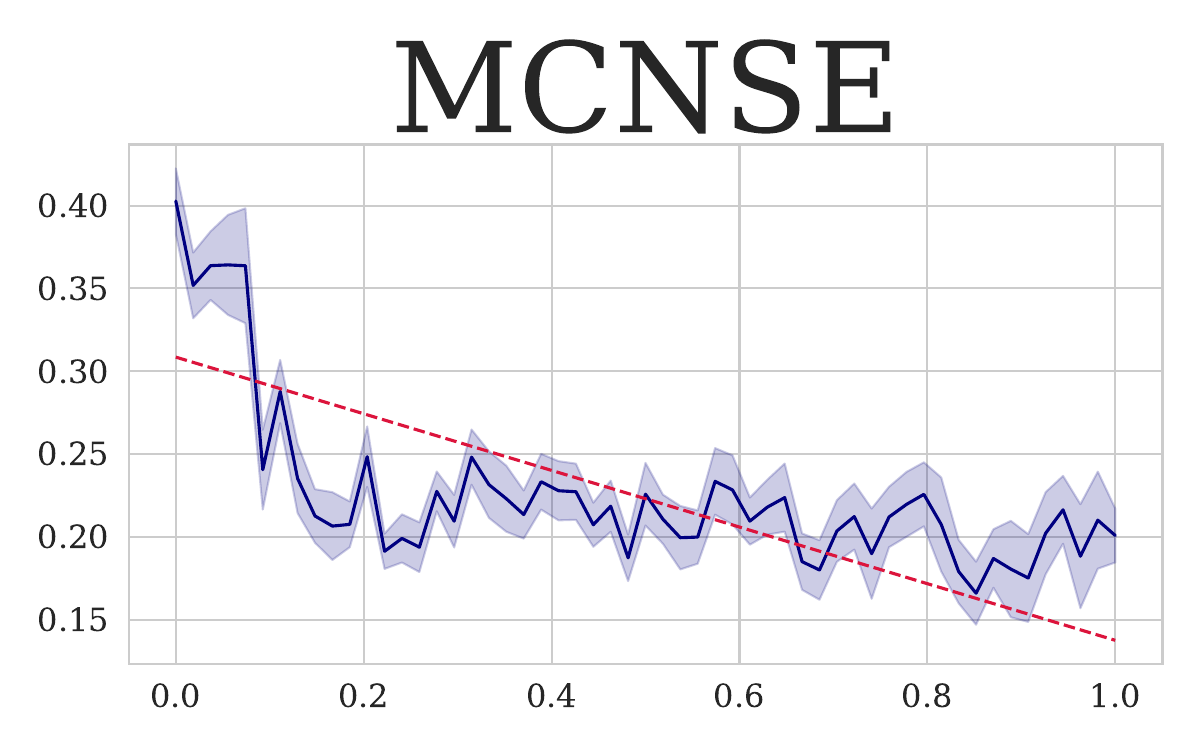}
    \end{subfigure}
    \begin{subfigure}{0.15\textwidth}
        \includegraphics[width=\linewidth]{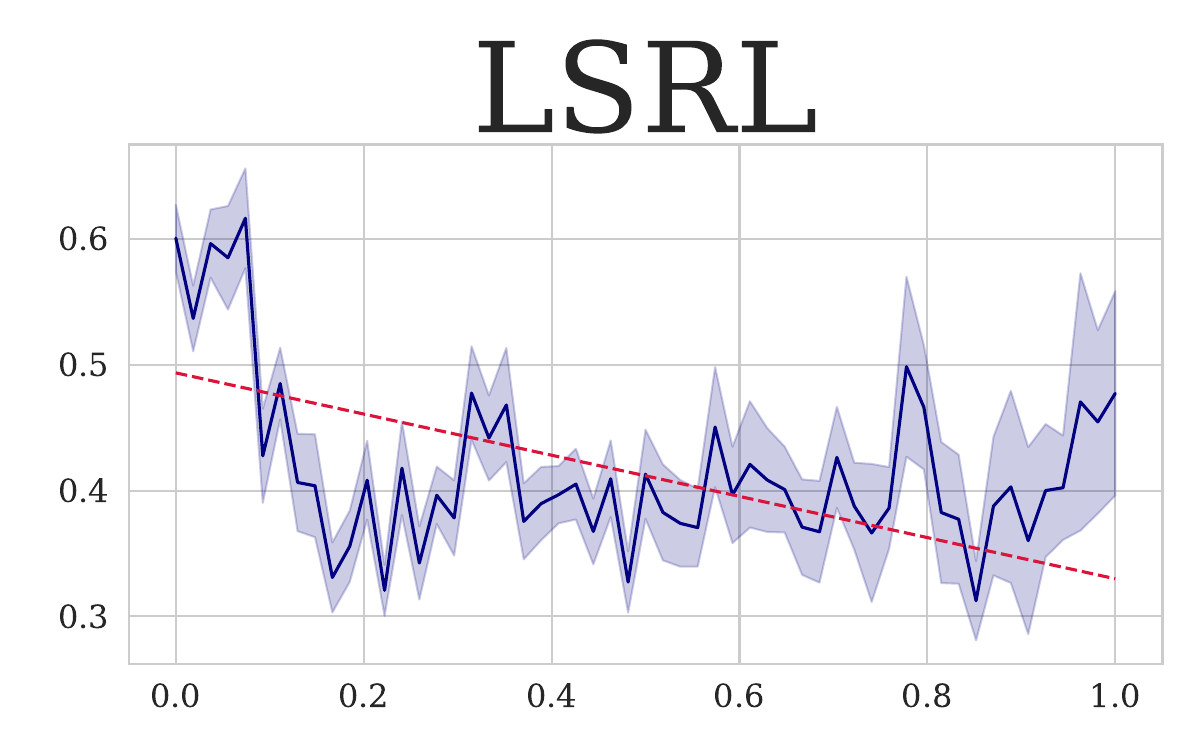}
    \end{subfigure}
    \vspace{0.3em}
    {\centering \textbf{\small WMT19 Ru-En} \par}
    \vspace{0.2em}
    \begin{subfigure}{0.15\textwidth}
        \includegraphics[width=\linewidth]{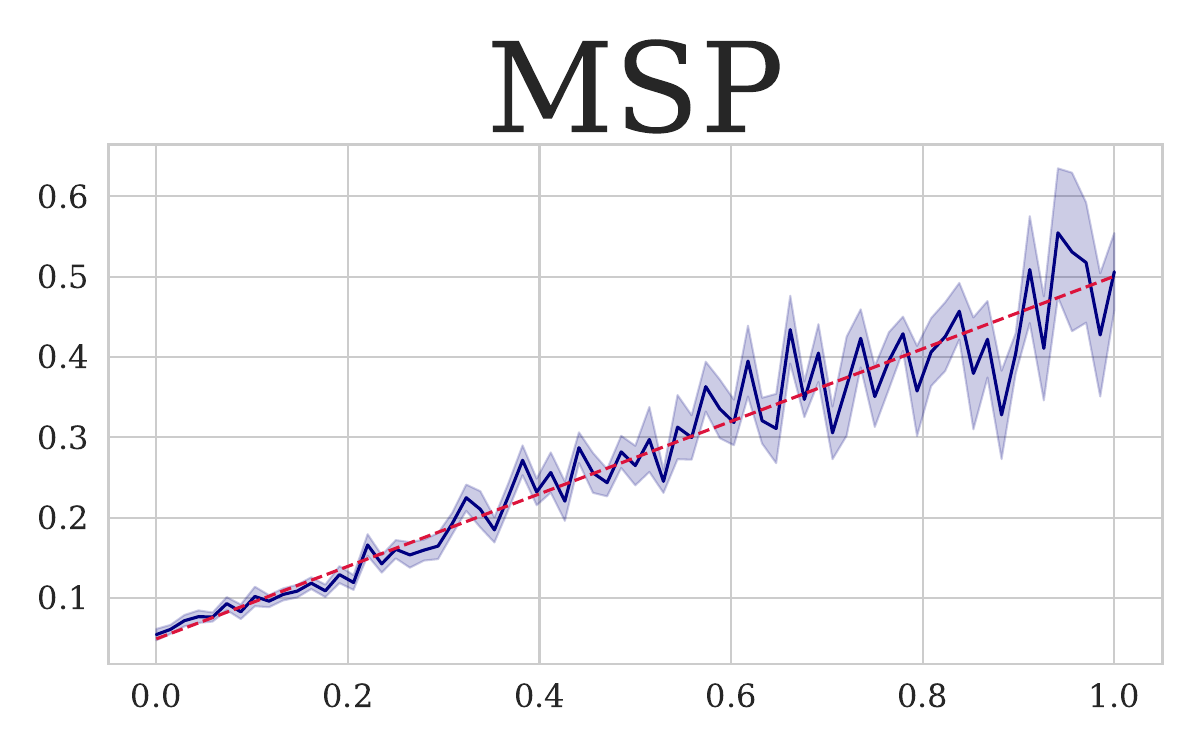}
    \end{subfigure}
    \begin{subfigure}{0.15\textwidth}
        \includegraphics[width=\linewidth]{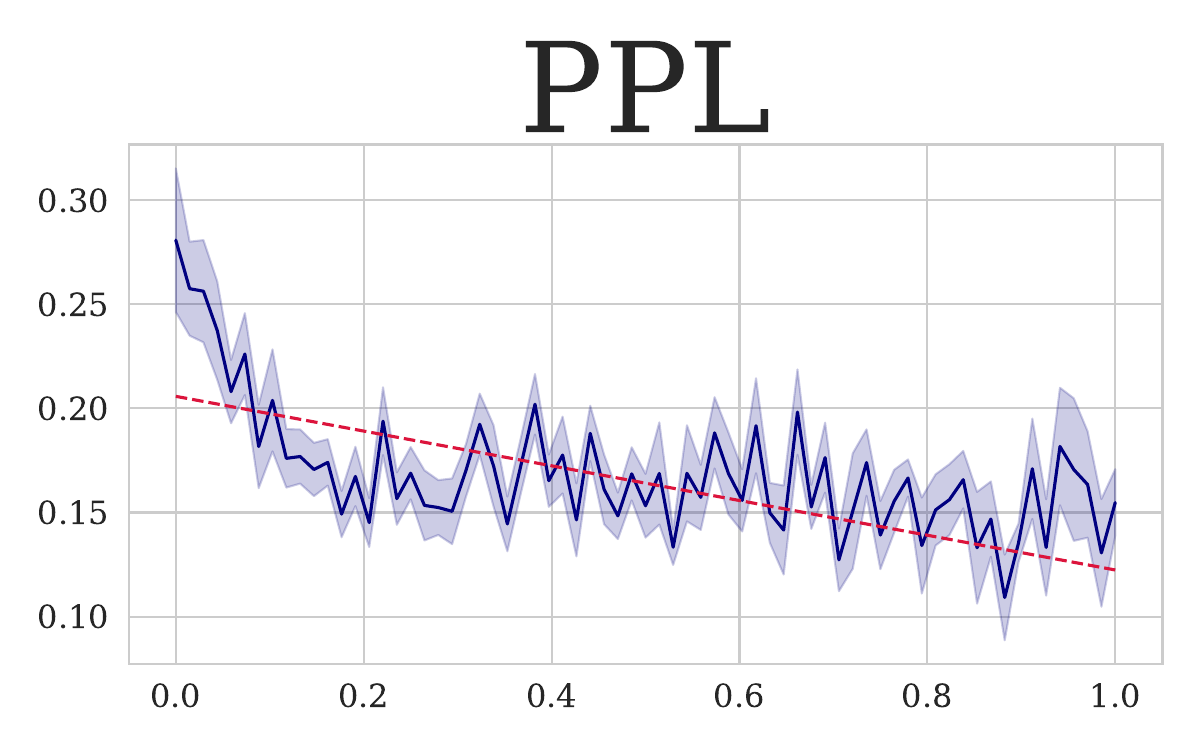}
    \end{subfigure}
    \begin{subfigure}{0.15\textwidth}
        \includegraphics[width=\linewidth]{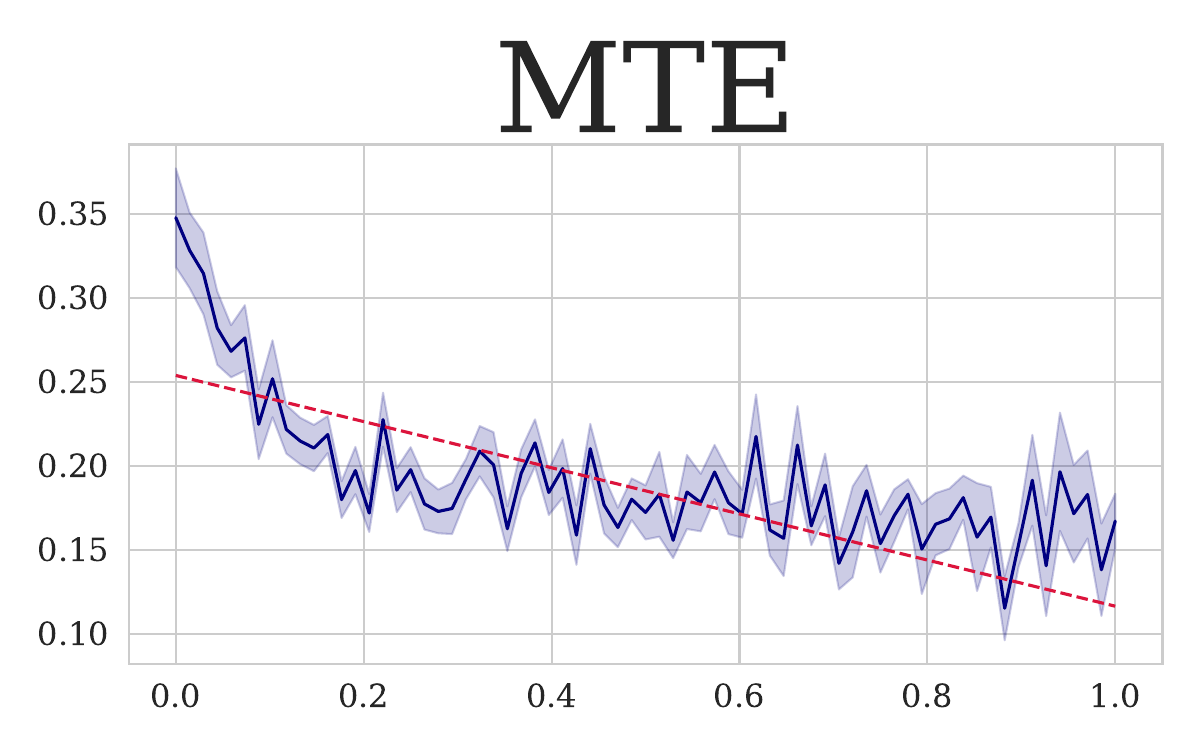}
    \end{subfigure}
    \begin{subfigure}{0.15\textwidth}
        \includegraphics[width=\linewidth]{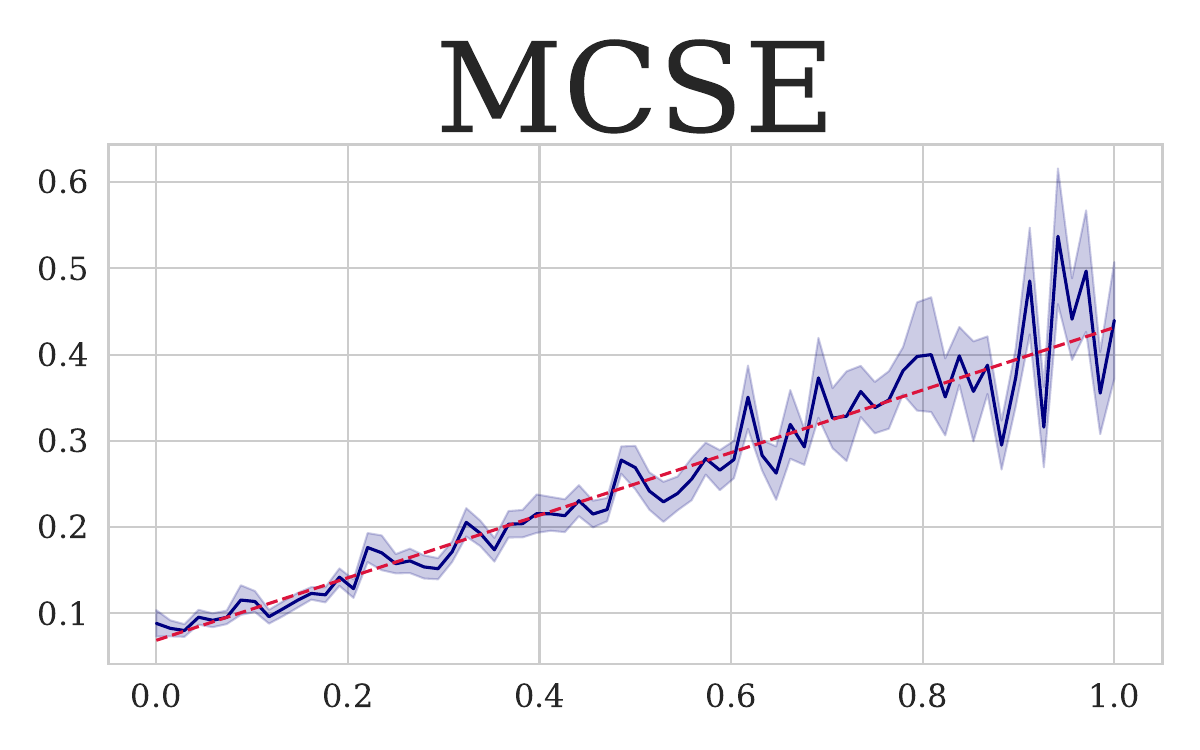}
    \end{subfigure}
    \begin{subfigure}{0.15\textwidth}
        \includegraphics[width=\linewidth]{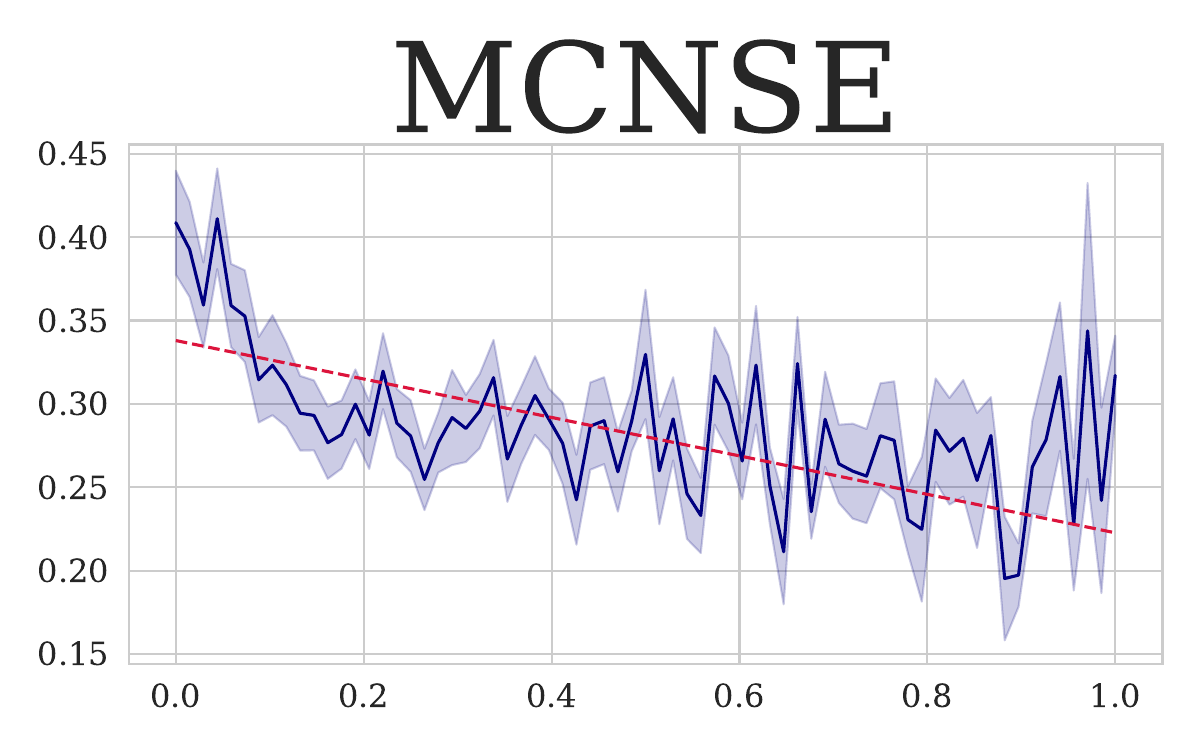}
    \end{subfigure}
    \begin{subfigure}{0.15\textwidth}
        \includegraphics[width=\linewidth]{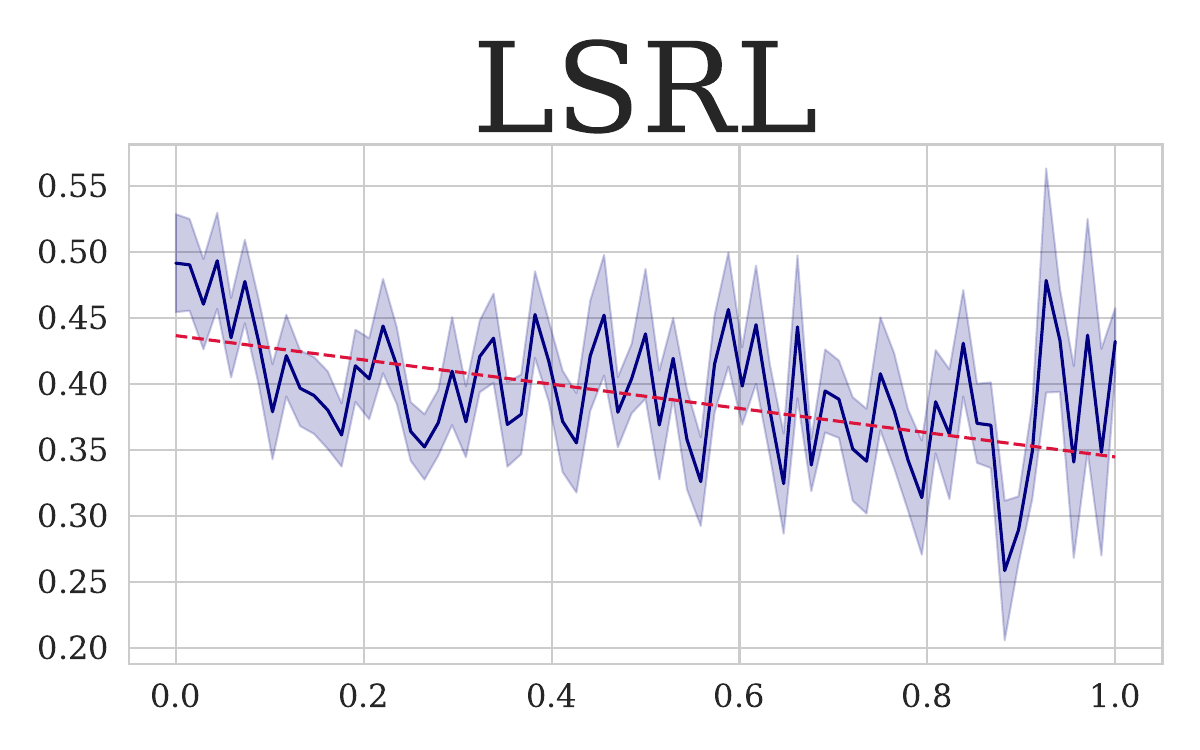}
    \end{subfigure}
    \vspace{0.3em}
    {\centering \textbf{\small WMT19 Fi-En} \par}
    \vspace{0.2em}
    \begin{subfigure}{0.15\textwidth}
        \includegraphics[width=\linewidth]{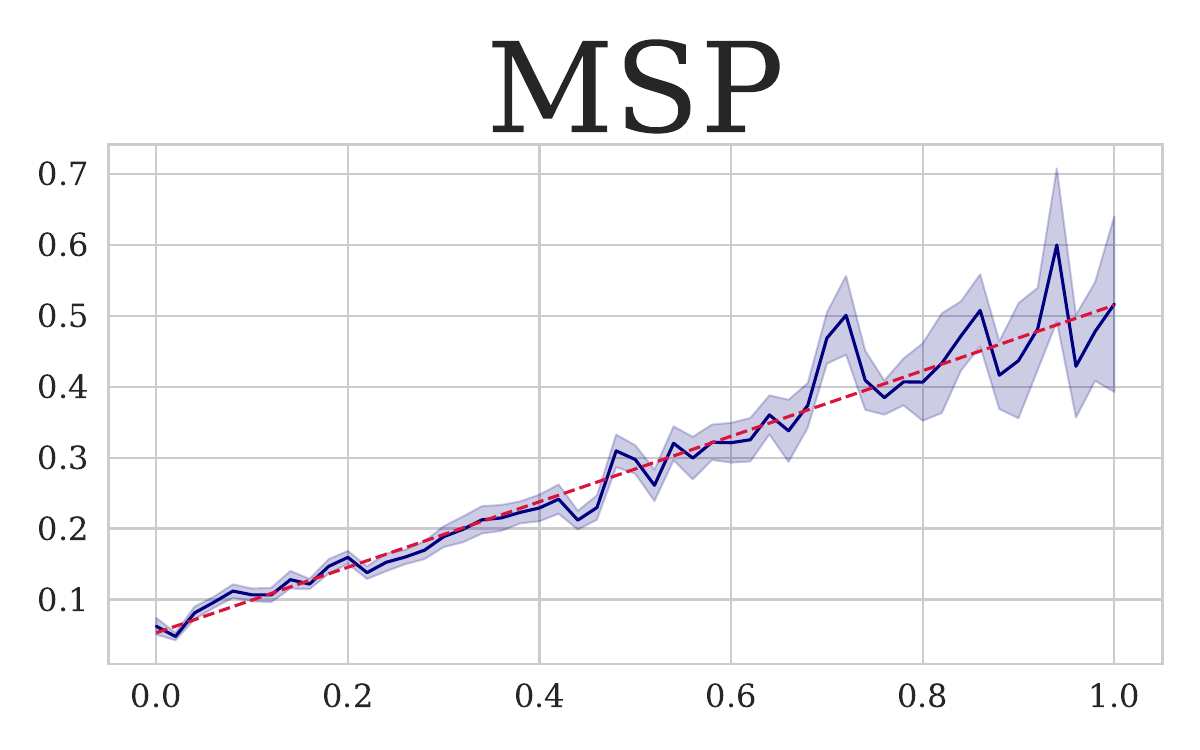}
    \end{subfigure}
    \begin{subfigure}{0.15\textwidth}
        \includegraphics[width=\linewidth]{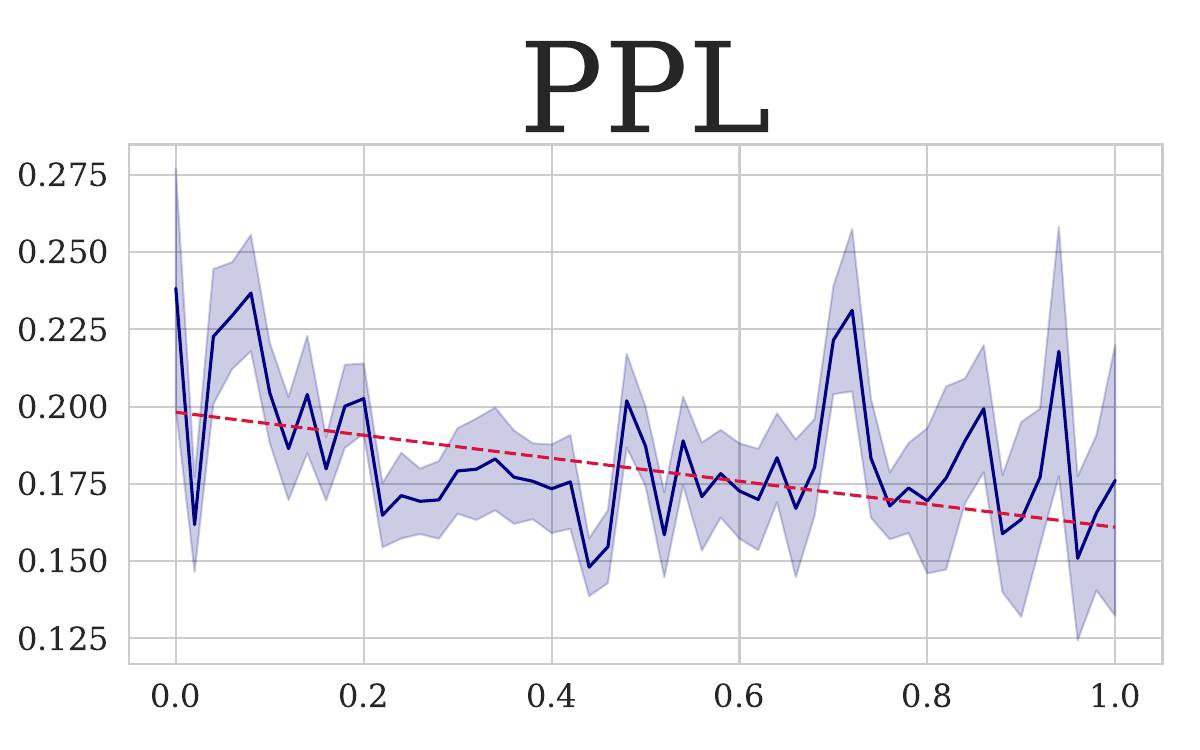}
    \end{subfigure}
    \begin{subfigure}{0.15\textwidth}
        \includegraphics[width=\linewidth]{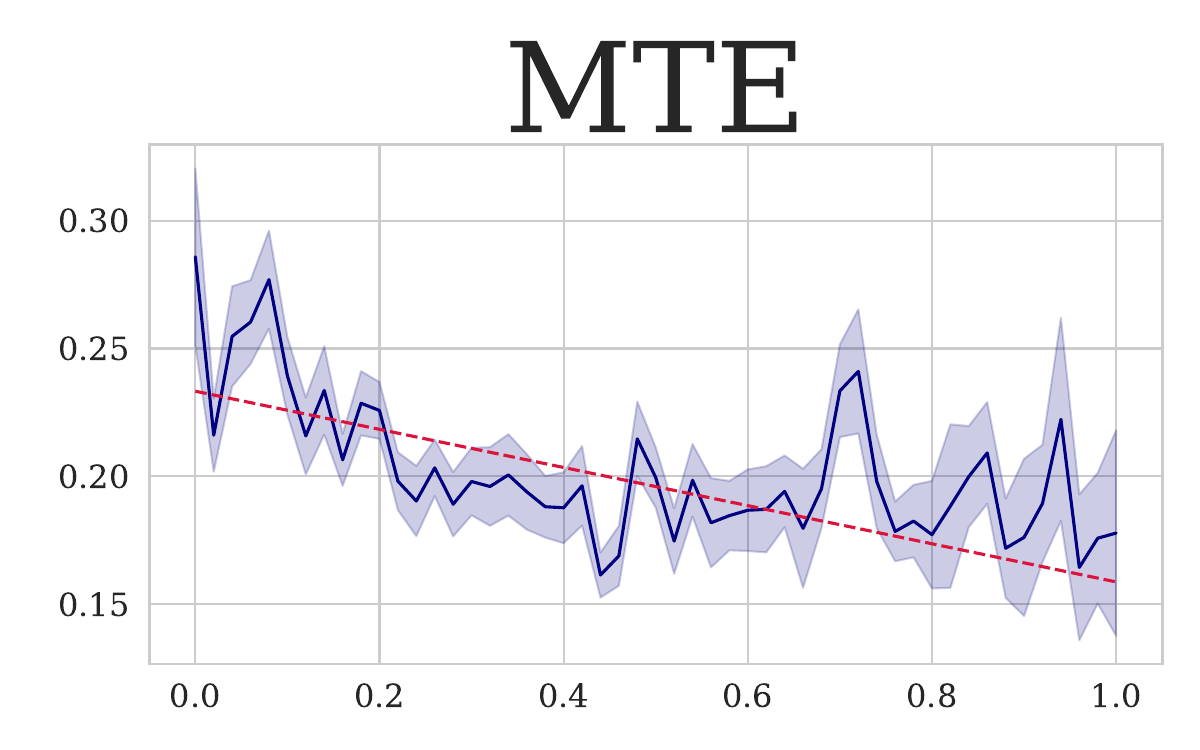}
    \end{subfigure}
    \begin{subfigure}{0.15\textwidth}
        \includegraphics[width=\linewidth]{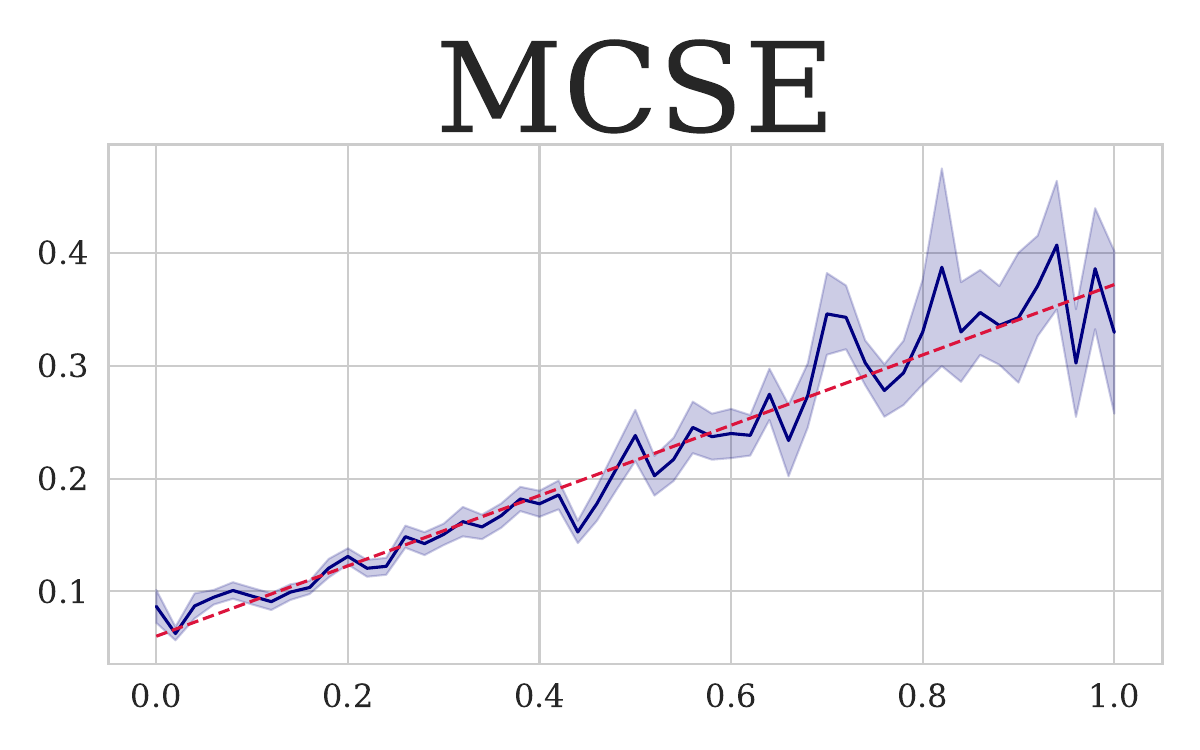}
    \end{subfigure}
    \begin{subfigure}{0.15\textwidth}
        \includegraphics[width=\linewidth]{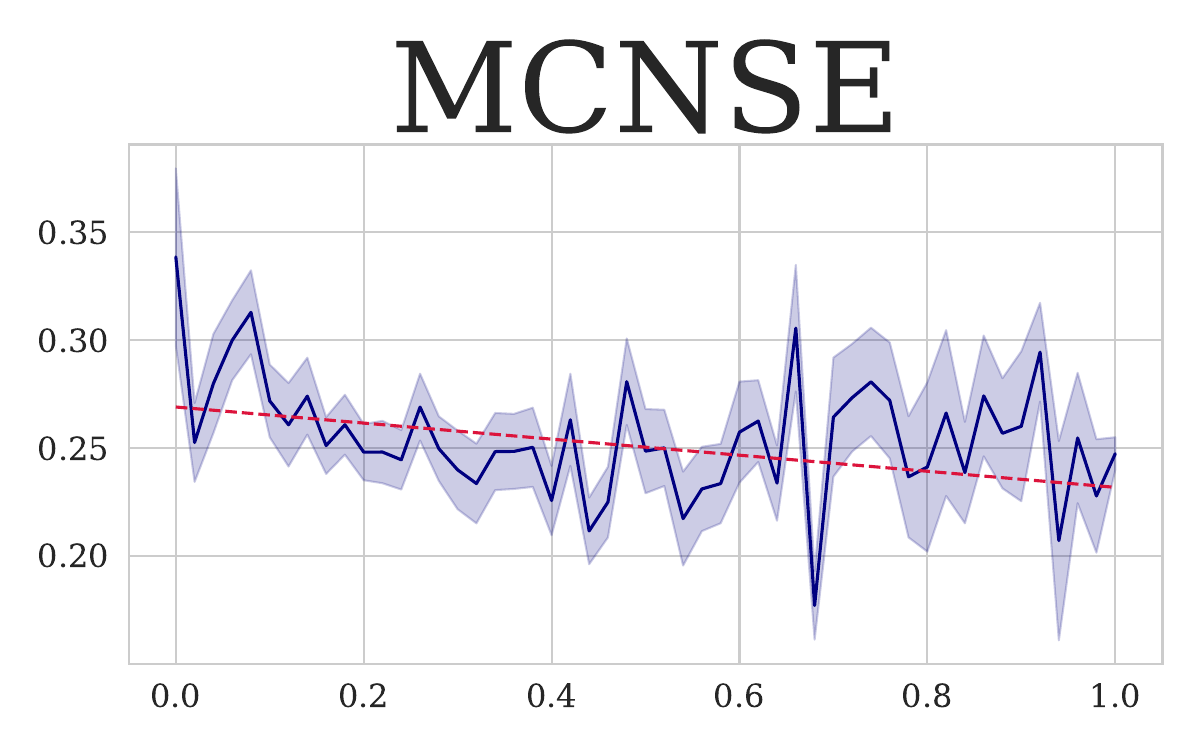}
    \end{subfigure}
    \begin{subfigure}{0.15\textwidth}
        \includegraphics[width=\linewidth]{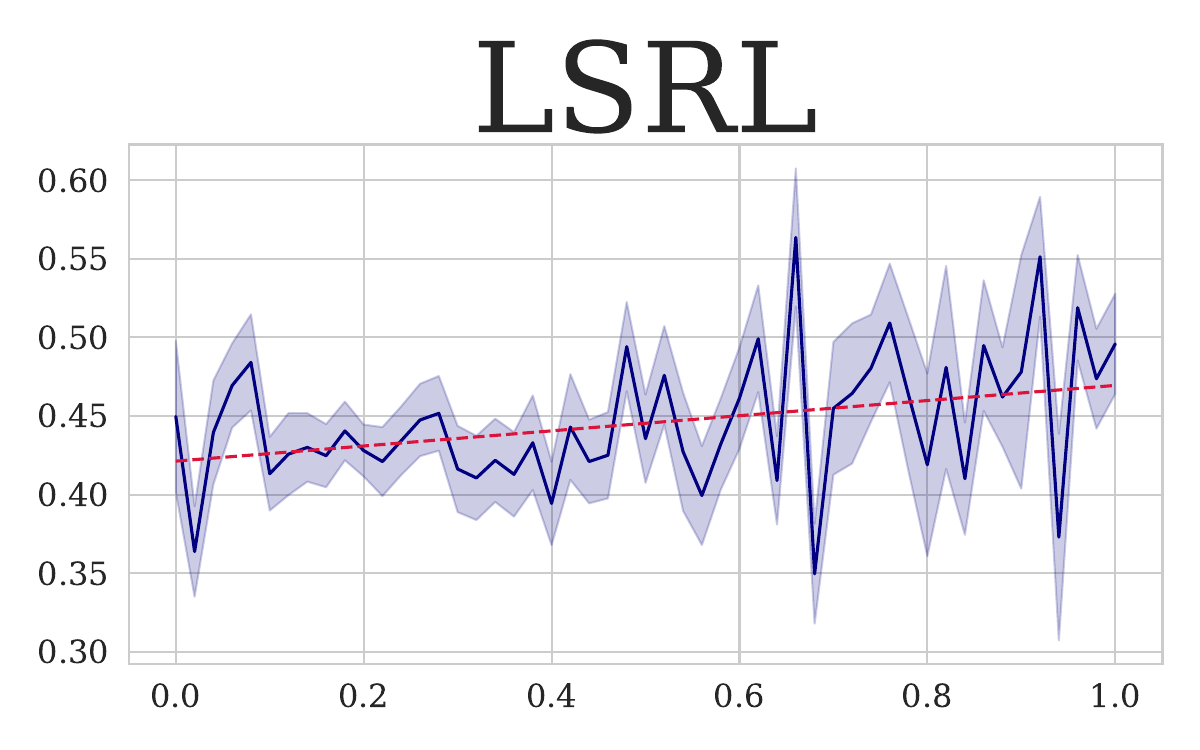}
    \end{subfigure}
    \vspace{0.3em}
    {\centering \textbf{\small WMT19 De-En} \par}
    \vspace{0.2em}
    \begin{subfigure}{0.15\textwidth}
        \includegraphics[width=\linewidth]{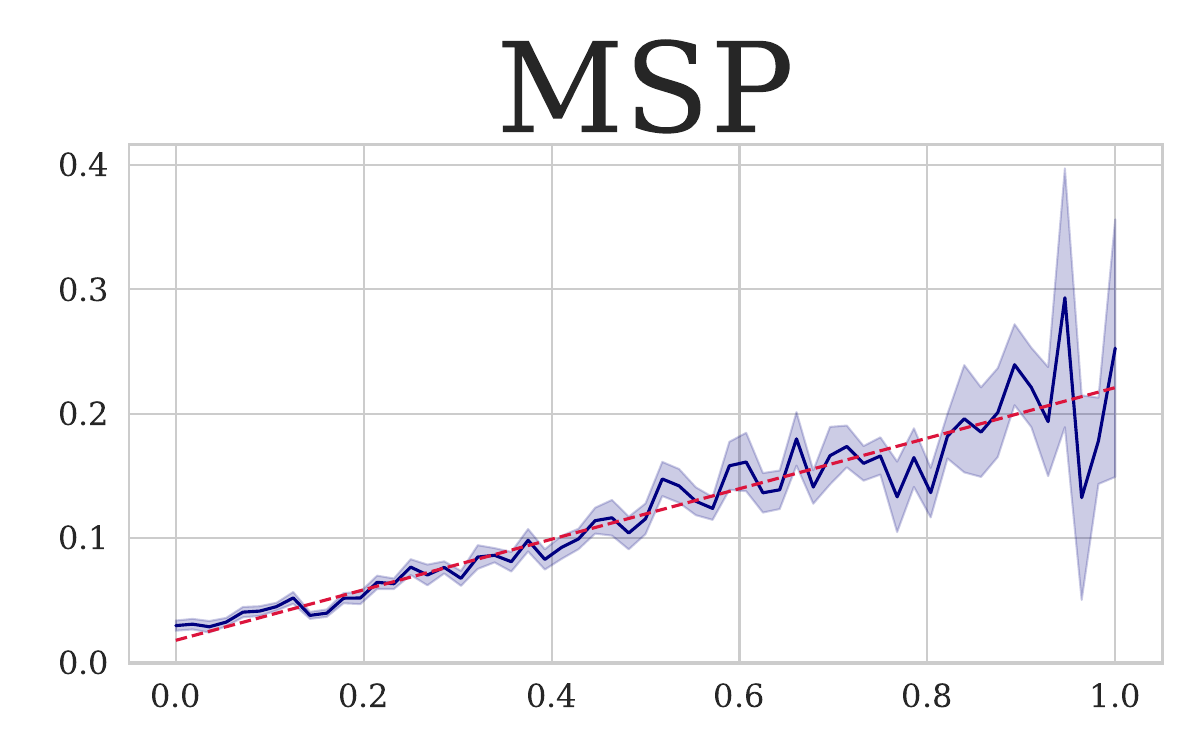}
    \end{subfigure}
    \begin{subfigure}{0.15\textwidth}
        \includegraphics[width=\linewidth]{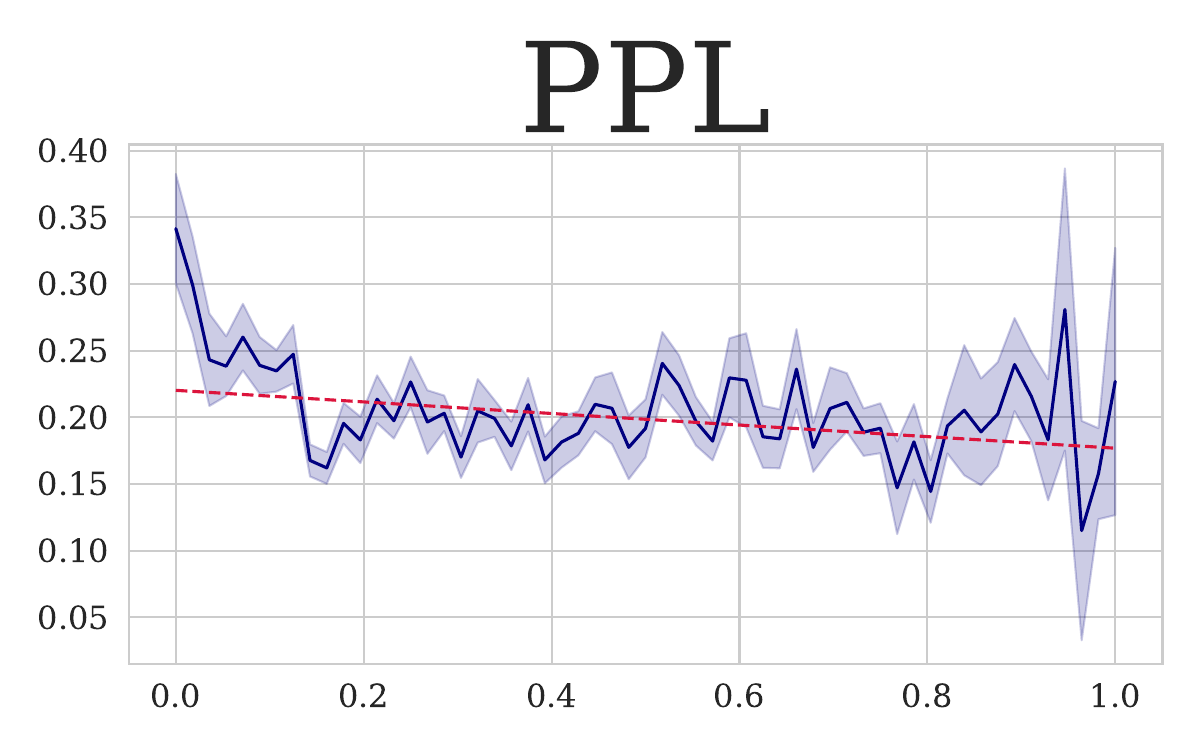}
    \end{subfigure}
    \begin{subfigure}{0.15\textwidth}
        \includegraphics[width=\linewidth]{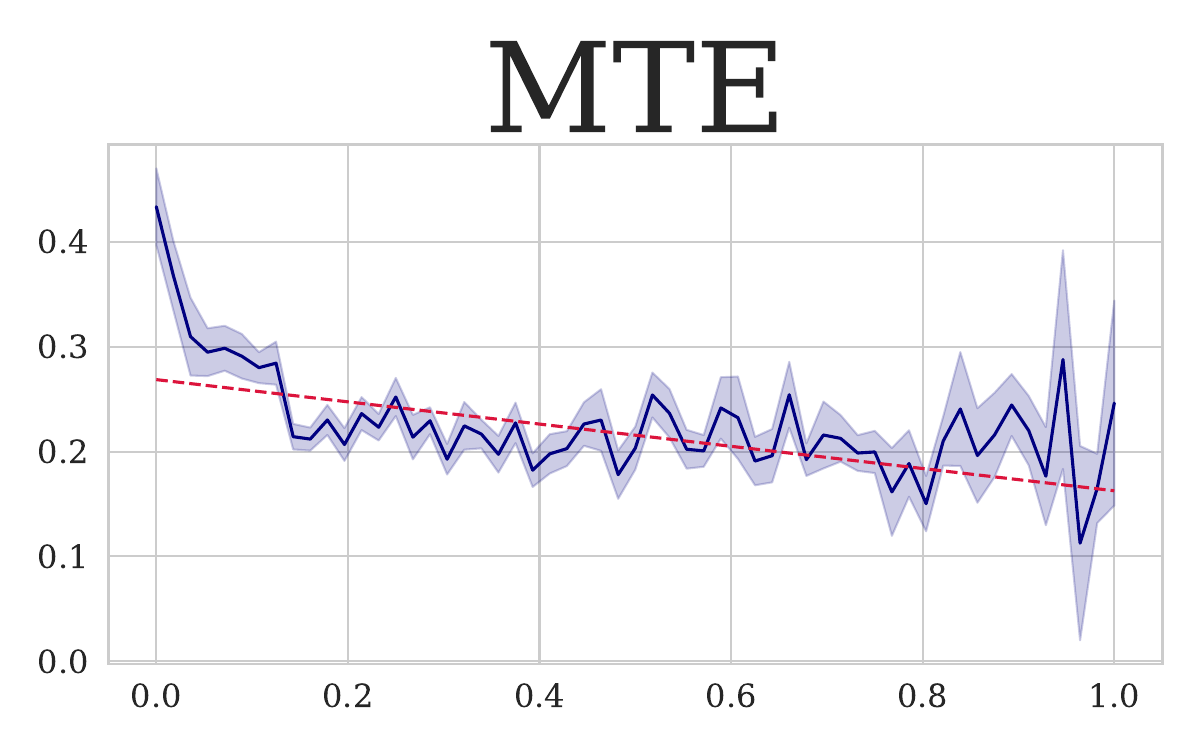}
    \end{subfigure}
    \begin{subfigure}{0.15\textwidth}
        \includegraphics[width=\linewidth]{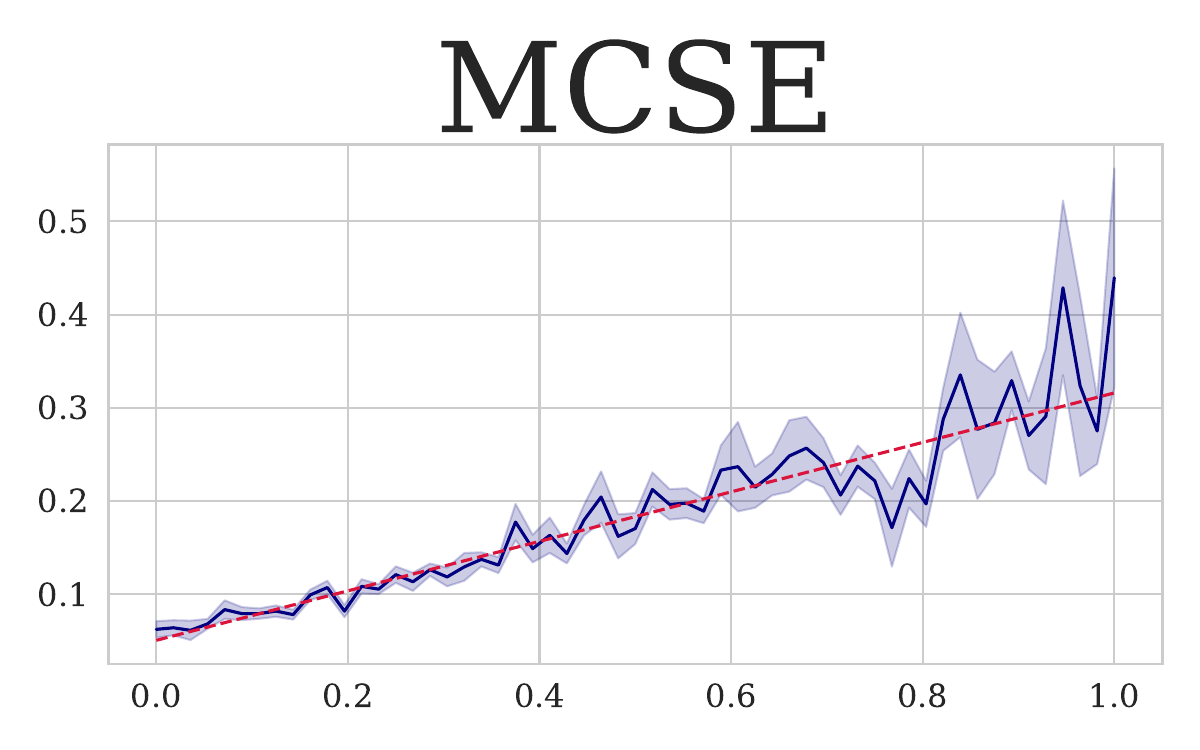}
    \end{subfigure}
    \begin{subfigure}{0.15\textwidth}
        \includegraphics[width=\linewidth]{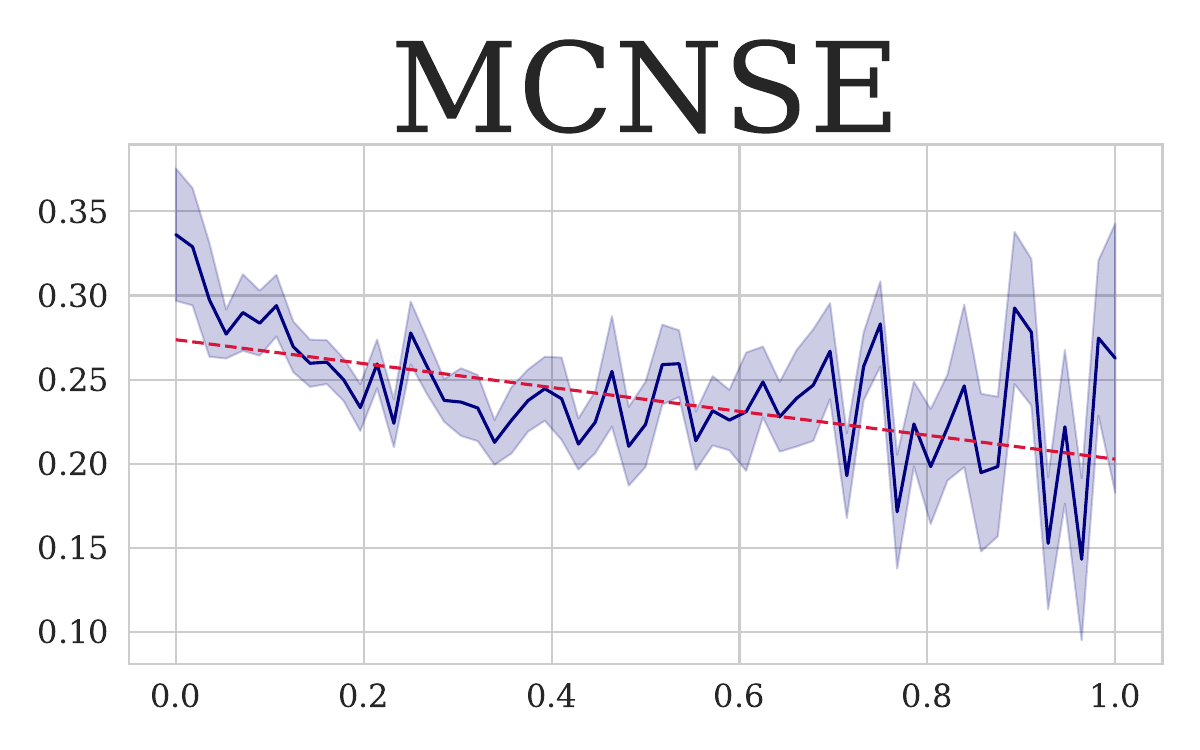}
    \end{subfigure}
    \begin{subfigure}{0.15\textwidth}
        \includegraphics[width=\linewidth]{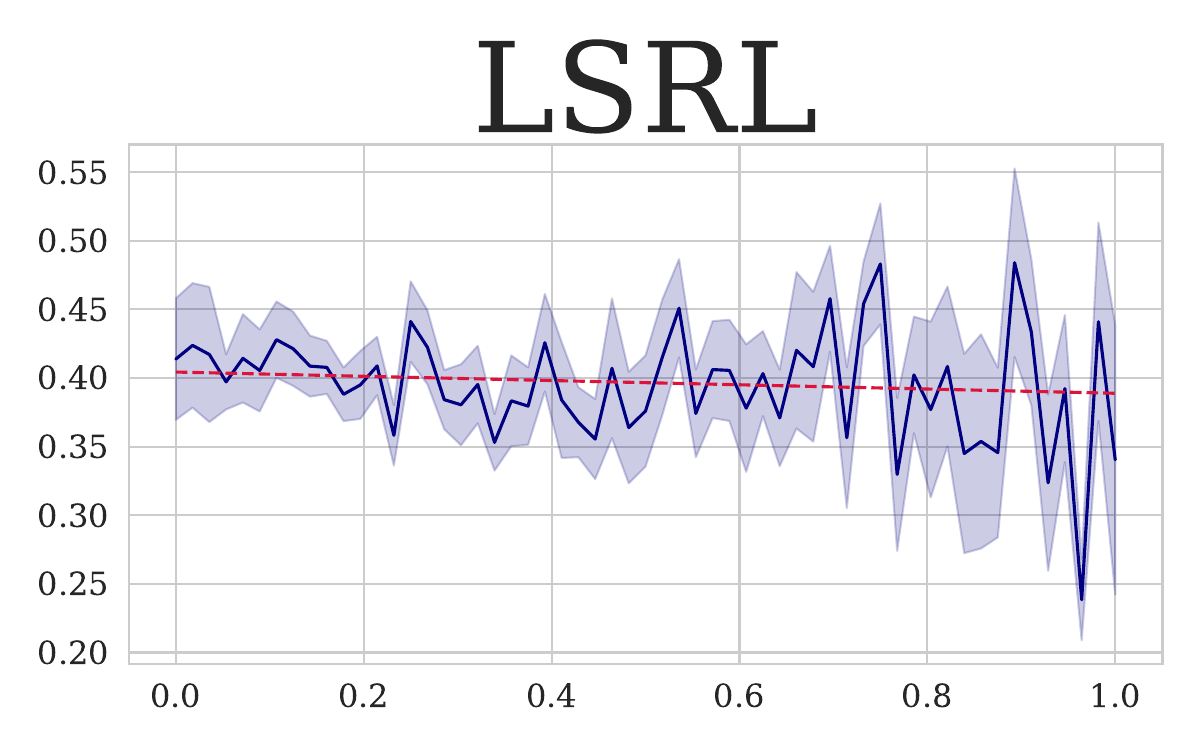}
    \end{subfigure}
    \vspace{0.3em}
    {\centering \textbf{\small WMT19 Lt-En} \par}
    \vspace{0.2em}
    \begin{subfigure}{0.15\textwidth}
        \includegraphics[width=\linewidth]{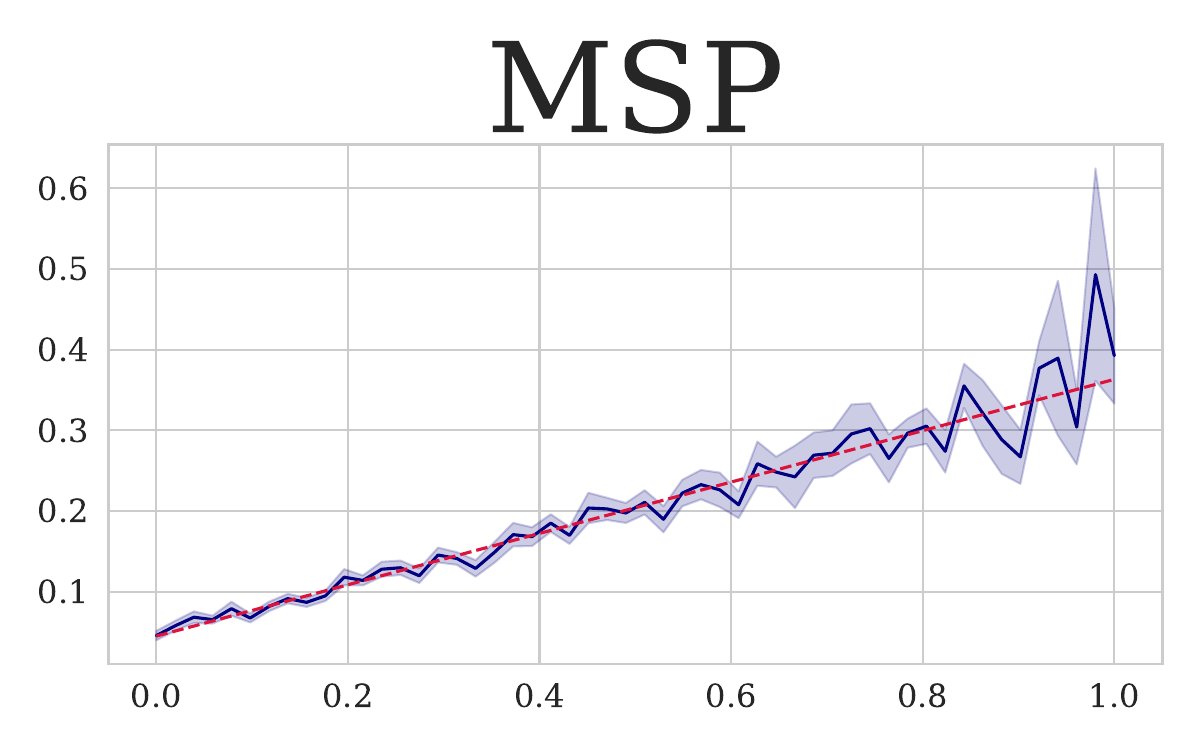}
    \end{subfigure}
    \begin{subfigure}{0.15\textwidth}
        \includegraphics[width=\linewidth]{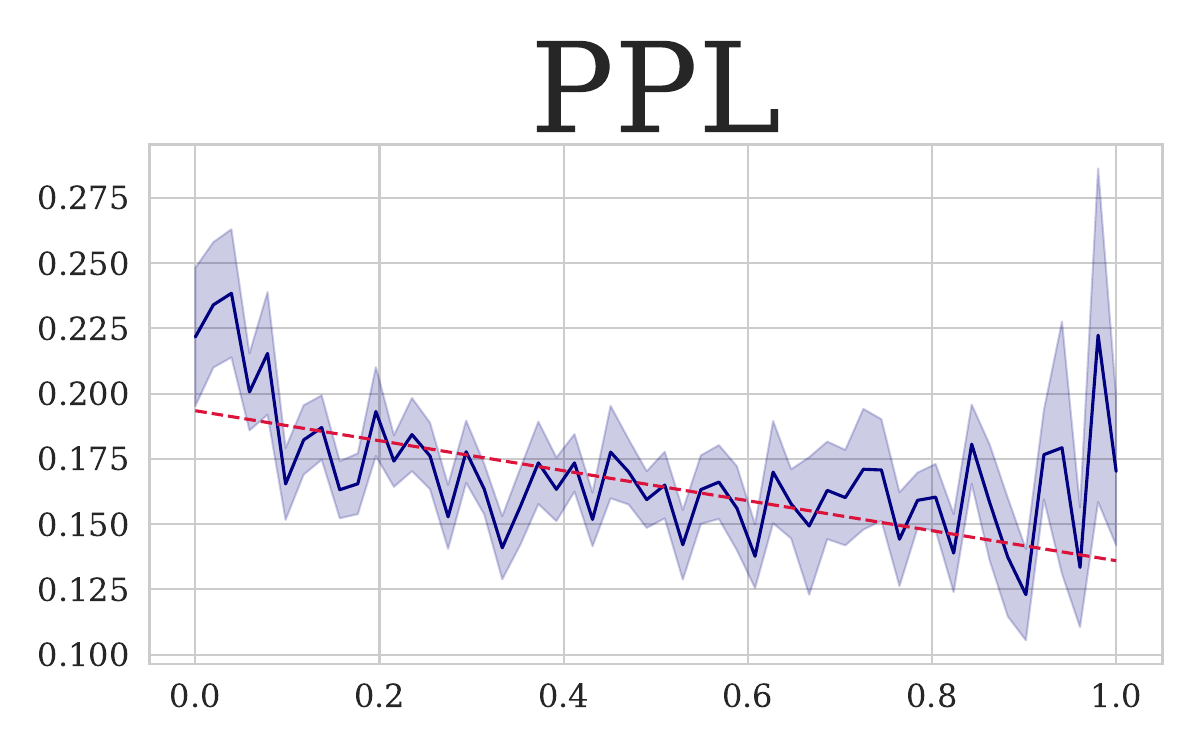}
    \end{subfigure}
    \begin{subfigure}{0.15\textwidth}
        \includegraphics[width=\linewidth]{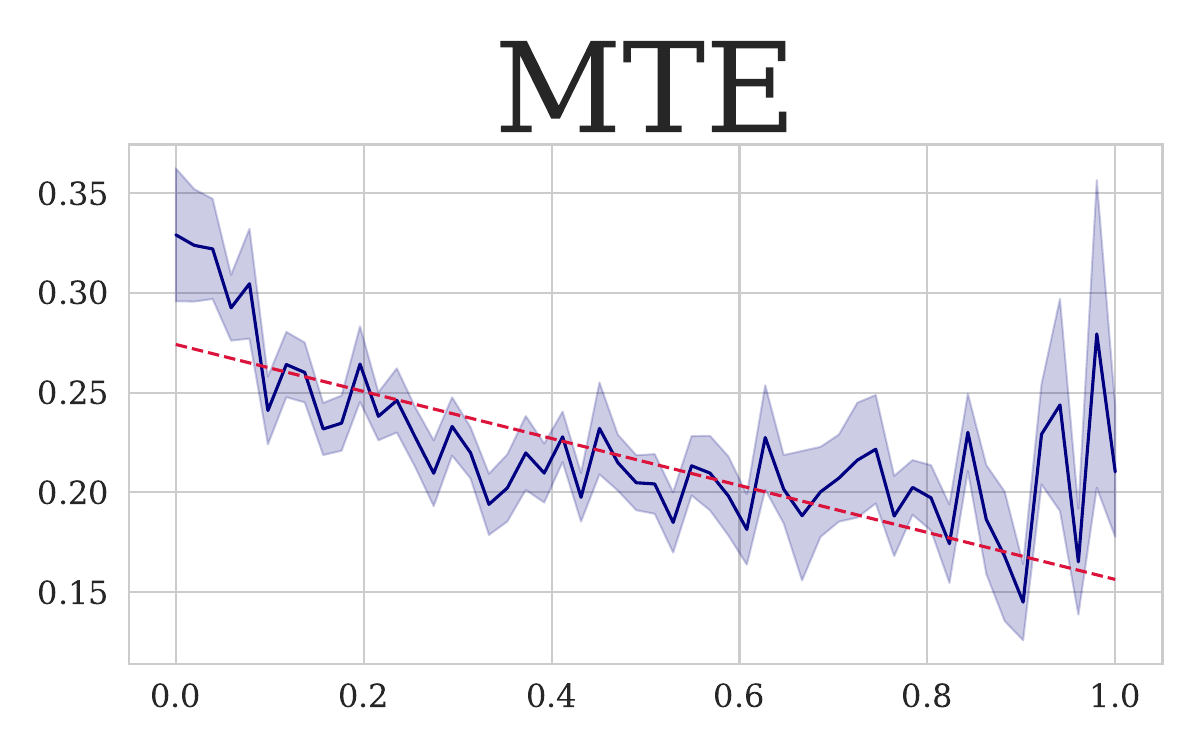}
    \end{subfigure}
    \begin{subfigure}{0.15\textwidth}
        \includegraphics[width=\linewidth]{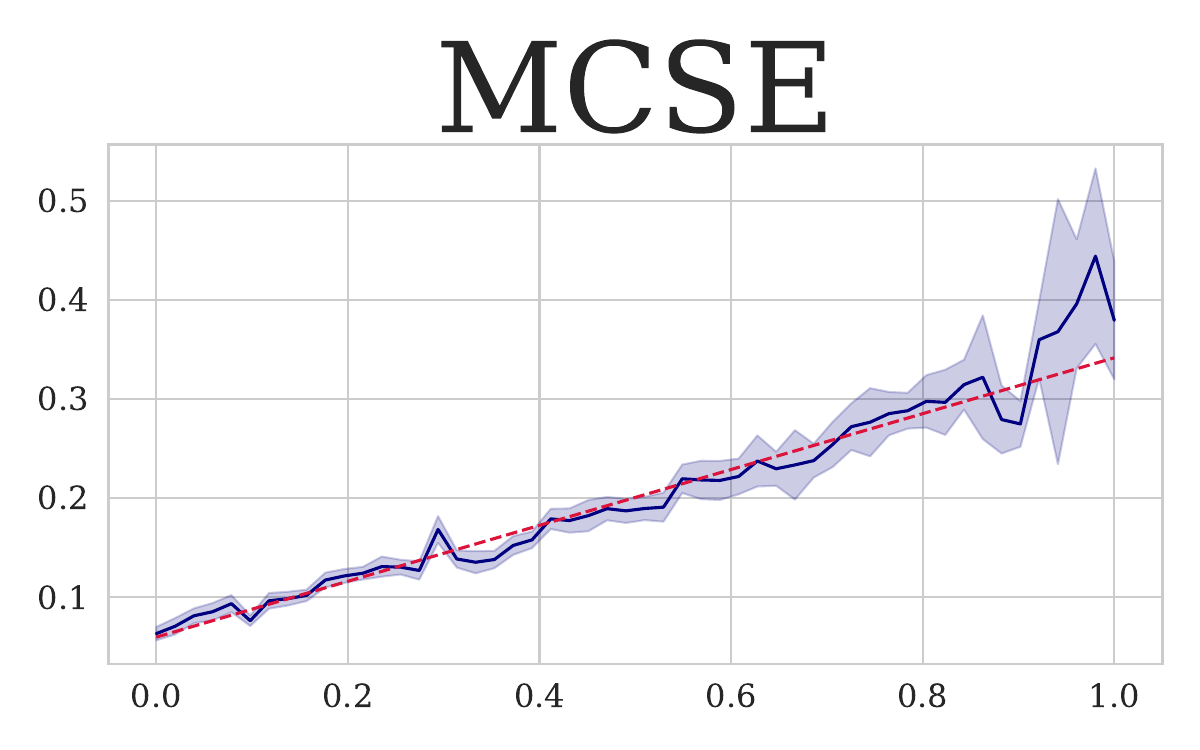}
    \end{subfigure}
    \begin{subfigure}{0.15\textwidth}
        \includegraphics[width=\linewidth]{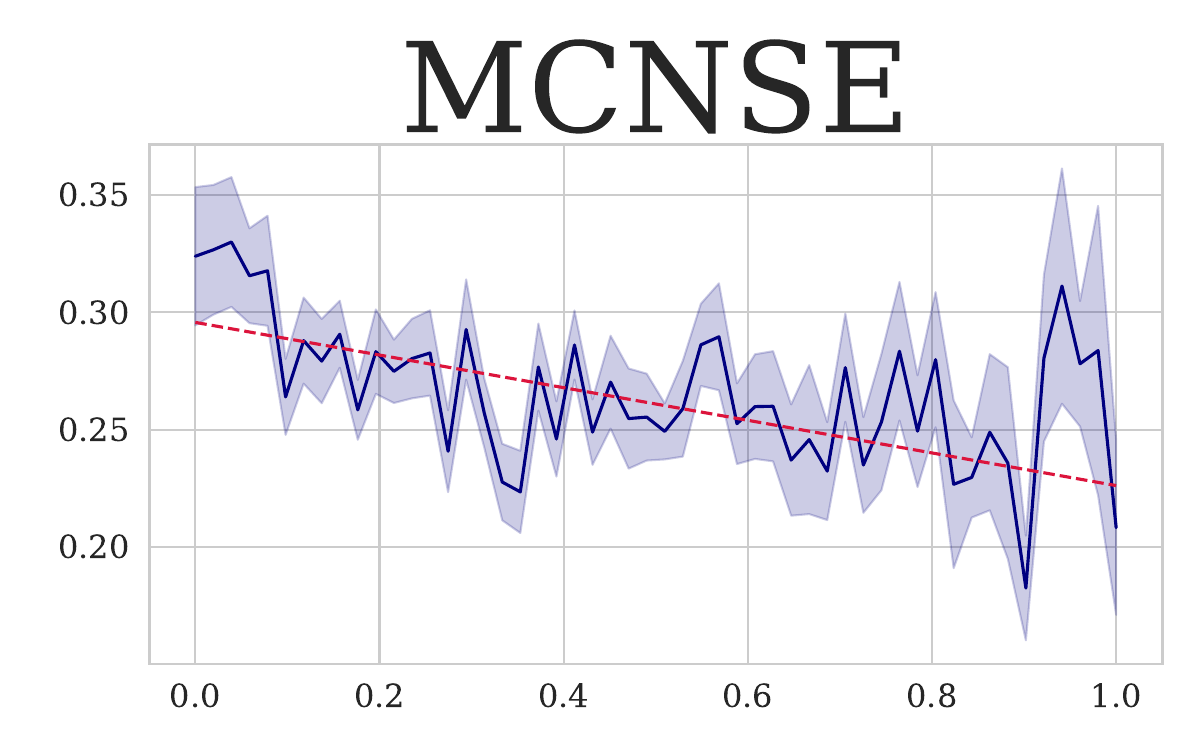}
    \end{subfigure}
    \begin{subfigure}{0.15\textwidth}
        \includegraphics[width=\linewidth]{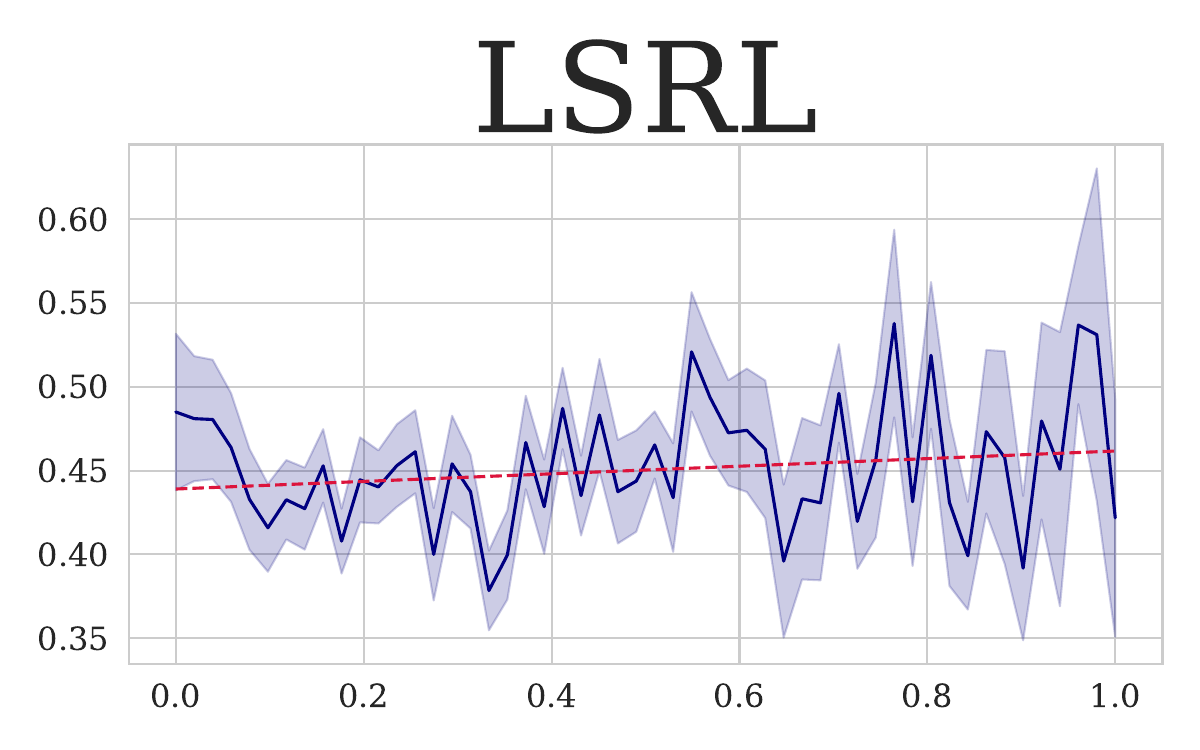}
    \end{subfigure}
    \vspace{0.3em}

        \caption{Uncertainty metric trends for model \textbf{EUROLLM} across all datasets.}
    \label{fig:ue_metrics_eurollm}
\end{figure*}

  Figures~\ref{fig:ue_metrics_llama}, \ref{fig:ue_metrics_gemma} and \ref{fig:ue_metrics_eurollm} depict the length bias of various UQ methods under consideration. Specifics of the charts are the same as in~\ref{sec:length_effects_quality}.

\newpage

\newpage

\newpage
\begin{table}[ht!]
    \centering
\footnotesize
\scalebox{0.85}{
\begin{tabular}{l|cc|cc|cc|cc|cc|cc|cc}
\toprule
\textbf{Dataset} & \multicolumn{2}{c}{MSP} & \multicolumn{2}{c}{PPL} & \multicolumn{2}{c}{MTE} & \multicolumn{2}{c}{MCSE} & \multicolumn{2}{c}{MCNSE} & \multicolumn{2}{c}{LSRL} & \multicolumn{2}{c}{TokenSAR} \\
 & slope & p-val & slope & p-val & slope & p-val & slope & p-val & slope & p-val & slope & p-val  & slope & p-val \\
 \midrule
\multicolumn{13}{c}{\textbf{\llama}} \\
\midrule
WMT14 CS-EN & 0.47 & 0.00 & -0.07 & 0.00 & -0.08 & 0.00 & 0.45 & 0.00 & -0.00 & 0.76 & 0.05 & 0.00 & -0.10 & 0.00 \\
WMT14 DE-EN & 0.45 & 0.00 & -0.08 & 0.00 & -0.11 & 0.00 & 0.43 & 0.00 & -0.01 & 0.67 & 0.06 & 0.00 & -0.08 & 0.00 \\
WMT14 FR-EN & 0.52 & 0.00 & -0.08 & 0.00 & -0.10 & 0.00 & 0.48 & 0.00 & -0.05 & 0.00 & 0.05 & 0.00 & -0.06 & 0.00 \\
WMT14 RU-EN & 0.33 & 0.00 & -0.07 & 0.00 & -0.12 & 0.00 & 0.39 & 0.00 & -0.09 & 0.00 & -0.15 & 0.00 & -0.07 & 0.00 \\
WMT19 DE-EN & 0.45 & 0.00 & -0.05 & 0.00 & -0.13 & 0.00 & 0.31 & 0.00 & -0.03 & 0.06 & 0.09 & 0.00 & -0.07 & 0.00 \\
WMT19 FI-EN & 0.48 & 0.00 & -0.01 & 0.15 & -0.08 & 0.00 & 0.39 & 0.00 & -0.00 & 0.74 & 0.08 & 0.00 & -0.03 & 0.00 \\
WMT19 LT-EN & 0.43 & 0.00 & -0.05 & 0.00 & -0.08 & 0.00 & 0.45 & 0.00 & -0.00 & 0.73 & 0.09 & 0.00 & -0.06 & 0.00 \\
WMT19 RU-EN & 0.46 & 0.00 & -0.09 & 0.00 & -0.14 & 0.00 & 0.36 & 0.00 & -0.10 & 0.00 & -0.06 & 0.00 & -0.09 & 0.00 \\

\midrule
\multicolumn{13}{c}{\textbf{\gemma}} \\
\midrule
WMT14 CS-EN & 0.55 & 0.00 & -0.05 & 0.00 & -0.10 & 0.00 & 0.59 & 0.00 & 0.02 & 0.23 & 0.04 & 0.00 & -0.08 & 0.00 \\
WMT14 DE-EN & 0.40 & 0.00 & -0.08 & 0.00 & -0.12 & 0.00 & 0.49 & 0.00 & 0.01 & 0.53 & 0.05 & 0.00 & -0.09 & 0.00 \\
WMT14 FR-EN & 0.49 & 0.00 & -0.09 & 0.00 & -0.14 & 0.00 & 0.57 & 0.00 & -0.05 & 0.00 & 0.03 & 0.04 & -0.09 & 0.00 \\
WMT14 RU-EN & 0.38 & 0.00 & -0.17 & 0.00 & -0.24 & 0.00 & 0.49 & 0.00 & -0.14 & 0.00 & -0.17 & 0.00 & -0.18 & 0.00 \\
WMT19 DE-EN & 0.38 & 0.00 & -0.03 & 0.01 & -0.09 & 0.00 & 0.32 & 0.00 & -0.03 & 0.03 & 0.05 & 0.00 & -0.06 & 0.00 \\
WMT19 FI-EN & 0.51 & 0.00 & -0.00 & 0.79 & -0.06 & 0.00 & 0.51 & 0.00 & 0.00 & 0.84 & 0.09 & 0.00 & -0.03 & 0.01 \\
WMT19 LT-EN & 0.57 & 0.00 & -0.02 & 0.03 & -0.06 & 0.00 & 0.40 & 0.00 & 0.00 & 0.96 & 0.09 & 0.00 & -0.03 & 0.00 \\
WMT19 RU-EN & 0.63 & 0.00 & -0.06 & 0.00 & -0.11 & 0.00 & 0.59 & 0.00 & -0.07 & 0.00 & -0.02 & 0.08 & -0.07 & 0.00 \\

\midrule
\multicolumn{13}{c}{\textbf{\eurollm}} \\
\midrule
WMT14 CS-EN & 0.30 & 0.00 & -0.06 & 0.00 & -0.09 & 0.00 & 0.42 & 0.00 & -0.06 & 0.00 & -0.01 & 0.43 & -0.06 & 0.00 \\
WMT14 DE-EN & 0.36 & 0.00 & -0.08 & 0.00 & -0.12 & 0.00 & 0.34 & 0.00 & -0.02 & 0.11 & 0.01 & 0.55 & -0.09 & 0.00 \\
WMT14 FR-EN & 0.37 & 0.00 & -0.12 & 0.00 & -0.14 & 0.00 & 0.17 & 0.00 & -0.04 & 0.00 & 0.00 & 0.99 & -0.12 & 0.00 \\
WMT14 RU-EN & 0.27 & 0.00 & -0.14 & 0.00 & -0.19 & 0.00 & 0.23 & 0.00 & -0.17 & 0.00 & -0.16 & 0.00 & -0.16 & 0.00 \\
WMT19 DE-EN & 0.20 & 0.00 & -0.04 & 0.00 & -0.11 & 0.00 & 0.27 & 0.00 & -0.07 & 0.00 & -0.02 & 0.43 & -0.05 & 0.00 \\
WMT19 FI-EN & 0.46 & 0.00 & -0.04 & 0.00 & -0.07 & 0.00 & 0.31 & 0.00 & -0.04 & 0.00 & 0.05 & 0.01 & -0.05 & 0.00 \\
WMT19 LT-EN & 0.32 & 0.00 & -0.06 & 0.00 & -0.12 & 0.00 & 0.28 & 0.00 & -0.07 & 0.00 & 0.02 & 0.23 & -0.06 & 0.00 \\
WMT19 RU-EN & 0.45 & 0.00 & -0.08 & 0.00 & -0.14 & 0.00 & 0.36 & 0.00 & -0.12 & 0.00 & -0.09 & 0.00 & -0.09 & 0.00 \\

\bottomrule
\end{tabular}
}
\caption{Regression slopes and p-values measuring the correlation between output length and various uncertainty metrics on machine translation datasets.}
\end{table}

\begin{table}[ht!]
    \centering
\footnotesize
\scalebox{0.85}{
\begin{tabular}{l|cc|cc|cc|cc|cc|cc|cc}
\toprule
\textbf{Dataset} & \multicolumn{2}{c}{MSP} & \multicolumn{2}{c}{PPL} & \multicolumn{2}{c}{MTE} & \multicolumn{2}{c}{MCSE} & \multicolumn{2}{c}{MCNSE} & \multicolumn{2}{c}{LSRL} & \multicolumn{2}{c}{TokenSAR} \\
 & slope & p-val & slope & p-val & slope & p-val & slope & p-val & slope & p-val & slope & p-val & slope & p-val \\
\midrule
\multicolumn{13}{c}{\textbf{\llama}} \\
\midrule
XSUM & 0.142 & 0.000 & -0.261 & 0.000 & -0.283 & 0.000 & 0.042 & 0.008 & -0.026 & 0.096 & 0.029 & 0.025 & -0.264	& 0.0
 \\
\midrule
\multicolumn{13}{c}{\textbf{\gemma}} \\
\midrule
XSUM & 0.054 & 0.001 & -0.295 & 0.000 & -0.320 & 0.000 & 0.036 & 0.007 & -0.041 & 0.001 & 0.019 & 0.067 & -0.307	& 0.0\\
\bottomrule
\end{tabular}
}
\caption{Regression slopes and p-values measuring the correlation between output length and various uncertainty metrics on XSUM dataset.}
\end{table}

\begin{table}[ht!]
    \centering
\footnotesize
\scalebox{0.8}{
\begin{tabular}{l|cc|cc|cc|cc|cc|cc|cc}
\toprule
\textbf{Dataset} & \multicolumn{2}{c}{MSP} & \multicolumn{2}{c}{PPL} & \multicolumn{2}{c}{MTE} & \multicolumn{2}{c}{MCSE} & \multicolumn{2}{c}{MCNSE} & \multicolumn{2}{c}{LSRL}  & \multicolumn{2}{c}{TokenSAR}\\
 & slope & p-val & slope & p-val & slope & p-val & slope & p-val & slope & p-val & slope & p-val & slope & p-val \\
\midrule
\multicolumn{13}{c}{\textbf{\llama}} \\
\midrule
GSM8K & 0.418 & 0.000 & -0.109 & 0.000 & -0.098 & 0.000 & 0.273 & 0.000 & 0.096 & 0.000 & 0.219 & 0.000 & -0.109 &	0.0 \\
\midrule
\multicolumn{13}{c}{\textbf{\gemma}} \\
\midrule
GSM8K & 0.322 & 0.000 & -0.100 & 0.000 & -0.092 & 0.000 & 0.266 & 0.000 & 0.085 & 0.000 & 0.277 & 0.000 & -0.1	& 0.0
 \\
\bottomrule
\end{tabular}
}
\caption{Regression slopes and p-values measuring the correlation between output length and various uncertainty metrics on GSM8k dataset.}
\end{table}

\newpage

\section{Detailed Description of Uncertainty Quantification Methods}
\label{sec:appendix_methods}

  Here, we provide details of the UQ methods used in the experiments omitted from the main part of the paper.

\noindent\textbf{Maximum Sequence Probability (MSP)}
  is one of the simplest and most direct methods for estimating uncertainty. It measures the negative log-likelihood of the most likely output sequence given a specific input. Under the assumption that the model is most confident in its most probable output, lower values indicate higher confidence:
  \begin{equation}
    U_{\text{MSP}}(\yv \mid \xv) = - \log P(\yv \mid \xv).
  \label{eq:msp}
  \end{equation}

\noindent\textbf{Perplexity (PPL)}
  is a widely used metric for evaluating uncertainty in autoregressive models~\citep{fomicheva-etal-2020-unsupervised}. It computes the negative average log-likelihood per token, making it explicitly length-normalized. Lower perplexity indicates higher model confidence:
  \begin{equation}
    U_{\text{PPL}}(\yv \mid \xv) = -\frac{1}{L} \log P(\yv \mid \xv),
  \label{eq:ppl}
  \end{equation}
  where $L$ is the length of the output sequence $\yv$.

\noindent\textbf{Mean Token Entropy (MTE)}
  captures the average uncertainty at the token level. It measures how peaked or flat the model's predicted distribution is at each decoding step:
  \begin{equation}
    U_{\text{MTE}}(\yv \mid \xv) = \frac{1}{L} \sum_{l = 1}^L \HC(y_l \mid \yv_{<l}, \xv),
  \label{eq:entropy}
  \end{equation}
  where $\HC(y_l \mid \yv_{<l}, \xv) = -\sum_{v} P(y_l = v \mid \yv_{<l}, \xv) \log P(y_l = v \mid \yv_{<l}, \xv)$ is the entropy of the token distribution at position $l$.

\noindent\textbf{Monte Carlo Sequence Entropy (MCSE)}
  estimates sequence-level uncertainty via sampling. We draw $M$ sequences ${\yv^{(i)}}{i=1}^M$ from the model's output distribution and compute their average negative log-likelihood:
  \begin{equation}
    U_{\text{MCSE}}(\xv) = -\frac{1}{M} \sum_{i = 1}^M \log P(\yv^{(i)} \mid \xv).
  \label{eq:mcse}
  \end{equation}

\noindent\textbf{Monte Carlo Normalized Sequence Entropy (MCNSE)}
  is a length-normalized variant of MCSE. For each sampled sequence $\yv^{(i)}$, we normalize the log-likelihood by its length $L^{(i)}$:
  \begin{equation}
    U_{\text{MCNSE}}(\xv) = -\frac{1}{M} \sum_{i = 1}^M \frac{1}{L^{(i)}} \log P(\yv^{(i)} \mid \xv).
  \label{eq:mcnse}
  \end{equation} 

\noindent\textbf{Lexical Similarity with ROUGE-L (LSRL)}  
  measures the average pairwise lexical similarity between all sampled sequences. Unlike the previous methods, which rely on model probabilities, LSRL captures diversity among generated hypotheses by comparing their surface forms:
  \begin{equation}
    U_{\text{LSRL}}(\xv) = 1 - \frac{2}{M(M-1)} \sum_{i < j} \text{ROUGE-L}(\yv^{(i)}, \yv^{(j)}).
  \label{eq:lsrl}
  \end{equation}

\noindent\textbf{TokenSAR}
  computes relevance-weighted average of the negative log probabilities of generated tokens: 
  \begin{equation}    
    U_{\text{TokenSAR}}(\mathbf{x})  =
    - \sum_{l=1}^{L} \tilde{R}_T(y_l, \mathbf{y}, \mathbf{x}) \log P(y_l \mid \mathbf{y}_{<l}, \mathbf{x}),
  \label{eq:tokensar}
  \end{equation}
  where the normalized relevance weight for each token $y_l$ is given by $\tilde{R}_T(y_k, \mathbf{y}, \mathbf{x}) = frac{R_T(y_k, \mathbf{y}, \mathbf{x})}{\sum_{l=1}^L R_T(y_l, \mathbf{y}, \mathbf{x})}.$ and $R_T(\cdot)$ denotes the token relevance function, derived from a sentence similarity function $g(\cdot,\cdot)$ as $R_T(y_k, \mathbf{y}, \mathbf{x}) = 1 - g(\mathbf{x} \cup \mathbf{y}, \mathbf{x} \cup \mathbf{y} \setminus y_k).$

\newpage

\section{Detailed Experimental Results}
\label{sec:experimental_results}
  Tables~\ref{tab:detailed_comet}, \ref{tab:detailed_xcomet} and \ref{tab:detailed_metricx} contain PRR scores for all UQ methods, along with their LINE counterparts for NMT datasets. Table~\ref{tab:detailed_gsm_xsum} contains the same data for summarization and mathematical reasoning.
  
  \FloatBarrier
  \begin{table*}[h!]
\footnotesize
\centering
\begin{tabular}{lcccccccc}
\toprule
&\multicolumn{4}{c}{\textbf{WMT14}}&\multicolumn{4}{c}{\textbf{WMT19}}\\
\cmidrule(lr){2-5}
\cmidrule(lr){6-9}
&Cs-En&De-En&Ru-En&Fr-En&De-En&Fi-En&Lt-En&Ru-En\\
\midrule
& \multicolumn{8}{c}{Llama 3.1 8B}\\
\midrule
MSP & 0.42 & 0.39 & 0.45 & 0.35 & 0.46 & 0.19 & 0.29 & 0.43 \\
MSP-LINE & 0.47 & 0.49 & 0.48 & 0.40 & \underline{0.51} & 0.47 & 0.47 & 0.41 \\
\midrule
PPL & 0.42 & 0.46 & 0.37 & 0.31 & 0.41 & 0.52 & 0.47 & 0.32 \\
PPL-LINE & \underline{0.52} & 0.51 & \underline{0.53} & \underline{0.41} & 0.46 & 0.52 & 0.49 & \underline{0.46} \\
\midrule
MTE & 0.44 & 0.48 & 0.41 & 0.37 & 0.42 & \underline{0.54} & \underline{0.52} & 0.33 \\
MTE-LINE & \textbf{0.58} & \textbf{0.56} & \textbf{0.59} & \textbf{0.48} & \textbf{0.55} & \textbf{0.56} & \textbf{0.56} & \textbf{0.53} \\
\midrule
MCSE & 0.36 & 0.32 & 0.35 & 0.30 & 0.36 & 0.08 & 0.20 & 0.36 \\
MCSE-LINE & 0.38 & 0.36 & 0.33 & 0.28 & 0.38 & 0.32 & 0.36 & 0.32 \\
\midrule
MCNSE & 0.48 & 0.44 & 0.40 & 0.36 & 0.43 & 0.46 & 0.48 & 0.35 \\
MCNSE-LINE & 0.49 & 0.44 & 0.47 & 0.39 & 0.45 & 0.46 & 0.48 & 0.44 \\
\midrule
LSRL & 0.45 & 0.44 & 0.38 & 0.35 & 0.46 & 0.37 & 0.42 & 0.35 \\
LSRL-LINE & 0.41 & 0.41 & 0.44 & 0.32 & 0.40 & 0.37 & 0.40 & 0.38 \\
\midrule
TokenSAR & 0.44 & 0.45 & 0.37 & 0.35 & 0.40 & 0.52 & 0.46 & 0.32 \\
TokenSAR-LINE & 0.51 & \underline{0.52} & 0.52 & 0.41 & 0.47 & 0.53 & 0.49 & 0.46 \\

\midrule
& \multicolumn{8}{c}{Gemma 2 9B}\\
\midrule
MSP & 0.40 & 0.37 & 0.43 & 0.29 & 0.49 & 0.18 & 0.35 & 0.40 \\
MSP-LINE & \underline{0.48} & 0.50 & 0.47 & 0.38 & \textbf{0.53} & 0.42 & \underline{0.36} & \textbf{0.41} \\
\midrule
PPL & 0.44 & 0.48 & 0.38 & 0.36 & 0.44 & 0.46 & 0.30 & 0.31 \\
PPL-LINE & 0.46 & \underline{0.51} & \underline{0.50} & 0.40 & 0.47 & 0.46 & 0.32 & 0.35 \\
\midrule
MTE & 0.44 & 0.49 & 0.38 & 0.37 & 0.44 & \underline{0.49} & 0.30 & 0.30 \\
MTE-LINE & \textbf{0.49} & \textbf{0.54} & \textbf{0.53} & \textbf{0.44} & \underline{0.51} & \textbf{0.49} & \textbf{0.36} & 0.40 \\
\midrule
MCSE & 0.32 & 0.31 & 0.35 & 0.28 & 0.41 & 0.09 & 0.29 & 0.36 \\
MCSE-LINE & 0.39 & 0.43 & 0.38 & 0.35 & 0.47 & 0.31 & 0.29 & 0.39 \\
\midrule
MCNSE & 0.44 & 0.50 & 0.42 & 0.37 & 0.47 & 0.41 & 0.35 & 0.37 \\
MCNSE-LINE & 0.44 & 0.50 & 0.48 & 0.39 & 0.49 & 0.41 & 0.35 & \underline{0.41} \\
\midrule
LSRL & 0.40 & 0.47 & 0.40 & 0.33 & 0.43 & 0.40 & 0.34 & 0.34 \\
LSRL-LINE & 0.38 & 0.46 & 0.45 & 0.32 & 0.41 & 0.40 & 0.28 & 0.35 \\
\midrule
TokenSAR & 0.41 & 0.46 & 0.36 & 0.37 & 0.42 & 0.45 & 0.29 & 0.28 \\
TokenSAR-LINE & 0.45 & 0.50 & 0.49 & \underline{0.41} & 0.46 & 0.45 & 0.35 & 0.34 \\

\midrule
& \multicolumn{8}{c}{EuroLLM 9B}\\
\midrule
MSP & 0.29 & 0.33 & 0.42 & 0.24 & 0.40 & 0.16 & 0.28 & 0.43 \\
MSP-LINE & 0.37 & 0.46 & 0.50 & 0.34 & 0.51 & 0.39 & 0.34 & \underline{0.44} \\
\midrule
PPL & 0.51 & 0.50 & 0.43 & 0.44 & 0.52 & 0.48 & 0.36 & 0.32 \\
PPL-LINE & \underline{0.53} & \underline{0.53} & \underline{0.54} & \underline{0.47} & \underline{0.54} & 0.49 & \underline{0.40} & 0.39 \\
\midrule
MTE & 0.52 & 0.52 & 0.46 & 0.47 & 0.51 & \underline{0.51} & 0.37 & 0.34 \\
MTE-LINE & \textbf{0.57} & \textbf{0.55} & \textbf{0.56} & \textbf{0.52} & \textbf{0.58} & \textbf{0.52} & \textbf{0.45} & \textbf{0.45} \\
\midrule
MCSE & 0.35 & 0.36 & 0.42 & 0.28 & 0.41 & 0.21 & 0.34 & 0.42 \\
MCSE-LINE & 0.46 & 0.47 & 0.46 & 0.40 & 0.49 & 0.40 & 0.37 & 0.39 \\
\midrule
MCNSE & 0.23 & 0.36 & 0.28 & 0.22 & 0.34 & 0.36 & 0.28 & 0.24 \\
MCNSE-LINE & 0.22 & 0.36 & 0.33 & 0.22 & 0.34 & 0.35 & 0.28 & 0.29 \\
\midrule
LSRL & 0.32 & 0.38 & 0.31 & 0.29 & 0.40 & 0.35 & 0.32 & 0.26 \\
LSRL-LINE & 0.32 & 0.37 & 0.36 & 0.29 & 0.41 & 0.35 & 0.31 & 0.29 \\
\midrule
TokenSAR & 0.42 & 0.47 & 0.42 & 0.40 & 0.44 & 0.46 & 0.31 & 0.34 \\
TokenSAR-LINE & 0.42 & 0.49 & 0.53 & 0.41 & 0.45 & 0.46 & 0.35 & 0.41 \\
\midrule
\end{tabular}
\caption{Detailed PRR scores for all methods and their LINE counterparts. Metric: Comet WMT22.}
\label{tab:detailed_comet}
\end{table*}

  \FloatBarrier
  \newpage
  \FloatBarrier
  \begin{table*}[h!]
\footnotesize
\centering
\begin{tabular}{lcccccccc}
\toprule
&\multicolumn{4}{c}{\textbf{WMT14}}&\multicolumn{4}{c}{\textbf{WMT19}}\\
\cmidrule(lr){2-5}
\cmidrule(lr){6-9}
&Cs-En&De-En&Ru-En&Fr-En&De-En&Fi-En&Lt-En&Ru-En\\
\midrule
& \multicolumn{8}{c}{Llama 3.1 8B}\\
\midrule
MSP & 0.21 & 0.22 & 0.31 & 0.19 & 0.23 & 0.08 & 0.12 & 0.26 \\
MSP-LINE & 0.40 & 0.42 & 0.44 & 0.31 & 0.40 & 0.41 & 0.38 & 0.35 \\
\midrule
PPL & 0.43 & 0.45 & 0.41 & 0.32 & 0.39 & 0.48 & 0.43 & 0.31 \\
PPL-LINE & 0.48 & 0.47 & \underline{0.49} & 0.36 & 0.40 & 0.48 & 0.44 & 0.40 \\
\midrule
MTE & 0.47 & \underline{0.48} & 0.46 & \underline{0.39} & \underline{0.43} & \textbf{0.52} & \underline{0.49} & 0.36 \\
MTE-LINE & \textbf{0.54} & \textbf{0.51} & \textbf{0.54} & \textbf{0.43} & \textbf{0.47} & \underline{0.51} & \textbf{0.49} & \textbf{0.45} \\
\midrule
MCSE & 0.16 & 0.14 & 0.20 & 0.16 & 0.14 & -0.00 & 0.07 & 0.21 \\
MCSE-LINE & 0.29 & 0.29 & 0.30 & 0.25 & 0.32 & 0.29 & 0.29 & 0.27 \\
\midrule
MCNSE & 0.42 & 0.38 & 0.40 & 0.32 & 0.38 & 0.39 & 0.44 & 0.32 \\
MCNSE-LINE & 0.42 & 0.38 & 0.43 & 0.34 & 0.39 & 0.39 & 0.44 & 0.36 \\
\midrule
LSRL & 0.39 & 0.35 & 0.37 & 0.30 & 0.36 & 0.32 & 0.42 & 0.31 \\
LSRL-LINE & 0.38 & 0.35 & 0.38 & 0.29 & 0.35 & 0.33 & 0.40 & 0.33 \\
\midrule
TokenSAR & 0.46 & 0.45 & 0.41 & 0.34 & 0.39 & 0.49 & 0.44 & 0.32 \\
TokenSAR-LINE & \underline{0.49} & 0.47 & 0.48 & 0.36 & 0.41 & 0.49 & 0.44 & \underline{0.40} \\

\midrule
& \multicolumn{8}{c}{Gemma 2 9B}\\
\midrule
MSP & 0.19 & 0.22 & 0.29 & 0.13 & 0.28 & 0.06 & 0.24 & 0.27 \\
MSP-LINE & 0.39 & 0.45 & 0.41 & 0.29 & \underline{0.45} & 0.35 & 0.35 & 0.39 \\
\midrule
PPL & 0.42 & 0.47 & 0.41 & 0.33 & 0.42 & 0.41 & 0.33 & 0.34 \\
PPL-LINE & 0.43 & \underline{0.48} & \underline{0.43} & 0.33 & 0.43 & 0.41 & 0.34 & 0.37 \\
\midrule
MTE & \underline{0.45} & 0.47 & 0.42 & \underline{0.36} & 0.44 & \textbf{0.45} & 0.34 & 0.35 \\
MTE-LINE & \textbf{0.46} & \textbf{0.49} & \textbf{0.46} & \textbf{0.37} & \textbf{0.47} & \underline{0.45} & \textbf{0.37} & \textbf{0.41} \\
\midrule
MCSE & 0.11 & 0.15 & 0.21 & 0.12 & 0.20 & -0.03 & 0.17 & 0.23 \\
MCSE-LINE & 0.31 & 0.36 & 0.33 & 0.28 & 0.40 & 0.27 & 0.25 & 0.35 \\
\midrule
MCNSE & 0.38 & 0.43 & 0.40 & 0.33 & 0.43 & 0.36 & 0.32 & 0.38 \\
MCNSE-LINE & 0.38 & 0.43 & 0.41 & 0.33 & 0.44 & 0.36 & 0.32 & \underline{0.40} \\
\midrule
LSRL & 0.35 & 0.38 & 0.36 & 0.26 & 0.36 & 0.34 & 0.34 & 0.34 \\
LSRL-LINE & 0.35 & 0.38 & 0.37 & 0.26 & 0.36 & 0.36 & 0.31 & 0.34 \\
\midrule
TokenSAR & 0.42 & 0.46 & 0.39 & 0.33 & 0.42 & 0.41 & 0.31 & 0.32 \\
TokenSAR-LINE & 0.43 & 0.47 & 0.41 & 0.34 & 0.44 & 0.41 & \underline{0.35} & 0.37 \\

\midrule
& \multicolumn{8}{c}{EuroLLM 9B}\\
\midrule
MSP & 0.13 & 0.23 & 0.30 & 0.11 & 0.23 & 0.04 & 0.15 & 0.28 \\
MSP-LINE & 0.32 & 0.44 & 0.45 & 0.27 & 0.43 & 0.34 & 0.31 & 0.38 \\
\midrule
PPL & 0.51 & 0.52 & 0.45 & 0.41 & 0.48 & 0.44 & 0.40 & 0.33 \\
PPL-LINE & 0.52 & 0.53 & 0.46 & 0.42 & 0.48 & 0.44 & 0.42 & 0.37 \\
\midrule
MTE & \underline{0.54} & \underline{0.54} & \textbf{0.48} & \underline{0.46} & \underline{0.50} & \textbf{0.49} & \underline{0.42} & 0.36 \\
MTE-LINE & \textbf{0.55} & \textbf{0.55} & \underline{0.47} & \textbf{0.46} & \textbf{0.51} & \underline{0.47} & \textbf{0.47} & \textbf{0.42} \\
\midrule
MCSE & 0.20 & 0.24 & 0.28 & 0.16 & 0.25 & 0.10 & 0.25 & 0.27 \\
MCSE-LINE & 0.42 & 0.42 & 0.39 & 0.36 & 0.40 & 0.36 & 0.36 & 0.34 \\
\midrule
MCNSE & 0.23 & 0.33 & 0.29 & 0.22 & 0.28 & 0.32 & 0.26 & 0.21 \\
MCNSE-LINE & 0.21 & 0.33 & 0.28 & 0.21 & 0.26 & 0.30 & 0.25 & 0.23 \\
\midrule
LSRL & 0.29 & 0.34 & 0.30 & 0.27 & 0.31 & 0.30 & 0.31 & 0.20 \\
LSRL-LINE & 0.29 & 0.34 & 0.30 & 0.27 & 0.30 & 0.32 & 0.31 & 0.21 \\
\midrule
TokenSAR & 0.44 & 0.49 & 0.45 & 0.39 & 0.43 & 0.43 & 0.35 & 0.35 \\
TokenSAR-LINE & 0.41 & 0.49 & 0.44 & 0.37 & 0.43 & 0.42 & 0.36 & \underline{0.39} \\

\midrule
\end{tabular}
\caption{Detailed PRR scores for all methods and their LINE counterparts. Metric: MetricX XXL.}
\label{tab:detailed_metricx}
\end{table*}

  \FloatBarrier
  \newpage
  \FloatBarrier
  \begin{table*}
\footnotesize
\centering
\begin{tabular}{lcccccccc}
\toprule
&\multicolumn{4}{c}{\textbf{WMT14}}&\multicolumn{4}{c}{\textbf{WMT19}}\\
\cmidrule(lr){2-5}
\cmidrule(lr){6-9}
&Cs-En&De-En&Ru-En&Fr-En&De-En&Fi-En&Lt-En&Ru-En\\
\midrule
& \multicolumn{8}{c}{Llama 3.1 8B}\\
\midrule
MSP & 0.25 & 0.35 & 0.41 & 0.33 & 0.31 & 0.04 & 0.15 & 0.37 \\
MSP-LINE & 0.33 & 0.38 & 0.41 & 0.33 & \underline{0.37} & 0.39 & 0.38 & 0.34 \\
\midrule
PPL & 0.36 & 0.35 & 0.30 & 0.24 & 0.33 & 0.49 & 0.49 & 0.27 \\
PPL-LINE & 0.42 & 0.43 & 0.47 & \underline{0.35} & 0.37 & 0.49 & 0.48 & 0.42 \\
\midrule
MTE & 0.40 & 0.37 & 0.33 & 0.30 & 0.34 & \textbf{0.51} & \textbf{0.53} & 0.32 \\
MTE-LINE & \textbf{0.48} & \textbf{0.47} & \textbf{0.53} & \textbf{0.42} & \textbf{0.44} & 0.49 & \underline{0.52} & \textbf{0.51} \\
\midrule
MCSE & 0.20 & 0.29 & 0.33 & 0.29 & 0.24 & -0.04 & 0.06 & 0.31 \\
MCSE-LINE & 0.26 & 0.27 & 0.28 & 0.24 & 0.29 & 0.27 & 0.28 & 0.28 \\
\midrule
MCNSE & 0.36 & 0.34 & 0.32 & 0.27 & 0.33 & 0.38 & 0.43 & 0.32 \\
MCNSE-LINE & 0.36 & 0.35 & 0.39 & 0.31 & 0.35 & 0.38 & 0.42 & 0.40 \\
\midrule
LSRL & 0.34 & 0.35 & 0.29 & 0.27 & 0.30 & 0.30 & 0.32 & 0.30 \\
LSRL-LINE & 0.33 & 0.31 & 0.38 & 0.24 & 0.25 & 0.33 & 0.34 & 0.33 \\
\midrule
TokenSAR & 0.38 & 0.34 & 0.31 & 0.27 & 0.32 & \underline{0.50} & 0.48 & 0.27 \\
TokenSAR-LINE & \underline{0.43} & \underline{0.43} & \underline{0.49} & 0.34 & 0.37 & 0.49 & 0.48 & \underline{0.43} \\

\midrule
& \multicolumn{8}{c}{Gemma 2 9B}\\
\midrule
MSP & 0.20 & 0.35 & 0.39 & 0.27 & 0.34 & 0.00 & 0.15 & 0.35 \\
MSP-LINE & 0.29 & \underline{0.38} & 0.38 & 0.29 & \textbf{0.38} & 0.29 & 0.22 & 0.33 \\
\midrule
PPL & 0.32 & 0.34 & 0.29 & 0.25 & 0.33 & 0.37 & 0.28 & 0.27 \\
PPL-LINE & 0.33 & 0.37 & 0.45 & \underline{0.30} & 0.35 & 0.37 & 0.28 & 0.33 \\
\midrule
MTE & \underline{0.35} & 0.33 & 0.29 & 0.25 & 0.32 & \textbf{0.42} & 0.29 & 0.27 \\
MTE-LINE & \textbf{0.37} & \textbf{0.38} & \textbf{0.48} & \textbf{0.34} & \underline{0.37} & \underline{0.40} & \underline{0.31} & \textbf{0.37} \\
\midrule
MCSE & 0.15 & 0.30 & 0.34 & 0.26 & 0.27 & -0.07 & 0.09 & 0.32 \\
MCSE-LINE & 0.23 & 0.32 & 0.34 & 0.26 & 0.33 & 0.20 & 0.12 & 0.32 \\
\midrule
MCNSE & 0.28 & 0.34 & 0.33 & 0.25 & 0.32 & 0.29 & 0.19 & 0.33 \\
MCNSE-LINE & 0.28 & 0.33 & 0.40 & 0.27 & 0.33 & 0.29 & 0.19 & \underline{0.36} \\
\midrule
LSRL & 0.26 & 0.30 & 0.30 & 0.20 & 0.28 & 0.29 & 0.19 & 0.27 \\
LSRL-LINE & 0.25 & 0.28 & 0.38 & 0.19 & 0.26 & 0.32 & 0.16 & 0.29 \\
\midrule
TokenSAR & 0.32 & 0.31 & 0.28 & 0.23 & 0.31 & 0.40 & 0.30 & 0.26 \\
TokenSAR-LINE & 0.34 & 0.36 & \underline{0.45} & 0.30 & 0.35 & 0.39 & \textbf{0.32} & 0.33 \\

\midrule
& \multicolumn{8}{c}{EuroLLM 9B}\\
\midrule
MSP & 0.13 & 0.31 & 0.38 & 0.21 & 0.27 & -0.03 & 0.06 & 0.36 \\
MSP-LINE & 0.24 & 0.38 & 0.42 & 0.25 & 0.38 & 0.26 & 0.19 & 0.37 \\
\midrule
PPL & 0.39 & 0.40 & 0.38 & 0.33 & 0.43 & 0.40 & 0.33 & 0.31 \\
PPL-LINE & 0.41 & \underline{0.43} & \underline{0.50} & \underline{0.37} & \underline{0.44} & 0.38 & 0.34 & 0.39 \\
\midrule
MTE & \underline{0.43} & 0.42 & 0.39 & 0.35 & 0.43 & \textbf{0.45} & \textbf{0.39} & 0.32 \\
MTE-LINE & \textbf{0.46} & \textbf{0.46} & \textbf{0.51} & \textbf{0.42} & \textbf{0.47} & \underline{0.42} & \underline{0.38} & \textbf{0.44} \\
\midrule
MCSE & 0.19 & 0.31 & 0.35 & 0.26 & 0.30 & -0.01 & 0.08 & 0.34 \\
MCSE-LINE & 0.31 & 0.36 & 0.38 & 0.33 & 0.38 & 0.28 & 0.20 & 0.32 \\
\midrule
MCNSE & 0.19 & 0.28 & 0.23 & 0.17 & 0.26 & 0.30 & 0.22 & 0.21 \\
MCNSE-LINE & 0.18 & 0.28 & 0.29 & 0.17 & 0.25 & 0.28 & 0.19 & 0.26 \\
\midrule
LSRL & 0.23 & 0.28 & 0.22 & 0.23 & 0.28 & 0.24 & 0.18 & 0.18 \\
LSRL-LINE & 0.23 & 0.28 & 0.26 & 0.23 & 0.28 & 0.26 & 0.18 & 0.22 \\
\midrule
TokenSAR & 0.34 & 0.38 & 0.37 & 0.30 & 0.37 & 0.40 & 0.30 & 0.34 \\
TokenSAR-LINE & 0.34 & 0.41 & 0.49 & 0.32 & 0.38 & 0.39 & 0.30 & \underline{0.41} \\

\midrule
\end{tabular}
\caption{Detailed PRR scores for all methods and their LINE counterparts. Metric: XComet XXL.}
\label{tab:detailed_xcomet}
\end{table*}
  \FloatBarrier
  \newpage
  \FloatBarrier

  \begin{table*}[h!]
\footnotesize
\centering
\begin{tabular}{lcc}
\toprule
& \textbf{XSum} & \textbf{GSM8k} \\
\midrule
& \multicolumn{2}{c}{Llama 3.1 8B} \\
\midrule
MSP & 0.33 & 0.32 \\
MSP-LINE & 0.36 & 0.33 \\
\midrule
PPL & \underline{0.37} & 0.30 \\
PPL-LINE & 0.37 & \underline{0.38} \\
\midrule
MTE & 0.36 & 0.34 \\
MTE-LINE & 0.35 & \textbf{0.40} \\
\midrule
MCSE & 0.03 & 0.35 \\
MCSE-LINE & 0.04 & 0.35 \\
\midrule
MCNSE & 0.02 & 0.34 \\
MCNSE-LINE & 0.03 & 0.36 \\
\midrule
LSRL & 0.09 & 0.36 \\
LSRL-LINE & 0.10 & 0.36 \\
\midrule
TokenSAR & \textbf{0.37} & 0.30 \\
TokenSAR-LINE & 0.37 & 0.38 \\

\midrule
& \multicolumn{2}{c}{Gemma 2 9B} \\
\midrule
MSP & 0.35 & 0.30 \\
MSP-LINE & \textbf{0.38} & 0.30 \\
\midrule
PPL & 0.35 & 0.25 \\
PPL-LINE & \underline{0.37} & 0.36 \\
\midrule
MTE & 0.33 & 0.29 \\
MTE-LINE & 0.36 & \underline{0.40} \\
\midrule
MCSE & 0.00 & 0.39 \\
MCSE-LINE & 0.03 & \textbf{0.40} \\
\midrule
MCNSE & 0.02 & 0.36 \\
MCNSE-LINE & 0.03 & 0.37 \\
\midrule
LSRL & 0.04 & 0.39 \\
LSRL-LINE & 0.09 & 0.39 \\
\midrule
TokenSAR & 0.32 & 0.24 \\
TokenSAR-LINE & 0.33 & 0.36 \\

\midrule
\end{tabular}
\caption{Detailed PRR scores for all methods and their LINE counterparts. Metrics: AlignScore (XSum) and Accuracy (GSM8k).}
\label{tab:detailed_gsm_xsum}
\end{table*}
  \FloatBarrier

 \newpage 

\section{Ablation}

\subsection{Polynomial Detrending}
\label{sec:poly_detr}
  Tables~\ref{tab:comet_poly}, \ref{tab:xcomet_poly}, \ref{tab:metricx_poly} contain comparison in PRR scores between first, second and third degree LINE correction to the considered base UQ methods.

  \begin{table*}[h!]
\footnotesize
\centering
\begin{tabular}{lcccccccc}
\toprule
&\multicolumn{4}{c}{\textbf{WMT14}}&\multicolumn{4}{c}{\textbf{WMT19}}\\
\cmidrule(lr){2-5}
\cmidrule(lr){6-9}
&Cs-En&De-En&Ru-En&Fr-En&De-En&Fi-En&Lt-En&Ru-En\\
\midrule
& \multicolumn{8}{c}{Llama 3.1 8B}\\
\midrule
MSP & 0.42 & 0.39 & 0.45 & 0.35 & 0.46 & 0.19 & 0.29 & 0.43 \\
$\text{MSP-LINE}_1$ & 0.47 & 0.49 & 0.48 & 0.40 & 0.51 & 0.47 & 0.47 & 0.41 \\
$\text{MSP-LINE}_2$ & 0.50 & 0.52 & 0.52 & 0.41 & 0.51 & 0.50 & 0.47 & 0.47 \\
$\text{MSP-LINE}_3$ & 0.53 & 0.53 & 0.51 & 0.34 & 0.53 & 0.48 & 0.47 & 0.48 \\
\midrule
PPL & 0.42 & 0.46 & 0.37 & 0.31 & 0.41 & 0.52 & 0.47 & 0.32 \\
$\text{PPL-LINE}_1$ & 0.52 & 0.51 & 0.53 & 0.41 & 0.46 & 0.52 & 0.49 & 0.46 \\
$\text{PPL-LINE}_2$ & 0.42 & 0.47 & 0.42 & 0.32 & 0.44 & 0.53 & 0.48 & 0.39 \\
$\text{PPL-LINE}_3$ & 0.54 & 0.52 & 0.51 & \underline{0.42} & 0.48 & 0.54 & 0.49 & 0.45 \\
\midrule
MTE & 0.44 & 0.48 & 0.41 & 0.37 & 0.42 & 0.54 & 0.52 & 0.33 \\
$\text{MTE-LINE}_1$ & \underline{0.58} & \underline{0.56} & \textbf{0.59} & \textbf{0.48} & \textbf{0.55} & \textbf{0.56} & \textbf{0.56} & \textbf{0.53} \\
$\text{MTE-LINE}_2$ & 0.46 & 0.51 & 0.48 & 0.38 & \underline{0.51} & 0.53 & 0.53 & 0.45 \\
$\text{MTE-LINE}_3$ & \textbf{0.59} & \textbf{0.57} & \underline{0.57} & \textbf{0.48} & \textbf{0.55} & \underline{0.55} & \underline{0.55} & \underline{0.51} \\
\midrule
MCSE & 0.36 & 0.32 & 0.35 & 0.30 & 0.36 & 0.08 & 0.20 & 0.36 \\
$\text{MCSE-LINE}_1$ & 0.38 & 0.36 & 0.33 & 0.28 & 0.38 & 0.32 & 0.36 & 0.32 \\
$\text{MCSE-LINE}_2$ & 0.42 & 0.40 & 0.36 & 0.28 & 0.39 & 0.32 & 0.35 & 0.32 \\
$\text{MCSE-LINE}_3$ & 0.48 & 0.40 & 0.39 & 0.28 & 0.42 & 0.31 & 0.36 & 0.38 \\
\midrule
MCNSE & 0.48 & 0.44 & 0.40 & 0.36 & 0.43 & 0.46 & 0.48 & 0.35 \\
$\text{MCNSE-LINE}_1$ & 0.49 & 0.44 & 0.47 & 0.39 & 0.45 & 0.46 & 0.48 & 0.44 \\
$\text{MCNSE-LINE}_2$ & 0.42 & 0.46 & 0.37 & 0.30 & 0.42 & 0.45 & 0.47 & 0.37 \\
$\text{MCNSE-LINE}_3$ & 0.53 & 0.46 & 0.46 & 0.40 & 0.47 & 0.47 & 0.49 & 0.43 \\
\midrule
LSRL & 0.45 & 0.44 & 0.38 & 0.35 & 0.46 & 0.37 & 0.42 & 0.35 \\
$\text{LSRL-LINE}_1$ & 0.41 & 0.41 & 0.44 & 0.32 & 0.40 & 0.37 & 0.40 & 0.38 \\
$\text{LSRL-LINE}_2$ & 0.45 & 0.44 & 0.31 & 0.29 & 0.42 & 0.39 & 0.41 & 0.31 \\
$\text{LSRL-LINE}_3$ & 0.46 & 0.38 & 0.43 & 0.35 & 0.44 & 0.39 & 0.41 & 0.37 \\
\midrule
TokenSAR & 0.44 & 0.45 & 0.37 & 0.35 & 0.40 & 0.52 & 0.46 & 0.32 \\
$\text{TokenSAR-LINE}_1$  & 0.51 & 0.52 & 0.52 & 0.41 & 0.47 & 0.53 & 0.49 & 0.46 \\
$\text{TokenSAR-LINE}_2$ & 0.39 & 0.47 & 0.41 & 0.31 & 0.44 & 0.54 & 0.47 & 0.38 \\
$\text{TokenSAR-LINE}_3$  & 0.52 & 0.47 & 0.51 & 0.41 & 0.48 & 0.54 & 0.48 & 0.45 \\
\midrule
\end{tabular}
\caption{PRR scores with linear and polynomial detrending -- Comet WMT22.}
\label{tab:comet_poly}
\end{table*}
  \newpage
  \FloatBarrier
  \begin{table*}
\footnotesize
\centering
\begin{tabular}{lcccccccc}
\toprule
&\multicolumn{4}{c}{\textbf{WMT14}}&\multicolumn{4}{c}{\textbf{WMT19}}\\
\cmidrule(lr){2-5}
\cmidrule(lr){6-9}
&Cs-En&De-En&Ru-En&Fr-En&De-En&Fi-En&Lt-En&Ru-En\\
\midrule
& \multicolumn{8}{c}{Llama 3.1 8B}\\
\midrule
MSP & 0.25 & 0.35 & 0.41 & 0.33 & 0.31 & 0.04 & 0.15 & 0.37 \\
$\text{MSP-LINE}_1$ & 0.33 & 0.38 & 0.41 & 0.33 & 0.37 & 0.39 & 0.38 & 0.34 \\
$\text{MSP-LINE}_2$ & 0.36 & 0.43 & 0.47 & 0.35 & 0.37 & 0.42 & 0.37 & 0.39 \\
$\text{MSP-LINE}_3$ & 0.37 & 0.43 & 0.46 & 0.29 & 0.39 & 0.39 & 0.38 & 0.41 \\
\midrule
PPL & 0.36 & 0.35 & 0.30 & 0.24 & 0.33 & \underline{0.49} & 0.49 & 0.27 \\
$\text{PPL-LINE}_1$ & 0.42 & 0.43 & 0.47 & 0.35 & 0.37 & \underline{0.49} & 0.48 & 0.42 \\
$\text{PPL-LINE}_2$ & 0.34 & 0.37 & 0.35 & 0.26 & 0.34 & 0.50 & 0.45 & 0.32 \\
$\text{PPL-LINE}_3$ & 0.43 & 0.44 & 0.45 & 0.35 & 0.38 & \underline{0.49} & 0.46 & 0.39 \\
\midrule
MTE & 0.40 & 0.37 & 0.33 & 0.30 & 0.34 & \textbf{0.51} & \textbf{0.53} & 0.32 \\
$\text{MTE-LINE}_1$ & \textbf{0.48} & \underline{0.47} & \textbf{0.53} & \textbf{0.42} & \textbf{0.44} & 0.49 & \underline{0.52} & \textbf{0.51} \\
$\text{MTE-LINE}_2$ & 0.38 & 0.40 & 0.38 & 0.33 & 0.39 & 0.46 & 0.48 & 0.39 \\
$\text{MTE-LINE}_3$ & \underline{0.46} & \textbf{0.48} & 0.48 & \underline{0.41} & \underline{0.43} & 0.46 & 0.49 & \underline{0.45} \\
\midrule
MCSE & 0.20 & 0.29 & 0.33 & 0.29 & 0.24 & -0.04 & 0.06 & 0.31 \\
$\text{MCSE-LINE}_1$ & 0.26 & 0.27 & 0.28 & 0.24 & 0.29 & 0.27 & 0.28 & 0.28 \\
$\text{MCSE-LINE}_2$ & 0.30 & 0.33 & 0.31 & 0.24 & 0.28 & 0.28 & 0.26 & 0.29 \\
$\text{MCSE-LINE}_3$ & 0.32 & 0.33 & 0.34 & 0.24 & 0.30 & 0.24 & 0.27 & 0.33 \\
\midrule
MCNSE & 0.36 & 0.34 & 0.32 & 0.27 & 0.33 & 0.38 & 0.43 & 0.32 \\
$\text{MCNSE-LINE}_1$ & 0.36 & 0.35 & 0.39 & 0.31 & 0.35 & 0.38 & 0.42 & 0.40 \\
$\text{MCNSE-LINE}_2$ & 0.31 & 0.39 & 0.27 & 0.23 & 0.31 & 0.38 & 0.41 & 0.31 \\
$\text{MCNSE-LINE}_3$ & 0.39 & 0.39 & 0.36 & 0.32 & 0.36 & 0.38 & 0.42 & 0.38 \\
\midrule
LSRL & 0.34 & 0.35 & 0.29 & 0.27 & 0.30 & 0.30 & 0.32 & 0.30 \\
$\text{LSRL-LINE}_1$ & 0.33 & 0.31 & 0.38 & 0.24 & 0.25 & 0.33 & 0.34 & 0.33 \\
$\text{LSRL-LINE}_2$ & 0.35 & 0.35 & 0.20 & 0.21 & 0.27 & 0.35 & 0.35 & 0.25 \\
$\text{LSRL-LINE}_3$ & 0.35 & 0.27 & 0.32 & 0.27 & 0.28 & 0.34 & 0.35 & 0.32 \\
\midrule
TokenSAR & 0.38 & 0.34 & 0.31 & 0.27 & 0.32 & \underline{0.50} & 0.48 & 0.27 \\
$\text{TokenSAR-LINE}_1$  & 0.43 & 0.43 & \underline{0.49} & 0.34 & 0.37 & 0.49 & 0.48 & 0.43 \\
$\text{TokenSAR-LINE}_2$  & 0.33 & 0.36 & 0.36 & 0.25 & 0.33 & 0.50 & 0.44 & 0.33 \\
$\text{TokenSAR-LINE}_3$  & 0.41 & 0.36 & 0.46 & 0.34 & 0.38 & 0.49 & 0.45 & 0.39 \\

\midrule
\end{tabular}
\caption{PRR scores with linear and polynomial detrending -- XComet XXL.}
\label{tab:xcomet_poly}
\end{table*}
    \FloatBarrier
  \newpage
  \FloatBarrier
  \begin{table*}[h!]
\footnotesize
\centering
\begin{tabular}{lcccccccc}
\toprule
&\multicolumn{4}{c}{\textbf{WMT14}}&\multicolumn{4}{c}{\textbf{WMT19}}\\
\cmidrule(lr){2-5}
\cmidrule(lr){6-9}
&Cs-En&De-En&Ru-En&Fr-En&De-En&Fi-En&Lt-En&Ru-En\\
\midrule
& \multicolumn{8}{c}{Llama 3.1 8B}\\
\midrule
MSP & 0.21 & 0.22 & 0.31 & 0.19 & 0.23 & 0.08 & 0.12 & 0.26 \\
$\text{MSP-LINE}_1$ & 0.40 & 0.42 & 0.44 & 0.31 & 0.40 & 0.41 & 0.38 & 0.35 \\
$\text{MSP-LINE}_2$ & 0.42 & 0.44 & 0.46 & 0.32 & 0.40 & 0.44 & 0.38 & 0.39 \\
$\text{MSP-LINE}_3$ & 0.43 & 0.44 & 0.46 & 0.29 & 0.41 & 0.41 & 0.38 & 0.39 \\
\midrule
PPL & 0.43 & 0.45 & 0.41 & 0.32 & 0.39 & 0.48 & 0.43 & 0.31 \\
$\text{PPL-LINE}_1$ & 0.48 & 0.47 & 0.49 & 0.36 & 0.40 & 0.48 & 0.44 & \underline{0.40} \\
$\text{PPL-LINE}_2$ & 0.40 & 0.46 & 0.41 & 0.29 & 0.40 & 0.49 & 0.42 & 0.35 \\
$\text{PPL-LINE}_3$ & 0.49 & 0.48 & 0.48 & 0.35 & 0.41 & 0.48 & 0.45 & 0.39 \\
\midrule
MTE & 0.47 & 0.48 & 0.46 & 0.39 & 0.43 & \textbf{0.52} & \underline{0.49} & 0.36 \\
$\text{MTE-LINE}_1$ & \textbf{0.54} & \underline{0.51} & \textbf{0.54} & \textbf{0.43} & \textbf{0.47} & \underline{0.51} & \underline{0.49} & \textbf{0.45} \\
$\text{MTE-LINE}_2$ & 0.45 & 0.50 & 0.47 & 0.36 & \underline{0.46} & 0.49 & 0.47 & \underline{0.40} \\
$\text{MTE-LINE}_3$ & \underline{0.53} & \textbf{0.52} & \textbf{0.54} & \underline{0.42} & \textbf{0.47} & 0.49 & \textbf{0.50} & \textbf{0.45} \\
\midrule
MCSE & 0.16 & 0.14 & 0.20 & 0.16 & 0.14 & -0.00 & 0.07 & 0.21 \\
$\text{MCSE-LINE}_1$ & 0.29 & 0.29 & 0.30 & 0.25 & 0.32 & 0.29 & 0.29 & 0.27 \\
$\text{MCSE-LINE}_2$ & 0.33 & 0.31 & 0.33 & 0.25 & 0.31 & 0.30 & 0.28 & 0.28 \\
$\text{MCSE-LINE}_3$ & 0.36 & 0.30 & 0.36 & 0.25 & 0.32 & 0.28 & 0.29 & 0.31 \\
\midrule
MCNSE & 0.42 & 0.38 & 0.40 & 0.32 & 0.38 & 0.39 & 0.44 & 0.32 \\
$\text{MCNSE-LINE}_1$ & 0.42 & 0.38 & 0.43 & 0.34 & 0.39 & 0.39 & 0.44 & 0.36 \\
$\text{MCNSE-LINE}_2$ & 0.38 & 0.39 & 0.35 & 0.28 & 0.38 & 0.39 & 0.43 & 0.31 \\
$\text{MCNSE-LINE}_3$ & 0.45 & 0.39 & 0.43 & 0.34 & 0.39 & 0.40 & 0.46 & 0.35 \\
\midrule
LSRL & 0.39 & 0.35 & 0.37 & 0.30 & 0.36 & 0.32 & 0.42 & 0.31 \\
$\text{LSRL-LINE}_1$ & 0.38 & 0.35 & 0.38 & 0.29 & 0.35 & 0.33 & 0.40 & 0.33 \\
$\text{LSRL-LINE}_2$ & 0.40 & 0.36 & 0.28 & 0.27 & 0.35 & 0.35 & 0.42 & 0.28 \\
$\text{LSRL-LINE}_3$ & 0.40 & 0.34 & 0.38 & 0.31 & 0.35 & 0.34 & 0.42 & 0.32 \\
\midrule
TokenSAR & 0.46 & 0.45 & 0.41 & 0.34 & 0.39 & 0.49 & 0.44 & 0.32 \\
$\text{TokenSAR-LINE}_1$ & 0.49 & 0.47 & 0.48 & 0.36 & 0.41 & 0.49 & 0.44 & 0.40 \\
$\text{TokenSAR-LINE}_2$  & 0.40 & 0.45 & 0.41 & 0.29 & 0.40 & 0.50 & 0.42 & 0.35 \\
$\text{TokenSAR-LINE}_3$ & 0.48 & 0.45 & 0.48 & 0.35 & 0.41 & 0.49 & 0.46 & 0.39 \\
\midrule
\end{tabular}
\caption{PRR scores with linear and polynomial detrending -- MetricX XXL.}
\label{tab:metricx_poly}
\end{table*}

  \FloatBarrier

\newpage
  
\subsection{Reducing number of quality labels}
\label{sec:small_quality_labels}
  Obtaining quality labels can be expensive in certain setups. To address this issue, we estimate the quality (metric)-vs–length regression (equation~\eqref{eq:debiasing:qualitytrend}) using a \emph{small, length-balanced} subset of the training data, rather than the full sample. The goal is to recover an effective quality trend with far fewer labels.

  We first remove length outliers by keeping the 5th–95th percentiles and rescale lengths to $[0,1]$. The length axis is then partitioned into $n$ adaptive bins via K-means (narrower in dense regions, wider in sparse ones). From each bin we sample about $S/n$ items without replacement, where $S$ is the target labeled size; bins with fewer items contribute all their points, and the shortfall is redistributed across the others. Finally, we fit a metric–vs–length regression on this subset and apply it to remove the length trend at test time.

  \begin{table*}[h!]
    \footnotesize
    \centering
    \scalebox{0.9}{
    \begin{tabular}{lcc}
      \toprule
       & \textbf{\gsm } & \textbf{\xsum} \\
      \midrule
      & \multicolumn{2}{c}{Llama 3.1 8B} \\
      \midrule
      MSP & 0.32 & 0.33 \\
      MSP-LINE (500 sample) & 0.33 & 0.36\\
      MSP-LINE (Full sample) & 0.33 & 0.36 \\
      \midrule
      PPL & 0.30 & \textbf{0.37} \\
      PPL-LINE (500 sample) & 0.38 & \underline{0.37} \\
      PPL-LINE (Full sample) & 0.38 & \underline{0.37} \\
      \midrule
      MTE & 0.34 & 0.36 \\
      MTE-LINE (500 sample) & \underline{0.39} & 0.35 \\
      MTE-LINE (Full sample) & \textbf{0.40} & 0.35 \\
      \midrule
      MCSE & 0.35 & 0.03 \\
      MCSE-LINE (500 sample) & 0.35 & 0.04 \\
      MCSE-LINE (Full sample) & 0.35 & 0.04 \\
      \midrule
      MCNSE & 0.34 & 0.02 \\
      MCNSE-LINE (500 sample) & 0.36 & 0.03 \\
      MCNSE-LINE (Full sample) & 0.36 & 0.03 \\
      \midrule
      LSRL & 0.36 & 0.09 \\
      LSRL-LINE (500 sample) & 0.36 & 0.10 \\
      LSRL-LINE (Full sample) & 0.36 & 0.10 \\
      \midrule
      TokenSAR	& 0.30	& 0.37\\
      TokenSAR-LINE (500 sample) &	0.38 &	0.36 \\
      TokenSAR-LINE (Full sample)	& 0.38 &	0.37 \\
      \midrule
      & \multicolumn{2}{c}{Gemma 2 9B} \\
      \midrule
      MSP & 0.30 & 0.35 \\
      MSP-LINE (500 sample) & 0.30 & \textbf{0.38} \\
      MSP-LINE (Full sample) & 0.30 & \textbf{0.38} \\
      \midrule
      PPL & 0.25 & 0.35 \\
      PPL-LINE (500 sample) & 0.36 & 0.37 \\
      PPL-LINE (Full sample) & 0.36 & \underline{0.37} \\
      \midrule
      MTE & 0.29 & 0.33 \\
      MTE-LINE (500 sample) & \textbf{0.40} & 0.36 \\
      MTE-LINE (Full sample) & \textbf{0.40} & 0.36 \\
      \midrule
      MCSE & 0.39 & 0.00 \\
      MCSE-LINE (500 sample) & \underline{0.40} & 0.03 \\
      MCSE-LINE (Full sample) & \textbf{0.40} & 0.03 \\
      \midrule
      MCNSE & 0.36 & 0.02 \\
      MCNSE-LINE (500 sample) & 0.37 & 0.03 \\
      MCNSE-LINE (Full sample) & 0.37 & 0.03 \\
      \midrule
      LSRL & 0.39 & 0.04 \\
      LSRL-LINE (500 sample) & 0.39 & 0.09 \\
      LSRL-LINE (Full sample) & 0.39 & 0.09 \\
      \midrule
      TokenSAR	& 0.24	& 0.32 \\
      TokenSAR-LINE (500 sample)	& 0.35 &	0.33 \\
      TokenSAR-LINE (Full sample) &	0.36 &	0.33 \\
      \bottomrule
    \end{tabular}}
    \caption{PRR scores on GSM8K and XSum using 500 samples for quality trend fitting.}
  \end{table*}

\end{document}